\documentclass[11pt, twoside, openright]{thesis}

\usepackage{psl-cover}
\usepackage{lmodern}
\definecolor{maincolor}{RGB}{36, 56, 141}
\colorlet{secondarycolor}{violet}

\usepackage{amsmath,amsfonts,bm}

\def\figref#1{figure~\ref{#1}}

\def\eqref#1{equation~\ref{#1}}

\def\1{\bm{1}}

\def\va{{\bm{a}}}

\def\vp{{\bm{p}}}

\def\vr{{\bm{r}}}
\def\vs{{\bm{s}}}

\def\vx{{\bm{x}}}
\def\vy{{\bm{y}}}

\def\eva{{a}}

\def\mW{{\bm{W}}}

\DeclareMathAlphabet{\mathsfit}{\encodingdefault}{\sfdefault}{m}{sl}
\SetMathAlphabet{\mathsfit}{bold}{\encodingdefault}{\sfdefault}{bx}{n}

\def\sA{{\mathbb{A}}}

\usepackage[utf8]{inputenc}
\usepackage[margin=0.5cm,labelfont=bf]{caption}
\usepackage{subcaption}
\usepackage{graphicx}
\usepackage[style=authoryear-comp,
url=false,
backref=true,
maxbibnames=12,
]{biblatex}

\DefineBibliographyStrings{english}{%
  backrefpage = {Cited on page},
  backrefpages = {Cited on pages},
}

\usepackage{setspace}

\usepackage{bold-extra}
\usepackage{amsmath}
\usepackage{csquotes}
\usepackage{amssymb}
\usepackage{dsfont}
\usepackage{array}
\usepackage{tikz}
\usepackage{booktabs}
\usepackage{nameref}
\usepackage{mathtools}
\usepackage[stretch=10]{microtype}
\usepackage{acronym}
\usepackage{xcolor}
\usepackage{colortbl}
\usepackage{multirow}
\usepackage{xspace}
\usepackage{url}
\usepackage{xhfill}
\usepackage[version=4]{mhchem}
\usepackage{anyfontsize}
\newcommand{\eg}{e.g.,\xspace}

\newcommand{\etc}{etc.\@\xspace}
\newcommand{\ie}{i.e.,\xspace}

\usepackage[symbol]{footmisc}

\usepackage{fancyhdr}
\usepackage{hyperref}
\usepackage{cleveref}
\hypersetup{
    colorlinks=true,
    linkcolor=red,
    filecolor=magenta,
    urlcolor=blue,
    citecolor=purple,
    pdfauthor={Hugo Cisneros},pdftitle={Unsupervised Learning in Complex
  Systems},pdfsubject={PhD thesis of Hugo Cisneros},pdfkeywords={Machine
  learning, Artificial Intelligence, Complex systems,
  Complexity},pdfproducer={LaTeX},pdfcreator={LaTeX},
    pdfpagemode=FullScreen,
    }

\acrodef{WADE}[WADE]{Weighted Average Data Efficiency}
\acrodef{RNN}[RNN]{recurrent neural network}
\newacro{ESN}{echo-state network}
\newacro{DS}{dynamical system}
\newacro{CA}{cellular automaton}
\newacroplural{CA}[CAs]{Cellular automata}
\newacro{ECA}{elementary cellular automaton}
\newacroplural{ECA}[ECAs]{Elementary cellular automata}

\newacro{NCA}{neural cellular automaton}
\newacroplural{NCA}[NCAs]{neural cellular automata}
\newacro{CNN}{convolutional neural network}
\newacro{RBN}{random boolean network}
\newacro{OEE}{open-ended evolution}
\newacro{AC}{artificial chemistry}
\newacroplural{AC}[AC]{artificial chemistries}
\newacro{NS}{novelty search}
\newacro{RC}{reservoir computing}

\AtBeginEnvironment{thebibliography}{\linespread{1}\selectfont}

\urldef\projecturl\url{https://hugocisneros.com/ALIFE-Paper-2020/}

\title{Unsupervised Learning in Complex Systems}
\institute{L'École normale supérieure de Paris}
\doctoralschool{Sciences Mathématiques de\\ Paris Centre}{386}
\specialty{Informatique}
\author{Hugo Cisneros}
\date{15 Mai 2022}

\jurymember{1}{Lenka Zdeborová}{École Polytechnique Fédérale de Lausanne}{Président du jury}
\jurymember{2}{Stefano Nichele}{Østfold University College}{Rapporteur}
\jurymember{3}{Hiroki Sayama}{Binghamton University}{Rapporteur}
\jurymember{5}{Josef Sivic}{École Normale Supérieure, Inria\\CIIRC, Czech Technical University}{Directeur de thèse}
\jurymember{6}{Tomas Mikolov}{CIIRC, Czech Technical University}{Co-encadrant de thèse}

\frabstract{

  Dans cette thèse, nous explorons l'utilisation de systèmes complexes pour
  étudier l'apprentissage et l'adaptation dans les systèmes naturels et
  artificiels. L'objectif est de développer des systèmes autonomes capables
  d'apprendre sans supervision, de se développer de manière autonome et de
  devenir de plus en plus complexes avec le temps. Les systèmes complexes
  apparaissent comme un cadre adapté pour comprendre ces phénomènes en raison de
  leur capacité à croître en complexité. Être en mesure de construire des
  algorithmes d'apprentissage qui nécessitent peu ou pas de supervision
  permettrait une plus grande flexibilité et adaptabilité dans diverses
  applications. En tentant de comprendre les principes fondamentaux de
  l'apprentissage dans les systèmes complexes, nous espérons améliorer notre
  capacité à concevoir et à mettre en œuvre des algorithmes d'apprentissage à
  l'avenir. Cette thèse apporte les contributions suivantes : le développement
  d'une métrique de complexité générale que nous appliquons pour rechercher des
  systèmes complexes qui présentent une croissance de complexité, l'introduction
  d'une méthode pour étudier les calculs dans des systèmes complexes à grande
  échelle, et le développement d'une métrique pour l'efficacité de
  l'apprentissage ainsi que l'introduction d'un jeu de données de référence pour
  évaluer la vitesse des algorithmes d'apprentissage.
Nos résultats contribuent significativement à notre compréhension de l'apprentissage 
et de l'adaptation dans les systèmes naturels et artificiels. 
 De plus, notre approche contribue à une
  nouvelle direction prometteuse pour la recherche dans ce domaine. 
Nous espérons que ces 
résultats inspireront le développement d'algorithmes d'apprentissage plus efficaces 
et plus performants à l'avenir.
}

\enabstract{

  In this thesis, we explore the use of complex systems to study learning and
  adaptation in natural and artificial systems. The goal is to develop
  autonomous systems that can learn without supervision, develop on their own,
  and become increasingly complex over time. Complex systems are identified as a
  suitable framework for understanding these phenomena due to their ability to
  exhibit growth of complexity.
  Being
  able to build learning algorithms that require limited to no supervision would
  enable greater flexibility and adaptability in various applications. By
  understanding the fundamental principles of learning in complex systems, we
  hope to advance our ability to design and implement practical learning
  algorithms in the future. This thesis makes the following key contributions:
  the development of a general complexity metric that we apply to search for
  complex systems that exhibit growth of complexity, the introduction of a
  coarse-graining method to study computations in large-scale complex systems,
  and the development of a metric for learning efficiency as well as a benchmark
  dataset for evaluating the speed of learning algorithms. Our findings add substantially to our
  understanding of learning and adaptation in natural and artificial systems. Moreover, our approach contributes to a
  promising new direction for research in this area. We hope these 
  findings will inspire the development of more effective and efficient learning algorithms in the future.}

\frkeywords{Apprentissage automatique {\Large\color{maincolor}$\star$} Intelligence
  artificielle {\Large\color{maincolor}$\star$} Systèmes complexes
  {\Large\color{maincolor}$\star$} Complexité}

\enkeywords{ Machine learning {\Large\color{maincolor}$\star$} Artificial
  Intelligence {\Large\color{maincolor}$\star$} Complex systems
  {\Large\color{maincolor}$\star$} Complexity}

\ifprintversion
\else
\fi

\begin{document}
\frontmatter
\hypersetup{pageanchor=false}
\maketitle
\hypersetup{pageanchor=true}


\ifprintversion
    \setboolean{@twoside}{true}
    \pagestyle{fancy}
    \fancyhead{} 
    \fancyhead[ER,OL]{\footnotesize \rightmark} \fancyhead[EL,OR]{\thepage}
    \fancyfoot{} 
    \fancyfoot[C]{---\quad \textbf{H. Cisneros: Unsupervised Learning in
        Complex Systems}\quad ---}
\else
    \setboolean{@twoside}{false}
    \pagestyle{fancy}
    \fancyhead{} 
    \fancyhead[R]{\footnotesize \rightmark} \fancyhead[L]{\thepage}
    \fancyfoot{} 
    \fancyfoot[C]{---\quad \textbf{H. Cisneros: Unsupervised Learning in
        Complex Systems}\quad ---}
\fi

\cleardoublepage
\chapter*{Résumé}
\addstarredchapter{Résumé}
\selectlanguage{french}

\thefrabstract{}
\vfill
\thefrkeywords{}
\cleardoublepage
\chapter*{Abstract}
\addstarredchapter{Abstract}
\selectlanguage{english}
\theenabstract{}
\vfill
\theenkeywords{}

\cleardoublepage
\chapter*{Acknowledgments}
\addstarredchapter{Acknowledgments}

I express my sincere gratitude to Professor Josef Sivic and Tomas Mikolov for
their invaluable guidance, enthusiasm, and endless encouragement, without which
this thesis would not have been possible.

I am very grateful to a number of people who have directly or indirectly
influenced and helped me in my work, both as colleagues and friends: Jelle, Barbora, Kateryna, Teven, and David.
Special thanks to Barbora Hudcová and Josef Sivic for their meticulous
proofreading of parts of this thesis.

I would also like to thank everyone at the Foundational AI Lab and the
Automated Reasoning Lab of the Czech Institute of Informatics, Robotics and
Cybernetics (CIIRC) for creating a welcoming and supportive work environment.

Most importantly, I would like to express my deep appreciation to my parents,
family, and friends for their support and understanding throughout my academic
journey.

Lastly, I would like to express my heartfelt thanks to Brune for her unwavering
patience, help, support, and constant inspiration.

\cleardoublepage
\hypertarget{contents}{}
\tableofcontents

\mainmatter
\chapter{Introduction}
\label{cha:introduction}

\section{Goal}

Complex systems are systems composed of many interconnected elements that 
interact with each other, often in non-linear ways, giving rise to emergent 
properties that cannot be explained by the properties of the individual 
elements alone. In many naturally occurring complex systems, learning is a key process that 
allows the system to adapt and evolve over time. 
In this thesis, we explore complex systems
as a framework for studying learning and adaptation in natural and artificial
systems. The aim of this thesis is to develop methods for studying and using
computations that take place in complex dynamical systems to eventually create
learning algorithms that require limited to no supervision. Complex systems often exhibit 
features such as self-organization, adaptation, feedback loops, and the ability
to undergo phase transitions, all of which can make them challenging to model and 
predict. Examples of complex systems that occur naturally include ecosystems, weather 
patterns, financial 
markets, social networks, and biological organisms.
The objective of this thesis is broken down into the following subgoals:

\begin{enumerate}
  \item The first subgoal is to identify complex dynamical systems that have the 
  potential to display emergent open-ended growth, which refers to the ability of 
  a system to spontaneously generate new levels of complexity over time without 
  reaching a state of equilibrium. This growth is often associated with 
  evolutionary-like properties, such as variation, selection, and inheritance.
  There are many ways to define complex systems, and as
        illustrated in Figure \ref{fig:comparison_ca}, some may exhibit more
        interesting and promising behaviors than others. A cellular automaton (CA) 
        is an instance of a complex system that can be described as a grid of cells that can 
        take on a finite set of states and change their state over time according to a set 
        of rules that depend on the states of their neighboring cells. Interesting \acp{CA}, such 
        as the one shown in Figure \ref{fig:structured_sys}
        may be hard to find depending on how the
        search space is defined. The work presented in Chapter
        \ref{cha:meas-compl-evolv} aims to construct a metric of complexity
        that can help to identify interestingly behaving complex
        dynamical systems.
\begin{figure}[htbp]
  \centering
\begin{subfigure}[t]{.4\linewidth}
  \centering
  \includegraphics[width=\linewidth]{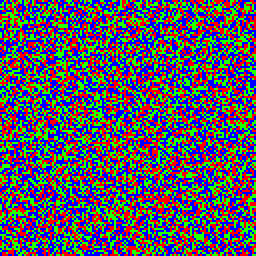}
  \caption{A disordered \acl{CA}.}
 \label{fig:disordered_sys}
\end{subfigure}
\hspace{30pt}
\begin{subfigure}[t]{.4\linewidth}
  \centering
  \includegraphics[width=\linewidth]{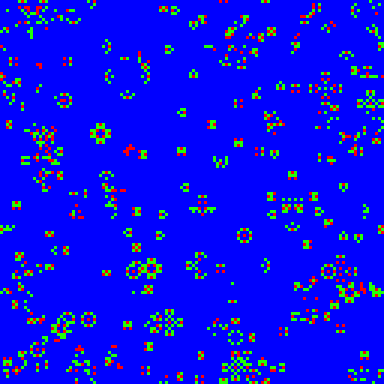}
  \caption{A \acl{CA} with visible emergent structures.}
  \label{fig:structured_sys}
\end{subfigure}
\caption{Two examples of complex systems (cellular automata) with different behavior
  types. \ref{fig:structured_sys} and \ref{fig:disordered_sys} show a single state 
  of a randomly initialized 2D \ac{CA} simulated for a fixed number of steps. Some \acp{CA} 
  appear more promising than others for the design of
  unsupervised learning systems because of their emergent complex structures
  (visible in \ref{fig:structured_sys}), whereas the \ac{CA} in
  \ref{fig:disordered_sys} seems to behave randomly.}
  \label{fig:comparison_ca}
\end{figure}

  \item The second subgoal is to measure the fraction of systems that have the most complex and rapidly
        evolving behavior. Defining these notions is also part of the goal.
        This sub-goal is distinct from the first one, which focuses on complex system design in 
        general. Instead, this sub-goal involves establishing approaches to evaluate growth 
        in complexity.
        We believe that systems with the most complex and rapidly
        evolving behavior are promising for further use since
        they may exhibit open-ended complexity growth. In Chapters
        \ref{cha:meas-compl-evolv} and \ref{cha:visu-comp-large}, we present
        different methods to measure the evolving complexity and understand when the
        complexity increases over time. Chapter \ref{cha:visu-comp-large} of this thesis is dedicated to investigating the importance of multiscale analysis for the complexity of
        cellular automata, which is the process of studying a system at different levels of detail or resolution.   Additionally, we identify complex systems with behavior that
        changes when we manipulate the scale of the system, either by increasing or decreasing its size.
\begin{figure}[htbp]
  \centering
 \includegraphics[width=.9\linewidth]{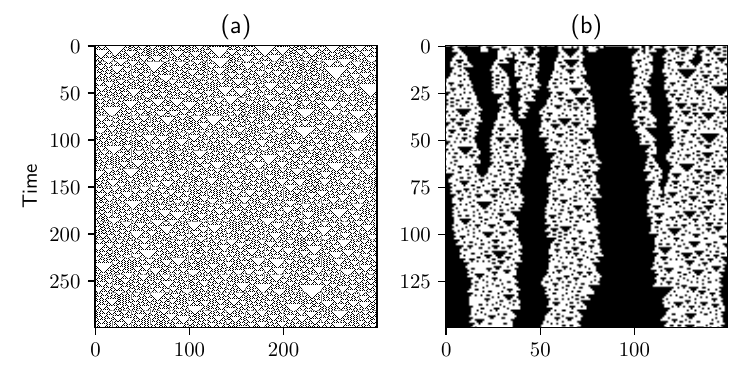}
 \caption{Filtering the behavior of elementary \acl{CA} rule 18. This
   uncovers a highly structured behavior with the propagation of area boundaries
   within the apparent randomness. (a) shows 300 timesteps
of a randomly initialized rule 18 simulation. Notice the complex structures made
visible in (b) with a filtering method. This Figure is discussed in more detail in Chapter \ref{cha:visu-comp-large}.}
  \label{fig:rule_18}
\end{figure}

  \item Apply promising systems to challenging learning tasks where classical
        machine learning models may fail or become less efficient. The goal is to
        define various ways to apply evolving complex dynamical systems to some
        standard learning tasks and to find out if it improves the performance
        or efficiency of the learning algorithm. An example application that we
        explore in Chapter \ref{cha:learn-effic-compl} is illustrated in
        Figure~\ref{fig:ca_lm}, where a \ac{CA} is used to implement a language
        model, a probabilistic model that is used to generate text to complete
        sentences. In Chapter \ref{cha:background}, we explore the similarities
        between \acp{CA} networks and highlight the numerous potential applications of \acp{CA}.
\begin{figure}[htbp]
  \centering
  \includegraphics[width=\linewidth]{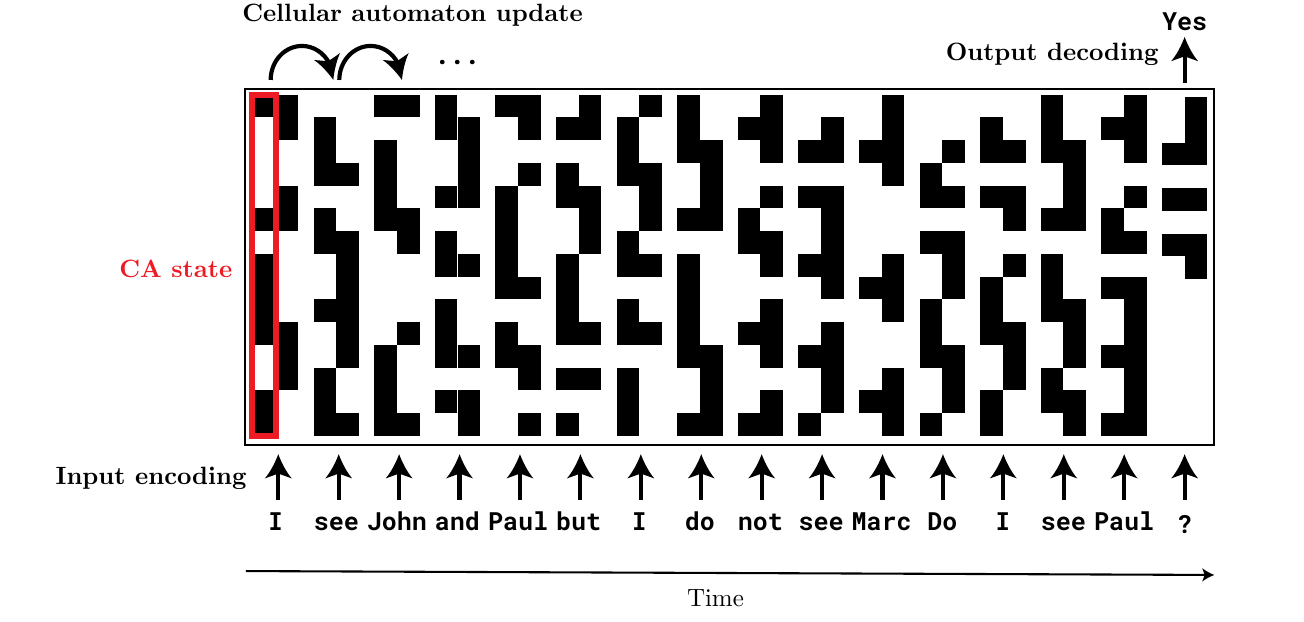}
  \caption{An example of using a \acl{CA} and reservoir computing to implement a
    language model. Tokens are encoded within the \acl{CA} internal state. The
    \acl{CA} update rule is then applied, and an output value is decoded from the
    final state. Each vertical bar represents two consecutive internal \ac{CA} states, which 
    are enriched with the encoded input.}
  \label{fig:ca_lm}
\end{figure}

\end{enumerate}

In this thesis, we do not focus our attention on what the machine learning
community commonly refers to as ``unsupervised learning'', that is, learning underlying
structures from unlabeled data
\parencite{hintonUnsupervisedLearningFoundations1999}. In machine learning,
supervision refers to a label, a number, or a collection of numbers that
represents an expected outcome associated with some input data. Our goal is to
construct models that can develop autonomously without necessarily needing any external data or
interactions. Therefore, we first seek systems that behave that way on their
own and postpone the issue of learning to optimize a particular objective
function to the end of the thesis (Chapter \ref{cha:learn-effic-compl}). 
We are particularly interested in systems that can develop through
internal evolution rules without any input, which is the case for many complex
systems.

Throughout the thesis, we work toward achieving these goals, focusing on one
particular complex system: the \acl{CA} \parencite{vonneumannTheorySelfreproducingAutomata1966}. This model has been extensively studied
due to its simple definition and its ability to simulate a wide range of complex
behaviors. We provide a detailed description of cellular automata in Section
\ref{sec:cellular-automata-sec}. This thesis provides new insights into the role
of learning in complex systems and open-ended evolution and demonstrates the
potential of cellular automata as a model for studying these phenomena.

\section{Motivation}\label{sec:motivation}

It is possible that some form of evolutionary mechanism may be necessary in
order to achieve advanced forms of artificial intelligence (AI), particularly 
if the goal is to create
intelligent systems that can adapt and learn in complex and changing
environments. Complex dynamical systems could be the key to overcoming the problems
with existing learning algorithms, such as difficulties in generalization,
robustness, or the ability to learn continuously
\parencite{parisiContinualLifelongLearning2019}. The natural intelligence of
biological systems seems to depend on emerging properties selected through
evolution. For example, biological life exhibits a pattern of major evolutionary
transitions in which autonomously replicating entities at the lower level merged to form a
single more complex corporate body
\parencite{lorenzEmergenceModularityBiological2011}. This is believed to have
occurred for the emergence of eukaryotic cells, as well as for multicellular
life \parencite{hammerschmidtLifeCyclesFitness2014}. All complex and diverse
biological entities have presumably emerged from a single common ancestor and,
even before, from inorganic components present on the surface of the Earth
\parencite{woeseUniversalAncestor1998, smithOriginsLifeBirth2000}.

So far, it remains uncertain which algorithmic characteristics might enable 
an artificial system to exhibit a path comparable to the natural evolutionary
process within its state space, which involves the transition from basic 
constituents to intricate entities. Producing emerging
phenomena similar to those of nature \emph{in silico} is a long-standing
challenge. Many algorithms try to mimic evolutionary
properties to solve particular tasks, called evolutionary algorithms
\parencite{fogelArtificialIntelligenceSimulated1966,
  ,millerDesigningNeuralNetworks1989, backOverviewEvolutionaryAlgorithms1993}.
However, most of these methods focus on searching the space of solutions using
high-level evolutionary mechanisms, such as genetic mutations and crossovers.
Although effective in solving precise tasks, these methods often obscure another
crucial component of natural evolution by using an explicit fitness function and 
hard-coded primitives.

Most existing machine learning algorithms rely on the choice of an objective
function: a clearly defined mapping from the current state and parameters of a
model to a real value, which indicates the performance of that model. The
function depends on the objective of the model. For a supervised learning
problem, we may count the number of misclassified objects or the distance
between the predictions and the expected results. Even in unsupervised learning,
the family of algorithms used for learning from unlabeled data, objectives are
still central. For example, the well-known K-means clustering algorithm minimizes
the sum of square distances of data points to cluster centers.

This reliance on objective functions creates two main issues: (i) The objective
is not always clearly defined or can be too broad for general-purpose
applications. For example, a possible objective function of a walking robot
could be ``not fall when stepping through its surrounding environment''. This
function is impractical to define and will vary greatly depending on the
parameters of the environment (e.g. depending on the terrain, malfunctioning or 
missing limbs of the robots). Indirect rewards such as ``head not touching the ground''
may help, but it does not cover all possible ways the model could fail. 
This problem of defining an objective is also present within our goal of designing
a general-purpose, autonomously developing learning algorithm.
Furthermore, (ii) using predefined functions as goals can be counterproductive
because, as many examples in nature demonstrate, robust paths to complex
objectives are often deceptive. They involve developing in unexpected directions
that may initially seem to be against the original goal
\parencite{stanleyWhyGreatnessCannot2015}.

In this thesis, the term \emph{unsupervised} refers to a form of learning with
no predefined objective. Like in natural evolution, we expect true unsupervised
algorithms to develop new features autonomously and become progressively more
complex over time. Such algorithms would regularly learn to solve problems on
their own without the need to be explicitly guided towards the problem solutions, 
thereby discovering robust
and diverse solutions to deceptive problems.

\section{Challenges}\label{sec:challenges}

\begin{figure}[htbp]
  \centering
  \includegraphics[width=.98\linewidth]{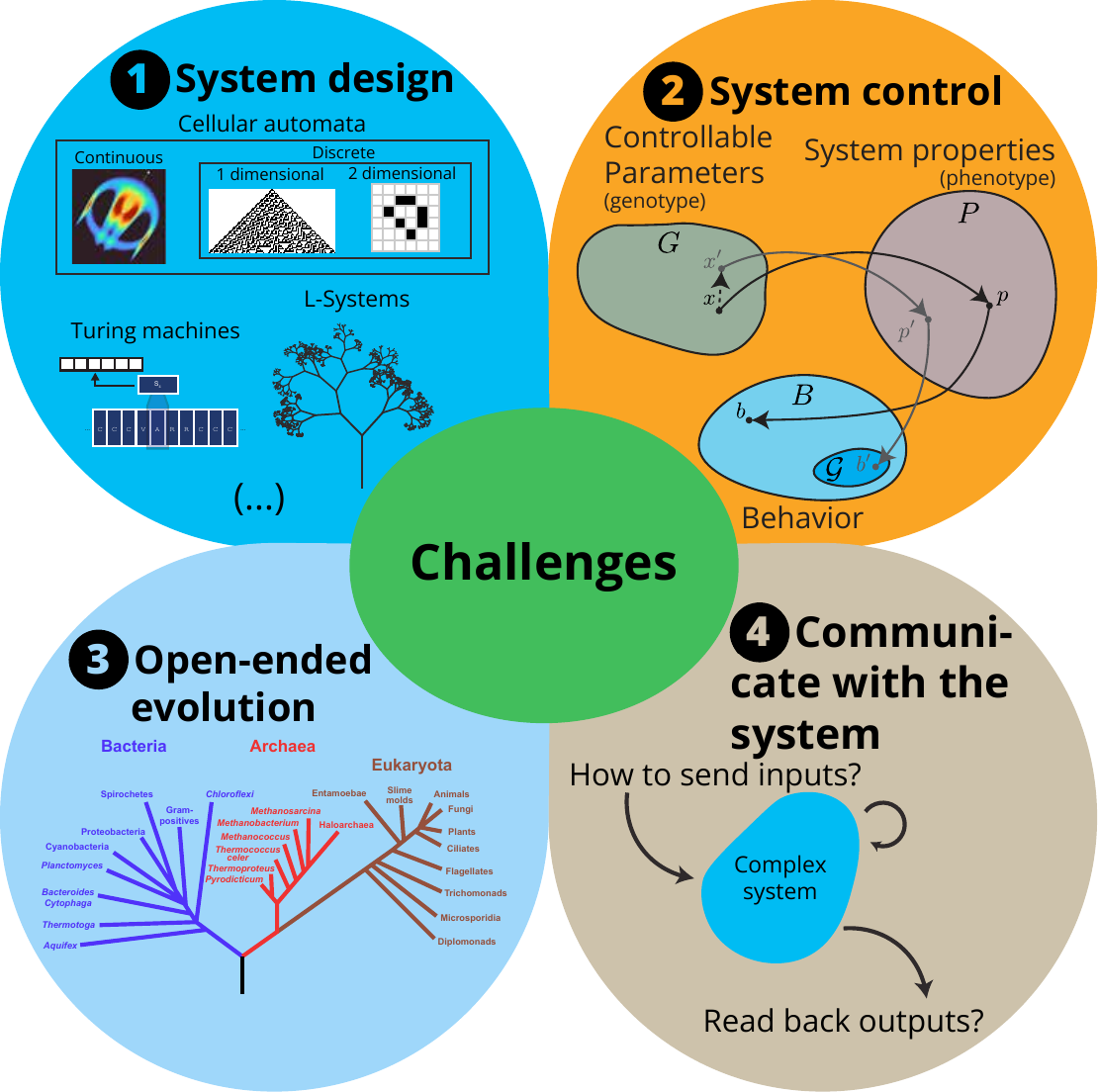}
  \caption{The challenges of working with complex systems encountered across
    this thesis can be broken down into several categories. \textbf{(1)} The choice
    and design of the system, \textbf{(2)} the control of the system which
    involves understanding the complex mapping between a parameter space and the
    possibly unpredictable behavior of a system, \textbf{(3)} the search for
    open-ended evolution properties similar to the ones found in nature,
    \textbf{(4)} the issue of sending inputs and reading outputs from the
    internal state of a complex system. This illustration uses
    \href{https://commons.wikimedia.org/wiki/File:Lenia_icon4.png}{Lenia
      icon4.png} by Bert Wang-Chak Chan, licensed under
    \href{https://creativecommons.org/licenses/by-sa/4.0/}{CC BY 4.0}. }
  \label{fig:challenges}
\end{figure}

The study of complex systems presents a range of difficult challenges by itself
\parencite{sanmiguelChallengesComplexSystems2012}. 
The complexity of a system is an emergent property that arises from various factors, 
including its intricate structure, the number of elements it contains, how it functions, 
and how it responds to different types of external influences. These factors contribute 
to the overall complexity of the system, which could be measured in various ways.
For many complex systems, their emergent mechanisms are
poorly understood. We identify four main challenges associated with the study of
complex systems in the context of this thesis: (i) the questions of the design choices in defining and sampling a
complex system (section \ref{sec:design-compl-syst}), (ii) which will in turn define
their potential to support a form of open-ended evolution (section
\ref{sec:open-ended-evolution}) and (iii) how we can expect to build an interface to
communicate with it (section \ref{sec:compl-syst-inputs}), (iv) which is essential to
achieve some form of control of that system (section \ref{sec:compl-syst-contr}).

\subsection{Design of a complex system\label{sec:design-compl-syst}}

The first challenge is to construct a suitable complex system. The definition of 
complex systems is broad, and several systems with
interesting dynamics have been studied under that name. For example, abstract models such as L-Systems (formal grammars used to model the growth and development of organic structures),
Random Boolean networks (dynamical systems consisting of a fixed number of binary nodes 
that are randomly connected to each other and updated based on the state of its neighboring 
nodes), Turing machines (abstract deterministic computing machines consisting of a tape and 
a read-write head) and \Acfp{CA}. In this thesis, we
focus mainly on the last listed element: the \acl{CA}. There are multiple
benefits to working with this model. It is very simple to define, and its high
parallelism makes its implementation straightforward. Furthermore, \acp{CA} have
shown the ability to simulate a wide range of complex behaviors
\parencite{wolframNewKindScience2002}.

Choosing the right complex system architecture is essential because it defines
the search space over which interesting and useful systems can be found.
Correctly parameterizing that space can also be challenging because too many
degrees of freedom make it difficult to search for and find good systems, while
too few might indicate a lack of expressivity. An example parameterization of
\acp{CA} with little expressivity is Langton's lambda parameter
\parencite{langtonComputationEdgeChaos1990}. At the other end of the spectrum, the
\ac{CA} rule is a simple parametrization with many degrees of freedom, which
makes it impractical.

Even when we limit ourselves to the study of \acp{CA}, many variants can be
considered (we list some of them in Chapter~\ref{cha:background}).
They can have continuous or discrete states and operate
in continuous or discrete space. For discrete state and space \acp{CA},
the number of states and the window of the update function can vary, as well as the
topology of the simulation space. We tackle this issue from various angles
throughout the thesis, exploring the question of rule definition in Chapter
\ref{cha:meas-compl-evolv}, the scale in Chapter \ref{cha:visu-comp-large}, and
the connectivity pattern in Chapter \ref{cha:learn-effic-compl}.

\subsection{Complex systems control}\label{sec:compl-syst-contr}

\begin{figure}[htbp]
  \centering
  \begin{subfigure}[t]{.04\linewidth}
    \centering
    \includegraphics[width=\linewidth]{figures/arrow}
    \caption*{}
  \end{subfigure}
  \begin{subfigure}[t]{.45\linewidth}
    \centering
    \includegraphics[width=.93\linewidth]{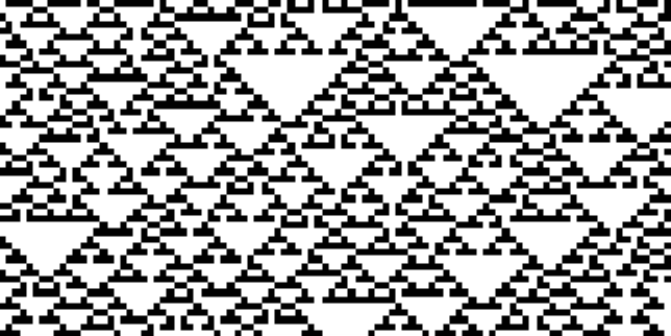}
    \caption{\ac{CA} rule 22 on a tape of size 100, ran for 50 steps from a
      random initial state.}
    \label{fig:ca_comp_a}
  \end{subfigure}
  \hspace{10pt}
  \begin{subfigure}[t]{.45\linewidth}
    \centering
    \includegraphics[width=.93\linewidth]{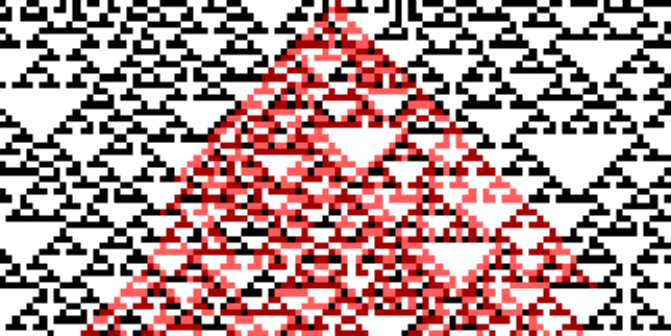}
    \caption{\ac{CA} rule 22 on a tape of size 100 ran from a slightly
      different initial state. The cells different from Figure \ref{fig:ca_comp_a} are
      overlayed in red.}
    \label{fig:ca_comp_b}
  \end{subfigure}

  \caption{Comparison of the same 1 dimensional \ac{CA} ran from two initial conditions
    differing only by one cell. Each row is the state of the \ac{CA} at one time
    step, with time increasing from top to bottom. The effect of the small perturbation 
    of initial conditions grows rapidly as the \ac{CA} evolves in time.}
  \label{fig:ca_comp}
\end{figure}

Even if we want complex systems that evolve in unexpected directions and grow in
an open-ended way without supervision, it may still be useful to steer them
locally toward specific targets. A major challenge that follows from this goal
is that many complex systems are unpredictable. Deterministic dynamical systems can 
exhibit behavior that is extremely sensitive to their initial conditions. This sensitivity 
is sometimes referred to as chaos, where even tiny perturbations in the initial conditions 
can result in significantly different outcomes or trajectories. This is illustrated 
with a simple \ac{CA} rule in Figure \ref{fig:ca_comp}. 
Because of this property, it is often very hard to predict how a
given system will evolve over time, and therefore hard to steer its evolution
towards a particular final state or along a chosen trajectory. Although it can be challenging 
to apply these systems to certain basic tasks, their ability to generate unexpected solutions 
to difficult problems is also an advantage for the development of open-ended systems. Such systems 
need to continuously produce new and diverse solutions without a 
predefined endpoint and their sensitivity to initial conditions can be helpful in 
achieving this goal.

This problem is also related to the issue of sending a control signal to a
complex system that depends on the choice of encoding for that signal. This
encoding can significantly impact the system's ability to respond to the control 
signal and generate the desired outcome, which we describe in Section \ref{sec:compl-syst-inputs}. 

\subsection{Open-ended evolution without
  objectives}\label{sec:open-ended-evolution}

Another challenge is posed by the lack of a clear objective in the design of
fully unsupervised learning systems. Even without an objective, it is still essential
to understand which complex systems to choose from all available options or how
to tune their parameters so that they behave in interesting ways. For example,
\acp{CA} such as the one shown in Figure \ref{fig:disordered_sys} is unlikely 
to be useful for building systems that demonstrate a growth of complexity 
due to their disordered nature in both space and time. Because of this disordered behavior, they cannot
preserve information about the past.

Our goal is to create a system that has the property of evolving in an
open-ended way, and we need metrics that can help select systems. However, these metrics 
should not be used as an additional objective function, as doing so would result 
in the problems described in Section \ref{sec:motivation}. We address this 
challenge in particular
in Chapter~\ref{cha:meas-compl-evolv}, by designing a complexity metric that can
help select interesting complex systems without relying on a specific
task-based performance score.

\subsection{Complex systems inputs and outputs\label{sec:compl-syst-inputs}}

Some of the complex systems we study in this thesis are closed systems. They do
not expect inputs or have well-defined outputs. For example, \ac{CA} and
\ac{RBN} do not have the notion of inputs and outputs built into the model. They
are standalone objects that evolve according to a set of internal rules.

A common challenge for such systems is to define the inputs and outputs in a way
that preserves their internal dynamics. This is also related to the control
problem of Section \ref{sec:compl-syst-contr}, because controlling a complex
system implies being able to send a control signal and read the current
state of the system.

Throughout this work, we use a framework called \emph{reservoir computing} (RC) for this 
purpose. Reservoir
computing allows one to harvest the internal computations of complex systems, or
\emph{read} information from its internal state by learning a linear regression
that maps that internal state to desired outputs (see Section
\ref{sec:res-models}).

\section{Contributions}

The main contributions of this thesis are as follows.
\begin{enumerate}
  \item Our first contribution is a review of the literature (Chapter \ref{cha:literature-review})
        describing the connections between cellular automata
        and other complex systems, open-ended evolution, and neural networks.
        Studying these fields from a single point of view is a relatively novel
        endeavor, and we consider a review necessary to place the rest of our
        work in its context.

  \item Our second contribution is the development of a general complexity metric that can help identify
        complex systems with interesting behavior (Chapter \ref{cha:meas-compl-evolv}). Our findings were published in \parencite{cisnerosEvolvingStructuresComplex2019}.

  \item The third contribution is the development of a coarse-graining method to visualize computations in
        cellular automata and other discrete systems with local interactions (Chapter \ref{cha:visu-comp-large}). Our findings were published in \parencite{cisnerosVisualizingComputationLargescale2020}.

  \item Our fourth contribution is the introduction of a learning efficiency metric for learning
        algorithms and a benchmark dataset of progressively harder
        language tasks (Chapter \ref{cha:learn-effic-compl}). Our findings were published in \parencite{cisnerosBenchmarkingLearningEfficiency2022}.
\end{enumerate}

\section{Thesis overview}

Chapter \ref{cha:background} presents some background notions about complex
systems, cellular automata, and reservoir computing.

In Chapter \ref{cha:literature-review}, we review relevant methods and tools for measuring
the complexity of complex systems and use their computations for
various tasks.

Chapter \ref{cha:meas-compl-evolv} introduces a complexity metric that allows us 
to select complex systems with interesting behavior. The metric measures the
``novelty'' of the temporal states of a system compared to a reference system. We
built a dataset of interesting cellular automata to validate the quality of the
metric.

Chapter \ref{cha:visu-comp-large} addresses the question of large-scale complex
systems and the applicability of complexity metrics at multiple scales. We
propose three algorithms for the coarse-graining of cellular automata. This allows
us to reduce the size of large-scale systems while retaining the interesting parts
of the behavior.

Chapter \ref{cha:learn-effic-compl} presents a learning efficiency metric and a
dataset to measure the learning speed of various systems. We show that
reservoir computing-based systems using cellular automata can be more efficient
than usual machine learning algorithms in constrained data and computation
settings.

Chapter \ref{cha:conclusion} summarizes our contributions to this thesis. 

Chapter \ref{cha:future_work} outlines potential avenues for further research in 
the field. It provides suggestions on how current work could be expanded or 
improved, highlighting areas that require additional research to advance 
understanding of the subject. We propose some experiments that could extend the 
potential applications of cellular automata, as well as a theoretical framework 
for learning in dynamical systems.

\section{Publications and software}

The thesis has led to the following publications.

\begin{itemize}
  \item \fullcite{cisnerosEvolvingStructuresComplex2019}
  \item \fullcite{cisnerosVisualizingComputationLargescale2020}
  \item \fullcite{cisnerosBenchmarkingLearningEfficiency2022}
\end{itemize}

The code to reproduce the experiments of all three publications is available on
GitHub: \url{https://github.com/hugcis/evolving-structures-in-complex-systems},
\url{https://github.com/hugcis/benchmark_learning_efficiency}. We also published a
dataset that we used for our benchmark in our last
publication. It is available at
\url{https://github.com/hugcis/incremental_tasks/}.

\chapter{Background}
\label{cha:background}

In this chapter, we provide background on some of the topics that we study in
the thesis. We begin by describing the \acf{CA} model, which is the complex
system we focus on in most of the thesis (Section
\ref{sec:cellular-automata-sec}). Next, we describe the similarities between
\acp{CA} and \acfp{RNN}, which is useful for understanding the characteristics
and applications these two models share (Section \ref{sec:cell-autom-rnns}).
\acp{CA} can be completely reformulated as a special case of \acp{RNN}. This
reformulation makes it natural to replace recurrent \acp{RNN} with \acp{CA} in
several use cases. We describe the \acf{RC} model (Section
\ref{sec:res-models}), which enables sending inputs to and harvesting the
computations of complex systems to perform machine learning tasks without
requiring explicit knowledge of the underlying system dynamics or the need to
solve complex optimization problems. We apply reservoir computing to sequential
language-like tasks in Chapter~\ref{cha:learn-effic-compl}. This model was
developed with \acp{RNN}, but we describe how \acp{CA} and other complex systems
can be a suitable replacement, and the benefits of this approach.

\section{Cellular automata}\label{sec:cellular-automata-sec}

Stanislaw Ulam and John von
Neumann proposed the \ac{CA} model in the 1940s as a means 
of modeling crystal growth and creating an autonomous self-replicating system
\parencite{vonneumannTheorySelfreproducingAutomata1966}.

\subsection{Definition}\label{sec:definition}
It is usually defined in a regular lattice in one or two dimensions. Each of its
components is called a cell and can be in a state $k \in \mathcal{S}$. $\mathcal{S}$ is the space of
available states for cells, usually chosen to be $\{0, 1\}$ for binary
\acp{CA} or $\{1, \ldots, n\}$ for \acp{CA} with $n$ states.

\begin{figure}[htbp]
  \centering
  \begin{subfigure}[c]{.3\linewidth}
    \centering
    \includegraphics[width=\linewidth]{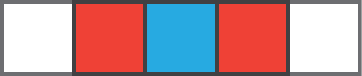}
    \caption{Standard 1D \ac{CA} neighborhood}
    \label{fig:1d_neigh}
  \end{subfigure}
  \begin{subfigure}[c]{.3\linewidth}
    \centering
    \includegraphics[width=\linewidth]{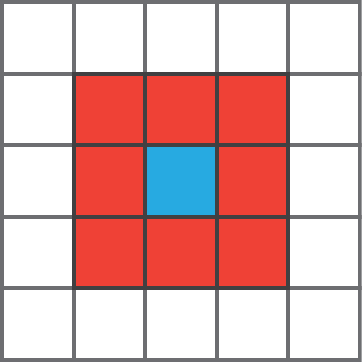}
    \caption{Moore neighborhood}
    \label{fig:moore}
  \end{subfigure}
  \begin{subfigure}[c]{.3\linewidth}
    \centering
    \includegraphics[width=\linewidth]{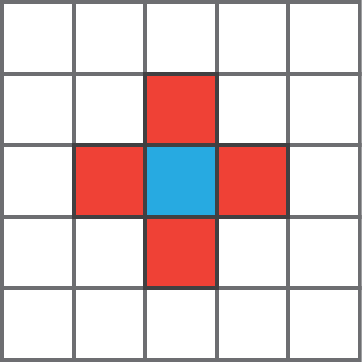}
    \caption{von Neumann neighborhood}
    \label{fig:von_neumann}
  \end{subfigure}

  \caption{Illustration of commonly used neighborhoods for 1D and 2D \ac{CA}.}
  \label{fig:neighborhoods}
\end{figure}

A neighborhood function $\boldsymbol{N}$ is defined that associates each cell
with the indexes of its neighbors in the grid. In general, a \ac{CA} can be
constructed on any space $\mathcal{L}$ where this function can be defined. The space $\mathcal{L}$
specifies an index and the relationship between the cells. In practice, regular
finite or infinite grids are chosen, $\mathcal{L} \subset \mathbb{Z}$ or $\mathcal{L} \subset \mathbb{Z}^{2}$. For example, the
grid could be a one-dimensional torus with 10 cells, that is,
$\mathcal{L}_{{T_{10}}} = \{1, 2, \ldots, 10 \}$. The neighborhood function has the following
general form:
\begin{equation}
  \begin{aligned}
\boldsymbol{N}_{\mathcal{L}} :\quad & \mathcal{L} \rightarrow \mathcal{L}^{s}\\
&{i} \mapsto [{c_{j}}]_{j\in \mathcal{N}_{c_{i}}},
  \end{aligned}
\end{equation}
where $\mathcal{N}_{c_{i}}$ is the neighborhood of cell $c_{i}$, $s$ is the
number of cells in the neighborhood, and the returned value is a finite set of
cells: the neighbors of cell $c_{i}$. For the torus $\mathcal{L}_{T_{10}}$
above, we can define the neighbors to be the cell itself and the two immediately
adjacent cells. This type of 1D neighborhood is usually associated with a
\emph{radius} parameter $r = 1$. The neighborhood $r=1$ is illustrated in Figure
\ref{fig:1d_neigh}. It corresponds to the following neighborhood function:
\begin{equation}
  \begin{aligned}
\boldsymbol{N}_{\mathcal{L}_{T_{10}}} :\quad & \mathcal{L} \rightarrow \mathcal{L}^{3} \\
&\boldsymbol{N}_{\mathcal{L}_{T_{10}}}({i}) = \begin{cases}
                      [c_{i - 1}, c_{i}, c_{i + 1}],& \text{if}\quad i \in \{2,\ldots , 9\}\\
                       [c_{10}, c_{1}, c_{2}], & \text{if} \quad i = 1 \\
                       [c_{9}, c_{10}, c_{1}], & \text{if} \quad i = 10. \\
                    \end{cases}
  \end{aligned}
  \label{eq:torus_index}
\end{equation}

A larger radius $r$ would correspond to including more cells in the
neighborhood, going in each direction from the initial cell. On two-dimensional
grids, there are multiple ways to define the neighborhood. Some common examples
are the Moore neighborhood (see Figure \ref{fig:moore}) and the von Neumann
neighborhood (see Figure \ref{fig:von_neumann}). In the rest, we omit the
subscript on the neighborhood function $\boldsymbol{N}$ as we almost always work
with regular grids on a torus, for which it is simpler to just mention the size
and dimension.

A \ac{CA} evolves in discrete time steps. An update rule
$\boldsymbol{\Phi}: \mathcal{S}^{s} \rightarrow \mathcal{S}$ defines the new state of a cell as a function of its
local neighborhood at the current time step. The local group of cells is said to
be in a particular \emph{configuration}. The function $\boldsymbol{\Phi}$ is
applied in parallel to all cells. For a \ac{CA} in its initial state at time
step 0 --- \ie a set of cells $\left(c_{i}^{(0)}\right)_{i \in \mathcal{L}} \in \mathcal{S}^{|\mathcal{L}|}$, and a
neighborhood function $\boldsymbol{N}$, we have the following update rule:
\begin{equation}
\begin{aligned}
\forall i \in \mathcal{L}, \quad c_{i}^{(t + 1)} = \boldsymbol{\Phi}\left(
\left(c_{j}^{(t)}\right)_{j \in \boldsymbol{N}(c_i)}\right).
\end{aligned}
\end{equation}

\begin{figure}[htbp]
  \centering
 \includegraphics[width=.7\linewidth]{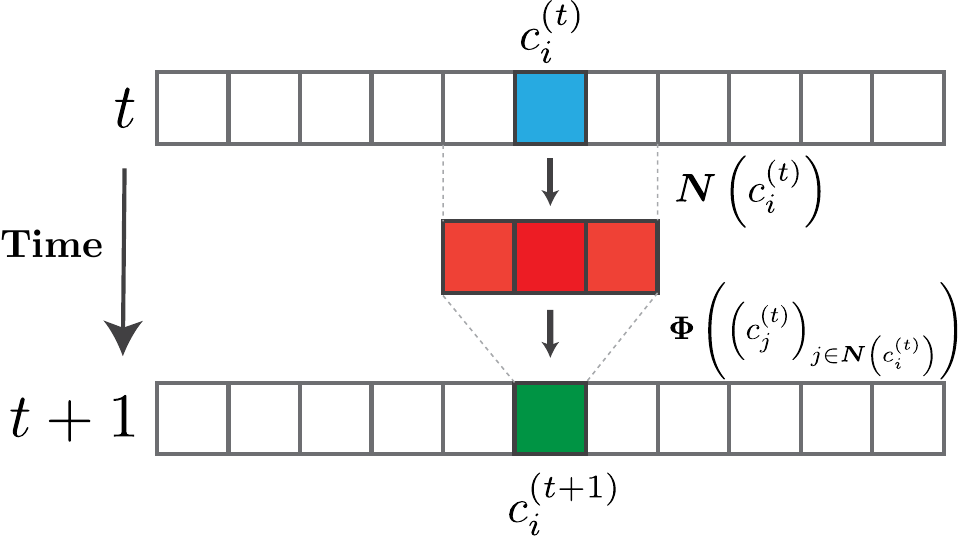}
  \caption{Illustration of a \acl{CA} update rule in 1 dimension. For each cell,
  we look up the neighboring cells and update their state according to the current
state of the neighbors. This operation is applied in parallel to the entire grid.}
  \label{fig:ca_base}
\end{figure}

The details of an update step in a 1-dimensional \ac{CA} is shown in Figure
\ref{fig:ca_base} for example. The neighborhood of the cell $c_{i}$ is $c_{i}$
itself, as well as its two immediately adjacent cells.

The function $\boldsymbol{\Phi}$ is sometimes called a \emph{rule table} because it
associates an output state for each possible combination of input neighbor
states. In an implementation of the \ac{CA} model, these output states can be
looked up from a table or array containing all possible transitions from input
to output.

There are many ways to define \acp{CA}, and we gave a broad
definition that includes many rarely used \ac{CA} variants. For another more straightforward
definition of \acp{CA}, see, for example, \autocite{kariBasicConceptsCellular2012}.

\paragraph{Rule representation.}
A useful rule representation can be obtained by listing all output states
corresponding to input neighborhood configurations in a predetermined order.
This results in a list of values $[o_1, \ldots, o_{s^{|\mathcal{S}|}}]$, with
$\forall i,\ o_{i} \in \mathcal{S}$, where $s$ is the number of cells in a neighborhood, $\mathcal{S}$ is
the space of available states, and $|\mathcal{S}|$ is the number of available states per
cell. Using $o_{i}$ as the digits of a base-$|\mathcal{S}|$ number, each rule is uniquely
represented by a number. For example, as explained in more detail in Section
\ref{sec:elem-cell-autom}, a one-dimensional cellular automaton with a
neighborhood of size three has eight possible configurations of neighboring cells (000,
001, 010, 011, 100, 101, 110, and 111), each of which can map to two states (0 or
1). To obtain the rule number, we concatenate the new states for each of the
input configurations, resulting in an 8-bit number. The 256 possible binary
rules in one dimension with neighborhood size 3 can be numbered from 0 to 255
with this standard method and are referred to by their number in the literature.

\paragraph{Boundary conditions.}
The grid of a \ac{CA} can be finite or infinite. In the infinite case, the grid
is assumed to be initialized to a uniform state, except for a few cells set to
other states. The simulation is then run on these few cells, while the rest of
the infinite grid does not have to be simulated from the start. For a finite
grid, an exhaustive simulation can be run, but one needs to define
boundary conditions. The boundaries can be set to wrap around the other side of the
grid, forming a torus. An example of the corresponding indexing is given in
\eqref{eq:torus_index}. Other choices of boundary conditions consist of adding
virtual padding cells outside of the main grid. They can be set to a fixed
state, a randomly chosen state, or they can mirror the cells inside the grid.
Each of these choices affects the evolution and properties of the \ac{CA}, but
the importance of these boundaries decreases for very large grids.

\subsection{Classification of cellular automata\label{sec:class-cell-autom}}

Stephen Wolfram approached \acp{CA} as discrete, spatially extended dynamical
systems \parencite{wolframUniversalityComplexityCellular1984}. The analogy is
only superficial since many concepts from dynamical systems theory, such as
``chaos'', ``attractors'' and ``sensitivity to initial conditions'' only admit a
rigorous definition in the continuous state and continuous time models. Wolfram
proposed a qualitative classification of \ac{CA} update rule behaviors roughly analogous to
classifications in dynamical systems theory, with four classes defined as
follows.

\begin{description}
  \item[Class 1] All initial configurations converge to a single fixed configuration, such as a configuration consisting entirely of 1s.
  \item[Class 2] All initial configurations relax after a transient period to some
        fixed point or some temporally periodic cycle of configurations, but
        which one depends on the initial configuration. (\ac{CA} defined on
        finite lattices always have periodic behavior because there
        is only a finite number of grid configurations. Class 2 does not refer to
        this type of periodic behavior but rather to cycles with periods much
        shorter than the total number of possible states).
  \item[Class 3] All initial configurations exhibit chaotic behavior
        after a short transient period. (The term “chaotic” here refers to
        apparently unpredictable space-time behavior, not the standard notion 
        of chaos in dynamical system theory.)
  \item[Class 4] Some initial configurations result in complex localized
        structures that can persist for a long time.
\end{description}

This classification being qualitative and not rigorous, there is not even a
clear consensus on which of the \acp{ECA} rules (one dimensional binary \acp{CA} with neighborhood size 3) belong to class 4 or class 3.
Class 4 \ac{CA} rules are speculated to be capable of universal computation
\parencite{wolframUniversalityComplexityCellular1984}, which refers to the ability of a system to simulate any other system or computation, given enough time and memory. For example,
\textcite{liStructureElementaryCellular1990} claimed that \ac{ECA} rule 110 has
class 4 behavior and was eventually proven to be universal
\parencite{cookUniversalityElementaryCellular2004}, but no other \ac{ECA} has
been proven universal since.

\textcite{zenilCompressionBasedInvestigationDynamical2010} studied the
compression size of the space-time diagrams of all \ac{ECA} for a fixed-length
simulation. Using a simple k-means clustering technique, he obtained two groups
that roughly match Wolfram’s classes 1 and 2 and classes 3 and 4. We reproduce
these results in Figure~\ref{subfig:comp_scores} in Chapter~\ref{cha:meas-compl-evolv}.
They offer an interesting
nonqualitative confirmation of the results of Wolfram's classification.
However, these results vary significantly if we modify the initial conditions, as
well as the grid size, data representation, or compression algorithm
\parencite{hudcovaClassificationComplexSystems2020}.

\textcite{wuenscheGlobalDynamicsCellular1992} studied the behavior of \acp{ECA}
when the simulations are reversed and computed the preimages of each
configuration. They introduced the Z-parameter, which is the probability that a
partial preimage can be extended by one symbol for each \ac{CA}. The authors speculate
that class 4 occurs at $\text{Z} \approx 0.75$. This is not a classification but a
class 4 membership test that can be computed from the rule directly, making it
practical compared to alternatives. However, when tested in practice, Wuensche's
hypothesis that $\text{Z} \approx 0.75$ corresponds to class 4 is rarely verified.

Hudcová defined a classification of \ac{CA} that is based on the form of the
asymptotic growth of its transients, obtained by repeated simulation of the
rules of \ac{CA} starting from various initial conditions and using the best
fitting function for the asymptotic growth of its transients
\parencite{hudcovaClassificationComplexSystems2020,
  hudcovaClassificationDiscreteDynamical2022}. For a 
detailed review of the current methods and challenges of \acp{CA} classifications, see
\autocite{vispoelProgressGapsObstacles2022}.

\subsection{Cellular automata variants}

Several \ac{CA} variants have been proposed, modifying or
restricting various parts of the definition in Section \ref{sec:definition}.
Here, we list some common ones.

\paragraph{Elementary cellular automata.\label{sec:elem-cell-autom}}
\Acp{ECA} are 1 dimensional \acp{CA} with two states per cell and
neighborhood size 3 --- the cell and its two direct neighbors on the 1D grid.
There are 8 possible configurations of a neighborhood with 3 cells and 2 states
per cell, which corresponds to 256 possible ways to define an \ac{ECA} --- two
possible outputs for each of these 8 possible configurations; hence
$2^{8} = 256$ possible \ac{CA}. This relatively small number of rules allows for an
exhaustive exploration of the rule space and mapping of the properties of \ac{ECA},
which would not be possible for general \acp{CA}.

\begin{figure}[htbp]
  \centering
  \includegraphics[width=.95\linewidth]{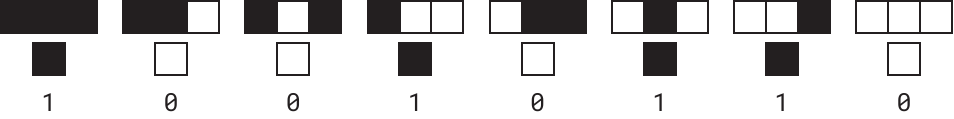}
  \caption{Illustration of the rule of \ac{ECA} number 150.}
  \label{fig:eca_150_rule}
\end{figure}

These \acp{CA} have been extensively studied and offer an interesting combination of
trivial definition and implementation, and complex and unpredictable properties.
One of the fundamental problems of \ac{CA} research is to classify the 256 rules
into well-defined behavior types and order them by complexity, which was
attempted in several previous works
\parencite{wuenscheGlobalDynamicsCellular1992,
  gutowitzTransientsCyclesComplexity1991,
  wuenscheClassifyingCellularAutomata1999, wolframNewKindScience2002,
  zenilCompressionBasedInvestigationDynamical2010,
  hudcovaClassificationComplexSystems2020,
  hudcovaComputationalHierarchyElementary2021}. We discuss these classifications
in more detail in Section \ref{sec:class-cell-autom}.

\begin{figure}[htbp]
  \centering
  \begin{subfigure}[t]{.04\linewidth}
  \centering
  \includegraphics[width=\linewidth]{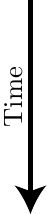}
  \caption*{}
\end{subfigure}
\begin{subfigure}[t]{.45\linewidth}
  \centering
  \includegraphics[width=\linewidth]{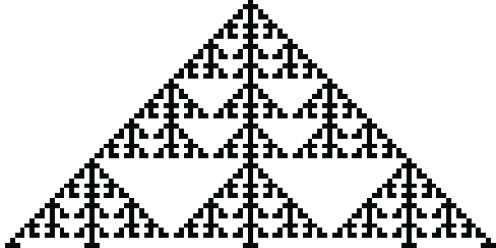}
  \caption{Single cell initialization}
  \label{fig:eca_150_single}
\end{subfigure}
\begin{subfigure}[t]{.45\linewidth}
  \centering
  \includegraphics[width=\linewidth]{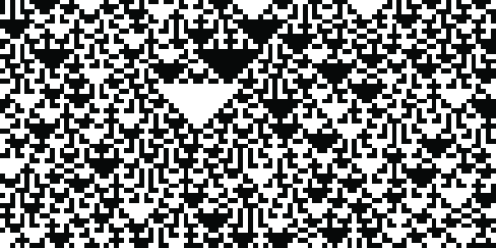}
  \caption{Random initialization}
  \label{fig:eca_150_random}
\end{subfigure}
\caption{Evolution of \ac{ECA} number 150 simulated on a grid of size 100 for 50
  steps. \ref{fig:eca_150_single} shows a simulation starting from a blank grid
  with only one cell set to 1. \ref{fig:eca_150_random} shows a simulation
  starting from a random initialization. Time flows from top to bottom, and each row of cells represents a single internal CA state at a single time step.}
  \label{fig:eca_150}
\end{figure}

\ac{ECA} rules are easily visualized because there are only eight possible
neighborhood configurations that need to be shown. For example, Figure
\ref{fig:eca_150_rule} shows the transition table for \ac{ECA} Rule 150. If we
assign 1 to the black state and 0 to the white state, the possible neighborhood
configurations are ordered in their binary order from right to left, with the
leftmost bit being the most significant. This is how the rule number is
computed, using the output states (last row in Figure \ref{fig:eca_150_rule}) as
a binary representation. Example simulations of \ac{ECA} rule 150 starting 
from various initial states are shown in Figure~\ref{fig:eca_150}. In these 
representations, time flows from top to bottom, and each row of cells represents 
a single internal \ac{CA} state at a single time step. This representation of an \ac{ECA} rule is often called a \emph{spacetime diagram}.

There are several well-known \ac{ECA} rules with particularly interesting
properties. For example, a universal computer has been constructed in rule 110
\parencite{cookUniversalityElementaryCellular2004}, making it the first (and
only to date, although \ac{ECA} Rule 54 is postulated to be as well) Turing complete \ac{ECA}.

\paragraph{Totalistic cellular automata.}
Totalistic \ac{CA} are a subset of \ac{CA} whose rules can be expressed as a
function of the sum of neighboring cell values. These \ac{CA} were introduced by
Stephen Wolfram \parencite{wolframStatisticalMechanicsCellular1983}. \ac{ECA} 150 is 
an example of a totalistic \ac{CA}, and its rule and typical space-time diagram are depicted in Figures~\ref{fig:eca_150_rule} and \ref{fig:eca_150}. Its rule could be summarized 
in natural language as 
``if exactly two states are 1 or all states are 0, the next state is 0. Otherwise, if 
one or three states are 1, the next state is 1''. Conway's Game of Life, presented 
in the following paragraph, is an example
of a totalistic \ac{CA} in two dimensions.

\paragraph{Game of Life.\label{sec:game-life}}
The game of life is one of the most famous \ac{CA}. It was proposed by the
mathematician John Conway in 1970 \parencite{gardnerMathematicalGames1970}. It
is a two-dimensional binary \ac{CA}. Its two states are often called ``alive''
and ``dead''. Its rule can be summarized in three sentences as follows.
\begin{itemize}
  \item Any live cell with two or three live neighbors survives.
  \item Any dead cell with three living neighbors becomes a live cell.
  \item Other live cells and already dead cells are dead in the next generation.
\end{itemize}
This \ac{CA} has a particularly active community dedicated to finding
interesting patterns with particular properties, such as long cycling periods or
particular speeds of movement through the grid. Some of these patterns have been
used as memory registers and communication channels to build a Universal
Computer \parencite{IgblanLifeUniversal}. This construction proved that the game
of life is Turing complete. This property is thought to be also shared by other
\acp{CA} --- it is proven for at least \ac{ECA} rule 110
\parencite{cookUniversalityElementaryCellular2004}. This makes \acp{CA}
models theoretically appealing for the design of a learning algorithm because
they can simulate any algorithm.

\begin{figure}[htbp]
  \centering
  \begin{subfigure}[t]{.31\linewidth}
    \centering
    \includegraphics[width=\linewidth]{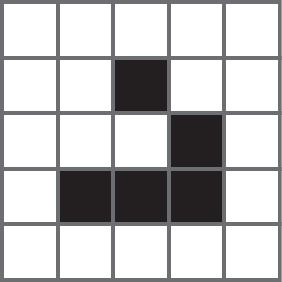}
    \caption{Moving oscillator\\ (\emph{glider}).}
    \label{fig:glider}
  \end{subfigure}
  \begin{subfigure}[t]{.31\linewidth}
    \centering
    \includegraphics[width=\linewidth]{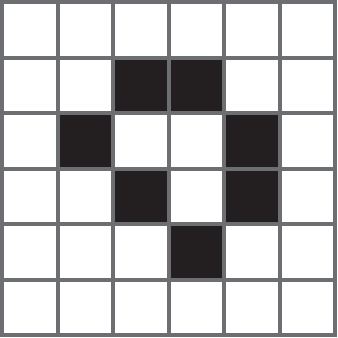}
    \caption{Fixed pattern\\ (\emph{still life}).}
    \label{fig:still_life}
  \end{subfigure}
  \begin{subfigure}[t]{.31\linewidth}
    \centering
    \includegraphics[width=\linewidth]{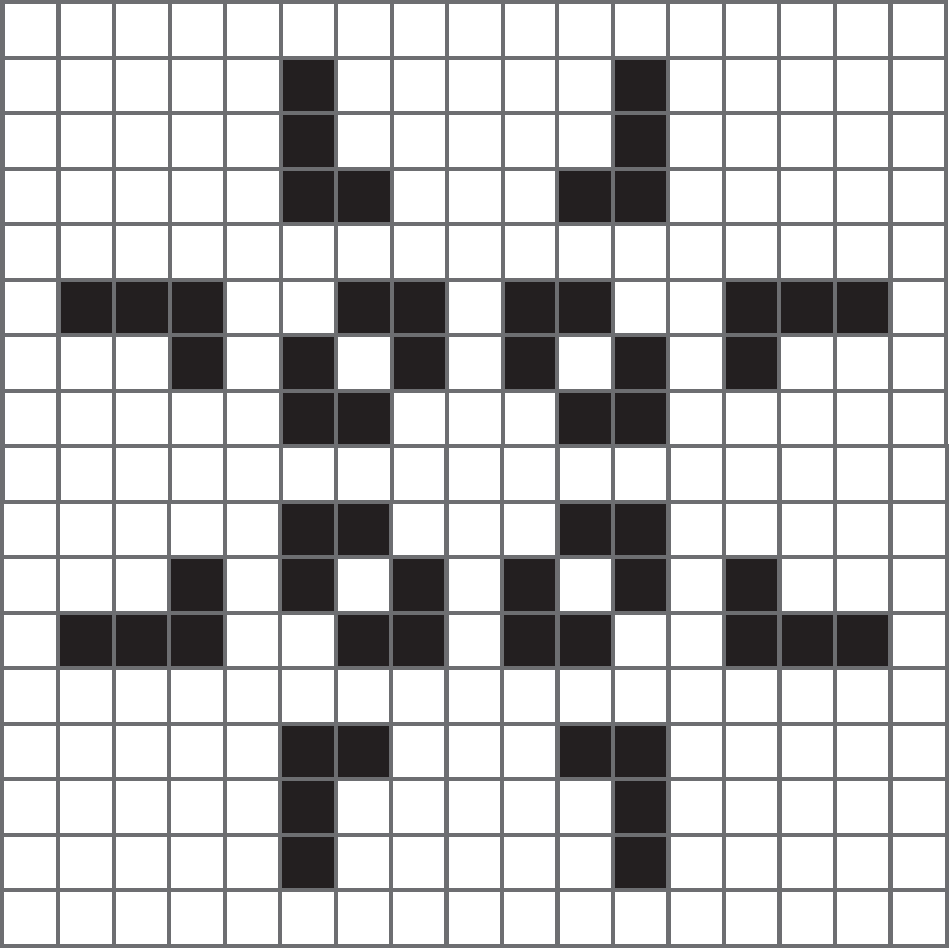}
    \caption{Period 3 oscillator\\ (\emph{pulsar}).}
    \label{fig:pulsar}
  \end{subfigure}

  \caption{Some game of life patterns with various properties. The moving
    oscillator \ref{fig:glider} moves 1 cell along the bottom right diagonal
    every 4 steps. The fixed pattern \ref{fig:still_life} never changes except
    if it interacts with others. The oscillator \ref{fig:pulsar} loops through
    three different configurations.}
  \label{fig:gol_patterns}
\end{figure}

\paragraph{Asynchronous cellular automata.}
A cellular automaton is said to be asynchronous when its cells are not
updated in parallel at each time step. Various cell update schemes can be
chosen. Asynchronous \acp{CA} have to define an
order of update of the cells, which is not easily defined on an infinite grid.
Therefore, most implementations are done on finite grids.
This update order can be random, follow a predefined order, or be decided by a
global controller. Groups of cells can be updated simultaneously, or updates can
be done cell by cell. The many effects of asynchronous updates can be difficult
to predict and can make these types of \acp{CA} relatively difficult to work with.

Asynchronous \acp{CA} are an attractive model when using large systems for which
parallel update is prohibitively expensive. The rule could adapt itself and
preferentially update cells in active or useful parts of the \ac{CA} while doing
slower updates to the less active parts of the \ac{CA}. An asynchronous \ac{CA}
with evolutionary properties was constructed in
\parencite{nehanivEvolutionAsynchronousCellular2003}. The authors argue that
asynchronicity is a strong advantage in the construction of systems that can evolve
within \acp{CA}.

\paragraph{Stochastic cellular automata.}
In a stochastic cellular automaton, the update function $\boldsymbol{\Phi}$ is
stochastic. This means that the next state of all cells in the grid is
sampled from a probability distribution that depends on the current neighborhood
configuration. We can write
\begin{equation}
\begin{aligned}
  \forall i \in \mathcal{L},\ s \in \mathcal{S} \quad p\left(c_{i}^{(t + 1)} = s \right) = \boldsymbol{\Phi} \left(\boldsymbol{N}\left(c_{i}^{(t)}\right)\right)(s).
\end{aligned}
\end{equation}
These \acp{CA} have had successful applications in physics
\parencite{vichniacSimulatingPhysicsCellular1984,
  ottaviSimulationIsingModel1989} or biology
\parencite{boasCellularPottsModel2018}, because their stochasticity allows
modeling complex physical phenomena.

\paragraph{Continuous cellular automata.}
Another family of \acp{CA} uses real-valued states, often restricted to the range
$[0, 1]$. The update function is then a real multivariate function, which we can
write
\begin{equation}
  \begin{aligned}
  \boldsymbol{\Phi}: [0, 1]^{s} \rightarrow [0, 1].
  \end{aligned}
  \label{eq:phi_cont}
\end{equation}

Examples of continuous \acp{CA} include Lenia, which uses convolutional operators
followed by thresholding for the update rule
\parencite{chanLeniaBiologyArtificial2019a}, or \acp{NCA}
\parencite{mordvintsevGrowingNeuralCellular2020} which represent \acp{CA} as
neural networks (more details in Sections \ref{sec:cell-autom-rnns} and
\ref{sec:neur-cell-autom}). \textcite{garzonRealComputationCellular1993}
explores the condition for a real-valued \ac{CA} to be able to compute
real-valued functions.

\paragraph{Higher dimensional cellular automata.}
\acp{CA} of dimension greater than 2 have been comparatively less studied for
several reasons, including the limits that arise when simulating and visualizing
systems in more than 2 dimensions on a computer screen. Another problem is with
the increasing number of possible rules in higher dimensions. The number of
neighbors per cell grows exponentially with dimension, and the number of
possible rules has a doubly exponential growth rate. This makes the
convenient representation of \acp{CA} rules as tables infeasible since the size
of that table quickly exceeds the available memory of most computers. For
example, there are $2^9 = 512$ possible $3\times 3$ configurations of a binary Moore neighborhood (see Figure~\ref{fig:moore}) in 2D that can each lead to 2 states, which means that there exists $2^{512} \approx 10^{154}$ distinct 2D rules with 2 states and a
Moore neighborhood. This is already an incomprehensibly large number, but there
are $2^{134,217,728} \approx 10^{40,403,562}$ such rules in 3D. In general, there are $2^{3^n}$ possible binary \ac{CA} rules with a Moore-like neighborhood in dimension $n$.

Despite these limitations, several works have studied higher dimensions \acp{CA},
although mostly in 3 dimensions
\parencite{tsalidesThreedimensionalCellularAutomata1989,
  sudhakaranGrowing3DArtefacts2021}. For example, there are several examples of
successful 3D \ac{CA} simulations applied to material sciences
\parencite{gandin3DCellularAutomaton1997, arataFreeformShapeModeling1999,
  panStudyFailureScale2009, dicaprio3DCellularAutomata2016}.

\paragraph{Hexagonal cellular automata.}
Hexagonal \acp{CA} are defined on grids tiled with hexagons, where each
cell has 6 direct neighbors.

\begin{figure}[htbp]
  \centering
  \begin{subfigure}[b]{.35\linewidth}
    \centering
    \includegraphics[width=\linewidth]{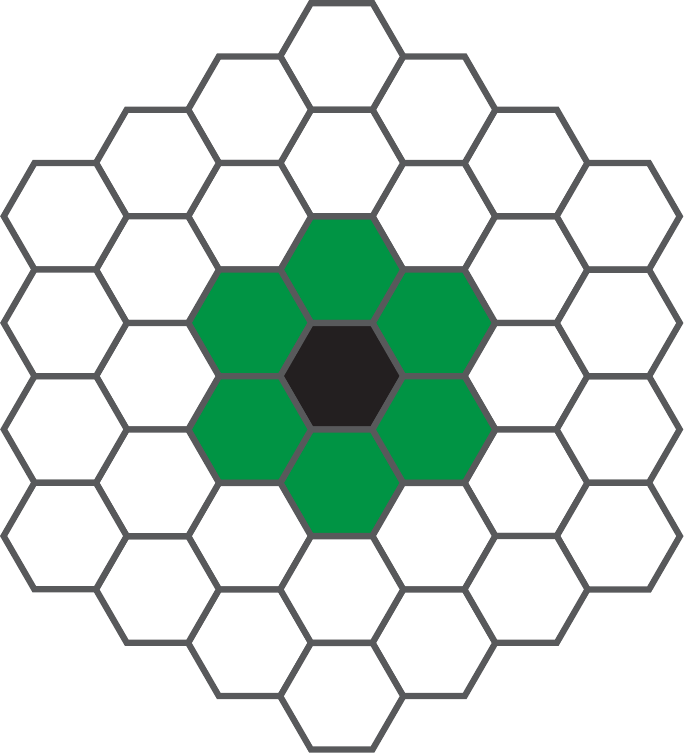}
    \caption{Range 1 neighborhood}
    \label{fig:hexagonal_1}
  \end{subfigure}
  \hspace{10pt}
  \begin{subfigure}[b]{.35\linewidth}
    \centering
    \includegraphics[width=\linewidth]{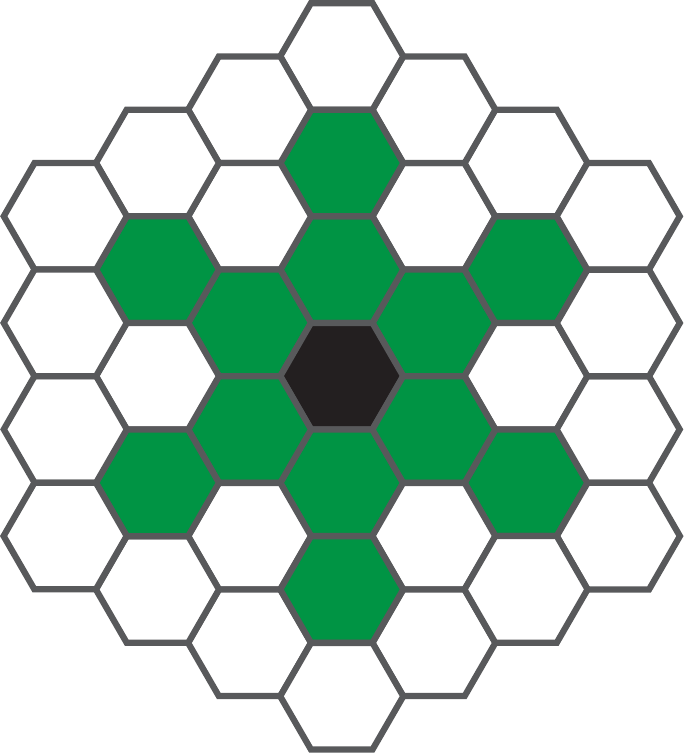}
    \caption{Ray neighborhood}
    \label{fig:hexagonal_2}
  \end{subfigure}
  \caption{Example definition of the neighborhood for hexagonal cellular automata.}
\label{fig:hexagonal}
\end{figure}

One advantage of this type of \acp{CA} over regular grids is the absence of
preferred direction since they are all equivalent. In a square grid, diagonally
adjacent cells span more distance than directly adjacent cells, whereas on a   
hexagonal grid, all neighbors are equivalent
\parencite{moreeHexagonalSquareLattice2004}.

\subsection{Parametrizing and sampling CA rules}
As explained in Section \ref{sec:challenges}, one of the main challenges of
working with complex systems is the parametrization and control of the behavior
of that system. \acp{CA} are no exception, and several parametrization schemes
have been proposed, with their respective benefits and drawbacks. There are
infinitely many ways to define rules for \acp{CA}, and for each of these
definitions, a very large number of rules are available, making it impossible to
sample a significant portion of the rule space. A good parametrization allows one 
to scan a wide range of behaviors of \acp{CA} by traversing the space in an
interesting way.

\paragraph{Langton's lambda\label{sec:langtons-lambda}.}

Christopher Langton proposed one of the first parameters that could be used to
control the behavior of \acp{CA} to a certain extent
\parencite{langtonStudyingArtificialLife1986, langtonComputationEdgeChaos1990}.
For a \ac{CA} with $K = |\mathcal{S}|$ states and a state arbitrarily chosen to
be the \emph{quiescent} state (this corresponds to the ``dead'' state in the
game of life, for example), if there are $n$ transitions to the quiescent state
within the rule table and $K^{s} - n$ remaining transitions that do not lead to
the quiescent state, where $s$ is the number of cells in the neighborhood, we
have

\begin{equation}
  \label{eq:langton}
  \begin{aligned}
    \lambda = \frac{K^{s} - n}{K^{s}}.
  \end{aligned}
\end{equation}

\begin{figure}[htbp]
  \centering
  \includegraphics[width=\linewidth]{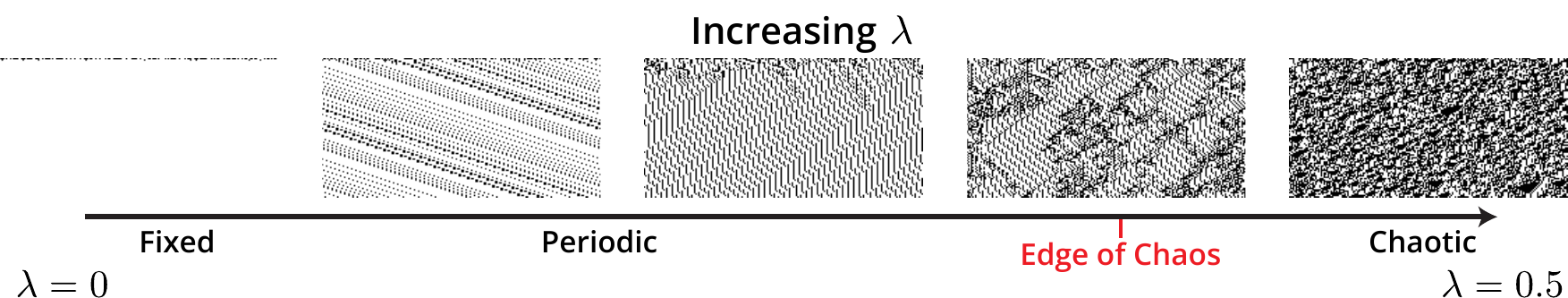}
  \caption{The effect of varying $\lambda$ for a 1D binary \ac{CA} rule. A \ac{CA}
    rule is progressively modified along a trajectory of increasing $\lambda$. The
    rule starts with a fixed behavior, becomes periodic, and then complex when
    hitting the ``edge of chaos''. Note the localized structures that spread
    over time (down along the y axis) and interact in the fourth figure. When
    $\lambda$ increases further the rule becomes chaotic.}
  \label{fig:langton_lambda}
\end{figure}

For Langton, the purpose of $\lambda$ is to search the space of \acp{CA} in an ordered
manner by varying the value of the parameter. His procedure starts with a rule
with all transitions leading to the quiescent state. The value of $\lambda$ is
increased in discrete steps to $1 - 1 / K^{s}$ by randomly changing the
transitions of the rule to lead to a different state. An illustration of the
behavior change of a \ac{CA} rule under this procedure is shown in Figure
\ref{fig:langton_lambda}. Langton collected various measures of the dynamical
behavior of \ac{CA} and studied them as a function of $\lambda$. He observes a phase
transition as $\lambda$ approaches its maximal value, where \acp{CA} seem to behave
in the most complex way, with long transients and large-scale propagating
structures. He calls this phase transition ``edge of chaos'' because it
corresponds to $\lambda$ values just before the generated \acp{CA} become
chaotic.

\paragraph{Dirichlet sampling.}
For an arbitrary number of states and an arbitrary neighborhood size, it can be
challenging to sample \acp{CA} while scanning the wide range of behavior these
models are capable of simulating. This is because the size of the space of
\acp{CA} rules is very large. The naive method of uniformly sampling each
transition output is not useful in large rule spaces with multiple states, large
neighborhoods, or grid dimensions larger than 1.

Rules with equal proportions of transitions leading to all states tend to be the
most chaotic. This is also what Langton observed when studying the $\lambda$
parameter. This is explained by the fact that sampling each transition output
uniformly is equivalent to sampling the rules from a multinomial distribution
with the number of possible states ($K$), number of possible transitions
($n = K^{K^{s}}$, with $s$ the number of cells in a neighborhood), and the uniform
probabilities $p_{1}= \ldots= p_{K}$ as parameters. A random variable
$X = (X_{1}, \ldots, X_{K})$ indicates the number of times each result is observed.
The probability mass function of this distribution is

\begin{equation}
  \label{eq:multinomial}
  \begin{aligned}
    P(X_{1}=x_{1}, \ldots , X_{K}=x_{k}) = \frac{n!}{x_{1}!\cdots x_{K}!} p_{1}^{x_{1}}\cdots p_{K}^{x_{K}}.
  \end{aligned}
\end{equation}

This quantity becomes vanishingly small for large $n$ (that is, rules with many
states or large neighborhoods) and rules with an output transition skewed
towards a specific state. For example, the binary Game of Life rule has 372
transitions leading to state 0 and 140 leading to state 1. The probability
of sampling a rule with these transition proportions is $8.24\mathrm{e}{-26}$
(compared to a $17\%$ probability of sampling a rule with between 254 and 257
transitions to 0). Using a uniform transition sampling method makes it highly 
improbable to obtain a rule resembling the Game of Life, let alone the rule itself.

We propose a Dirichlet sampling method as an alternative way to sample rules
with fixed ratios of transitions leading to each output state. For a rule with
$K$ states, we sample a $K$-uple from a Dirichlet distribution of order $K$ with
parameters $(\alpha_{k})_{k\in [1, K]}$ where $\alpha_{0} = \alpha_{1} = \ldots = \alpha_{K} = \alpha < 1$. The
result is a quantile $K$-uple $(q_{1}, \ldots, q_{K})$. For $\alpha < 1$, the distribution
is concentrated around the corners of a simplex of dimension $K$, which means
that it preferably samples a tuple with one of its values dominating the others.

\begin{figure}[htbp]
  \centering
  \includegraphics[width=.5\linewidth]{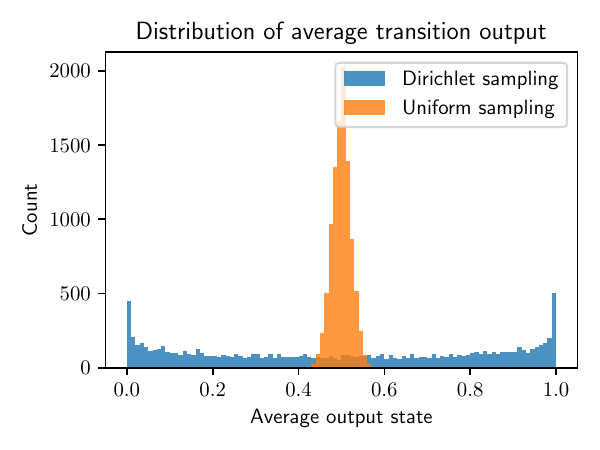}
  \caption{An illustration of the difference between naive uniform rule sampling
    and Dirichlet sampling on binary rules. 10000 2D binary CA rules were
    sampled with each method, the plot displays a histogram of their average
    output state. Uniform sampling leads to more rules with an average output
    state close to 0.5, meaning approximately as many transitions toward both
    states. Dirichlet sampling can control, through the choice of parameter $\alpha$,
    how many rules with more skewed transitions should be sampled, which is
   shown by the spikes on the blue histogram close to 0 and 1.}
  \label{fig:ca_rule_sampling}
\end{figure}

Rules are sampled so that the number of transitions to each output state matches
the quantile generated from the Dirichlet distribution. The samples will be more
likely to have a dominant quantile, which can be associated with the quiescent
state in Langton's $\lambda$ calculation. The resulting sampling of cellular automata
is much better for scanning the entire space of rules and generating \ac{CA}
rules that would be impossible to reach with the naive sampling method, as illustrated
in Figure \ref{fig:ca_rule_sampling}.

\paragraph{Smooth sampling with the recurrent convolutional neural network
  analogy.}

The recurrent convolutional network analogy of \acp{CA} (see Section
\ref{sec:cell-autom-rnns} for details) formulates a \ac{CA} as a neural network.
In this paradigm, the rule of a \ac{CA} is completely determined by the parameters
of that neural network. This allows us to modify the \ac{CA} rules as we would
change the weights of a neural network. For example, as long as all simulation
steps are differentiable, the backpropagation algorithm can be applied and used
to update a \ac{CA} rule by updating the weights of the neural network with
gradient descent. Dimensionality reduction methods applied to the parameter
space of the neural network can help to understand the structure of that space and
the \acp{CA} behaviors produced by different sets of parameters.

\section{Cellular automata and Recurrent neural networks}\label{sec:cell-autom-rnns}

The purpose of this section is to show that cellular automata and recurrent
convolutional neural networks have very strong connections and to draw this
parallel as clearly as possible. We express the \ac{CA} update function as a set
of convolutional operations that can be derived from any \ac{CA} rule.

This connection yields interesting consequences for both the theoretical
properties of these models and the potential applications of \acp{CA} and
recurrent networks. It shows, for instance, that emergent properties with
increasing complexity and perhaps open-ended development that we expect from
complex systems such as \acp{CA} are possible within the hidden state of a
recurrent neural network.

The conceptual similarity between \acp{CA} and neural networks is not surprising. 
Fundamentally, \acp{CA} represent the paradigm of emergence, in which
the origins of complex behaviors are searched for in an assemblage of possibly
very simple parts rather than being viewed as a sum of complex building blocks. It is
often argued that there is no better-known example of a truly emergent
phenomenon than that of the emergence of consciousness out of the large network
of functionally simple (and certainly unconscious) neuronal components that make
up the human brain. A biological neural network consists of a
large space of interconnected nodes with dynamic behavior that is a local
function of the other nodes to which it is connected, which is strikingly similar
to the principle of a \ac{CA}. Artificial neural networks can
be thought of as being a set of biologically inspired CA rules
\parencite{ilachinskiCellularAutomataDiscrete2001}. The more precise parallel
between cellular automata and a form of recurrent/convolutional network has also
been drawn by several other researchers
\parencite{wulffLearningCellularAutomaton1993,
  gilpinCellularAutomataConvolutional2018,
  mordvintsevGrowingNeuralCellular2020}.

In this section, we formally describe the relation between \acp{CA} and \acp{RNN}
and discuss the extensions this implies for the \ac{CA} model, as well as some of
the consequences of that relation.

\subsection{RNN formalism}

\begin{figure}[htbp]
  \centering
  \includegraphics[width=\linewidth]{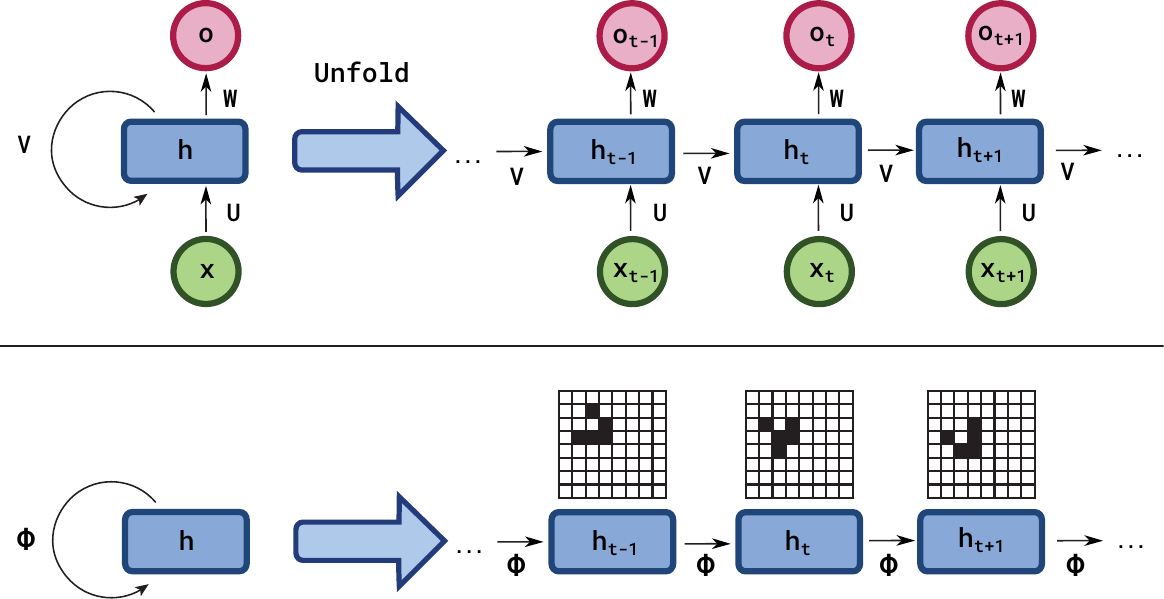}
  \caption{\label{fig:standard_rnn} Standard RNN architecture (top) and Game of
    Life seen as a RNN with no inputs and outputs (bottom). For the \ac{CA}, the
    ``hidden'' state $h_t$ is the current state of the grid at timestep $t$. The
    operator $\boldsymbol{\Phi}$ is the \ac{CA} update rule which is
    equivalent to a \ac{CNN} (see Section~\ref{sec:transition-rule-as}). This
    illustration is based on
    \href{https://commons.wikimedia.org/wiki/File:Recurrent_neural_network_unfold.svg}{Recurrent
      neural network unfold} by
    \href{https://commons.wikimedia.org/wiki/User:Ixnay}{fdeloche}, licensed
    under \href{https://creativecommons.org/licenses/by-sa/4.0/}{CC BY 4.0}.}
\end{figure}

\begin{figure}[htbp]
  \centering
  \includegraphics[width=.3\linewidth]{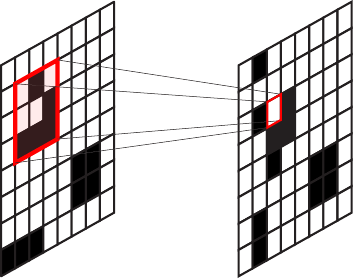}
  \caption{\label{fig:ca_cnn}The CA update rule is local and can be represented
    by a set of \acp{CNN}: linear local uniform transformations
    followed by the application of a non-linear function.}
\end{figure}

We write the definition of a \ac{CA} with a formalism and notation inspired by \acp{RNN}. This parallel is illustrated in Figures~\ref{fig:standard_rnn} and \ref{fig:ca_cnn}. The state of the grid at time $t$ is denoted $h_t$ and
corresponds to the hidden state in a RNN\@. In the case of classical \acp{CA},
it is a 1 or 2D vector of discrete values (see Section \ref{sec:definition}),
but~\parencite{mordvintsevGrowingNeuralCellular2020} and other \acp{CA}
extensions use continuous values, much like usual \acp{RNN}.

The mapping $\Phi$ operates only on the hidden state. Because it is local,
it can be shown to be equivalent to a convolutional layer, as explained below in
Section~\ref{sec:transition-rule-as}. The inputs and outputs
$(\mathbf{x}, \mathbf{o})$ in Figure \ref{fig:standard_rnn} are not included
in the classical definition of \acp{CA}, but are an easy-to-implement extension, as
discussed in Section~\ref{sec:adding-inputs-outp}.

\subsection{Transition rule as a set of convolutions\label{sec:transition-rule-as}}

\begin{figure}[htbp]
  \centering
  \includegraphics[width=.4\linewidth]{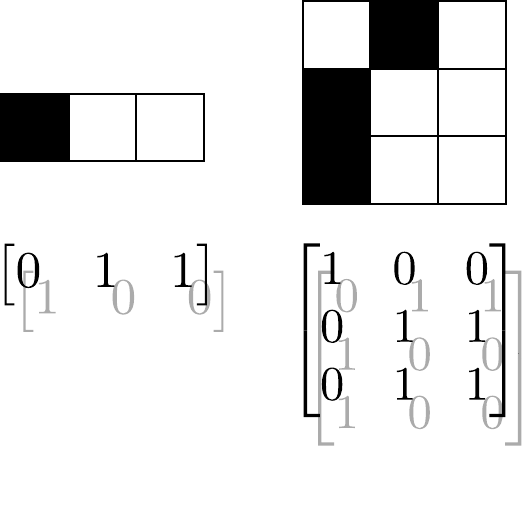}
  \caption{\label{fig:cell}A \ac{CA} neighborhood vector representation example
    with 2 states. A $3\times 3$ square of cells with two states can be represented
    by a $3\times 3 \times 2$ tensor. Left: 1D 3-neighbors representation. Right: 2D $3\times3$
    neighbors representation. The last dimension of the vector encodes a
    ``one-hot'' representation of the state of the cell.}

\end{figure}

Each cell $c_i$ from the \ac{CA} definition can be in one of the $k$ states. We
represent the cells as vectors of size $k$, where $k$ is the number of states, with only
zeros excepted for a one in the $k$-th position, which can be interpreted as a 
one-hot encoding of the cell state. A neighborhood of size $3$ in a
1D CA can be represented as a $3 \times k$ vector
$\mathbf{u}_i = [u_{i-1}, u_{i}, u_{i+1}]$. Each $u_{i}$ is a vector of size $k$
with a $1$ in position $s_i$. This is illustrated in the 1D (resp. 2D) case on
the left (resp. right) of Figure~\ref{fig:cell}. For a CA with only two states,
it is redundant to have a $3 \times 2$ vector, but the ``one-hot'' encoding
becomes useful when working with more states.

With this representation, we can express the transition rule $\Phi$ as a simple
convolutional neural network. This network would be made up of two layers, which
are shown in Figure~\ref{fig:ca_cnn_2dims} for the 2D case. We describe a simple construction of a \ac{CNN} representing any binary \ac{CA} rule below:

\begin{figure}[htbp]
  \centering
  \begin{subfigure}[b]{\linewidth}
    \centering
  \includegraphics[width=.9\linewidth]{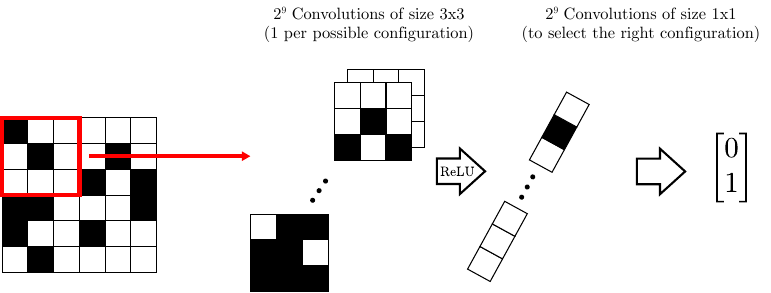}
  \caption{\label{fig:global_schema}A representation of any binary 2D CA rule
    operating on $3\times 3$ neighborhoods as a 2-layer \ac{CNN}. There are 512 filters
    receptive to each $3\times 3$ neighborhood states in the first layer. The second
    layer determines the output value for each neighborhood state.}
  \end{subfigure}
  \vspace{20pt}
  \begin{subfigure}[b]{\linewidth}
    \centering
  \includegraphics[width=.9\linewidth]{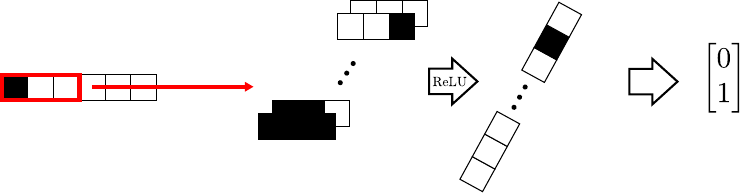}
  \caption{\label{fig:one_d_global_schema}A representation of any binary 1D CA
    rule operating on neighborhoods of size 3 as a 2-layer \ac{CNN}. There are 8
    filters of size $3\times 3$ in the first convolutional layer, and 8 filters of size $1\times 1$
    in the second one.}
  \end{subfigure}
  \caption{Two examples of \ac{CA} rules represented as a simple CNN. Padding
    ensures the grid obtained after the CNN step is of the same size as the
    original grid.}\label{fig:ca_cnn_2dims}
\end{figure}

\paragraph{The first convolutional layer.} It is receptive to each possible
neighborhood configuration. In the 1D case with neighborhood size 3, this is all
eight configurations $000, 001, 010, \ldots, 111$. For a radius $r$ in 1D, this layer
is composed of $k^{2r + 1}$ filters of dimension $k \times (2r + 1)$ with only ones
and zeros, which are the mirror of a possible configuration of size $(2r+1)$ of
the neighborhood.

The product of each filter with an input neighborhood in the grid will be an
integer between $0$ and $2r + 1$. Applying a ReLU nonlinearity preceded by a
constant vector of value $2r$ as bias, we obtain the input to the second layer,
a vector of size $k^{2r + 1}$ for each cell containing all zeros except for
one corresponding to the detected neighborhood. This vector plays the role of an
indicator of the current input configuration.

\paragraph{A second convolutional layer.} It has $k^{2r + 1}$ filters of size $1$
that are $0$s or $1$s depending on the desired output value of that transition.
When applied to the indicator vector above, the output obtained will be the
output value of the rule corresponding to the detected neighborhood state.

With some circular padding that wraps around the edges of the grid or applies
constant values outside of the main grid, the size of the output is the same as
the input. This means that the CNN step can be applied multiple times, simulating a
single \ac{CA} step every time.



\subsection{Adding inputs and outputs\label{sec:adding-inputs-outp}}

Figure ~\ref{fig:standard_rnn} shows the locations where inputs and 
outputs can be incorporated into the Game of Life or any other \ac{CA}. 
We provide a variety of examples of how input and output can be integrated 
and how they can interact with the rule of the \ac{CA}.

\subsubsection{Inputs}
There are many ways to add inputs to the recurrent \ac{CNN} model presented
above. We divide our proposed methods into two classes: inputs that directly
modify the hidden state during the CA evolution and inputs that augment the
hidden state without changing it directly and affect the rule.

\paragraph{Hidden-state augmentation and rule modulation.}
Inputs are usually added to the hidden state just before the application of the
non-linearity in standard RNNs.

In the case of cellular automata, it can also be done in the following ways:

\begin{figure*}[ht]
  \centering
  \includegraphics[width=.9\linewidth]{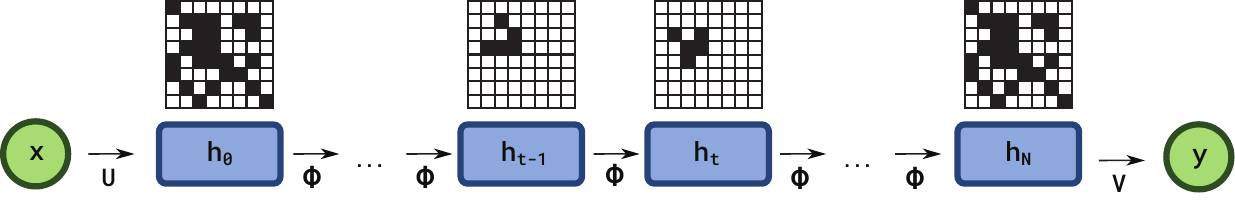}
  \caption{\label{fig:encode_decode} Inputs $x$ and outputs $y$ can be directly
    encoded within the hidden state. $U$ and $V$ are the operators mapping $x$ 
    to a hidden state tensor and decoding a hidden state tensor to output $y$.}
\end{figure*}

\begin{itemize}
  \item \textbf{Vector state.} A straightforward way to add inputs to a \ac{CA} is to
        consider the grid as having an additional (possibly read-only)
        dimension. For example, a 1D automaton of size $N$ would be represented by
        a vector of dimension $N\times 2$ where the first vector of size $N$ is the
        state of the grid and the other is the input. The update rule would be
        changed to take into account this new component. One can see this model
        as using multiple \ac{CA} rules at the same time, with the input state
        conditioning the rule chosen for a given update step.

  \item \textbf{Variable update rule.} Inputs can also influence a CA's evolution by
        changing the update rule. With this configuration, the update rule $\Phi$
        is now a function of the input $x$. We now have $\Phi_x = G(x)$, where

            \[G: \mathcal{X} \rightarrow \left({\{ 1, \ldots, k \}}^{2r+1} \to \{1, \ldots, k\}\right),\]

        and $\mathcal{X}$ is the input space.
        
        A similar approach was used in
        \parencite{adamsFormalDefinitionsUnbounded2017}. It showed that
        conditioning a CA update rule on another CA's evolution could enable a
        higher diversity of behavior.

  \item \textbf{Concatenation.} One way to incorporate input is by concatenating 
  it with the hidden state, such as at the boundaries of the grid, before applying the
  update rule. This creates a fixed read-only memory space, as shown in 
  Figure~\ref{fig:concat}. However, a potential drawback of this method is that 
  information must propagate through the grid in order to be processed in locations 
  far from the read-only memory's position. 
        
        This approach is more reminiscent of computer architecture, where designated 
        areas are reserved for specific data functions.

\begin{figure}[ht]
  \centering \includegraphics[width=.5\linewidth]{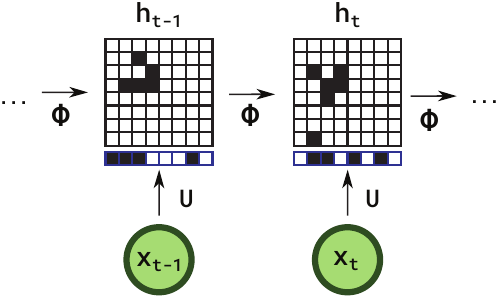}
  \caption{\label{fig:concat} Input $x_t$ is projected and concatenated to
    hidden state $h_t$, affecting the boundary conditions of the 
    \ac{CA}. For the \ac{CA}, this vector is just like another part of the
    hidden state except that it is not writable}
\end{figure}
\end{itemize}

\paragraph{Hidden-state manipulation.}
Another approach is to directly modify the hidden state to ``communicate''
information to the system. These methods are used
in~\parencite{mordvintsevGrowingNeuralCellular2020,
  randazzoSelfclassifyingMNISTDigits2020} to make interactive demos and allow
users to directly modify that hidden state. A user can interact with the system
by drawing with its mouse. This sets parts of the internal state to a fixed cell
state.

\begin{itemize}
  \item \textbf{Masking.} Input data can serve as a mask on top of the current
        grid state, forcing some cells into states that depend on the input
        values. If we view the state activations in the CA as some neural
        pattern, this approach can be seen as a form of \emph{neuromodulation},
        which has previously been used in machine learning
        \parencite{soltoggioEvolutionaryAdvantagesNeuromodulated2008,
        ishiguroNeuromodulatedControlBipedal2003,
        beaulieuLearningContinuallyLearn2020}.

        A mask can be constructed from an input vector $x$ by linearly
        transforming $x$ and applying an activation function (\eg~step
        function). It controls which channels are activated and where
        information can flow or not. This kind of mask can be applied with
        element-wise multiplication or addition but also more elaborate
        operations. It is illustrated in Figure~\ref{fig:mask}.

\begin{figure}[ht]
  \centering
  \includegraphics[width=.5\linewidth]{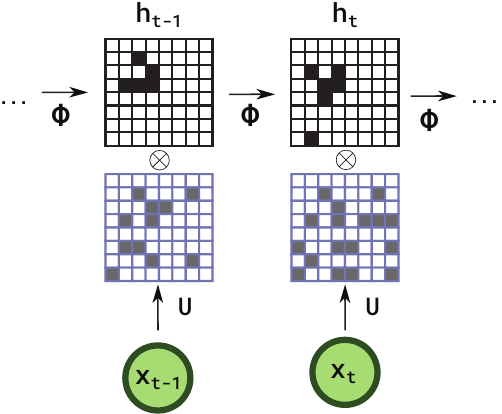}
  \caption{\label{fig:mask} Input is converted into a binary mask for the hidden
    state. The mask is applied through element-wise multiplication here.}
\end{figure}

\end{itemize}

\subsubsection{Outputs}

Similarly, inputs can be extracted from the hidden state in several different
ways. One might read outputs from the boundaries of the grid, from a
transformation of the grid using a neural network, etc. The output could also be
stored within an additional dimension of the ``extended'' grid state presented
above.

\subsubsection{Cellular automata and computations}

From a computational point of view, the hidden state can be seen as a working
tape on which some kind of \emph{parallel Turing machine} (the cellular automaton)
performs computations. In this framework, the input is encoded as the
initial state of the tape and decoded from the last state of the tape (as
illustrated in Figure~\ref{fig:encode_decode}). The recurrent convolutional
neural network (RCNN), or \ac{CA}, then plays the role of a fixed computer
program that is executed. This is the setting adopted in previous work on
constructing computations with cellular automata: a task is chosen, and rules that 
can execute that task on input/output pairs
are searched~\parencite{mitchellComputationCellularAutomata2005}.

Another interesting point of view can be taken when we consider the
hidden state (or tape) as encoding both some data and a computer program, as is
the case with a Turing machine. We can then expect a \ac{CA} (or \ac{RNN}) to not
only compute the result of a particular function but to be equivalent to a
general-purpose computer capable of computing the result of any chosen algorithm
without any need for optimization. Making a universal \ac{CA} compute some
program amounts to encoding the program in the right language and
communicating it by embedding it within the hidden state. In practice, figuring
out the right encoding is often the biggest challenge. This is also the reason
that the few Turing complete \acp{CA} are not so useful in practice since we do
not know of any efficient encoding for programming them.

\subsection{Consequences}

Viewing cellular automata as recurrent convolutional neural networks, as described above, has several interesting consequences, of which we list a few
here.

\subsubsection{Turing-completeness of the system}

Because we can simulate rule 110 ECA in the above RCNN system, it follows from
the Turing completeness of this CA rule that the RCNN is Turing complete. This
is an interesting result, although not very significant since \acp{RNN} have
already been proven to be Turing
complete~\parencite{siegelmannComputationalPowerNeural1992}, with very few
practical implications. Our model is relatively far from a real-world CNN with a
fixed number of layers independent from one another --- compared to a variable
number of steps and shared layers for the automaton-RCNN. Thus, this property
does not seem to bear any consequence on \acp{CNN}.

\subsubsection{Differentiable cellular automata}

Each step of the computations involved in computing a step of a cellular
automaton represented as a RCNN is differentiable. Therefore, we can
theoretically couple this framework with backpropagation to create
\emph{learnable} cellular automata that can adapt their rules to minimize a
target loss function.

This is the direction taken by~\textcite{mordvintsevGrowingNeuralCellular2020}.
The authors use supervised learning on a \ac{CA} and train it to obtain a stable
self-repairing target shape. However, because supervised \ac{CA} can only do as
much as they have been trained to do, it may defeat the purpose of working with
a model capable of spontaneous complex emergent behavior. An open-ended
complexity increase could very unlikely be achieved through pure supervision.

\textcite{gilpinCellularAutomataConvolutional2018} takes the reverse approach
and tries to train RCNNs to simulate a fixed \ac{CA} rule, using the statistics
of that training process as a way to help understand the structure of the
\ac{CA} rule space.

\subsection{Beyond the naive rule representation}

The representation of \ac{CA} rules that we built in Section
\ref{sec:transition-rule-as} is a one-to-one mapping. Each rule has its
R\ac{CNN} counterpart, and vice versa. However, there are several ways to make
that representation of \ac{CA} rules more efficient, using a neural network with
fewer parameters. For example, we can make the first layer of filters receptive
to both a given configuration and its inverse (\eg~$(1, 0, 1)$ and $(0, 1, 0)$
in an \acl{ECA}) by using negative values in the filter (use a filter $(1, -1, 1)$
instead of $(1, 0, 1)$ and $(0, 1, 0)$ separately).

For example, the rule of the game of life does not need more than two convolutional
filters on the first layer to be represented by a \ac{CNN} because it is
totalistic. The new state of a cell depends only on its number of neighbors in
the alive state and the current state of the cell.

\section{Reservoir computing \label{sec:res-models}}
\begin{figure}[htbp]
  \centering
  \includegraphics[width=.8\linewidth]{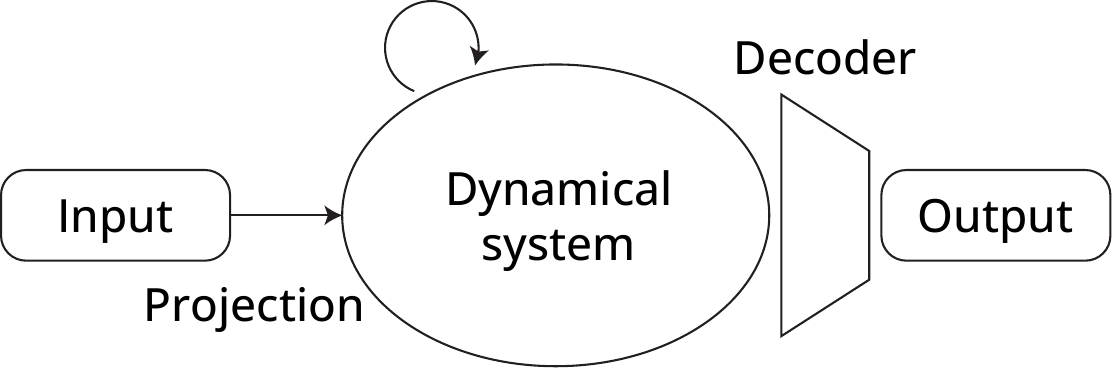}
  \caption{Diagram of the general functioning of a \acf{RC} system. Input values
    are projected in the state or evolution function of a dynamical system which
    is run according to its internal rule. Output values are decoded from the
    internal state of the dynamical systems with a trainable layer.}
  \label{fig:reservoir_diagram}
\end{figure}

\Acf{RC} is a computational framework that aims to exploit the states of a
complex dynamical system to perform some target task
\parencite{tanakaRecentAdvancesPhysical2019}. It relies on a \emph{reservoir} of
computations, that is, a dynamical system performing some computations on its
internal state. An input signal is fed into this reservoir that behaves as a
black box from the point of view of the algorithm. Input values can be
sequential or nonsequential. They are projected and combined with the dynamical
system using a suitable mechanism. This projection can be learned, set randomly,
or chosen. The internal state of the dynamical system evolves according to its
update rule, which is defined by the chosen system. Finally, a decoder model
(usually a linear regression) is trained in a supervised manner to extract
necessary information from the internal state of the dynamical systems to
predict the right output. A general diagram describing this process is shown in
figure \ref{fig:reservoir_diagram}.

Some early known formulations of the \ac{RC} idea were done by
\textcite{kirbyContextDynamicsNeural1991} and
\textcite{schomakerNeuralNetworkModels1990,
  schomakerSimulationRecognitionHandwriting1991,
  schomakerNeuralOscillatornetworkModel1992}. In another early work,
\textcite{buonomanoTemporalInformationTransformed1995} used a random spiking
neural network with excitatory and inhibitory elements. The probabilities of neuron 
connection was inspired by the real rat brain data
\parencite{masonSynapticTransmissionIndividual1991}. The authors observed an interval-sensitive response of these neurons when subjected to pulses spaced by varying
time intervals. This indicates an ability to encode temporal information. To
measure this phenomenon, they trained an output linear layer to recognize specific
patterns in the activation of the last layer of neurons. This idea of using a
randomly initialized \ac{RNN} with fixed weights and training a simple linear layer
to decode outputs was reinvented and popularized later under the names
``Echo-state networks'' \parencite{jaegerEchoStateApproach2001} and ``Liquid
state machines'' \parencite{maassRealTimeComputingStable2002}.

Another advantage of reservoir computing is that reservoirs can be
implemented using a variety of physical systems, substrates, and devices. These
types of reservoir are sometimes called ``exotic''
\parencite{lukoseviciusReservoirComputingApproaches2009} in contrast to
\ac{RNN}-based reservoirs. Physical \ac{RC} is a promising area of research
that may help us develop cheap and efficient machine-learning hardware and can
inform us about the nature and behavior of the underlying physical systems.
\textcite{tanakaRecentAdvancesPhysical2019} identifies four characteristics that
make a reservoir suitable for solving a computational task.

\begin{enumerate}
  \item The inputs must be mapped to a high-dimensional space, which
        corresponds to the internal state of the reservoir. This
        high-dimensionality ensures that the inputs are separated with high
        probability.

  \item The reservoir should be non-linear. This will allow inputs that are not
        linearly separable to be separated as they are projected within the
        reservoir.

  \item The reservoir should have the echo-state property
        \parencite{jaegerEchoStateApproach2001, yildizRevisitingEchoState2012},
        or fading memory \parencite{boydFadingMemoryProblem1985,
        maassRealTimeComputingStable2002, maassFadingMemoryKernel2004}. This
        property relates the asymptotic properties of the excited reservoir dynamics
        to the driving signal. A reservoir with the echo-state property will
        asymptotically forget all input values. This makes some reservoirs unable
        to solve tasks where unbounded memory is possibly required, such as
        context-free grammar parsing
        \parencite{schmidhuberTrainingRecurrentNetworks2007}. Later, reservoir
        models that address this issue were developed
        \parencite{pascanuNeurodynamicalModelWorking2011}.

  \item A reservoir should separate responses from different signals into
        different parts of the space while being insensitive to small
        perturbations of the input. For a given system, this trade-off is often
        obtained at the transition between chaotic and non-chaotic regimes
        \parencite{bertschingerRealtimeComputationEdge2004,
        legensteinEdgeChaosPrediction2007}.
\end{enumerate}

We list common types of \ac{RC} that are based on various families of reservoirs
below:

\begin{description}
  \item[Dynamical systems.] This refers to a broad category of \ac{RC} models
        based on other dynamical systems than the random \acp{RNN}. This
        includes reservoir computing with \aclp{CA}, which we give more detail
        about in section \ref{sec:app-ca-res}.
  \item[Electronic \ac{RC}.] These models are implemented on electronic circuits,
        such as artificial neural networks implemented on physical circuits or
        other neuromorphic circuits. For example, \ac{RC} models can be build on
        FPGAs \parencite{antonikApplicationFPGAReal2018,
        verstraetenReservoirComputingStochastic2005,
        alomarLowcostHardwareImplementation2014,
        antonikFPGAImplementationReservoir2015} or memristive circuits
        \parencite{yangInvestigationsStaircaseMemristor2016,
        merkelMemristiveReservoirComputing2014,
        donahueDesignAnalysisNeuromemristive2015}.
  \item[Mechanical \ac{RC}.] Soft mechanical bodies have complex nonlinear
        dynamics that can be used as a reservoir to build a \ac{RC} model
        \parencite{pfeiferHowBodyShapes2007}. For example, a reservoir using a
        mass-spring network was proposed by
        \textcite{hauserTheoreticalFoundationMorphological2011}.
  \item[Biological \ac{RC}.] There have been multiple hypotheses on whether the
        parts of the brain behave like a \ac{RC} system
        \parencite{yamazakiCerebellumLiquidState2007}. The goal of this area of
        research is to study the \ac{RC}-like properties of
        biological systems rather than build reservoir computers. For example, a
        \ac{RC} model was proposed to understand the mechanism of
        context-dependent eye movement
        \parencite{domineyComplexSensorymotorSequence1995,
        domineyModelCorticostriatalPlasticity1995}.
  \item[Photonic \ac{RC}.] The principle of photonic \ac{RC} is to use optical
        reservoirs to generate the computations. A first example using
        semiconductor optical amplifiers (SOAs) was proposed by
        Vandoorne and colleague \parencite{vandoorneOpticalSignalProcessing2008,
        vandoorneParallelReservoirComputing2011} and subsequently built \parencite{vandoorneExperimentalDemonstrationReservoir2014}.
        The photonic implementation of reservoir computing is based on the idea that
        photonic accelerators \parencite{kitayamaNovelFrontierPhotonics2019} can
        realize fast information processing with low
        learning costs \parencite{paquotOptoelectronicReservoirComputing2012,
        martinenghiPhotonicNonlinearTransient2012,
        suganoReservoirComputingUsing2020, antonikHumanActionRecognition2019}.
        Some of the implementations use the scattering of light in complex media
        \parencite{dongOpticalReservoirComputing2020,
        rafayelyanLargeScaleOpticalReservoir2020}.
  \item[Spintronics \ac{RC}.] Some \ac{RC} works used nanoscale electronics
        involving the charge and spin of electrons called \emph{spintronics}
        \parencite{wolfSpintronicsSpinBasedElectronics2001}. Spintronics has
        been used in multiple ways to build \ac{RC} models, using spin torque
        oscillators (STO)
        \parencite{torrejonNeuromorphicComputingNanoscale2017,
        williameChaoticDynamicsMacrospin2019}, or spin waves
        \parencite{nakaneReservoirComputingSpin2018} for example.
\end{description}

\subsection{Reservoir computing applications}

\ac{RC} has had successful applications in various fields. It has been applied
for pattern classification; in audio with spoken digit recognition
\parencite{verstraetenIsolatedWordRecognition2005} and waveform classification
\parencite{paquotOptoelectronicReservoirComputing2012}; in computer vision with
written digits recognition \parencite{jalalvandRealTimeReservoirComputing2015},
and human motion classification \parencite{sohIterativeTemporalLearning2012,
  antonikHumanActionRecognition2019}. Another popular application of \ac{RC} is
time-series forecasting \parencite{jaegerEchoStateApproach2001,
  jaegerAdaptiveNonlinearSystem2002, wyffelsComparativeStudyReservoir2010}.
\ac{RC} has also been used in reinforcement learning to control agents in
artificial environments \parencite{kannoPhotonicReinforcementLearning2022} or
learn radio spectrum access \parencite{changDistributiveDynamicSpectrum2019}.

\subsection{Echo-state network}

\begin{figure}[htbp]
  \centering
  \includegraphics[width=.8\linewidth]{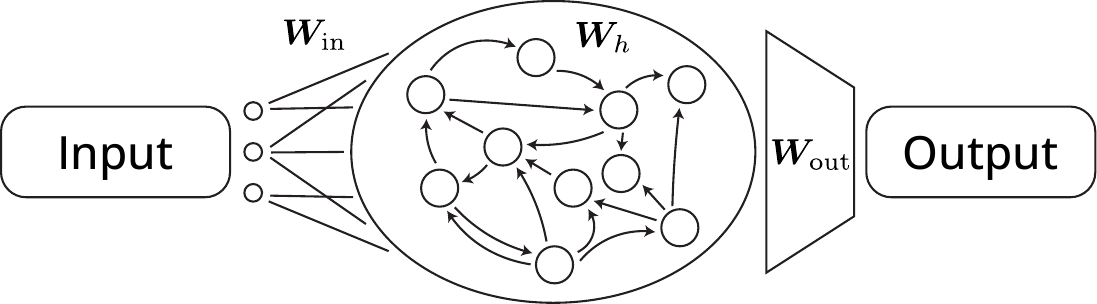}
  \caption{Diagram of an echo-state network. The \ac{RNN} is reprensented
    ``flattened'' in time. The circles represent the units of the hidden state
    of the \ac{RNN}. An arrow between unit $i$ and $j$ corresponds to a non-zero
    entry in the matrix, $\mW_{h, ij} \neq 0$. Non-linear operations are not
    represented.}
  \label{fig:echo_state_network}
\end{figure}

The \acf{ESN} (and simultaneously developed model \emph{Liquid state machine}
(LSM)) is the most well-known implementation of the \ac{RC} principle
\parencite{tanakaRecentAdvancesPhysical2019}. The main idea behind this model is
to drive a \ac{RNN} with some input signal, which will excite the neurons within
the reservoir and induce several nonlinear response signals. This is combined
with a trainable linear regression to extract a desired output from a
combination of these response signals. The \ac{ESN} model is illustrated in 
Figure \ref{fig:echo_state_network}.

In practice, the input projection is often fixed, but it can also be trainable.
The output can also be fed back to the reservoir as an input, and the
structure of the \ac{RNN} reservoir can be adapted to various applications.

We define an input sequence as $(\vx_{t})_{t=1}^{T} \in 
\mathbb{R}^{ n \times T}$, where $n$ is the input dimension and $T$ is the 
number of input time steps. The
initial state of the reservoir at $t = 0$ is $\vr_{0}$. The echo-state network
is based on the following update equation:
\begin{equation}
  \label{eq:esn}
  \vr_{t + 1} = (1 - \beta) \vr_{t}
  +
  \beta \tanh(\mW_{h}\: \vr_{t} + \mW_{\text{in}}\: \vx_{t + 1}),
\end{equation}
where $\vr_{t}$ is the $K$-dimensional state vector corresponding to
hidden neurons --- at time $t$, $\beta$ is the leak rate,
$\mW_{h} \in \mathbb{R}^{K \times K}$ is a sparsely connected random hidden
layer matrix and $\mW_{\text{in}} \in \mathbb{R}^{L \times K}$ is the input
projection matrix. The matrices $\mW_{h}$ and $\mW_{\text{in}}$ have
random values.

For the random initialization of $\mW_{h}$ and $\mW_{\text{in}}$,
\textcite{jaegerLongShortTermMemory2012} recommends: $\mW_{h}$ should have an
average of 10 non-zeros entries per row, all sampled uniformly in $[-1, 1]$. The
matrix is then scaled to achieve a set spectral radius $\rho$, which is optimized
for each experiment in \parencite{jaegerLongShortTermMemory2012}.
$\mW_{\text{in}}$ has its entries uniformly sampled in $[-1, 1]$. The
columns of the matrix are then individually scaled with factors $\sigma_{1}, \ldots, \sigma_{L}$
specific to each experiment. In this formulation, there are multiple parameters
to optimize:

\begin{itemize}
  \item The size of the reservoir $K$
  \item The spectral radius of $\mW_{h}$, $\rho$
  \item The input weight scaling parameters $\sigma_{1}, \ldots, \sigma_{L}$
  \item The leaking rate $\beta$.
\end{itemize}

\textcite{jaegerLongShortTermMemory2012} explores three
optimization schemes for these parameters:
\begin{description}
  \item[Blind.] The input weight scaling parameters are set to a single
        value $\sigma$. The parameters $K$, $\rho$, $\sigma$, and $\beta$ are
        optimized. This corresponds to a search space of dimension 4.
  \item[Basic.] Each $\sigma_{i}$ is optimized individually or in
        groups. The other parameters are also optimized.
  \item[Smart.] All parameters are optimized as in the \textbf{Basic}
        scheme, and the weights of $\mW_{h}$ are also designed specifically for
        the target task.
\end{description}

The $L$-dimensional outputs are computed at times $t > 0$ as
\begin{equation}
  \label{eq:esn-res}
\tilde{\vx}_{t+ 1} = D(\vr_{t}),
\end{equation}
where $D: \mathbb{R}^{K} \rightarrow \mathbb{R}^{L}$ is a (trained) decoding
function. For echo state networks, the decoder is often a linear transformation
$D(\vr_{t}) = \mW_{\text{out}} \vr_{t}$ where $\mW_{\text{out}}$ is a
matrix of dimensions $K \times L$. In our experiments in Chapter \ref{cha:learn-effic-compl}, 
we set $\beta = 0$, which
was empirically observed to yield the best results on our tasks. Parameter
$\beta$, as well as the randomly sampled weight matrix, are sometimes adjusted for
each task \parencite{jaegerLongShortTermMemory2012}. We only use default values
to obtain a task-independent setup with the least possible assumptions and to make
the methods comparable.

\subsection{Reservoir cellular automata\label{sec:app-ca-res}}

Cellular automata can also be used as a dynamical system in reservoir
computing. Yilmaz originally proposed this in 2014 and later was named ReCA
(Reservoir Cellular Automata) \autocite{yilmazReservoirComputingUsing2014,
  margemExperimentalStudyCellular2017}.

Several implementations of ReCA systems have been proposed and evaluated on
various tasks \autocite{yilmazReservoirComputingUsing2014,
  nicheleDeepLearningCellular2017, nicheleDeepReservoirComputing2017,
  nicheleReservoirComputingUsing2017, margemReservoirComputingBased2018,
  kleykoCellularAutomataCan2020, babsonReservoirComputingComplex2019,
  ,mcdonaldReservoirComputingExtreme2017a, moranReservoirComputingHardware2018}.
In this section, we lay out a general description of reservoir computing with
cellular automata based on these previous works and present our proposed
extensions. We apply reservoir cellular automata to language-like tasks in 
    Chapter \ref{cha:learn-effic-compl}.

\paragraph{Encoding\label{sec:encoding}.}
Input data is assumed to be categorical and sequential,

\begin{equation}
  X = [X_{1}, \ldots, X_{t}, \ldots],\quad t\in \mathbb{N}, \quad \forall t \ X_{t} \in \mathcal{X} \subset \mathbb{N}.
\end{equation}

A \ac{CA} is a discrete system not specifically designed to
handle input. The purpose of the encoding step in ReCA is to convert the input
data points to vectors that can be embedded in the state space of the cellular automaton. Such encoding should translate --- or embed --- the input space
$\mathcal{X}$ into a new one that can be combined with the current state of the
cellular automaton. The goal is to make a cellular automaton ``react'' to its
input and leverage the resulting computation to solve a problem.

\paragraph{Input projection.}
First, each categorical input vector is one-hot encoded into a vector of size
$L_{in}$, where $L_{in} = |\mathcal{X}|$ is the number of input categories. We have
\begin{equation}
\forall t,\quad \vx_{t} \in {\{0, 1\}}^{L_{in}},
\end{equation}

with
$\sum_{i = 1}^{L_{in}}{(\vx_{t})}_{i} = 1$.

Next, the vectors $\vx_{t}$ are projected onto the vectors
$\vp_{t} = P(\vx_{t}) \in \mathcal{P} = {\{0, 1\}}^{L_{d}}$ of fixed size $L_{d}$, where
$L_{d}$ is the size of the internal state of \ac{CA}. Usually, we have
$L_{d} > L_{in}$, which means that we embed the input in a space of higher
dimension. In our work, we compute this projection in one of three ways:

\begin{description}
  \item[One-to-one] each input bit is assigned to a single index in $\vp$.
        This is the projection function first proposed in
        \textcite{yilmazReservoirComputingUsing2014}.
\end{description}

\begin{figure}[htbp]
  \centering
  \includegraphics[width=.6\linewidth]{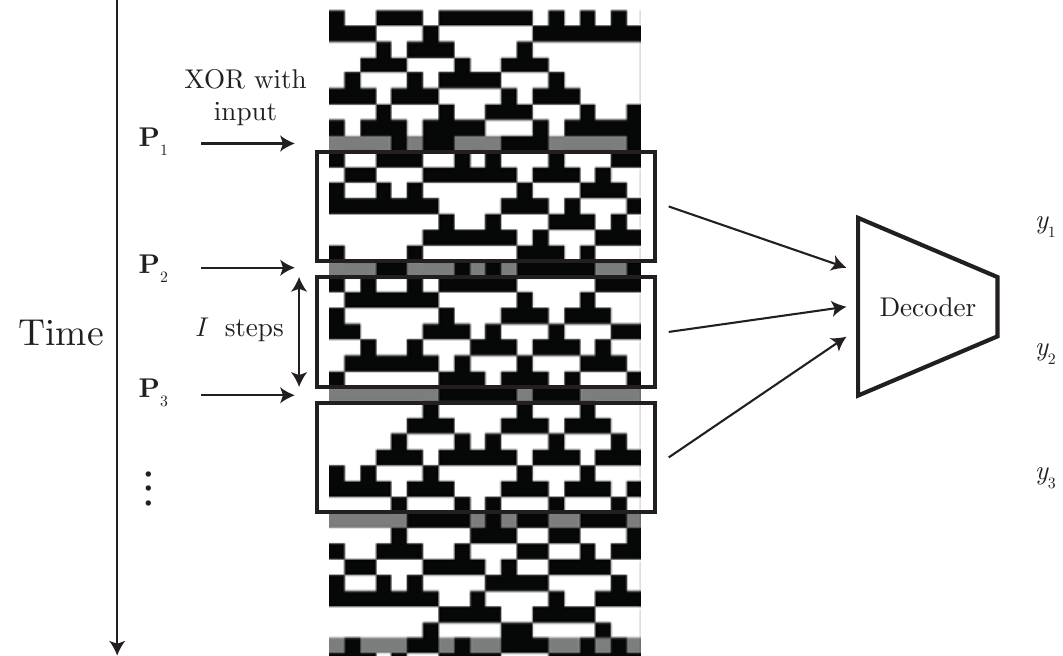}
  \caption{Illustration of the 1D ReCA model with the XOR-based input and
    state combination method. The highlighted boxes correspond to $I$ stacked \ac{CA} 
    states that make up a single reservoir state.}\label{fig:reca-schema}
\end{figure}

\paragraph{Building the reservoir}
The final state of the reservoir $\vr_{t + 1}$ is obtained by stacking $I$ consecutive
CA states obtained from a combination of inputs of a single state $\vs'$ by applying $\Phi$
again. The parameter $R$ can be understood as a space redundancy parameter,
while $I$ is a time redundancy parameter. If the size of the state is $n$, the
resulting reservoir vector $\vr$ has dimension $K = I \times n$, where $I$ is
defined above. Similar to the \ac{ESN}, the $L$-dimensional output tokens are
calculated at times $t > 0$ as

\begin{equation}
  \begin{aligned}
    \tilde{\vx}_{t+ 1} & = D(\vr_{t}),
  \end{aligned}
\end{equation}

where $D: \mathbb{R}^{K} \rightarrow \mathbb{R}^{L}$ is the (trained) decoding function. Here, we predict
$\tilde{\vx}_{t+1}$, which corresponds to a language-modeling-like task of
predicting the next input. It would also be possible to predict another sequence
$(\vy_{t})_{t}$. Usually a linear decoding function is used such that
$D(\vr_{t}) = \mW_{\text{out}}\vr_{t}$. The complete pipeline of a \ac{RC} model
with \ac{CA} is summarized in Figure \ref{fig:reca-schema}.

\chapter{Literature Review}
\label{cha:literature-review}

In the previous chapter, we gave a detailed definition of cellular 
automata, neural networks, and reservoir computing, and described the interplay 
between them. The current chapter aims to 
provide an overview of several fields associated with learning and complex
systems that have informed and influenced our work in this thesis. We begin by reviewing related
work on complexity measures for dynamical and complex systems (Section
\ref{sec:measuring-complexity}). We also review works on defining and
understanding emergence (Section \ref{sec:emergence}), evolutionary methods for
searching solutions in complex spaces (Section \ref{sec:evol-algor}), methods
for constructing open-ended evolving systems (Section
\ref{sec:open-ended-evolution-1}), and methods for extracting and using the
computations happening within complex systems (Section
\ref{sec:comp-with-compl}).

\section{Measuring complexity\label{sec:measuring-complexity}}

\begin{figure}[htbp]
  \centering
  \scalebox{1}{
\begin{tikzpicture}
\def\normaltwo{\x,{2*1/exp(((\x-3)^2)/2)}}

\def\ya{4.9}
\def\fya{2*1/exp(((\ya-3)^2)/2)}

\def\yb{1.4}
\def\fyb{2*1/exp(((\yb-3)^2)/2)}

\def\yc{3}
\def\fyc{2*1/exp(((\yc-3)^2)/2)}

\draw[color=blue,domain=0:6,smooth,samples=50, line width=0.25mm] plot (\normaltwo) node[right] {};

\draw[dashed] ({\ya},{\fya}) -- ({\ya},0) node[below] {\ref{fig:random}};
\draw[dashed] ({\yb},{\fyb}) -- ({\yb},0) node[below] {\ref{fig:simple}};
\draw[dashed] ({\yc},{\fyc}) -- ({\yc},0) node[below] {\ref{fig:complex}};

\draw (-.2,1.5) node[left] {Complexity};
\draw (3,-.5) node[below] {Randomness};

\draw[->,line width=0.1mm] (0,0) -- (6.2,0) node[right] {};
\draw[->,line width=0.1mm] (0,0) -- (0,2) node[above] {};

\end{tikzpicture}}
\caption{\emph{Ideal} ``complexity'' curve as a function of the ``randomness''
  of a system. The three \acp{CA} from figure \ref{fig:three_eca_complex} are
  displayed at their approximate expected location on that curve. This curve is
  just for illustrative purposes and does not correspond to any true function
  linking randomness and complexity.}
\label{fig:complexity_curve}
\end{figure}
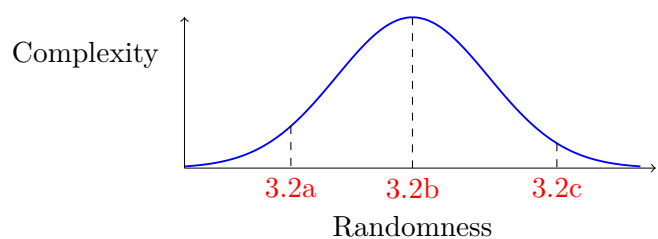
\begin{figure}[htbp]
  \centering
   \begin{subfigure}[b]{.03\linewidth}
    \centering
    \includegraphics[width=\linewidth]{figures/arrow.pdf}
    \caption*{}
  \end{subfigure}
  \begin{subfigure}[b]{.31\linewidth}
    \centering
    \includegraphics[width=\linewidth]{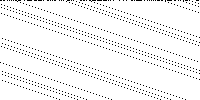}
    \caption{}
   \label{fig:simple}
  \end{subfigure}
  \begin{subfigure}[b]{.31\linewidth}
    \centering
    \includegraphics[width=\linewidth]{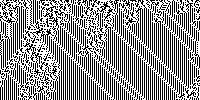}
    \caption{}
   \label{fig:complex}
  \end{subfigure}
  \begin{subfigure}[b]{.31\linewidth}
    \centering
    \includegraphics[width=\linewidth]{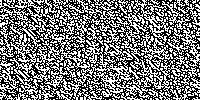}
    \caption{}
   \label{fig:random}
  \end{subfigure}
  \caption{Three 1D \aclp{CA} with qualitatively different behavior. The left
    (\ref{fig:simple}) and right (\ref{fig:random}) \acp{CA} are usually not
    defined as ``complex'', whereas the middle \ac{CA} (\ref{fig:complex})
    appears to be more complex than the others. The Figures show \acp{CA} space-time 
    diagrams. Each row represents the state of the \ac{CA} at a single time step, with 
    the time increasing from top to bottom.}
  \label{fig:three_eca_complex}
\end{figure}

Measuring the complexity of a system is a fundamentally difficult task. Many
complex systems exhibit what Peter Grassberger calls \emph{self-generated
  complexity} \parencite{grassbergerQuantitativeTheorySelfgenerated1986}. This
means that the formulation of the system is translationally invariant and the
observed structure arises from a spontaneous breakdown of translational
invariance. For example, \ac{CA} has a uniform update rule, but the complexity 
arises locally in some of them. Unfortunately, there is no universally accepted and formalized notion
of ``complexity'', although most intuitively agree that it exists. For
example, Figure \ref{fig:three_eca_complex} shows three examples of behaviors
generated by 1D \acp{ECA}. Most people would consider the leftmost figure
\ref{fig:simple} to be not complex. However, depending on one's definition of
complexity, the last picture could be labeled as complex.

No clear observable quantities and protocols have been proposed that would give a
quantitative notion of complexity.
\textcite{grassbergerProblemsQuantifyingSelfgenerated1989} argued that no single
quantity is sufficient to measure complexity since it depends on exactly how the
meaning is assigned to this term. Even when some meaning has been fixed by the
definition of an observable quantity, the statistics of the measurement of this
observable are also crucial to its
interpretation~\parencite{gutowitzCellularAutomataSciences1995}.

Despite these drawbacks, we seek a measure of complexity that matches intuition
as well as possible. Such a measure would assign low complexity to both simple
objects and random objects and high complexity to objects in between. For
illustration, we place the three \acp{ECA} from Figure
\ref{fig:three_eca_complex} on this ideal complexity curve, shown in Figure
\ref{fig:complexity_curve}. For example, in Chapter \ref{cha:meas-compl-evolv},
we develop a complexity metric that matches the human intuition of complexity on a
dataset of cellular automata.

One of the most widely known concepts of the complexity of symbol
sequences, ``algorithmic complexity,'' should also be called a measure of
information. This is the position held by one of its co-creators
\parencite{chaitinInformationRandomnessIncompleteness1990}. In the literature,
information measures are often used as complexity metrics. These quantities
are better measures of ``randomness'' than complexity, but they are nonetheless
important for the study of complex systems. In this section, we list several
complexity measures, including information measures that have influenced our
research.

\subsection{Information content}
For an event $E$ with probability $P$, the information content of this event is
defined as the negative logarithm of its probability, that is,

\begin{equation}
  I(E) :=  -\log(P).
\end{equation}

This metric quantifies how unlikely an event is and depends on how the
probability $P$ was estimated in practice. This notion is at the foundation of
information theory. The information content of a finite set of $N$ events is maximal 
when all events are equally probable, that is $I(E) = \log(N)$.

\subsection{Shannon Entropy}
Shannon entropy is defined as the expected information content of the input
\parencite{shannonMathematicalTheoryCommunication1975}. We have

\begin{align*}
  H(X) := \mathbb{E}[-\log(P(x))]_{x\in X}.
\end{align*}

This measure is a lower bound on the number of bits that the input could be
compressed to.

In the case of a 1D \ac{ECA}, the Shannon entropy could be computed cell-wise
over the time distribution of the states, producing an entropy per cell score
that can be averaged over an entire automaton. This measure is, for example, one
of the measures used by Wolfram in
\parencite{wolframStatisticalMechanicsCellular1983} and by Langton in
\parencite{langtonComputationEdgeChaos1990} to study how the parameter $\lambda$
affects the behavior of the automaton. There are several ways to compute it when
dealing with a CA, depending on which part of the CA is considered as the main
random variable. If the CA is finite, the state at timestep $t$ can be seen as a
random variable that can take one of $2^N$ possible values (with $N$ the width
of the automaton state). In that case, the probability of a state can easily be
estimated by counting its number of occurrences during evolution. For too large
state spaces, this becomes challenging, as it is very unlikely that a given state will
be seen again.

\subsection{Rényi Entropy}
The Rényi entropy is a generalization of Shannon entropy that gives different
weights to events of various probabilities
\parencite{renyiMeasuresEntropyInformation1961}. It is formally defined for an
$\alpha \geq 0, \alpha \neq 1$ and a random variable $X$ with possible outcomes $0, 1, ..., n$ as
\begin{align*}
  H_\alpha(X) = \frac{1}{1-\alpha} \log\left(\sum_{i=0}^np_i^\alpha\right).
\end{align*}

In the limit $\alpha \rightarrow 1$ it is equal to the Shannon entropy, where the probabilities of each event are equally important. $\alpha \rightarrow \infty$ yields the Min-entropy, and we have
$H_{\infty} (X) = \min_{i}(\log(p_{i}))$. With $\alpha \rightarrow 0$ Rényi entropy is the same as
Max-entropy, or topological entropy. Rényi entropy was used by
\textcite{wolframStatisticalMechanicsCellular1983} and
\textcite{lindgrenComplexityMeasuresCellular1988} to estimate the complexity in
infinite \acp{CA}.

\subsection{Mutual information}
The mutual information is a measure of the dependence between two variables. Two
independent variables will have mutual information of zero. If the two are
strongly dependent, for example, if one is a function of another, the mutual
information between them will be larger.

\textcite{chaitinMathematicalDefinitionLife1987} proposes to split a system into
fixed blocks and calculate mutual information among its components.
He uses the maximum value of the mutual information in all possible partitions
to define "life" in a mathematical way. \textcite{shawDrippingFaucetModel1984}
and \textcite{grassbergerQuantitativeTheorySelfgenerated1986} also use the
mutual information measure between two semi-infinite blocks in a sequence to
define complexity. Mutual information has also been used to study the
complexity of \acp{CA} as the parameter $\lambda$ is changed (for details
on $\lambda$ in the context of \ac{CA}, see section \ref{sec:langtons-lambda})
\parencite{gutowitzMethodsDesigningCellular1988,
  liTransitionPhenomenaCellular1990}.

\subsection{Computational complexity}
Algorithms can be described by their space and time complexity, which
respectively correspond to the amount of memory storage and CPU time necessary
to run them \parencite{traubInformationUncertaintyComplexity1983,
  packelRecentDevelopmentsInformationbased1987,
  hopcroftIntroductionAutomataTheory2007}. A problem that depend on a parameter
$N$ will be said to be NP-hard if the time needed to find a solution to them
increases exponentially with the parameter $N$. NP-hard problems can be said
to be complex as they require a supposedly irreducible amount of computation to
compute their output. In practice, this description only applies to computations
that were generated by a hand-designed algorithm and not the kind of
computations self-generated by a complex system that we are interested in. For
example, the computations generated by one \ac{CA} cannot be said to be
computationally more complex than those of another \ac{CA} because the algorithm
that implements them is fundamentally the same and computationally simple.

\subsection{Solomonoff–Kolmogorov-Chaitin complexity (algorithmic
  complexity)}\label{sec:algo-complexity}

Introduced by Solomonoff \parencite{solomonoffPreliminaryReportGeneral1960},
Kolmogorov \parencite{kolmogorovThreeApproachesQuantitative1968} and Chaitin
\parencite{chaitinLengthProgramsComputing1969,
  chaitinAlgorithmicInformationTheory1977,
  chaitinInformationRandomnessIncompleteness1990}, this complexity measure is
defined for a string $s$ of characters and a universal description language (\eg
a programming language) as the length of the shortest program that can generate
the string $s$.

This number $K(s)$ is called the minimum description length of
$s$. We note that Chaitin prefers to call his field "algorithmic information
theory." His position is that "algorithmic complexity" is a measure of
randomness rather than a measure of complexity
\parencite{chaitinInformationRandomnessIncompleteness1990}.

From the invariance theorem \parencite{solomonoffFormalTheoryInductive1964,
  solomonoffFormalTheoryInductive1964a,
  kolmogorovThreeApproachesQuantitative1968,
  chaitinLengthProgramsComputing1969}, the difference in algorithmic complexity
of the same string $s$ in two different description languages is bounded,
although this bound might be very large in practice.

The algorithmic complexity is uncomputable
\parencite{solomonoffFormalTheoryInductive1964,
  kolmogorovThreeApproachesQuantitative1968}, and there exists strings of
arbitrarily large complexity, which makes it hopeless to use this exact measure
in practice. It can be approximated from above, but no accuracy guarantee can be
given, and the runtime of the approximators can grow arbitrarily large.
Moreover, algorithmic complexity tells us how much information is required to
encode a number, but does not tell us how difficult it is to recreate the number
from that code \parencite{gell-mannSimplicityComplexityDescription1988}.

A range of resource-bounded approximations of algorithmic complexity were
developed to obtain computable complexity measures
\parencite{daleyMinimalprogramComplexitySequences1973,
  daleyInferenceOptimalDescriptions1977,
  ,federUniversalPredictionIndividual1992, koNotionInfinitePseudorandom1986,
  schmidhuberSpeedPriorNew2002}. One rather straightforward way of approaching
the algorithmic complexity of an arbitrary string is to use a compression
algorithm and use the length of the decompression program plus the length of
the compressed string as an upper bound to the algorithmic complexity. This is
similar to the optimal symbol compression ratio (OSCR) algorithm proposed by
\parencite{evansNewUniversalTwo2003}.
\textcite{zenilCompressionBasedInvestigationDynamical2010} used the
compression-based method to classify 1D \ac{ECA}. This method, which often
uses the popular LZ algorithm, can be seen as an independent complexity
measure, also closely related to Lempel-Ziv complexity, described in more
detail in the next section ~\ref{subsection:lempel-ziv}.

However, it is worth noting that when a complex system is described by an
algorithm (such as for a \ac{CA}), the algorithmic complexity is also easily
upper bounded by a constant value entirely defined by the algorithm that
simulated the complex system. For example, for a \ac{CA}, the algorithmic
complexity is lower than the size of the rule table of the automaton, its
characteristics (size, boundary conditions, \etc), its initial state, and the number
of simulation steps. This approximation does not inform us about the difference in
complexity between the two \acp{CA}.

\subsection{Lempel-Ziv complexity}\label{subsection:lempel-ziv}
The Lempel-Ziv complexity, as defined in
\parencite{lempelComplexityFiniteSequences1976} is the number of steps in the LZ
algorithm, directly related to the number of repeated substrings in the
input string. The main idea of this algorithm is to scan the input string while
trying to find some repetition of the previous input in the incoming data. This
builds over time a set of basic components called the exhaustive history of the
string, from which the complete string can be constructed. The number of
components in that set is the Lempel-Ziv complexity of that string.

The compressed length method for measuring complexity uses a compression
algorithm to reduce the size of the input. An input with regularities and
repetitive patterns will produce a small output, whereas a completely random
input will be irreducible to a simpler, shorter string.

\subsection{Logical Depth}

Bennett's logical depth is a measure of complexity based on the algorithmic
complexity \parencite{bennettDissipationInformationComputational1988,
  bennettLogicalDepthPhysical1995}. The main difference is that it takes into
account the computation time (or number of steps) along with the length of the
program used to generate the sequence. It is a combination of algorithmic
complexity and computational complexity. For a universal computer $U$, the
logical depth of a string $x$ at a significance level $s$ is defined as

\begin{equation}
  \label{eq:2}
  \min\{T(p): (|p| - |p^{*}| < s ) \wedge (U(p) = x) \},
\end{equation}
where $T$ is the number of steps that the universal computer runs the program
$p$ for. This quantity corresponds to the least time required to compute the
string $x$ by a $s$-incompressible program, that is a program $p$ whose length
is within $s$ symbols from the optimal program $p^{*}$ of the algorithmic
complexity metric.

\textcite{antunesComputationalDepthConcept2006} considered logical depth to be
one instance of a more general concept: \emph{computational depth}; and proposed
several other variants.

\subsection{Thermodynamic Depth}

The thermodynamic depth, developed by
\textcite{lloydComplexityThermodynamicDepth1988} is another measure of
complexity based on the intuitive notion that complex systems lie somewhere in
the continuum between order and chaos
\parencite{chaitinInformationRandomnessIncompleteness1990,
  ceccattoComplexityHierarchicalSystems1988, deutschQuantumTheoryChurch1985}.
Similar to computational complexity and logical depth, this metric is designed
to be a measure of how the system came to be in its final state. The complexity
corresponds to how difficult it is to put the system together in that state.

By the definition of thermodynamic depth, the average complexity of a state must be
proportional to the Shannon entropy
\parencite{shannonMathematicalTheoryCommunication1975} of the set of
trajectories that the experiment determines can lead to that state. If we write $p_i$ 
the probability of reaching the target state with the $i$-th possible trajectory, we have 

\begin{equation}
  \label{eq:3}
  S = -\left(\sum_{i} p_{i} \log p_{i}\right).
\end{equation}

The measure of complexity of a macroscopic state $s$ of a system that has
reached that state by the i-th possible trajectory is $-k(\log p_{i})$, where
$p_{i}$ is the probability that the system has reached $s$ by the i-th
trajectory, and $k$ is an arbitrary positive constant. The thermodynamic depth of
the state $s$ is defined as

\begin{equation}
  \label{eq:4}
  \mathcal{D}(s) = -k(\log p_{i}).
\end{equation}

The quantity $\mathcal{D}$ can be thought of as the amount of information
required to specify the trajectory that the system has followed to its present
state. The thermodynamic depth of the whole system is then

\begin{equation}
  \label{eq:5}
  \mathcal{D} = \sum_{s}\mathbb{P}(s)D(s),
\end{equation}
where $\mathbb{P}(s)$ is the probability that the system followed the
trajectory $s$.

\textcite{crutchfieldThermodynamicDepthCausal1999} argued that the thermodynamic
depth is a fundamentally flawed structural complexity measure because it relies
on a set of chosen macroscopic states for the system, which are difficult to
choose and need to be defined separately for each system.
\textcite{lloydComplexityThermodynamicDepth1988} does not mention how these states
are supposed to be chosen, nor do any follow-up work, making this complexity
metric difficult to use in practice.

\subsection{Epsilon-machines}

The $\epsilon$-machine is an approach to complexity that seeks to construct a metric
more suitable for physical systems that also addresses some issues of other
existing complexity metrics \parencite{crutchfieldOrderChaos2012}. It is the
result of the field of computational mechanics, an extension of statistical
mechanics that describes not only the statistical properties of a system but
also how it stores information and how it computes
\parencite{crutchfieldInferringStatisticalComplexity1989,
  crutchfieldCalculiEmergenceComputation1994,
  feldmanMeasuresStatisticalComplexity1998, crutchfieldOrderChaos2012}.
Computational mechanics algorithms take as input the time series being analyzed
and output a minimal, optimized model that can reproduce a time series that is
statistically equivalent to the input time series. The size of the model
produces a metric known as statistical complexity. The models are known as
$\epsilon$-machines. Three optimality theorems say that $\epsilon$-machines capture all the
properties of a process
\parencite{crutchfieldInferringStatisticalComplexity1989,
  crutchfieldThermodynamicDepthCausal1999,
  shaliziComputationalMechanicsPattern2001}: prediction: the $\epsilon$-machine is its
optimal predictor; minimality: compared to all other optimal predictors, the
$\epsilon$-machine of a process is its minimal representation; uniqueness: any minimal
optimal predictor is equivalent to the $\epsilon$-machine.

This model has many attractive properties and was successfully applied to study
the symbolic dynamics of chaotic systems
\parencite{crutchfieldCalculiEmergenceComputation1994}, molecular dynamics
\parencite{ryabovComputationalMechanicsMolecular2011}, single-molecule
microscopy \parencite{kellyNewMethodInferring2012}, and the spatiotemporal
complexity of \acp{CA} \parencite{crutchfieldTurbulentPatternBases1993,
  hansonComputationalMechanicsCellular1997,
  shaliziQuantifyingSelfOrganizationOptimal2004}. The main drawback is the
construction of the $\epsilon$-machine itself, for which there is no
general-purpose approach.

\subsection{Sophistication}

Intuitively, \emph{sophistication} is the complexity in a set of strings of which
the string is a ``typical'' member.
\textcite{motaSophisticationRandomnessDeficiency2013} defines sophistication
based on the original definition of \textcite{koppelStructure1988,
  koppelAlmostMachineindependentTheory1991a}. It measures the amount of structural
information contained in a string. It also uses a follow-up result by
\parencite{vitanyiMeaningfulInformation2006} that shows that Koppel's definition
is equivalent to measuring the complexity of a good model for a string up to
low order terms. The definition uses an intermediate quantity called
\emph{discrepancy} that measures how far a set $S$ is from being a good model
of a string $x$.

Formally, the discrepancy $\Delta$ is defined for a string $x$ and a set $S$ that contains $x$ as
\begin{equation}
  \label{eq:6}
  \Delta(x|S) := \log |S| - K(x) + K(S),
\end{equation}
where $K$ is the Solomonoff–Kolmogorov-Chaitin complexity function.

The sophistication of $x$ is defined as the complexity of the simplest model of
$x$ with a limited discrepancy:

\begin{equation}
  \label{eq:7}
  \text{soph}_{c}(x) := \min_{S}\left\{ K(S): \Delta(x|S) \leq c \right\}
\end{equation}
where the significance level $c$ tells us how much discrepancy the set $S$ is allowed to have.

Koppel’s definition of sophistication, Definition 3.1, may not be stable. Small
changes in c could cause large changes in $\text{soph}_{c}(x)$. For this reason,
\textcite{antunesSophisticationRevisited2009} introduces a new notion of \emph{coarse
sophistication} that incorporates the "constant" $c$ as a penalty in the formula
to obtain a more robust measure.
In the formalism of \textcite{motaSophisticationRandomnessDeficiency2013} the
definition is

\begin{equation}
  \label{eq:8}
  \text{csoph}(x) = \min_{S}\left\{ K(S) + \Delta(x|S) \right\}
\end{equation}

which is also equivalent to
\begin{equation}
  \label{eq:8b}
  \text{csoph}(x) = \min_{c}\left\{ \text{soph}_{c}(x) + c \right\}.
\end{equation}

Like algorithmic complexity, sophistication is a useful theoretical notion to
model ideas of entropy and complexity, but it cannot be directly applied in
numerical simulations and can only be approximated because it is also
uncomputable.



\section{Emergence}\label{sec:emergence}
The notion of emergence is central to the study of complex systems. It is a 
broad concept that could be phrased as \emph{the properties a composite entity
  acquires that its composing parts did not possess}. Emergence is often
described as being similar to self-organization. Although these two concepts are
closely related, they are fundamentally different ideas
\parencite{dewolfEmergenceSelfOrganisationDifferent2005}. Both properties can
exist independently within a dynamical system. Some examples of emergent
behavior include:

\begin{itemize}
  \item The spontaneous formation of clusters of randomly distributed objects -
        a behavior common in ant colonies forming bridges out of individual
        ants or birds flocking into ``murmurations'' - that naturally emerges
        out of a simple set of autonomous actions having nothing to do with that
        clustering. This was demonstrated by
        \textcite{beckersFomLocalActions2000} in the context of exploring
        collective robotics). The macroscopic behavior in each of these examples
        is unexpected even though the details of the microscopic dynamics are
        well-defined.
  \item The spirals of the Belousov- Zhabotinsky chemical reaction
        \parencite{tysonBelousovZhabotinskiiReaction2013}.
  \item The Navier-Stokes-like macroscopic behavior of a lattice gas that
        consists, of simple unit-bit billiards moving back and forth between
        discrete nodes along discrete links at the microscopic scale
        \parencite{hasslacherDiscreteFluids1987}.
\end{itemize}

There seems to be qualitatively different types of emergence, which
\textcite{dehaanHowEmergenceArises2006} describes as Discovery, Mechanistic
emergence, and Reflective emergence. These correspond to different levels, or
``strengths'', of emergence, from fractal patterns to complex social systems and
ecosystems.

A subset of emergence research has focused on complex adaptive systems. In
such systems, emergence is explicitly used to refer to macro-level patterns
arising from interacting agents \parencite{hollandEmergenceChaosOrder2000,
  kauffmanHomeUniverseSearch1995, langtonStudyingArtificialLife1986}.

\section{Evolutionary algorithms}\label{sec:evol-algor}
\begin{figure}[htbp]
  \centering
  \includegraphics[width=.8\linewidth]{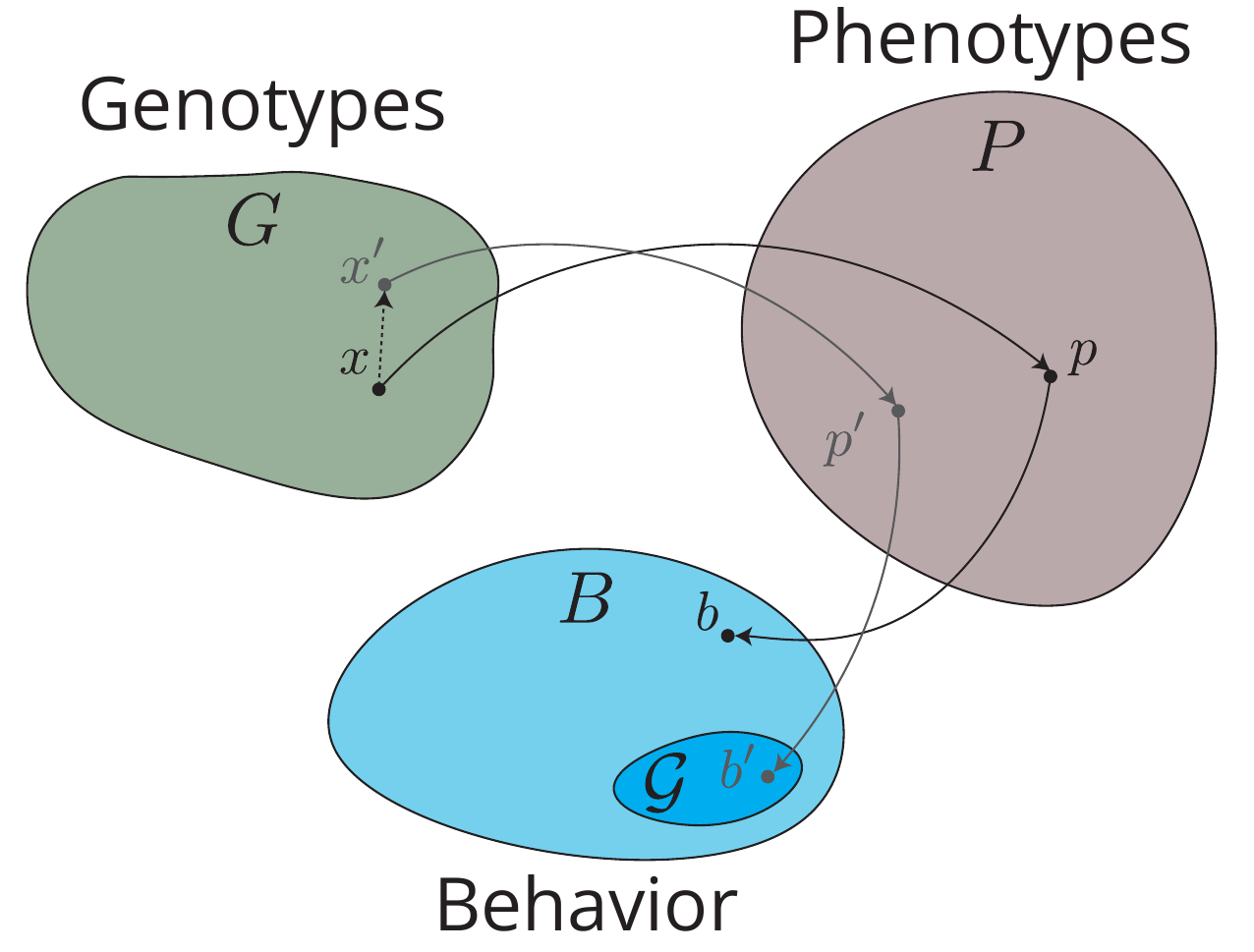}
  \caption{Illustration of the general principle common to several evolutionary
    algorithms. There are three virtual spaces: ($G$) the genotype space where
    the candidate solutions are generated through some encoding. This is the
    space where the search process is happening. ($P$) the phenotype space,
    which corresponds to the decoded solution from a genotype $x$. It can be the
    policy of an agent, the rule of a \ac{CA}, etc. ($B$), the behavior space,
    which is the space from which candidate solutions are evaluated. The smaller
    subspace $\mathcal{G}$ is the goal space, corresponding to the set of
    behaviors that would be considered successful. A small perturbation in the
    genotype will have potentially large consequences on the behavior of the
    candidate solution. In the case of a \ac{CA} rule, the genotype could be a 
    vector ($x$) that encodes a rule ($p$) that will translate into the \ac{CA} behavior $b$.}
  \label{fig:evolutionary_diagram}
\end{figure}

Evolutionary algorithms are the class of algorithms inspired by natural
biological processes, in particular evolution. There are many types of
evolutionary algorithms, from genetic algorithms to quality diversity and
novelty search \parencite{lehmanAbandoningObjectivesEvolution2011,
  lehmanEvolvingDiversityVirtual2011}. Evolutionary algorithms are used as
search methods, in particular when the search space is hard to parametrize or
very large \parencite{poliRelationsSearchEvolutionary1996}. Evolutionary
algorithms facilitate the search for complex optimization spaces by using intermediate
representations, as illustrated in Figure \ref{fig:evolutionary_diagram}.
\ac{CA} rules are a good example of such search spaces. The \ac{CA} rule space is
very large, and it is difficult to predict how perturbations of the rule
translate to changes in the behavior of a \ac{CA}. Genetic algorithms were
successfully used to evolve rules to perform complex computations
\parencite{mitchellEvolvingCellularAutomata1996}.

\subsection{Genetic pr\label{sec:genetic-programming}ogramming}
Many of the problems that machine learning and artificial intelligence are
attempting to solve require the discovery of a computer program that produces
some desired mapping between inputs and outputs. Solving the problems therefore
amounts to searching the space of computer programs in order to find a suitable
individual with high enough fitness
\parencite{langdonFoundationsGeneticProgramming2002,
  kozaGeneticProgrammingMeans1994, banzhafGeneticProgrammingIntroduction1998,
  ,bookerClassifierSystemsGenetic1989}.

In genetic programming, candidate solutions to a problem are encoded as
\emph{chromosomes}, which is a structured data representation that can be
modified incrementally. In practice, it is often chosen to be a string of bits.
Every generation, a population of solutions is evaluated. The best solutions are
kept and combined to form the candidates for the next generation. Random
mutations may also be applied to introduce randomness in the search.

\subsection{Novelty search}
\begin{figure}[htbp]
  \centering
  \includegraphics[width=.7\linewidth]{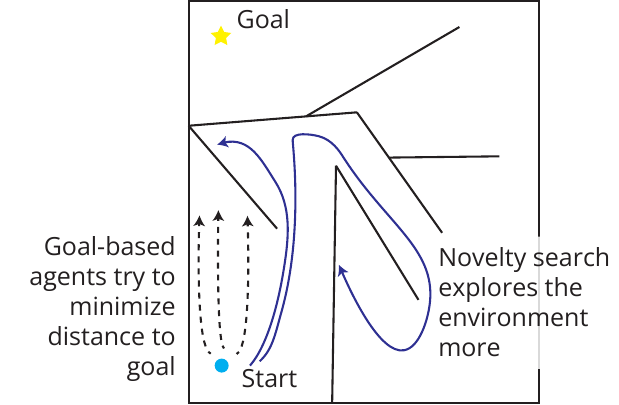}
  \caption{Illustration of the advantage of \acf{NS} on a maze environment with
    deceptive dead-ends. Dotted trajectories illustrate the behavior of a
    goal-based search, which will start by minimizing the distance to the goal.
    Blue trajectories are from a novelty search algorithm exploring the space
    more efficiently. This type of maze environment was proposed by
    \textcite{lehmanAbandoningObjectivesEvolution2011}.}
\label{fig:novelty_search}
\end{figure}

The idea behind \ac{NS} is to drive a search algorithm only by the novelty of
produced behavior \parencite{lehmanAbandoningObjectivesEvolution2011}. The most
counter-intuitive feature of this algorithm is that the actual objective of the
task is not taken into account at all during the search process. Generated
agents are evaluated solely on the novelty of their behavior. Despite this, it
appears to be at least as efficient as other search processes that are focused on goals, such as maze navigation and biped locomotion
\parencite{lehmanAbandoningObjectivesEvolution2011}, swarm robotics
\parencite{gomesEvolutionSwarmRobotics2013}, or neural network design
\parencite{risiEvolvingPlasticNeural2010}. \ac{NS} can lead to the discovery 
of innovative and creative solutions to complex and deceptive objectives that 
would be difficult or impossible to achieve using traditional optimization methods. 
We illustrate this property in Figure~\ref{fig:novelty_search}.

The principle of guiding search by novelty alone is closely related to the goal
of this thesis, and especially with the challenge of parametrizing and searching
the space of available complex systems without using any explicit goal function.
The purpose of \ac{NS} is to perform a search without a goal function, and we draw
inspiration from this algorithm throughout our work.

This method has some limits, which are caused by a relative lack of theoretical
understanding of the effects of the parameters of \ac{NS}. The choice of
behavior descriptors for determining the novelty of agents is also not
straightforward. Several empirical studies have investigated the impact of a
range of parameters on the performance of \ac{NS}
\parencite{gomesDevisingEffectiveNovelty2015,
  kistemakerCriticalFactorsPerformance2011}, while other works have begun to
shed light on the theory \parencite{doncieuxNoveltySearchTheoretical2019}. In
practice \ac{NS} algorithms are expected to cover the exploration space well
\parencite{cullyQualityDiversityOptimization2017,pughQualityDiversityNew2016},
in some cases uniformly \parencite{gomesDevisingEffectiveNovelty2015}. \ac{NS}
algorithms are related to genetic algorithms since they maintain a population,
that is, a set of individuals used to measure the novelty of current solutions.
There is also an archive of the behavior of previous individuals to ensure that the
novelty is measured against all previously generated behaviors. There exists
multiple strategies for maintaining this archive
\parencite{gomesDevisingEffectiveNovelty2015}. Some use individuals whose
novelty was above a threshold when first evaluated
\parencite{lehmanAbandoningObjectivesEvolution2011}, the most novel individuals
of each generation \parencite{liapisConstrainedNoveltySearch2015}, random
individuals from the past generations
\parencite{lehmanEfficientlyEvolvingPrograms2010}, or none at all
\parencite{mouretEncouragingBehavioralDiversity2012}. An alternative to
explicitly measuring novelty in \ac{NS} is \emph{fitness sharing}. The goal of
fitness sharing is to encourage the diversity of a population by making them
share virtual ``resources'' \parencite{goldbergSimpleGeneticAlgorithms1987,
  hollandAdaptationNaturalArtificial1992}. A particular implementation of
\ac{NS} using this idea, with rewards physically spread across the
environment, was done by \textcite{herelEmergenceNoveltyEvolutionary2022}.

Typically, novelty search algorithms use one of a few types of maps from
genotypes to phenotypes, that is, a map from the space of encoded solution to
agent and policy implementation
\parencite{mouretEncouragingBehavioralDiversity2012}. Some common ones are:

\begin{itemize}
  \item A \ac{RNN} (of type \cite{elmanFindingStructureTime1990} or
        \cite{jordanSerialOrderParallel1997}), for which the weights are
        directly searched. For input size $n_{i}$ output size $n_{o}$, and
        hidden size $n_{h}$, $n_{i}n_{h} + n_{h}n_{o} + n_{h}^{2}$ weights must
        be evolved.
  \item Neuro-evolution of augmenting topology (NEAT,
        \cite{stanleyEvolvingNeuralNetworks2002}), which is an encoding of both
        neural network weights and architectures that has a built-in mechanism
        to sustain diversity.
\end{itemize}

\section{Open-ended evolution\label{sec:open-ended-evolution-1}}
The field of Artificial Life research has worked to figure out what the
fundamental conditions for the emergence of living systems are, and how to
create a process that can display analogous levels of creativity and complexity
as natural evolution \parencite{eigenHypercycle1979,
  langtonArtificialLifeProceedings1989, dysonOriginsLife1999,
  stanleyWhyOpenEndednessMatters2019, packardOverviewOpenEndedEvolution2019,
  sorosOpenendednessLastGrand2017}. It builds upon the data and understanding we
collected about the process of life, but abstracts from any specific living
process and attempts to integrate various approaches into one unified research
to extract the first principles of life. A major assumption underpinning this
research is that this natural evolutionary process can be implemented equally
well in different media \parencite{dennettDarwinDangerousIdea1996}.

A system that behaves like natural evolution, producing a seemingly endless
amount of novelty and complexity starting from elementary building blocks is
called \emph{open-ended}. The main challenge of \ac{OEE} research is that there is
no single simple test for the phenomenon, but instead different kinds of
open-ended evolution exist. Systems can exhibit more than one kind at a time. In
the report from a workshop on \ac{OEE} at York
\parencite{taylorOpenEndedEvolutionPerspectives2016}, the authors summarized the
different types of \ac{OEE}, which were further refined in a follow-up work
\parencite{packardOverviewOpenEndedEvolution2019}. They are:

\begin{enumerate}
  \item Interesting new kinds of entities and interactions
  \item Evolution of evolvability
  \item Major transitions
  \item Semantic evolution.
\end{enumerate}

The first category describes the ability of a system to construct new entities
with different properties, behaviors, or interactions with other entities. For
example, the Tierra simulation sees entities emerge that exploit the computing
power of others, acting as parasites \parencite{rayApproachSynthesisLife1991}.
The second type is related to how open-ended evolution itself can be evolved
through the emergence of multiple ``stepping stones'' that allow individuals to use higher-level evolutionary units or through interactions
\parencite{patteeEvolvedOpenEndednessNot2019}. ``Major transitions'' refer to
the emergence of new levels of hierarchy in evolving populations. These
transitions are characterized by groups of reproducing entities that
interact tightly and form a new population of higher-level reproducing entities.
The process can repeat, with certain groups in the new population forming even
higher-level entities. This process of emergence of major transitions was
explored in \parencite{sayamaCardinalityLeapOpenEnded2019,
  morenoOpenEndedFraternalTransitions2019} for example. Lastly, semantic
evolution is related to the work of
\textcite{ikegamiOpenEndedEvolutionMechanism2019}, who propose a new category of
open-ended evolution called semantic evolution, which refers to the evolution of
semantic relationships within a system. This can be observed in the evolution of
web services, where the meaning of tags changes as new combinations of tags are
created. It is not biological in nature and is also present in the analysis of
technological evolution \parencite{bedauOpenEndedTechnologicalInnovation2019},
where the meaning and importance of keywords shift over time.

\subsection{Defining open-endedness}

Several necessary conditions and requirements have been identified in the
\ac{OEE} literature, forming an overlapping set of potential research
directions for developing open-ended systems. We list a few of these conditions
here. First, \textcite{maleyFourStepsOpenended1999} identifies four requirements:

\begin{enumerate}
  \item An open-ended evolutionary system must demonstrate unbounded diversity
        during its growth phase.
  \item An open-ended evolutionary system must embody selection.
  \item An open-ended evolutionary system must exhibit continuing new adaptive
        activity.
  \item An open-ended evolutionary system must have an endogenous implementation
        of niches.
\end{enumerate}
Next, \textcite{sorosIdentifyingNecessaryConditions2014} identified four more necessary
conditions, which are

\begin{enumerate}
  \item A rule should be enforced that individuals must meet some minimal
        criterion (MC) before they can reproduce, and that criterion must be
        nontrivial.
  \item The evolution of new individuals should create novel opportunities for
        satisfying the MC.
  \item Decisions about how and where individuals interact with the world should
        be made by the individuals themselves.
  \item The potential size and complexity of the individuals' phenotypes should
        be (in principle) unbounded.
\end{enumerate}
Then, \textcite{taylorRequirementsOpenEndedEvolution2015} also stated five
requirements for an open-ended system:

\begin{enumerate}
  \item Robustly reproductive individuals.
  \item A medium allowing the possible existence of a practically unlimited
        diversity of individuals and interactions, at various levels of
        complexity.
  \item Individuals capable of producing more complex offspring.
  \item An evolutionary search space that typically offers mutational pathways
        from one viable individual to another viable (and potentially fitter )
        individuals.
  \item Drive for continued evolution.
\end{enumerate}
Taylor also states a more general condition that should be sufficient for
creating an open-ended system as ``evolutionary dynamics in which new,
surprising, and sometimes more complex organisms continue to appear''
\parencite{taylorRequirementsOpenEndedEvolution2015,
  taylorOpenEndedEvolutionPerspectives2016}.

All these conditions illustrate a major challenge of \ac{OEE} research: the lack
of a clear definition or notion of what exact conditions make a system
open-ended. We seem to agree about what is not open-ended, but whenever a
constraint or requirement for \ac{OEE} is identified, subsequent evidence forces
us to refine them later. This is related to the problem of complexity, for which
no single definition exists \parencite{johnsonSimplyComplexityClear2009}.

A common process to produce systems that behave analogously to natural evolution
is to start from evolvable units or building blocks
\parencite{rayApproachSynthesisLife1991, simsEvolvingVirtualCreatures1994,
  ofriaAvidaSoftwarePlatform2004, yaegerComputationalGeneticsPhysiology1994,
  channonImprovingStillPassing2003, spectorDivisionBlocksOpenended2007,
  sorosIdentifyingNecessaryConditions2014}. The reason is that starting from
higher-level primitive units whose emergence would be hard to characterize may
be easier and faster than starting from lower-level components. The results obtained
from these systems are often surprising, as they bear some key similarities
to natural evolutionary processes. For example, we note the emergence of
parasitic entities within the Tierra simulation
\parencite{rayApproachSynthesisLife1991}. The process of emergence of these
building blocks from simple rules and components has also been investigated
significantly \parencite{bagleySpontaneousEmergenceMetabolism1991,
  huttonEvolvableSelfReproducingCells2007, flammEvolutionMetabolicNetworks2010,
  sayamaSeekingOpenendedEvolution2011}. It appears harder to bridge the gap and
create high-level evolutionary-like processes and behaviors from elementary
rules and substrates.

\subsection{Open-endedness in cellular automata}
One class of systems that has rich interactions between each of its components, as
well as no predefined evolvable units or assumptions about individuality is the
\ac{CA}. One of the very first automata, Von Neumann's self-reproducing machine,
was designed with goals that align with open-ended evolution, which is to build
a machine with no central controller and limited local interaction that can
self-reproduce as a whole
\parencite{vonneumannTheorySelfreproducingAutomata1966,
  pesaventoImplementationNeumannSelfReproducing1995}. Later, other
self-replicating structures such as Langton's loop
\parencite{langtonSelfreproductionCellularAutomata1984} and evoloops
\parencite{sayamaNewStructurallyDissolvable1999,
  salzbergComplexGeneticEvolution2004} showed that more properties of open-ended
systems can be included in a \ac{CA}. A potential limitation of \acp{CA} is the
absence of the notion of conservation of ``matter''. For example, the game of life
can start from a configuration with very few live cells and create many more at
no cost during its evolution. Some authors believe that this conservation
property is essential to the construction of an open-ended evolving system
\parencite{taylorChapterCreativityEvolution2002}.

\subsection{Artificial chemistries}
Artificial chemistries, are computational models that aim to simulate the
behavior of chemical systems, which are typically used to study the emergence of
complex behavior from simple interactions between individual molecules. Most of
the time, \acp{AC} do not try to accurately model chemical processes (as in
\parencite{ostrovskyCellularAutomataPolymer2001,
  buligaChemlambdaUniversalitySelfmultiplication2014,
  bedauLessAbstractArtificial2000, sayamaSeekingOpenendedEvolution2011} for
example). Instead, they build models of the dynamics of complex molecular
processes that lead to evolutionary behavior
\parencite{dittrichArtificialChemistriesReview2001}. By abstracting away from natural molecular processes, \acp{AC} tries to uncover fundamental
conditions for the emergence of organization, self-maintenance, or
self-construction with basic building blocks. There are various approaches for
building these \aclp{AC}, of which we review a few here.

\paragraph{Rewriting systems.}
Rewriting systems are a class of computational models that use a set of
rewriting rules to transform and manipulate strings of symbols. They are
composed of entities or symbols that get modified according to their set of
syntactic rules. Patterns of symbols or entities are replaced according to these
rules. In AlChemy \parencite{fontanaWhatWouldBe1994}, molecules are represented
as expressions in the $\lambda$-calculus. The $\lambda$-calculus is a mathematical formalism
that, like Turing machines, is capable of describing any computable function. In
AlChemy, pairs of randomly selected expressions are joined using function
application, evaluated, and the resulting value is added back to the population
of molecules. This process is repeated to simulate the behavior of a chemical
system. \textcite{kruszewskiEmergenceSelfReproducingMetabolisms2022}
hypothesized that self-reproducing metabolisms emerge as a subset of
autocatalyzed reactions within a Turing-complete set. They introduced
\emph{combinatory chemistry}, an artificial chemistry designed to capture the
core computational properties of living systems that is based on the rewriting
rules of \emph{combinatory logic} \parencite{curryCombinatoryLogic1958,
  schoenfinkelUeberBausteineMathematischen1924}. The generated structures
exhibit behaviors that are similar to natural metabolisms. The system is able to
represent patterns of arbitrary complexity, and it is able to produce diverse
self-reproducing patterns.

\paragraph{Cellular automata.}
\Acfp{CA} can be seen as a particular case of lattice molecular systems. For
example, the autopoietic system (referring to the ability of an organism to continuously 
sustain itself through the exchange of matter and energy with its environment, 
while maintaining its own boundaries and identity) introduced by
\textcite{varelaAutopoiesisOrganizationLiving1991} is a 2D square grid with
sites that be occupied by \emph{atoms} which is similar to a \ac{CA} with 4
states. Each of these states is analogous to a chemical component of the system:
($\emptyset$) empty site, (S) substrate, (K) catalyst, and (L) monomer. Basic rules are
applied asynchronously and define how neighboring atoms interact with each
other. Remarkably, stable self-repairing cells spontaneously arise from these
basic rules. Their membrane is composed of a chain of monomers, which is
maintained by the substrate and catalyst reacting according to the rules. Some
key components of that model were investigated in other works, showing that they
are crucial for this emergence autopoeisis to be possible
\parencite{zelenySelforganizationLivingSystems1977,
  mcmullinRediscoveringComputationalAutopoiesis1997}.

\section{Computing with complex systems}\label{sec:comp-with-compl}

The problem of computing within complex systems is closely related to the
question of decentralized parallel computation, in general, for which there exists
abundant literature. Different paradigms exist for controlling and harvesting
the computations within complex systems. Several other names have been used for
closely related topics, such as organic computing
\parencite{muller-schloerOrganicComputingParadigm2011} which is the study of
systems with life-like properties such as self-organization or the ability to
adapt to a dynamically changing environment. Agent-based computing
\parencite{jenningsAgentBasedComputingPromise1999} focuses on computing systems
composed of several relatively simple autonomous agents. Amorphous computing
\parencite{abelsonAmorphousComputing2000,
  nagpalProgrammablePatternFormationScaleIndependence2008,
  nagpalProgrammableSelfassemblyUsing2002} is about making large amounts of
individual computing elements work and ensuring ``the cooperation of large
numbers of unreliable parts interconnected in unknown, irregular, and
time-varying ways''.

\subsection{Computing in cellular automata}\label{sec:comp-cell-autom}

\Acfp{CA} are decentralized parallel systems with many identical components with
local connectivity. Because of these properties, they have the potential to
carry out robust and efficient computations. \ac{CA}-based computing machines could
recover from perturbations or carry out computations in stochastic environments.
Moreover, they are also interesting for modeling the behavior of natural complex
systems. For more details on \acp{CA}, see Section~\ref{sec:cellular-automata-sec}.

\paragraph{Von Neumann's self-reproducing \ac{CA}.}
The early developments of \acp{CA} were guided by Von Neumann's question ``What
kind of logical organization is sufficient for an automaton to be able to
reproduce itself?'' Since this question may admit very simple solutions,
additional constraints were added so that the problem does not admit trivial
solutions. For example, Von Neumann required that the automaton in question be
equivalent in power to a universal Turing machine while having minimal
complexity \parencite{vonneumannTheorySelfreproducingAutomata1966}. The final
model he proposed was an intricate \ac{CA} composed of a tape containing the
instructions to construct the next automaton and a construction unit that
would progressively build that new automaton cell by cell. This was an early
example of complex computations being carried out by a relatively simple
decentralized system such as a \ac{CA}.

\paragraph{Universal Computation in Cellular Automata.}
It is not difficult to see that a \ac{CA} can be capable of universal
computation. The basic approach is to show that it can simulate a Turing
machine, which we assume has an infinite tape. A \ac{CA} rule can be constructed
by reproducing all the steps of a Turing machine's behavior while adding a state
to each cell, indicating whether it is active or not, and enforcing that only one
cell is active at a time. This turns the parallel \ac{CA} into a sequential
object that simulates a Turing machine. A similar construction was carried out
by \parencite{smithSimpleComputationUniversalCellular1971}, making a \ac{CA}
with one-to-one correspondence between its time steps and that of a target
Turing machine.

\begin{figure}[htbp]
  \centering
  \includegraphics[width=.8\linewidth]{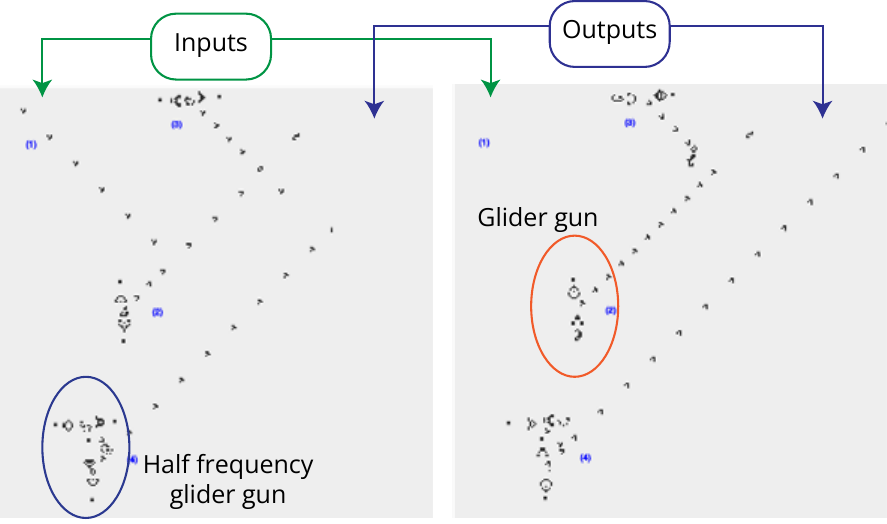}
  \caption{Construction of a NOT gate in game of life. The left image represents
    an input equal to 1, while the second shows an input equal to 0. Output
    streams behave like a NOT gate. A glider gun and a modified half-frequency
    glider gun are highlighted. They are necessary for the construction of the
    gate. Images are adapted from a blog post
    \parencite{carliniDigitalLogicGates2020}.}
  \label{fig:gol_not_gate}
\end{figure}

Another approach was used to show the universality of the game of life (see
details in Section \ref{sec:game-life}). Instead of simulating a Turing machine,
basic logical functions are built from gliders (see figure \ref{fig:glider})
produced by a game of life structure called \emph{glider gun}. Two gliders that
collide within the simulation will either annihilate or keep moving depending on
the exact position in which they collide. This allows the construction of NOT, AND and
OR gates. For example, see Figure \ref{fig:gol_not_gate} for a construction of a
NOT gate in the game of life.

Universal computation is an intriguing yet impractical property, and proving
that a system is a universal computers often has limited usability, but it
serves as a proof of principle that the studied system is theoretically as
powerful as any computer. In the case of the game of life, this was illustrated
by the implementation of the OTCA metapixel which is a tileable square cell
object that can be used to simulate another instance of game of life within the
game. It is important to note that such embedded computers and Turing machines
are often very slow and inefficient, and the game of life has brittle structures
which would be destroyed by the slightest perturbation, making these
constructions unusable for anything else than demonstration.

\paragraph{Evolving cellular automata with genetic algorithms.}
An important advancement beyond the clever yet challenging manual design of 
computational rules in \acp{CA} is the utilization of learning algorithms 
for automated rule design \parencite{mitchellEvolvingCellularAutomata1996}. 
Genetic algorithms, which are search methods inspired 
by biological evolution, are one such type of
learning algorithm \parencite{bookerClassifierSystemsGenetic1989}. See Section
\ref{sec:genetic-programming} for more details about genetic algorithms.

Some early work on evolving \acp{CA} was done by Norman Packard and colleagues
\parencite{packardAdaptationEdgeChaos1988,
  richardsExtractingCellularAutomaton1990}.
\textcite{richardsExtractingCellularAutomaton1990} used genetic algorithms to
learn the rules of \ac{CA} to match experimental data for the patterns created
by dendritic solidification of \ce{NH4Br}. In
\parencite{packardAdaptationEdgeChaos1988}, the author evolves \acp{CA} to
perform a specific computation, and observes that the evolution process tends to
select \acp{CA} with a $\lambda$ parameter close to the critical value observed by
Langton that corresponds to the edge of chaos (see Section
\ref{sec:langtons-lambda} for details about the parameter). However, other
authors were unable to replicate this experiment later
\parencite{mitchellRevisitingEdgeChaos1993}.
\textcite{kozaEvolutionSubsumptionUsing1992} applied genetic algorithms to
\acp{CA} to generate random numbers.

\begin{figure}[htbp]
  \centering
  \begin{subfigure}[b]{.4\linewidth}
    \centering
    \includegraphics[width=\linewidth]{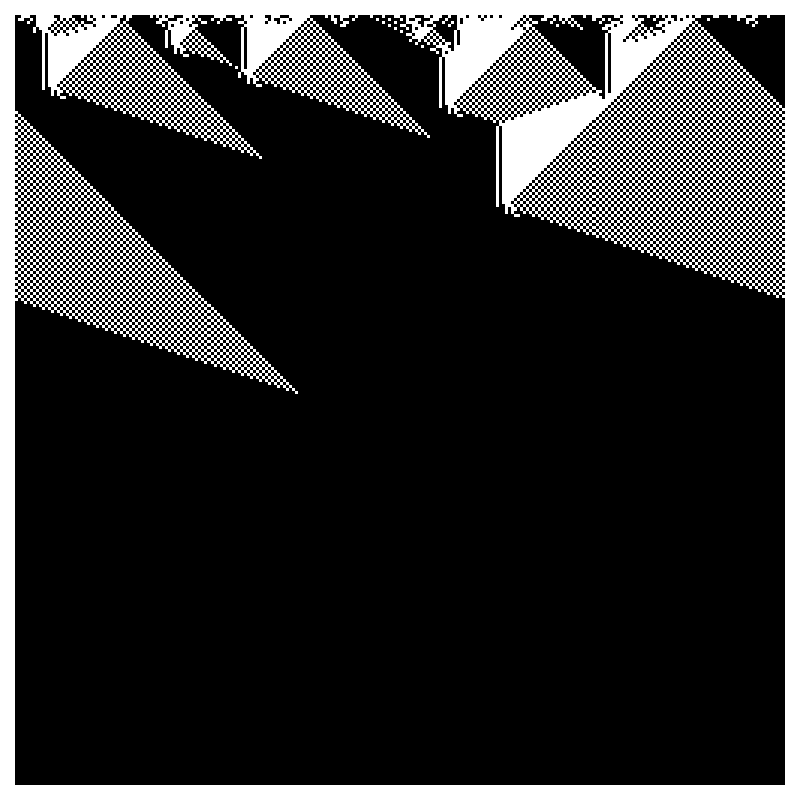}
    \caption{Density $> 0.5$}
    \label{fig:particle_ca_full}
  \end{subfigure}
  \hspace{10pt}
  \begin{subfigure}[b]{.4\linewidth}
    \centering
    \includegraphics[width=\linewidth]{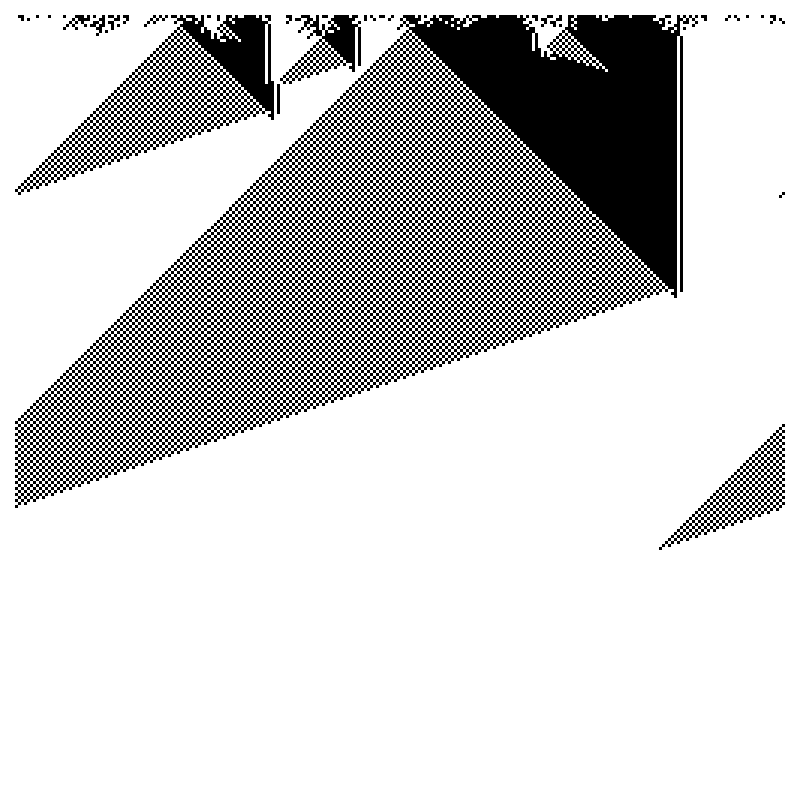}
    \caption{Density $< 0.5$}
    \label{fig:particle_ca_empty}
  \end{subfigure}
  \caption{The particule-based solution to the density classification problem
    \parencite{mitchellEvolvingCellularAutomata1996}. When the density is above
    0.5, the \ac{CA} quickly reaches a state with only black cells, whereas when
    the density is lower, it goes to a state with white cells only.}
  \label{fig:particle_ca}
\end{figure}

The problem of density classification was thoroughly explored since it fits the
\ac{CA} paradigm particularly well \parencite{mitchellRevisitingEdgeChaos1993,
  mitchellEvolvingCellularAutomata1994,
  crutchfieldEvolutionEmergentComputation1995, dasGeneticAlgorithmDiscovers1994,
  sipperCoevolvingNonuniformCellular1996, andreDiscoveryGeneticProgramming1996}.
The goal of this task is to have \acp{CA} figure out what the most abundant cell
state in an initial binary configuration in one dimension is. This task is trivial
for a central controller that can read the state of all cells in the initial
grid, but it is much more complex for decentralized systems like \acp{CA}
that have a small radius of interaction with neighbors. The evolved \acp{CA}
developed various strategies to solve the task. Some are unsophisticated, such
as expanding the 1s or 0s, which does not work on all validation examples. We
show one of the more sophisticated solutions discovered during some runs of the
genetic algorithm in figure \ref{fig:particle_ca}. That solution uses the
propagation of structures within the \ac{CA} state space to communicate the
local state to the rest of the grid, which usually ends with the right answer
from the \ac{CA}.

\paragraph{Reliable computation in cellular automata.}
The problem of carrying out reliable computations in \acp{CA} is studied early
on in its history because of the potential of that model for a real hardware
implementation. The first studies used probabilistic approaches for solving the
error detection and correction problems of automata --- which includes other
models than \acp{CA} \parencite{neumannProbabilisticLogicsSynthesis1956,
  winogradReliableComputationPresence1963,
  ,tsertsvadzeStochasticAutomataProblem1964,
  ,haraoConsiderationCellularAutomata1973}. Other methods make use of the
special structure of \acp{CA}, but rely on strong assumptions about the
likelihood of errors \parencite{haraoFaultTolerantCellular1975,
  nishioFaultTolerantCellular1975}. Peter Gács has shown how to construct a 1D
\ac{CA} which can perform arbitrarily large reliable computations, assuming that each
cell gets an error with a positive probability
\parencite{gacsReliableComputationCellular1986}. The fault model he describes is
important from the point of view of ergodic theory because Gács’ result
invalidates the ``positive probability conjecture'' in statistical physics,
which states that all one dimensional infinite particle systems with positive
transition probabilities are ergodic, which means that it will visit all parts of
its state space eventually. For more recent work on fault tolerance by Gács, see
\parencite{gacsReliableCellularAutomata2001}.

\paragraph{Neural cellular automata\label{sec:neur-cell-autom}.}
The principle of \ac{NCA} is based on the analogy between \acp{CA} and two types
of neural networks: \acp{RNN} and \acp{CNN} (see Section
\ref{sec:cell-autom-rnns} for more details about the analogy). Following this
analogy, \acp{CA} can be formulated as a continuous recurrent and convolutional
neural network system called a \acf{NCA}. \acp{NCA} can be manipulated like
neural networks, using automatic differentiation, backpropagation, and
optimization algorithms to make these models learn some target task. The first
example used this neural network representation to learn a \ac{NCA} version of
existing \ac{CA} rules from recorded examples of their evolution
\parencite{gilpinCellularAutomataConvolutional2018}. The structure of the
training process and the final error of the resulting model are used by the author as
a tool to understand the complexity and properties of the original \ac{CA}.

\acp{NCA} were also used to learn to produce stable self-repairing patterns that
resemble a target image \parencite{mordvintsevGrowingNeuralCellular2020}. The
training process is structured in such a way that the \ac{NCA} needs to learn to
maintain the pattern stable across time and space, but also recover from random
perturbations that are introduced at arbitrary points in time. This results in
an interesting demo where a pattern is generated and maintained in real-time by
a \ac{NCA} while a user can directly apply
perturbations\footnote{\url{https://distill.pub/2020/growing-ca/}}.

\begin{figure}[htbp]
  \centering
  \begin{subfigure}[t]{.49\linewidth}
    \centering
    \includegraphics[width=\linewidth]{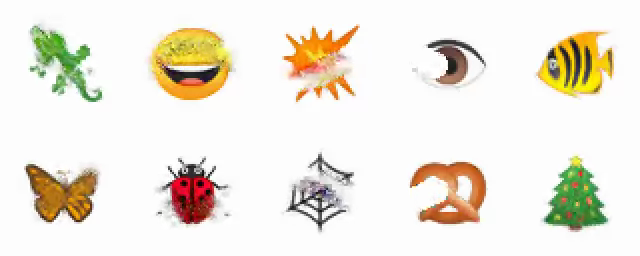}
    \caption{Perturbed patterns}
    \label{fig:nca_perturb}
  \end{subfigure}
  \begin{subfigure}[t]{.49\linewidth}
    \centering
    \includegraphics[width=\linewidth]{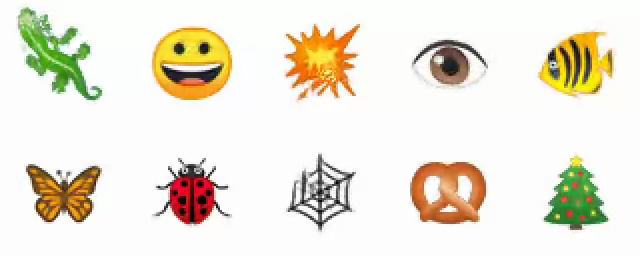}
    \caption{Recovered patterns after a few hundred \ac{NCA} steps.}
    \label{fig:nca_recover}
  \end{subfigure}
  \caption[Resistance of patterns]{Resistance of patterns to perturbations with
    \acp{NCA}. \ac{NCA} can robustly maintain patterns, even under relatively
    strong perturbations (~10-20\% of the pattern's size). This image was
    generated with an online notebook based on code from Alexander
    Mordvintsev\footnotemark.}
  \label{fig:nca}
\end{figure}
\footnotetext{\url{https://colab.research.google.com/github/google-research/self-organising-systems/blob/master/notebooks/growing_ca.ipynb}}

The same authors also used \acp{NCA} to learn to classify digits from the MNIST
database of handwritten digits
\parencite{randazzoSelfclassifyingMNISTDigits2020}. The particularity of this
model is its ability to perform the classification in a decentralized way, by
having the \ac{NCA} change the color of the digits directly on the \ac{CA} grid.
Cells can only communicate with their direct neighbors, so the model has to
reach a consensus by spreading information across the grid. This is done in
parallel, so multiple digits can be classified simultaneously through this
progressive consensus
process\footnote{\url{https://distill.pub/2020/selforg/mnist/}}.
Another work used the same model to learn to continuously generate images with
the same texture as a target \parencite{niklassonSelfOrganisingTextures2021}.
Again, an advantage of this method is the ability to recover from perturbations.

\begin{figure}[htbp]
  \centering
  \includegraphics[width=.8\linewidth]{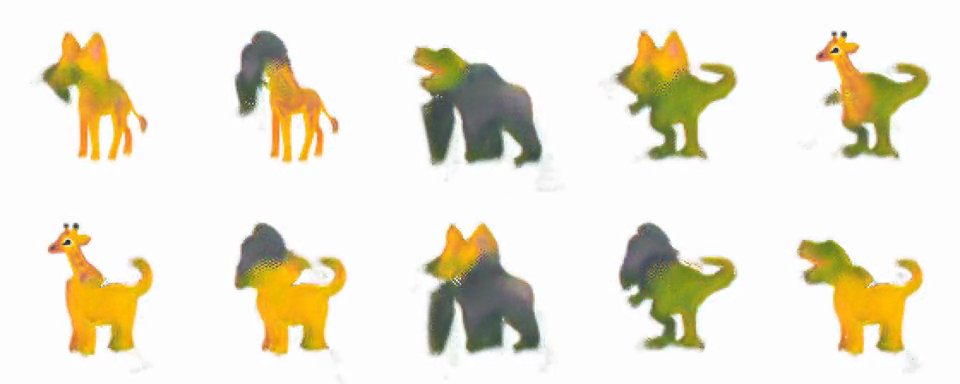}
  \caption{Examples of stable hybrids obtained with \acp{NCA} trained with the
    method of \parencite{cisnerosOpenendedCreationHybrid2021}.}
  \label{fig:nca_hybrid}
\end{figure}

We used \acp{NCA} as part of our submission to the Minecraft open-endedness
challenge 2021\footnote{\url{https://evocraft.life/}}, a competition organized
at the Gecco 2021 conference competition
track\footnote{\url{https://gecco-2021.sigevo.org/Competitions}}
\parencite{cisnerosOpenendedCreationHybrid2021}. The goal was to build an
open-ended system within the computer game Minecraft. This means building a
systems that fulfill the following requirements:

\begin{itemize}
  \item Divergence: Open-ended algorithms are not expected to slow down or
        converge but rather keep expanding and generating more complex outputs
        over time.

  \item Diversity: Does the algorithm produce entities with strong phenotypic
        diversity?

  \item Complexity: Can the algorithm produce complex entities or entities
        interactions that give rise to complex systems? Are hierarchical and
        modular structures present?

  \item Ecological interactions: Do the created entities interact with each
        other?

  \item Life-Like properties: Inspiration may be taken from other attributes of
        living systems.
\end{itemize}

In each of the examples in Figure \ref{fig:nca}, a separate \ac{CA} was trained
on a pattern starting from a single black pixel. We extend this setup by
training a single \ac{NCA} on multiple patterns from different seeds. Instead of
a single black cell, a small number of black cells are arranged in simple
shapes. We call these patterns \emph{seeds}. \acp{NCA} are neural networks with many redundant parameters. This allowed us to learn (and encode) more than one
pattern (or larger patterns) without adding parameters in the architecture. The
resulting patterns are slightly lower quality but still stable under
perturbations and capable of growing from their corresponding seed pattern.

We turned the process of combining seed patterns into a ``game'' by allowing user
interaction with the evolving CA through the Minecraft video game, using an API
to communicate between the program and the game. Our system is not open-ended by
itself, but serves as an exploration space and uses human interactions to play
with the shapes and seed patterns. This is in the spirit of open-ended game-like
systems like Picbreeder \parencite{secretanPicbreederCaseStudy2011,
  woolleyDeleteriousEffectsPriori2011} where players choose which generated
image to mutate or evolve from, collectively constructing surprisingly complex
images. The algorithms can generate endless novelty with the help of human
interactions that provide the missing step for making the system potentially
open-ended.

\subsection{Amorphous computing}
The goal of amorphous computing is to study and define the problem of computing
with multiple interconnected components that are unreliable, irregular,
time-varying, and with limited computational capacity
\parencite{abelsonAmorphousComputing2000}. The motivation for this area of
research is the development of biological substrates that can compute, function
as sensors and actuators, and robustly self-organize to compute with little
cost. The precise manufacturing of chemical or biological substrates with the ability
to communicate locally through chemical or physical interactions is within
reach, making their use as computing vessels particularly
attractive. We can embed millions of chips with sensors
\parencite{abelsonAmorphousComputing2000} or program biological cells to serve
as logic gates \parencite{weissProgrammingBiologicalCells1998,
  weissVivoDigitalCircuits2002} while relying on cheaper, decentralized parts
\parencite{buteraProgrammingPaintableComputer2002}.

\begin{figure}[htbp]
  \centering
  \includegraphics[width=.8\linewidth]{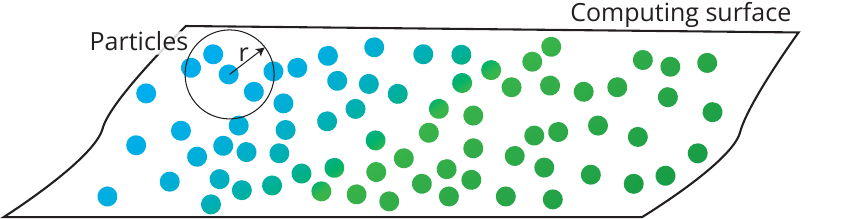}
  \caption{An example of amorphous computing on a surface. Multiple computing
    particles are arranged randomly. A wave (represented by the color of the
    cells) propagates within the medium. They can communicate with neighbors
    within a radius $r$.}
  \label{fig:amorphous_computing}
\end{figure}

In amorphous computing systems, a medium comprises ``computational
particles'' spread on a surface or mixed in a volume. The particles have 
limited computing power, with no knowledge of their position or orientation. The
particles can communicate with neighbors via some specified mechanism, which may
be analogous to biological ones. The two main components of this research are 
\emph{synthetic biology} and the choice of computing paradigm.
\begin{description}
  \item[Synthetic biology.] The design of a substrate from chemical processes or
        genetically engineered biological cells
        \parencite{weissVivoDigitalCircuits2002}. It should behave correctly and
        follow predefined rules.
  \item[Computing paradigm.] How to program individual agents to follow a
        predefined global goal
        \parencite{nagpalProgrammableSelfassemblyConstructing2001}?
\end{description}

The growing point language (GPL) is an example of a programming language that
enables programmers to specify complex patterns for computational particles,
which are internally represented in the particles as state machines
\parencite{cooreBotanicalComputingDevelopmental1999}.

\textcite{nagpalProgrammableSelfassemblyUsing2002} developed another
programming language inspired by biological cell differentiation
\parencite{lawrenceMakingFlyGenetics1992,
  wolpertPositionalInformationSpatial1969} that can be compiled into individual agent
programs to follow some global specifications.

\chapter{Measuring complexity in evolving complex systems}
\label{cha:meas-compl-evolv}

In this chapter, we propose an approach to measure the growth of the complexity of
emerging patterns in complex systems such as \acfp{CA}. We discuss several ways
in which a metric to measure complexity growth can be defined. This includes
approaches based on compression algorithms and artificial neural networks. We
believe that such a metric can be useful for designing systems that could exhibit
open-ended evolution, which itself might be a prerequisite for the development of
general artificial intelligence. We conduct experiments on 1D and 2D grid worlds
and demonstrate that using the proposed metric, we can automatically construct
computational models with emerging properties similar to those found in
Conway's Game of Life, as well as many other emergent phenomena. Interestingly,
some of the patterns we observe resemble forms of artificial life 
\parencite{langtonEditorIntroduction1993}. We test our
metric of structural complexity growth on cellular automata, but it can also be
applied to a wide range of complex systems.

\section{Introduction}
Recent advances in machine learning and deep learning have had success in
reproducing some very complex feats traditionally thought to be only achievable
by living beings. However, making these systems adaptable and capable of
developing and evolving on their own remains a challenge that could be crucial to 
eventually developing AI with general learning capabilities (for example, as
is further discussed in \parencite{mikolovRoadmapMachineIntelligence2018}).
Building systems that mimic some key aspects of the behavior of existing
intelligent organisms (such as the ability to evolve, improve, adapt, etc.)
might represent a promising path. Intelligent organisms --- for example, human beings but
also most living organisms, if we consider a broad definition of intelligence ---
are a form of spontaneously occurring, ever-evolving complex systems that
exhibit these kinds of properties
\parencite{bookerPerspectivesAdaptationNatural2004}. The ability to sustain
open-ended evolution appears to be a requirement in order to enable the emergence of
arbitrarily complex adaptive systems.

Although a rigorous attempt at defining intelligence or life is beyond the scope
of this work, we assume that a system we might identify as evolving, with the
potential of developing intelligence, should have the property of
self-preservation and the ability to grow in complexity over time. These
properties can be observed in living organisms
\parencite{bookerPerspectivesAdaptationNatural2004} and should also be part of
computational models that aim to mimic them.

To recognize self-preservation and growth in complexity, one should be able to
detect emerging macro-structures composed of smaller elementary components. For
the purpose of obtaining computational models that grow in complexity over time,
one should also be able to determine the amount of complexity these systems
contain. In this chapter, we propose and discuss several ways to estimate complexity and detect the presence of emerging and stable patterns in complex
systems such as cellular automata. We show that such metrics are useful when
searching the space of cellular automata with the objective of finding those
that seem to evolve over time.

\section{Related work}

\subsection{Artificial life and open-ended evolution}

Several works have attempted to artificially create open-ended evolution. A
nonexhaustive list of some well-known systems include Tierra
\parencite{rayApproachSynthesisLife1991}, Sims
\parencite{simsEvolvingVirtualCreatures1994}, Avida
\parencite{ofriaAvidaSoftwarePlatform2004}, Polyworld
\parencite{yaegerComputationalGeneticsPhysiology1994}, Geb
\parencite{channonImprovingStillPassing2003}, Division Blocks
\parencite{spectorDivisionBlocksOpenended2007} and Chromaria
\parencite{sorosIdentifyingNecessaryConditions2014}. Designs that focus on an objective and make use of reinforcement learning methods to drive evolution
are also being studied, e.g. in
\parencite{pathakLearningControlSelfAssembling2019}. Most of these simulated
``worlds'' have had some success reproducing key aspects of evolving
artificial life, enabling the emergence of complex behavior from simple
organisms. However, they still work within constrained simulated environments
and usually consider organisms composed of elementary building blocks, while
they do not work outside of this usually very constrained framework. The divergent
and creative evolutionary process could be happening at a much lower conceptual
level with fewer assumptions. For this reason, we consider cellular automata in
the rest of the chapter because they rely on very few assumptions while
offering a very large expressive power and a potentially wide range of behaviors
that can be discovered. However, the metrics defined in this work have the
potential to be applied to other types of complex system, as discussed in
Section~\ref{sec:conclusion}.

\subsection{Cellular automata}

Cellular automata are very simple systems, usually defined in one or two
dimensions, composed of cells that can be in a set of states. The cells are
updated in discrete time steps using a transition table that defines the next
state of a cell given the states of its neighbors. They were originally proposed
by Stanislaw Ulam and studied by Von Neumann
\parencite{vonneumannTheorySelfreproducingAutomata1966}, who was interested in
designing a computational system that can self-reproduce itself in a non-trivial
way. The trivial self-reproducing patterns were then those that do not have the
potential to evolve, for example, the growth of crystals.

Stephen Wolfram later took a more bottom-up approach, beginning with the study
of simple 1D binary cellular automata (CA), and identifying four qualitative
classes of cellular automaton behavior
\parencite{wolframUniversalityComplexityCellular1984}:

\begin{description}
\item[Class 1] evolves to a homogeneous state.
\item[Class 2] evolves to simple periodic patterns.
\item[Class 3] produces aperiodic disordered patterns.
\item[Class 4] produces complex aperiodic and localized structures, including
  propagating structures.
\end{description}

Wolfram and his colleagues also studied 2D CA using tools from information
theory and dynamical systems theory, describing the global properties of these
systems in terms of entropies and Lyapunov exponents
\parencite{packardTwodimensionalCellularAutomata1985}.

Christopher Langton and colleagues also studied CA dynamics
\parencite{liTransitionPhenomenaCellular1990} --- e.g. using the $\lambda$ parameter
\parencite{langtonComputationEdgeChaos1990} --- and designed a self-replicating
pattern much simpler than Von Neumann's
\parencite{langtonSelfreproductionCellularAutomata1984}, now known as Langton
loops. The main issues with his system and Von Neumann's universal replicator is
the fact that they are very fragile and based on a large amount of human design.
As a consequence, although they self-replicate, they cannot increase in
complexity and are not robust to perturbations or unexpected interactions with
the environment.

A genetic algorithm-based search for spontaneously occurring self-replicating
patterns was carried out in 2D cellular automata with several states was undertaken in
\parencite{bilottaARTIFICIALMICROWORLDSPART2011} using entropy measures of the
frequency distribution of $3\times 3$ patterns.

\subsection{Compression and complexity}
Compression has often been used as a measure of complexity. Lempel and Ziv have
introduced in \parencite{lempelComplexityFiniteSequences1976} the now widely used
Lempel-Ziv (LZ) algorithm as a method for measuring the complexity of a
sequence. By constructing back-references to previous parts of a string, the LZ
algorithm is capable of taking advantage of duplicate patterns in the input to
reduce its size. The DEFLATE algorithm that we use in the following section
combines LZ with Huffman coding for an efficient representation of the symbols
obtained after the first step. It is one of the most widespread compression
algorithms and is, for instance, used in gzip and PNG file compression standards.

The PAQ compression algorithm series \parencite{mahoneyFastTextCompression2000}
is an ensemble of compression algorithms initially developed by Matt Mahoney
with state-of-the-art compression ratio on several compression benchmarks.
Better compression of an input means a better approximation of the minimum
description length and implicit understanding of more of the underlying patterns
in input data. The usefulness of a better compressor is that it can detect much
more complex and intricate patterns that aren't simple repetitions of previous
patterns.

In \parencite{zenilCompressionBasedInvestigationDynamical2010}, H. Zenil
investigates the effects of a compression-based metric to classify cellular
automata and observes that it results in a partitioning of the space of 1D CA
into several clusters that match Wolfram's classes of automata. He also used
this approach in the output of simple Turing machines and some 1D automata with
more than two states and larger neighborhoods. Extensions of this work include
an asymptotic sensitivity analysis of compressed length for input
configurations of increasing complexity, as introduced in
\parencite{zenilAsymptoticBehaviorRatios2013,
  zenilWhatNatureLikeComputation2014}.

Additionally, the image decompression time as an approximation of Bennet's logical
depth \parencite{bennettLogicalDepthPhysical1995,
  zenilImageCharacterizationClassification2012} and the output distribution of
simple Turing machines combined with block decomposition of CA to approximate
their algorithmic complexity have also been investigated
\parencite{zenilTwodimensionalKolmogorovComplexity2015,
  soler-toscanoCalculatingKolmogorovComplexity2014}. However, the possible
extent to which such measures of complexity could be applied to more complex
automata and other complex systems has not yet been extensively studied. For a
review of several measures of complexity and their applications, see
\parencite{grassbergerRandomnessInformationComplexity1989}.

\section{Compression-based metric}\label{sec:compr-based-metr}

A cellular automaton of size $n$ in 1D can be represented at time $t$ by its
grid state $S^{(t)} = \{c_1^{(t)}, ..., c_n^{(t)}\}$ where each $c_i$ (also
called a cell) can take one of $k$ possible values (representing the possible
states) and a transition rule $\phi$. The transition rule is defined with
respect to a neighborhood radius $r$ with the mapping $\phi(c^{(t)}_{i-r}, ...,
c^{(t)}_i ..., c^{(t)}_{i+r} ) = c^{(t+1)}_i$ that maps $\{1, ..., k\}^{2r+1}$
to $\{1, ..., k\}$. The quantity $2r + 1$ corresponds to the number of
cells taken into account to calculate the next state of a cell, namely the
cell itself and the $r$ neighboring cells in both directions.

Simulating a CA amounts to the recursive application of this mapping $\phi$ to
an initial state $S^{(0)} = \{c_1^{(0)}, ..., c_n^{(0)}\}$.

In the rest of the chapter, we consider cyclic boundary conditions for the
automata, meaning that the indices $i-r, ..., i+r$ above are taken modulo $n$
the size of the automaton in 1D. Boundary conditions can have some effect on a
CA's evolution, but cyclic boundaries have been empirically observed to have
limited effect on the complexity of automata in 1D
\parencite{luvalleEffectsBoundaryConditions2019}.

The definition given in the equation above can be extended to higher dimensional
automata by modifying the neighborhood and the definition of $\phi$. A 2D
neighborhood of radius 1 can be defined as the 3 by 3 square around the center
cell --- also called the Moore neighborhood --- or by only considering the four
immediate horizontal and vertical neighbors of the center cell --- the Von Neumann
neighborhood.

Elementary cellular automata (ECA) are 1D CA with $k = 2$ and $r = 1$. There are
$2^3$ elements in $\{1, ..., k\}^{2r+1}$ and $2^{2^3} = 256$ possible different
set transition rules that are often compactly represented as a binary string
with 8 bits. The relatively low number of rules of this type makes it possible
to appreciate the performance of a metric and compare it with others.

We define the compressed length $C$ of a 1D cellular automaton at time $t$ as
\begin{align}
  C(S^{T}) = \text{length}\left(\texttt{comp}(c_1\ ||\ c_2\ ||\ ...\ c_n)\right)
\end{align}

where $||$ denotes the string concatenation operator, and cells $c_i$ are
implicitly converted into string characters (with one symbol per unique state).
\texttt{comp} is a compression algorithm that takes a string as input and
outputs a compressed string, and length is the operator that returns the length
of an input string.

\begin{figure}[tbp]
  \begin{subfigure}[b]{.47\linewidth}
    \centering
    \includegraphics[width=\linewidth]{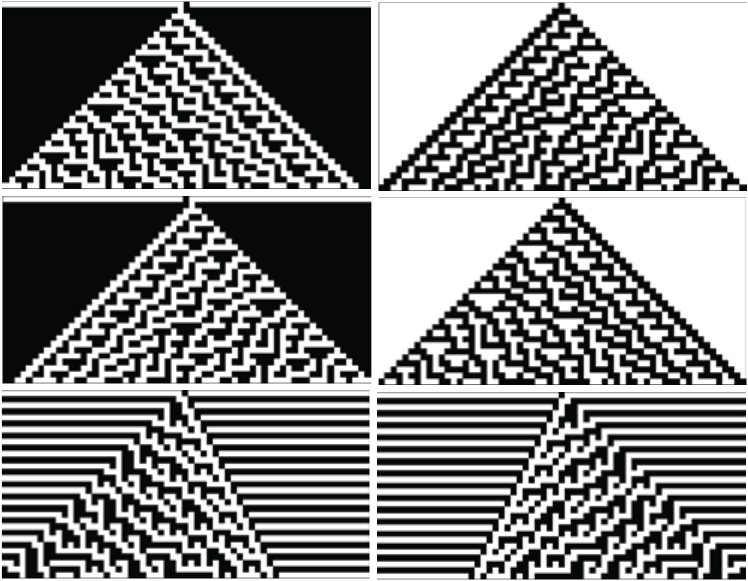}
    \caption{6 highest scoring automata. Only the first 30 timesteps are shown
      for readability.}
    \label{subfig:highest_6}
  \end{subfigure}
  \hspace{10pt}
  \begin{subfigure}[b]{.47\linewidth}
    \centering
    \includegraphics[width=\linewidth]{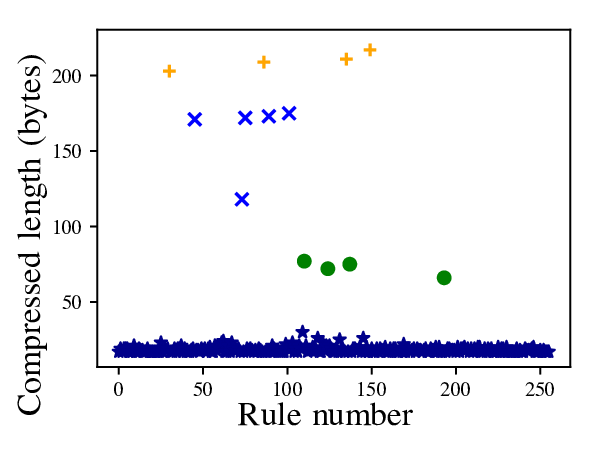}
    \caption{All 256 compressed length scores.}
    \label{subfig:comp_scores}
  \end{subfigure}
  \caption{Compression-based metric on 1D ECA. \ref{subfig:highest_6} represents
    the 1D ECA evolution with each line being the state of the automaton at a
    given timestep, starting from a single cell set to 1. Cells which are in
    state 1 are represented in black and cells in state 0 are represented in
    white. Time increases downwards. Figure~\ref{subfig:comp_scores} represents
    the compressed length of the 256 ECA rules, with different marker and colors
    corresponding to the obtained KMeans clusters.}

  \label{fig:comp_eca}
\end{figure}

Similarly to \parencite{kowaliwMeasuresComplexityArtificial2008,
  zenilCompressionBasedInvestigationDynamical2010}, we use zlib’s C
implementation of DEFLATE to compress the final state of the automaton. If we
apply the above metric to the 256 ECA run for 512 timesteps and initialized with
one activated cell in the middle, we obtain the plot of
Figure~\ref{subfig:comp_scores}. This example is re-used in the rest of the
chapter as a way to easily visualize and check that the defined complexity
measures are coherent with one another. The colors in
Figure~\ref{subfig:comp_scores} were obtained with a KMeans clustering algorithm
applied on the compressed length of the automata states.

As visible on Figure~\ref{subfig:comp_scores}, rules are clearly separated into
several clusters that turn out to match Wolfram’s classification of ECA. Class 3
behavior can be found at the top of the plot (highest compressed length, orange
and blue clusters), Class 1 and 2 are clearly separated at the bottom part (not
detailed here) and Class 4 rules (colored in green) lie in between the other
types of behavior. The 6 highest scoring rules are shown on
Figure~\ref{subfig:highest_6} and correspond to Class 3 behavior in Wolfram's
classification. Among the classes of behavior, some sub-clusters can be found
that correspond to similarly behaving rules.

Ultimately, the theoretical goal of using compression algorithms is to approach
the theoretical minimum description length of the input
\parencite{grunwaldMinimumDescriptionLength2007}. For very regular inputs, this
length should be relatively small and inversely for random inputs. However, gzip
and PAQ are crude approximations of the minimum description length and may only
approach it in a given context. As an example, compressing text data (a task
often performed with gzip in practice) is much more efficient with a language
model that can assign a very low probability to non grammatically correct
sentences. The Kolmogorov complexity
\parencite{kolmogorovThreeApproachesQuantitative1968} of a cellular automaton is
upper bounded by a value that is independent of the chosen rule, as it is
entirely determined by the transition table, the grid size, initial
configuration and number of steps.

\section{Predictor-based metric}

One obvious limit of using compression length as a proxy for complexity is the
fact that interesting systems mostly have intermediate compressed length.
Compressed length increases with the amount of disorder in the string being
compressed. Therefore, extreme lengths correspond either to systems that do not
increase in complexity on the lower end of the spectrum, or systems that produce
a maximal amount of disorder on the higher end. Neither of them have the
potential to create interesting behavior and increase in complexity.
Intermediate values of compressed length are also hard to interpret, since
average lengths might either correspond to interesting rules or slowly growing
disordered systems.

To cope with this limitation, one should also take into account the dynamics of
complexity, that is, how the system builds on its complexity at a given time as
it keeps evolving, while retaining some of the structures it had acquired
earlier. Compression leverages the amount of repetitions in inputs to further
compress, and this may also be used as an estimate of structure overlap, as
explained in the following section.

\subsection{Joint compression}\label{sec:joint-compression}

As a way to both measure the complexity and the amount of overlap between two
automata states apart in time, we define a joint compressed length metric for a
delay $\tau$ as
\begin{align}
  C'\left(S^{(T + \tau)}, S^{(T)}\right) =
  C\left(S^{(T)}\ ||\ S^{(T + \tau)}\right)
\end{align}
where $||$ represents the concatenation operator. This quantity is simply the
compressed length of a pair of global states --- defined at the beginning of
\ref{sec:compr-based-metr}, represented by the letter $S$ --- at two timesteps
separated by a delay $\tau$. In 1D, concatenation means chaining the two string
representations before compressing, and in 2D we can chain two flattened
representations of the 2D grid. This introduces several issues which we discuss
in Section~\ref{sec:count-based-pred}.

To quantify the amount of overlap between the two global states, we can compute
the ratio of this joint compressed length with the sum of the two compressed
lengths $C(S^t)$ and $C(S^{t-\tau})$, thereby forming the joint compression
score
\begin{align}
  \mu = \dfrac{C\left( S^t \right) +
  C\left( S^{t - \tau} \right)}{C'\left( S^t, S^{t - \tau} \right)}
\end{align}
defined for an automaton $S$, time $t$ and delay $\tau$.

\begin{figure}[htbp]
  \centering
  \includegraphics[width=\linewidth]{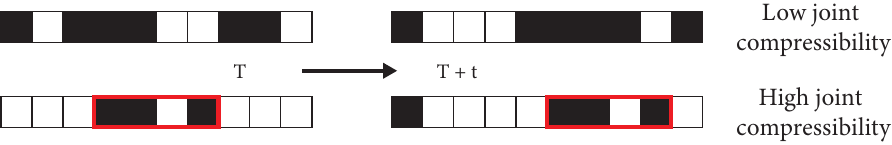}
  \caption{Joint compression method illustration. If a structure persists
    through time, this will decrease the joint compressed length compared to the
    sum of compressed lengths. A persistent structure is circled in red.}
  \label{fig:joint_schema}
\end{figure}

This metric is based on the intuition that if patterns occur at step $T - \tau$
of the automaton's evolution and are still present at step $T$, the joint
compressed length will be lower than the sum of the two compressed length. The
idea is illustrated in Figure~\ref{fig:joint_schema}, where it is pointed out
that a stable moving structure (sometimes called \emph{glider} or
\emph{spaceships} in Game of Life) will yield lower joint compressed lengths.
This also applies to structures that self-replicate, grow from a stable seed or
maintain the presence of some sub-structures. Bigger structures yield a higher
compression gain.

\begin{figure}[htbp]
  \centering
  \begin{subfigure}[b]{.45\linewidth}
    \centering
    \includegraphics[width=\linewidth]{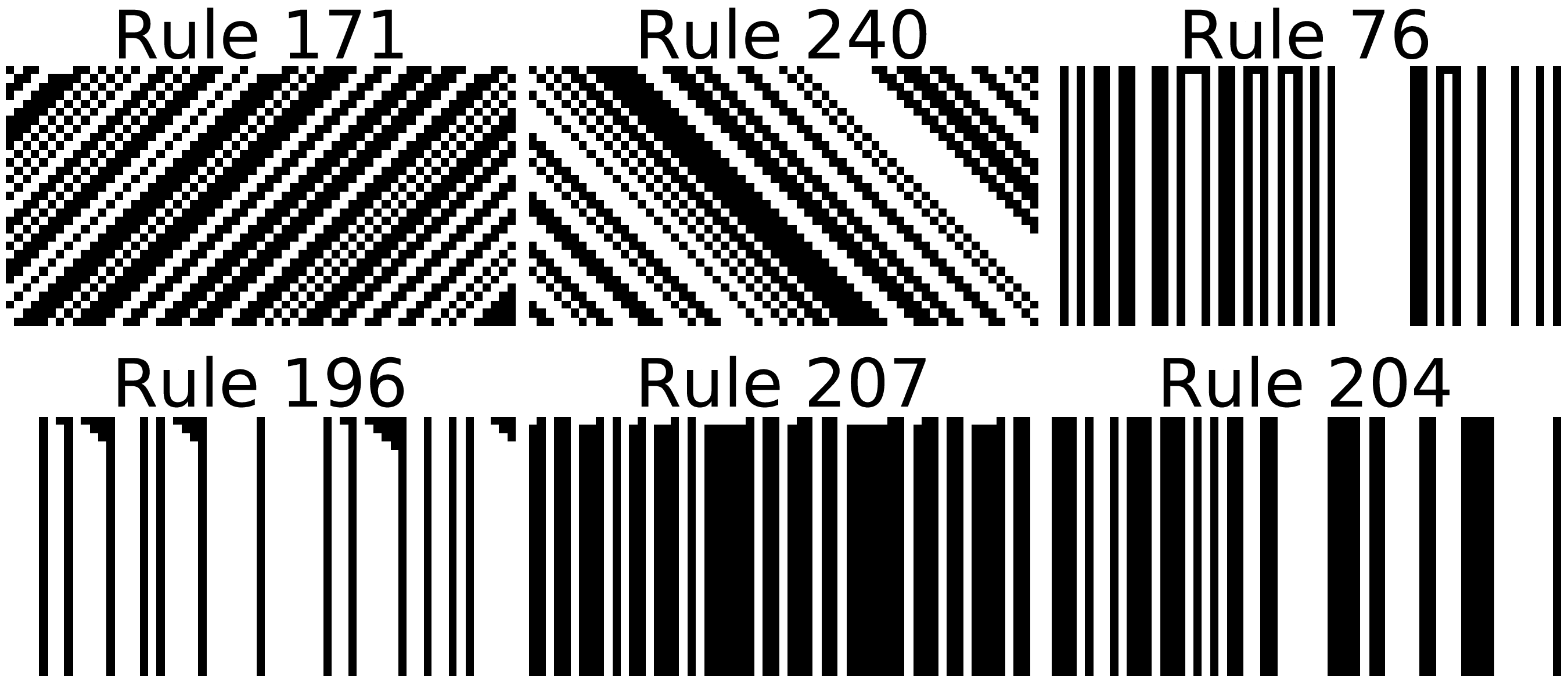}
    \caption{Highest joint compression score among the 256 ECA.}
    \label{fig:high_eca_joint}
  \end{subfigure}
  \begin{subfigure}[b]{.45\linewidth}
    \centering
    \includegraphics[width=\linewidth]{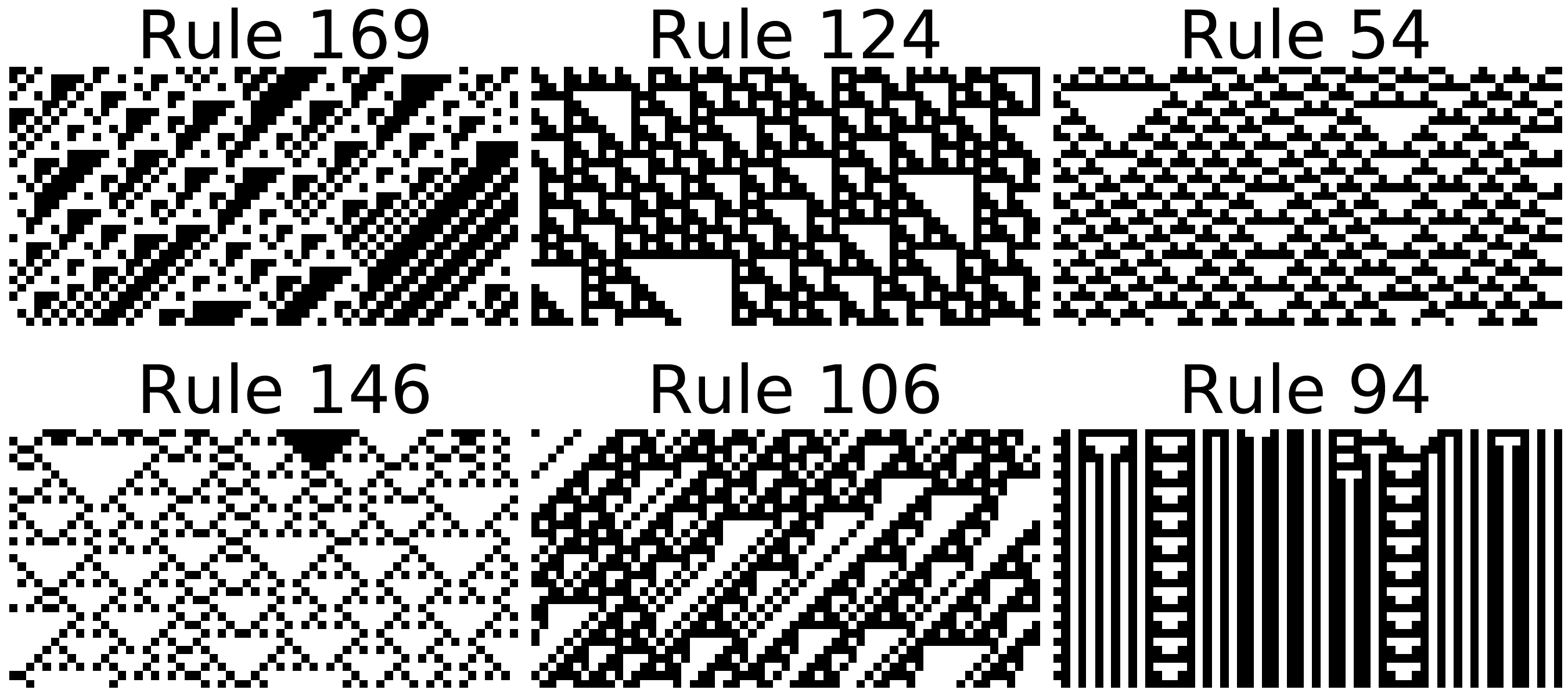}
    \caption{With condition on compressed length increase.}
    \label{fig:high_eca_joint+comp}
  \end{subfigure}

  \caption{Comparison of the raw joint compression score and the addition of a
    complexity increase condition. The high overlap in structures is not enough
    to get interesting rules a shown in \ref{fig:high_eca_joint}, but the
    addition of a complexity threshold allows to retrieve rules with complex but
    still structured behavior, as shown in \ref{fig:high_eca_joint+comp}.
    Figures are from the same slice of 60 cells over 30 timesteps taken from
    larger automata with random initial states. The top row corresponds to $t =
    0$ and time increases downwards.}
  \label{fig:joint_highest}
\end{figure}

Joint compression alone is not sufficient since it only selects rules that
either behave like identity or shift input because they maximize the
conservation of structures through time --- as illustrated in
Figure~\ref{fig:high_eca_joint}. However, one may combine the joint compression
score with another complexity measure to only select rules that exhibit some
disorder, or growth in complexity --- as Figure~\ref{fig:high_eca_joint+comp}
shows (the condition here was a threshold on the difference of compressed length
between initial and final states).

\subsection{Count-based predictor}\label{sec:count-based-pred}

A major issue with the joint compression metric is the fact that it is designed
to compress a linear stream of data. This is not ideal when considering higher
dimensional automata. Larger sets of transformations have to be
considered such as translations, rotations, flips, etc. Theoretically this
should not be a problem for a good enough linear compression algorithm, but
hardware and software limitations make it impractical to work with existing
algorithms on higher dimensional structures --- with e.g DEFLATE's upper limit
on dictionary size.

These higher dimensional automata might be better at generating complex
dynamics, and the large size of their rule spaces makes it a challenge to
explore. There has been at least one attempt to deal with these higher
dimensional systems \parencite{zenilTwodimensionalKolmogorovComplexity2015} that lacks the scalability
to work with large inputs.

An alternative to the linear compression-based method presented above would be
to use compressors optimized for n-dimensional data (e.g. PNG compression for 2D
automata) to take advantage of spatial correlation for compressing. However,
these compressors are rare for higher dimensional data, and are usually
optimized for one type of input --- e.g. images with PNG.

Another way to tackle the problem is to use a prediction based approach to
compression. Similarly to methods described in
\parencite{schmidhuberSequentialNeuralText1996} and one of the first steps of the PAQ
compression algorithm \parencite{mahoneyFastTextCompression2000}, we learn a statistical model of
input data to predict the content of a cell given its immediate predecessors.
For compression, this is often followed by an encoding step --- using Huffman or
arithmetic coding --- that encodes data which contains the least information
(least ``surprising'' data) with the most compact representation. This approach
can also be related to the texture synthesis method described in
\parencite{efrosTextureSynthesisNonparametric1999}, where the authors learn a non parametric model to
predict the next pixel of a texture given a previously synthesized neighborhood.
Additionally, because we do not need the operation to be reversible as in regular
compression, it is not necessary to limit the prediction model to making
prediction with predecessors only.

For a global state $S = (c_{1}, ... c_i, ..., c_{n})$, the neighborhood of cell
$i$ with radius $r$, denoted $n_{r,i}$ is defined as the tuple $n_{r,i} =
(c_{i-r}, ... c_{i-1}, c_{i+1} ..., c_{i+r})$ --- without the middle cell. The
goal of this method is to estimate the conditional probability distribution $p(s
| n_r)$ of the middle states at timestep $T$ given a neighborhood of radius $r$.
Assuming cell states given their neighborhood can be modeled by mutually
independent random variables, the log-probability of global state $S^{(T)}$ is
written
\begin{align}
  \log p(S^{(T)}) = \log \prod_{i=1}^N p(c_i | n_{r,i})  =
  \sum_{i=1}^N \log  p(c_i | n_{r,i})
\end{align}

If the automaton has a very ordered behavior, a model will predict with high
confidence the state of the middle cell given a particular neighborhood. On the
other hand, in the presence of maximal disorder, the middle cell will have an
equal probability of being in every state no matter the neighborhood. In the
latter case, a predictive model minimizing $-\log p(S^{(T)})$ would yield a high
negative log-probability.

A simple possible predictor for such purpose is a large lookup table that maps
all visited neighborhoods to a probability distribution over the states that the
middle cell can be in. State distributions for each neighborhood are obtained by
measuring the frequency of cell states given some observed neighborhoods. We
denote by $\Lambda$ this lookup table, defined for a window of radius $r$, which
maps all possible neighborhoods of size $2r + 1$ (ignoring the middle cell) to a
set of probabilities $p$ over the possible states $\{s_1, ..., s_n\}$, and $p$
can be written $[p_{s_1}, p_{s_2}, ... , p_{s_n}]$. $\Lambda$ is defined by
\begin{equation}
  \begin{aligned}
    \Lambda :&& \{s_1, ..., s_n\}^{2r} &\to&& \Delta_n\\
    && n_{r,i} &\mapsto&& p
  \end{aligned}
\end{equation}
where $\Delta_n$ denotes the probability simplex in dimension $n$.

\begin{figure}[htbp]
  \centering
  \includegraphics[width=.8\linewidth]{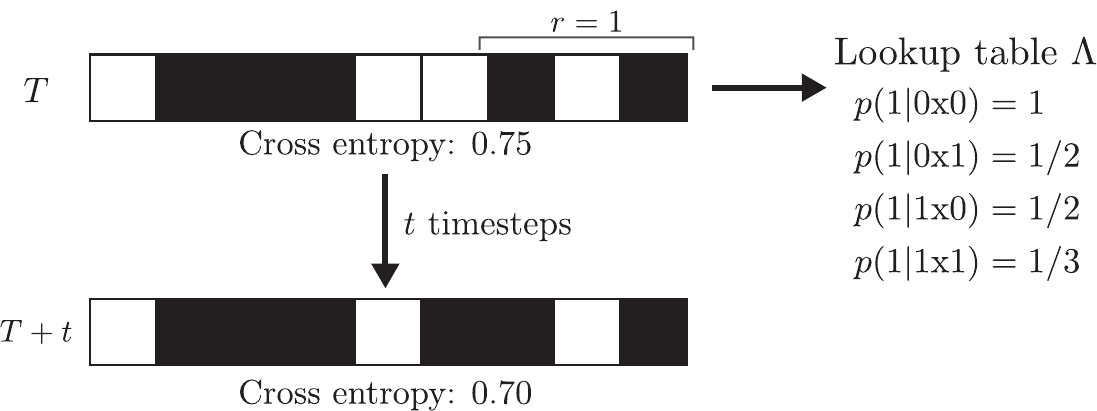}
  \caption{Count-based predictor method for a radius $r=1$. A
    frequency lookup table is computed from the global state at time $T$ by
    considering all neighborhoods with radius $r=1$ (3 consecutive cells
    but ignoring the middle cell). Cross-entropy with the automaton at time $T$
    quantifies the overall complexity. This can be compared to the cross-entropy
    at time $T + t$ for the amount of overlap.}
  \label{fig:schema-count}
\end{figure}

To measure the uncertainty of that predictor, we can compute the cross-entropy
loss between the data distribution it was trained on and its output. We compute
the log probability of the observed data given the model, or the quantity
\begin{align}
  L = - \frac{1}{N}\sum_{i=1}^N \sum_{k=1}^n \mathds{1}_{\{ s_k \}}(c_i)
  \log\Lambda(n_{r,i})_{s_k}
  \label{eq:loss_count}
\end{align}
where $\mathds{1}_{\{s_k\}}$ denotes the indicator function of the singleton set
$\{s_k\}$. An illustration of the counting process is represented in
Figure~\ref{fig:schema-count}. The quantity $L$ is minimal when the
$\Lambda(n_{r,i})_{s_k}$ is always equal to one, which means that the state of every cell
is entirely determined by its neighborhood.

\begin{figure}[htbp]
  \centering
  \begin{subfigure}[b]{.48\linewidth}
    \centering
    \includegraphics[width=\linewidth]{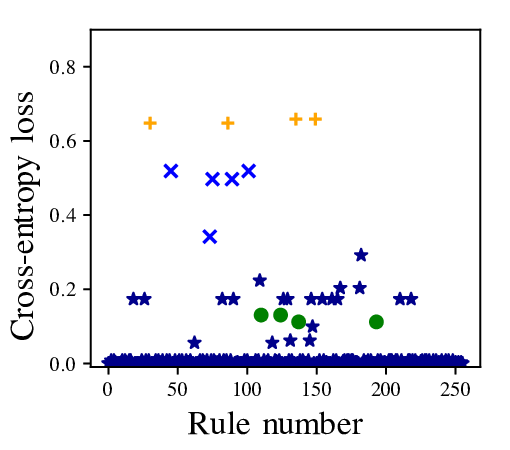}
    \caption{Count-based predictor}
      \label{subfig:cross_ent_one}
  \end{subfigure}
  \begin{subfigure}[b]{.4503\linewidth}
    \centering
    \includegraphics[width=\linewidth]{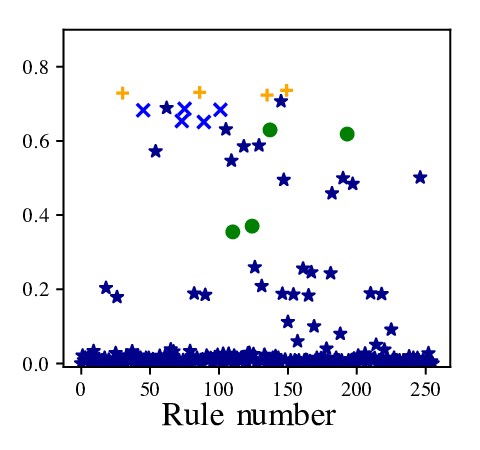}
    \caption{Neural network-based predictor}
      \label{subfig:nn_ent_one}
  \end{subfigure}

  \caption{Average cross entropy loss for the two predictor-based methods on all
    256 ECA. Rules are separated in several clusters. The
    count-based predictor (left plot) and neural network-based predictor (right
    plot) were applied with a neighborhood radius $r=1$ and $3$.}
\end{figure}

We apply this metric to all 256 ECA, with a window radius of size 3 (the 6
closest neighbors are used for prediction), and the same settings as for
Figure~\ref{subfig:comp_scores}. Cross-entropy loss of the lookup table gives
the results of Figure~\ref{subfig:cross_ent_one}. Colors are the same as in
Figure \ref{subfig:comp_scores} for comparison purposes.

We note the similarity between this plot and the one from Figure
\ref{subfig:comp_scores}, with a roughly equivalent resulting classification of
ECA rules, with the exception of rules with low score. Rules that produce highly
disordered patterns are on top of the plot whereas the very simply behaving
rules are at the bottom. This indicates coherence between the two metrics.

\subsection{Neural network based predictor}

\begin{figure}[htbp]
  \centering
  \includegraphics[width=.7\linewidth]{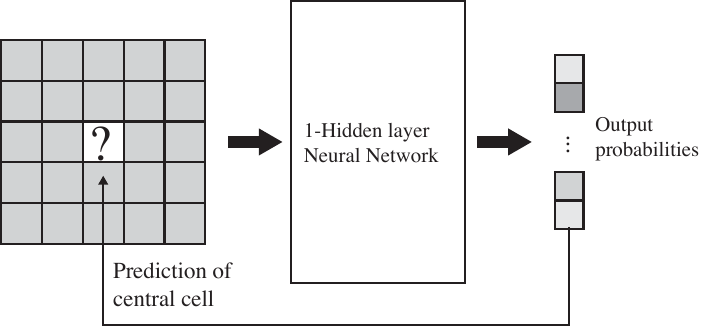}
  \caption{Neural network architecture for predicting a central cell given its
    neighbors. Output probabilities are defined for all possible states of the
    central cell.}
  \label{fig:nn_archi}
\end{figure}

The frequency-based predictor described above still has limitations:
\begin{itemize}
\item It does not take into account any redundancy in the input which may lead to
  suboptimal predictions (in a CA, very similar positions might have similar
  center cell state distribution, e.g. a glider in Game of Life should be
  recognized by the model no matter the rest of the neighborhood).
\item For the same reasons, when considering large window sizes, the number of
  possible neighborhood configurations grows much larger than the number of
  observed ones, leading to an input sparsity problem.
\end{itemize}
More sophisticated models can cope with above limitations by dealing with high-dimensional inputs without sparsity problems, and taking into account redundancy
of inputs and potential interactions between states for prediction.

We measure the cross-entropy loss of this simple model on the training set after
a standard learning procedure, which is the same for all rules. The procedure is
applied to a one-hidden-layer neural network with a fixed hidden-layer size. We
use a ReLU nonlinearity for the hidden layer and a softmax to obtain the output
probabilities.

For $n$ possible states ${s_1, ..., s_n}$, a cell in state $s_k$ is represented
as a vector of 0s of size $n$ with a 1 in position $k$. The input to the network
is the concatenation of these cell vectors for all cells in the neighborhood.
The output of the network is a vector of size $n$ with the output probability
for each state.

Gradient updates are computed during training to minimize the cross-entropy loss
between outputs and target examples. For a timestep $T$, we use the training
procedure in order to minimize with respect to $\theta$ the following quantity,
\begin{align}
  L_\theta^{(T)} = - \frac{1}{N}\sum_{i=1}^N \sum_{k=1}^n
  \mathds{1}_{\{ s_k \}}\left(c_i^{(T)}\right)
  \log\left[f_\theta\left(n_{r,i}^{(T)}\right)_{s_k}\right]
  \label{eq:loss_network}
\end{align}
where the neural network depending on parameter $\theta$ is denoted $f_\theta$,
and $n_{r,i}^{(T)}$, the neighborhood of cell $i$ with radius $r$ at time $T$ is
defined in the same way as in eq.~\eqref{eq:loss_count}. Loss is calculated with
respect to the testing set at time $T + \tau$ by computing the same quantity at
this subsequent timestep.

The training procedure is selected to achieve reasonable convergence of the loss
for the tested examples. It must be well defined and stay the same to allow for
comparison of the results across several rules. The score in timestep $T$ for a
delay $\tau$ is computed with the following formula
\begin{align}
  \mu_\tau = \dfrac{L^{(T)}}{L^{(T + \tau)}}
  \label{eq:main_metric}
\end{align}
where $L^{(T + \tau)}$ is the log probability of the automaton state at timestep
$T + \tau$ (defined in eq.~\eqref{eq:loss_network}) according to a model with
parameters learned during training at timestep $T$ and $L^{(T)}$ is the same as
in eq.~\eqref{eq:loss_network}. The value $\mu_\tau$ will be lower for easily
``learnable'' global states that do not translate well in future steps --- they
create more complexity or disorder --- thereby discarding slowly growing very
disordered structures. Higher values of $\mu_\tau$ correspond to automata that
have a disordered global state at time $T$ that can be transposed to future
timesteps relatively well. Those rules will tend to have interesting spatial
properties --- not trivially simple but not completely disordered because the
model transposes well --- as well as a large amount of overlap between a given
step and the future ones, indicating persistence of the spatial properties from
one state to another. We also selected the metric among other quantities
computed from $L^{(T)}$ and $L^{(T+\tau)}$ because it gave the best score in our experimental datasets.

\section{Experiments}\label{sec:experiments}

We carried out several experiments on a dataset of 500 randomly generated 3
states ($n=3$) rules with radius $r=1$. Those rules were manually annotated for
interestingness, defined as the presence of visually detectable non-trivial
structures. 
The annotation was carried out by five persons, using a majority rule to assign a binary label to each rule. 
The dataset contains 46 rules labeled as interesting and 454
uninteresting rules. Ranking those rules with the metrics introduced above
allows us to study the influence of parameters and the adequacy between
interestingness as we understand it and what the metric measures.

The task of finding interesting rules can be framed as either a classification
problem or a ranking problem with respect to the score we compute on the
dataset. The performance of our metric can be measured with the usual evaluation
metrics used on these problems, and, notably, the average precision (AP) of the
resulting classifier.

The average precision scores for the neural network and the count-based methods for time
windows of 5, 50, and 300 timesteps are given in Table~\ref{experiments_table}.
Scores were calculated in automata of size $256\times 256$ cells, ran for 1000
timesteps ($T + \tau = 1000$). Scores were computed for radii ranging from 1
cell (8 nearest neighbors) to 5 cells (120 neighbors), with a one-layer neural
network containing 10 hidden units trained for 30 epochs with batches
of 8 examples. The best AP for each time window is shown in bold.
Results for the frequency lookup table predictor are only shown for $r=1, 2$
because of the sparsity issues with the lookup table from $r=2$ and above, making it
unpractical to use the table --- $3^{24}$ possible entries for the lookup table
with $r=2$ against only $256^2$ observed states.

\begin{table}[t!]
  \renewcommand{\arraystretch}{1}
  \caption{Experimental results - AP scores}
  \label{experiments_table}
  \centering
  \begin{tabular}{c|c|c|c|c|c}
    \toprule
    \bfseries Neural network& $r=1$ &  2  & 3  & 4 & 5\\
    \midrule
    5 steps & 0.387 & 0.448 & \bfseries 0.541 & 0.525 & 0.534\\
    50 steps & 0.377 & 0.433 & 0.517 & 0.491 & \bfseries 0.542\\
    300 steps &0.358 & 0.454 & 0.488 & \bfseries 0.527 & 0.525\\
    \midrule
    \bfseries Lookup table& $r=1$ &  2  & & &\\
    \midrule
    5 steps & 0.092 & 0.070&&&\\
    50 steps & 0.102 & 0.070&&&\\
    300 steps & 0.093 & 0.069&&&\\
    \bottomrule
  \end{tabular}

  \begin{flushleft}{This table shows the average precision (AP)
      scores obtained on the dataset of section \ref{sec:experiments} with the neural network-based and lookup table-based methods. Results are shown for delays $\tau = 5, 50, 300$ and several radii values $r$.}\end{flushleft}
\end{table}

From these experiments, larger radii appear to perform slightly better, although
not in a radical way. Since the number of neighbors scales with the square of
the radius, reasonably small radii might be a good trade-off between performance
and computational cost of the metric.

We also study the performance of our metrics --- lookup table and neural
network-based --- as inputs of a binary classifier against two simple baselines
on a random 70/30 split of our dataset. The first baseline classifies all
examples as negative. The second baseline is based on compressed length as
defined in \parencite{zenilCompressionBasedInvestigationDynamical2010} and computed by choosing a pair
of thresholds that minimize mean square error when classifying examples in
between as positive --- this is based on the observation made in
Section~\ref{sec:compr-based-metr} that interesting rules have intermediate
compressed lengths. Results are in Table~\ref{experiments_table2} where only the
best radius is shown. The lookup table performs better than the baselines, but
the neural network gives the best score.

\begin{table}[t!]
  \renewcommand{\arraystretch}{1.2}
  \centering
  \begin{tabular}{lm{.1\linewidth}m{.19\linewidth}m{.1\linewidth}m{.11\linewidth}}
    \toprule
    \bfseries  Metric & Baseline & Compressed length
    \parencite{zenilCompressionBasedInvestigationDynamical2010} &
    Lookup Table & Neural network \\ \midrule
    \bfseries Accuracy & 0.90 & 0.913 & 0.926 & \bfseries 0.953 \\ \bottomrule
  \end{tabular}
  \\[+8pt]
  \caption{Experimental results - Accuracy of each metric of complexity when
    used to classify which automatons do evolve interestingly, compared against
    the trivial all-negative baseline and the compressed length
    metric~\parencite{zenilCompressionBasedInvestigationDynamical2010}.}
  \label{experiments_table2}
\end{table}

Above experiments demonstrate the capability of our proposed metric to match a
subjective notion of interestingness of our labeling. For instance, the top 5
and top 10 scoring rules of the best performing configuration ($r=3$, $\tau =
5$) are all labeled as interesting, and top 100 scores contain 41 of the 46
rules labeled as interesting.

\section{Discussion}

In this section, we discuss the results obtained by using the metric of
\eqref{eq:main_metric} and the way they can be interpreted.

\paragraph{One dimensional cellular automata}
By applying the metric on the same example as before, we again obtain a
plot with a rule classification that matches a visual appreciation of
complexity of 1D CA. Results are shown on Figure~\ref{subfig:nn_ent_one}.
Similarly to the previous cases, rules we might label as interesting are
unlikely to be either at the top or bottom of the plot.

\paragraph{Two dimensional cellular automata}

Simulations conducted with 2D CA used grids of size 256$\times$256. Automata
were ran for 1000 steps (the metric is measured with respect to the reference
time $T = 1000$). Rules are defined with a table of transitions from all
possible neighborhood configurations with radius $r=1$ (3$\times$3 squares) to a
new state for the central cell. Unbiased sampling of rules, obtained by
uniformly sampling the resulting state for each transition independently,
overwhelmingly produces rules with a similar amount of transitions towards each
state and fails to produce rules without completely disordered behavior more
than 99\% of the time.

Therefore, we adopt a biased sampling strategy of the rules, selecting the
proportion of transitions towards each state uniformly on the simplex --- e.g
for 3 states we might get the triple $(0.1, 0.5, 0.4)$ and sample transitions
according to these proportions. This parametrization can be related to Langton's
lambda parameter \parencite{langtonComputationEdgeChaos1990} that takes into account the
proportion of transitions towards a transient (inactive) state and all the other
states. We obtain approximately 10\% interesting rules with this sampling as the
proportions of our experimental dataset show.

\begin{figure}[t]
  \centering
  \subfloat{
    \includegraphics[width=.38\linewidth]{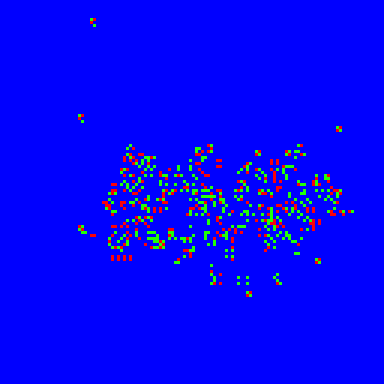}
  }\hfil
  \subfloat{
    \includegraphics[width=.38\linewidth]{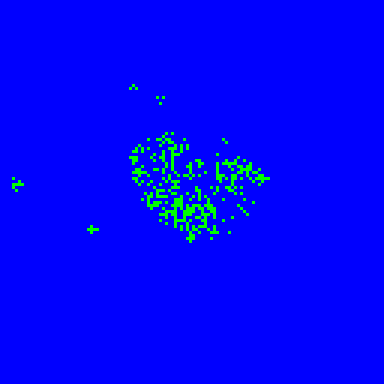}
  }
  \caption{Rules with 3 states that have spontaneously occurring glider
    structures. The gliders are the small structures that are outside of the
    center disordered zone. Some of them move along the diagonals while some
    others follow horizontal or vertical paths. Note that some repeating
    patterns occur also in the more disordered center zone.}
  \label{fig:gliders}
\end{figure}

\begin{figure}[t]
  \centering
  \subfloat{
    \includegraphics[width=.38\linewidth]{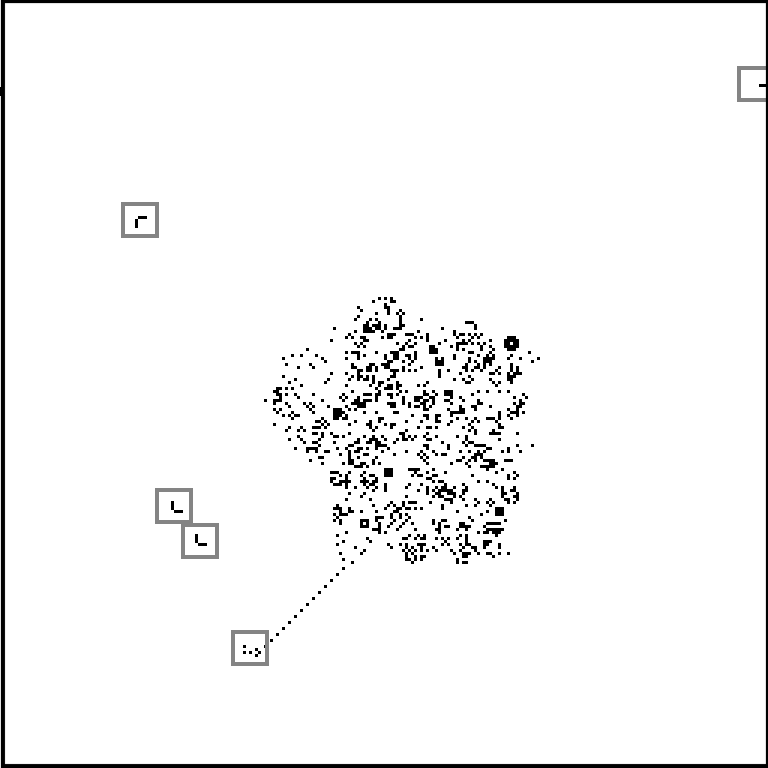}
  }\hfil
  \subfloat{
    \includegraphics[width=.38\linewidth]{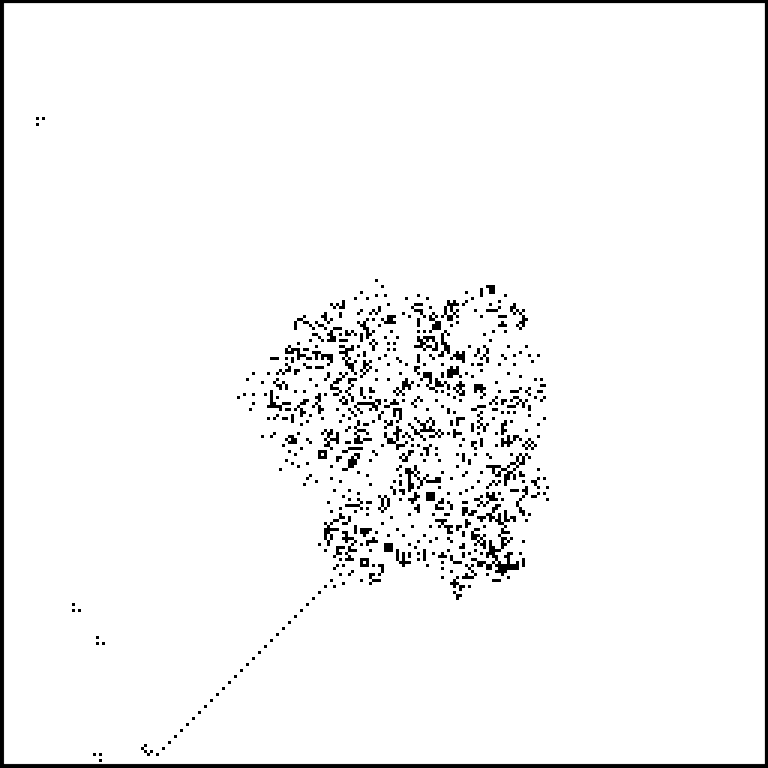}
  }\hfil\setcounter{subfigure}{0} 
  \subfloat[Timestep $T$]{
    \includegraphics[width=.38\linewidth]{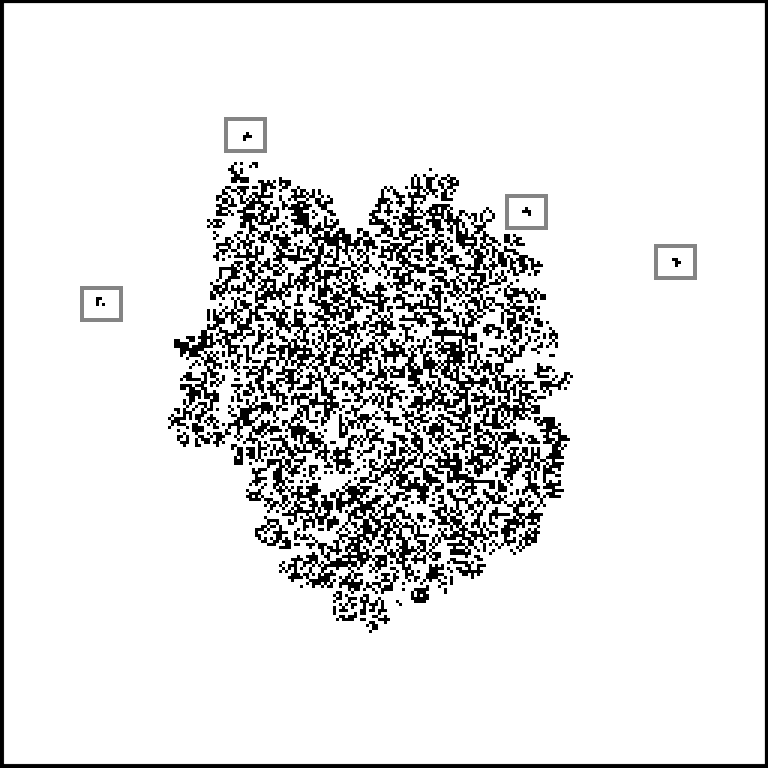}
  }\hfil
  \subfloat[Timestep $T + 50$]{
    \includegraphics[width=.38\linewidth]{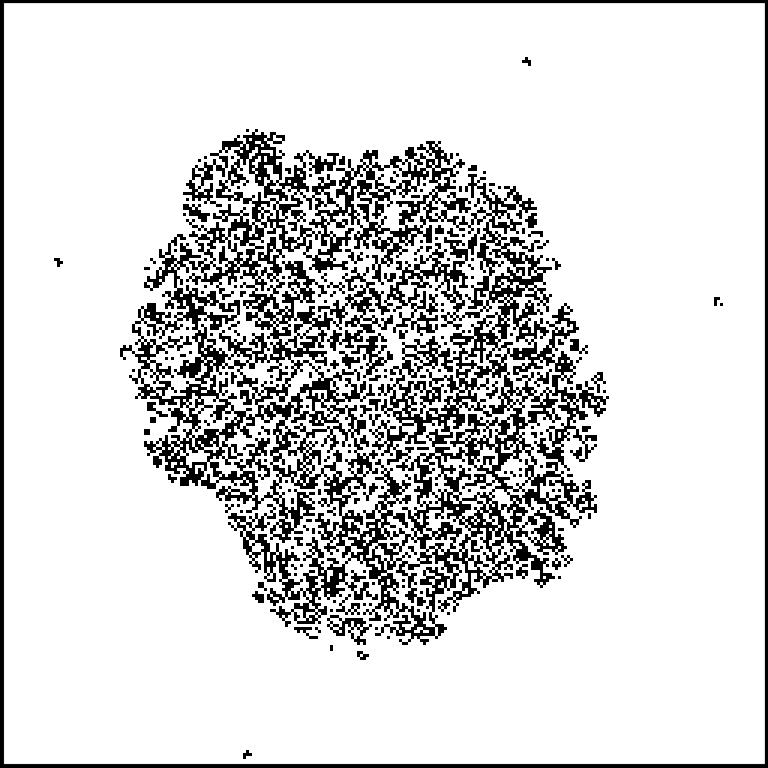}
  }
  \caption{Spontaneous glider formation and evolution is observed for some high
    scoring 2 states rules. Each row corresponds to a rule, with a 50 timesteps
    difference between the two columns. Gliders are marked with a gray square.
    Runs were initialized with a small 20 by 20 disordered square (uniformly
    sampled among possible configuration) in the center simulated for up to 400
    steps.}
  \label{fig:automat_glider}
\end{figure}

Using the neural network-based complexity metric, we were able to find rules
with interesting behavior among a very large set through random sampling. Some
of these rules are shown here. Figure~\ref{fig:automat_glider} displays three 2D
rules that were selected manually upon visual inspection among the 20 highest
scoring for metric $\mu_{50}$ (defined in eq.~\eqref{eq:main_metric}) of a sample
of 1700 randomly generated 2-states 3 by 3 neighborhood rules. For comparison,
Conway's Game of Life rule (GoL) ranks in the top 1\% of the 2500 rules
mentioned above for runs that do not end in a static global state. We observe
that spontaneous glider formation events appear to be captured by our metric.
Although gliders in cellular automata are a simple process that can be created manually,
detection of their spontaneous emergence within a random search setting
is a first step towards finding more complex macro structures that can emerge
out of simple components. Rules with low scores are overwhelmingly of the
disordered kind.

\begin{figure}[t]
  \centering
  \subfloat{
    \includegraphics[width=.38\linewidth]{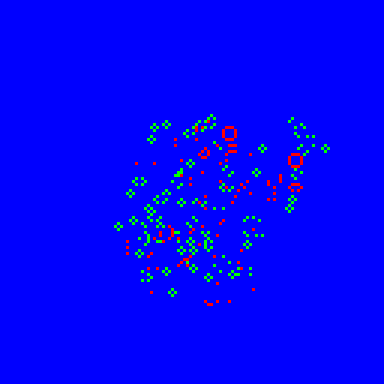}
  }
  \hfil
  \subfloat{
    \includegraphics[width=.38\linewidth]{figures/micro4}
  }
  \caption{Rules with 3 states that generate cell-like interacting structures.
    These patterns are either static or moving and can interact with one another
    to generate copies of themselves and other patterns. Note the very similar
    micro-structures that are repeated at several places in the space.}
  \label{fig:micro}
\end{figure}

\begin{figure}
  \centering
  \subfloat{
    \includegraphics[width=.38\linewidth]{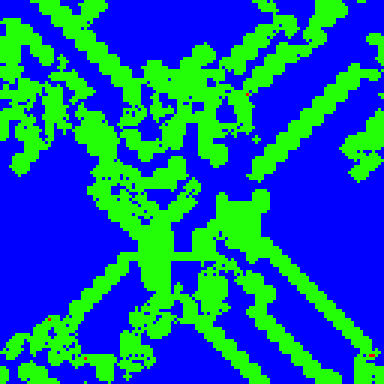}
  }
  \hfil
  \subfloat{
    \includegraphics[width=.38\linewidth]{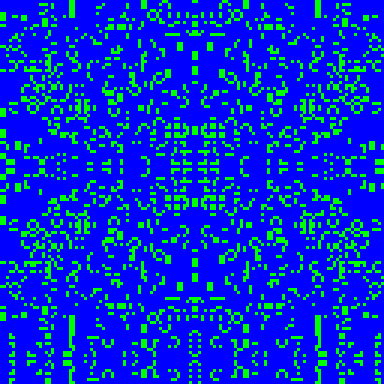}
  }
  \caption{Rules with surprising behaviors that are highly structured but
    complex. Those rules were selected among high-ranking rules for the
    neural-network based complexity metric. They all exhibit structurally non
    trivial behavior.}
  \label{fig:odd}
\end{figure}

Figures~\ref{fig:gliders}, \ref{fig:micro} and \ref{fig:odd} show some three
states rules that were selected through random sampling on the simplex with the
neural network-based metric. They were selected among the 30 highest scoring
rules out of 2500 randomly selected 3 states rules. All of their behaviors involve
the growth and interaction of some small structures made of elementary cells.

All automata were initialized with a random disordered square of 20 by 20 cells
in the center. In the Figures mentioned above, colors were normalized with the
most common state set to blue. Figure~\ref{fig:gliders} shows rules that
spontaneously emit gliders that go through space in a direction until they
interact with some other active part of the automaton. Figure~\ref{fig:micro}
shows rules that generate small structures of between four and thirty cells that
are relatively stable and interact with each other. These elementary components
could be a basis for the spontaneous construction of more complex and bigger
components. Figure~\ref{fig:odd} shows some other rules from this set of high
ranking automata. They highlight the wide range of behaviors that can be
obtained with these systems. Interesting rules from our search process can be
found, along with other examples, in the form of animated
GIFs\footnote{\url{https://bit.ly/interesting_automata}}.

For some of these rules interesting patterns appear less frequently in smaller
grids, indicating that the size of the space might impact the ability to
generate complex macro-structures. Increasing the size of the state space to
very large grids might therefore make it easier to generate very complex patterns.

\section{Conclusion}\label{sec:conclusion}

In this chapter, we have proposed compression-inspired metrics for measuring a type
of complexity occurring in complex systems. We demonstrated its usefulness
for selecting CA rules that generate interesting emergent structures from very
large sets of possible rules. In higher dimensions where linear compression, as in gzip,
is not sufficient to find complex patterns, our metric is also
useful.

We study 2 and 3 states automata here and we plan to investigate the effects of
additional states or larger neighborhoods on the ability to evolve more
structures and obtain more interesting behaviors.

The dataset and code to reproduce our experiments and improvement on the reported results
 are published on
GitHub\footnote{\url{https://github.com/hugcis/evolving-structures-in-complex-systems}}.
The metrics we introduce in this chapter could be used to design organized
systems of artificial developing organisms that grow in complexity through an
evolutionary mechanism. A possible path toward such systems could start by
creating an environment where computational resource allocation favors the
fraction of subsystems that perform the best according to our measure of
complexity.

The proposed metric is theoretically applicable to any complex system where a
notion of state of an elementary component and locality can be defined. With
these requirements satisfied, we can build a similar prediction model that uses
information about local neighbors to predict the state of a component, and
thereby assess the structural complexity of an input.

We believe that the capability of creating evolving systems out of such
elementary components and with few assumptions could be a step towards AGI. By
devising ways to guide this evolution in a direction that we find useful, we would be
able to find an efficient solution to hard problems while retaining the adaptability of
the system. It might be suitable to avoid over-specialization that can happen in
systems designed to solve a particular task --- e.g. reinforcement learning
algorithms that can play games, and supervised learning --- by staying away from
any sort of objective function to optimize and by leaving room for open-ended
evolution.

\chapter{Visualization of computations in large-scale complex systems}
\label{cha:visu-comp-large}

Emergent processes in complex systems such as cellular automata can perform
computations of increasing complexity, and could possibly lead to artificial
evolution. Such a feat would require scaling up current simulation sizes to
allow for enough computational capacity. Understanding complex computations
happening in cellular automata and other systems capable of emergence poses many
challenges, especially in large-scale systems. We propose methods for
coarse-graining cellular automata based on frequency analysis of cell states,
clustering and autoencoders. These innovative techniques facilitate the
discovery of large-scale structure formation and complexity analysis in those
systems. They emphasize interesting behaviors in elementary cellular automata
while filtering out background patterns. Moreover, our methods reduce large 2D
automata to smaller sizes and enable identifying systems that behave
interestingly at multiple scales.

\section{Introduction}
Cellular automata (CA) have been extensively studied since the 1960s. Originally
designed and studied to create artificial evolution from self-replication
\parencite{vonneumannTheorySelfreproducingAutomata1966,
  langtonSelfreproductionCellularAutomata1984}, previously studied cellular
automata simulations were often of relatively modest sizes. Only specific rules
with repetitive or predictable dynamics such as John Conway's Game of Life
\parencite{gardnerMathematicalGames1970} have been scaled up to larger grid
sizes ($10^4 \times 10^4$ or more cells).

For complex phenomena such as artificial evolution to exist and be open-ended
within those simulated worlds, there needs to be sufficient ``capacity'' --- a
large enough state-space. In nature, complex and significantly different
dynamics often arise from uniform laws at a smaller
scale~\parencite{andersonMoreDifferent1972}. It seems unlikely that such complex
processes, like artificial evolution, could happen in too small CAs because
higher order dynamics do not have enough capacity to emerge. However, several
issues arise when scaling CAs to large sizes:

\begin{itemize}
\item Time complexity rapidly becomes a bottleneck. Updating a large number of
  cells is costly. Tricks such as caching of some of the computations can help,
  but do not always improve performance
  significantly~\parencite{gosperExploitingRegularitiesLarge1984}.

\item Memory complexity can also become an issue when dealing with numerous
  states, and especially grids in 3 dimensions and more. In that case, even the
  underlying rule of the system cannot be stored within reasonable memory
  capacity.

\item Visual inspection of these large grids is infeasible. Studying CA
  complexity is rendered difficult by the highly variable nature of emergent
  processes. It is especially the case for large-scale systems.

\end{itemize}
When working with such large systems, it is less relevant to focus on the local
behaviors at the single cell level. This is similar to other complex systems
like the weather, in which behaviors of individual atoms in a cloud are
irrelevant to large-scale air mass movements. Much richer behaviors can be
observed from studying large patterns' formation and their evolution. This
should also hold true for CAs; we further discuss this question in
\nameref{sec:conclusion-vc}.

\begin{figure}[th]
  \centering
  \includegraphics[width=.93\linewidth]{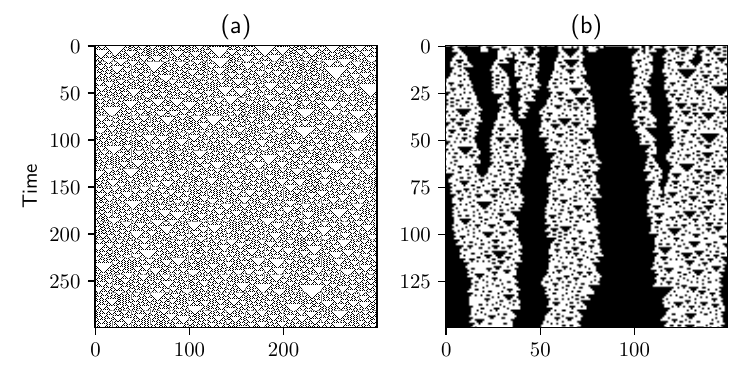}
  \caption{\label{fig:rule18_small} Hidden structures in rule 18 are uncovered
    by filtering the space-time diagram with our frequency histogram-based
    method. \textbf{(a)} shows 300 timesteps of a randomly initialized rule 18
    simulation. Notice the complex structures made visible in \textbf{(b)} with
    our method.}
\end{figure}

In this chapter, we investigate techniques that can help us visualize large
space-time diagrams of CAs. We demonstrate that simple clustering and
coarse-graining techniques can be used in order to perceive structures that
cannot emerge on smaller grids. This is also useful for disordered cellular
automata with hidden structures as is the case for the elementary cellular
automaton rule 18, illustrated in Figure~\ref{fig:rule18_small} --- more details
in~\nameref{sec:results}.

Reducing large grids to smaller sizes while preserving interesting behaviors
such as pattern formation is essential to apply to these CAs complexity metrics
designed to work on modestly sized
grids~\parencite{grassbergerQuantitativeTheorySelfgenerated1986,
  zenilCompressionBasedInvestigationDynamical2010,
  soler-toscanoCalculatingKolmogorovComplexity2014,
  zenilTwodimensionalKolmogorovComplexity2015}. Common metrics of complexity are
often limited by the number of components in the systems (number of cells in a
CA grid, timesteps, etc.) or may not be effective when small-scale patterns are
less relevant than large-scale ones.

\section{Related work}

Previous work on coarse-graining cellular automata focused either on conserving
the main computational properties of CA rules through exact coarse-graining or
on filtering interesting behaviors without reducing the amount of computations.
Our work both highlights interesting behaviors and compresses the
representation, which we argue are necessary to study complexity in large
cellular automata.

\subsection{Coarse-graining in cellular automata}

Coarse-graining is an approximation procedure used to speed up computations in
systems made of many components. It originated
in~\parencite{levittComputerSimulationProtein1975} and is now widely used in physics
to model complex systems at various granularity levels and is successful at
modeling bio-molecules \parencite{potoyanRecentSuccessesCoarsegrained2013,
  ingolfssonPowerCoarseGraining2014, kmiecikCoarseGrainedProteinModels2016}.

The exact coarse-graining of elementary cellular automata (ECA) has been
investigated extensively
in~\parencite{israeliComputationalIrreducibilityPredictability2004,
  israeliCoarsegrainingCellularAutomata2006}. Authors found ways of rewriting
one-dimensional CA rules into each other through coarse-graining of the
transition rule. They built a graph of equivalence of all 256 ECA and identified
some rules that do not admit any computational reduction. This indicates that
some cellular automata are accomplishing fundamentally more computations than
others.

\subsection{Filtering}

Filtering cellular automata (CA) was introduced to reduce a CA's behavior to its
most relevant parts. The goal is to extract relevant irregularities from a CA's
space-time diagram. Seminal work
by~\parencite{hansonAttractorbasinPortraitCellular1992,
  hansonComputationalMechanicsCellular1997} formalized the notion of domains and
coherent structures in cellular automata. They used a set of regular languages to
represent cellular automata dynamics and extract relevant behaviors such as
discontinuities between regular domains or ``particles''.
Figure~\ref{fig:rule110} shows a filtering example for cellular automaton rule
110 --- in Wolfram's numbering.

A filtering method similar to our proposed frequency-based coarse-graining ---
originally presented as a complexity metric for cellular automata --- is
introduced in~\parencite{wuenscheClassifyingCellularAutomata1999}. The author
proposes to progressively filter out cells in cellular automata's space-time
diagrams according to read frequency of the rule table. Cells that originated
from frequent rule table lookups are set to a quiescent or null state. The
choice of threshold has to be decided by a user for each rule. Another notable
difference is the method aims at making visualization of gliders easier without
reducing the size of the grid or making more compact representations.

More recent work by~\parencite{shaliziAutomaticFiltersDetection2006} uses the
combination of a modified Lyapunov exponent approach with \emph{statistical
  complexity} \parencite{shaliziQuantifyingSelfOrganizationOptimal2004} to underline
complex behaviors. However, the first method requires repeated perturbations and
simulations of the system to study its sensitivity.

\subsection{Scaling-up cellular automata}

Hashlife \parencite{gosperExploitingRegularitiesLarge1984} and other Game of
Life-specific optimizations enable simulating a large number of cells for
numerous timesteps. Nonetheless, these algorithms essentially exploit input
redundancy. The regularity in patterns allowing such optimizations might
indicate a lack of novel patterns being generated by the system.

This also means that Game of Life-based simulations are computationally
reducible to a much simpler system, indicating that its computations are
inefficient~\parencite{wolframNewKindScience2002}. An optimally complex-behaving
computational model should be impossible to predict except when computing its
actual evolution step by step.

In the following, we used coarse-graining as a method for scaling down CAs in
both time and space in order to make visualization of larger patterns and
complex behaviors easier. The underlying fine-scale computations may be
essential for these larger patterns to appear, hence the necessity to keep them.
However, analogous to many natural processes (swarms, chemistry, cells in an
organism, DNA), interesting behaviors might not be observable at the level of
individual components --- or small groups of components (individuals, single
cells or molecules in the examples above). We view coarse-graining as a way to
reduce a cellular automaton's space-time diagram to its most relevant parts
while keeping primary dynamics in the background. The resulting diagram would
ideally be an irreducible system.

\section{Proposed coarse-graining of cellular automata}

For reasons stated above, we introduce coarse-graining methods for cellular
automata that are not reversible --- information is discarded in the process.
This process does not attempt to find shortcuts for the computations of a
cellular automaton, but rather to select relevant parts of the space-time
diagram and discard information irrelevant to the core behavior. For example, a
standard glider in Game of Life spanning $3\times 3$ cells could be replaced
with a single cell moving diagonally when coarse-graining by a factor 3. This is
because the actual oscillator's dynamics might not be relevant at this coarser
scale.

Coarse-graining is akin to constructing \emph{supercells} from blocks of individual
cells. These supercells are assigned a new state and form a coarser partitioning
of the initial grid which can be studied as its own system. In particular,
complexity metrics or further coarse-graining can be applied to this new grid.

\subsection{Frequency histogram coarse-graining}\label{sec:simple-hier-coarse}

A simple coarse-graining is achieved by mapping blocks to a single
\emph{supercell} state according to the probability of this configuration
appearing, given a previously constructed model. The easiest way to think of it
is with a simple frequency counting model of the distribution of $2\times 2$
blocks in a 2D CA\@. For a 2-state automaton, there are 16 possible supercell
configurations. The simplest model for the occurrence of these blocks is their
empirical frequency. Let us consider a CA with $N$ blocks of $2\times 2$ cells,
let $S^{(in)} = \{\mathtt{0000}, \mathtt{0001}, \mathtt{0010}, \ldots,
\mathtt{1111}\}$ be the set of $2\times 2$ blocks and $s_i \in S^{(in)}$ be a
given supercell. The probability $p_i$ of observing supercell $i$ on a grid $G$
is estimated with
\begin{equation}
  p_i = \dfrac{\text{count}_G(s_i)}{\sum_{s_j\in S^{(in)}}\text{count}_G(s_j)}
  \label{eq:stat_est}
\end{equation}
where $\text{count}_G(s_i)$ is the number of blocks matching $(s_i)$ in $G$.

Supercells can then be assigned a particular state. We call the corresponding
mapping $f: S^{(in)} \mapsto S^{(out)}$. $S^{(out)}$ can be chosen depending on
the desired output or use. For instance, with $S^{(out)} = \{0, 1\}$ we can
define $f$ to map each supercell $s_i$ as follows:
\begin{align}
  f(i) = \begin{cases}
    \mathtt{0} &\quad\text{if }\ p_i\geq \alpha\\
    \mathtt{1} &\quad\text{if }\ p_i< \alpha
  \end{cases}
        \label{eq:alpha}
\end{align}
where $\alpha$ is a chosen threshold.

\subsubsection{Partitioning the histogram}
This method can be understood as partitioning the histogram of supercell
frequency. In equation~\eqref{eq:alpha}, supercells with low probability
--- with higher self-information --- are mapped to state \texttt{1} whereas
commonly occurring states are mapped to \texttt{0}.

Choosing a partition of the histogram is equivalent to selecting a suitable
$\alpha$ --- scalar for two output states, or vector $\mathbf{\alpha} =
(\alpha_1, \ldots, \alpha_n)$ for $n$ output states. Therefore, one can map
supercells to any number of target states (three or more) by partitioning the
frequency histogram into any number of bins. Supercell distribution can be
anything between uniform and very unbalanced, with a few supercells being
overwhelmingly represented (background) and only a few occurrences of other
configurations. The chosen partitioning has to deal with both situations equally
well. In the following, we use a uniform partitioning of the area under the
negative log-histogram for elementary cellular automata --- supercells are
divided into two bins of equal summed negative logarithmic probability. For 2D
CAs, we use the same method but with quadratic partition of the histogram
($1/k^2$ instead of $1/k$, with $k$ the number of output states, chosen because
of better visual results).

\subsubsection{Dithering}\label{sec:dithering}
\begin{figure}[th]
  \centering
  \begin{subfigure}{.32\linewidth}
    \fbox{\centering
    \includegraphics[width=\linewidth]{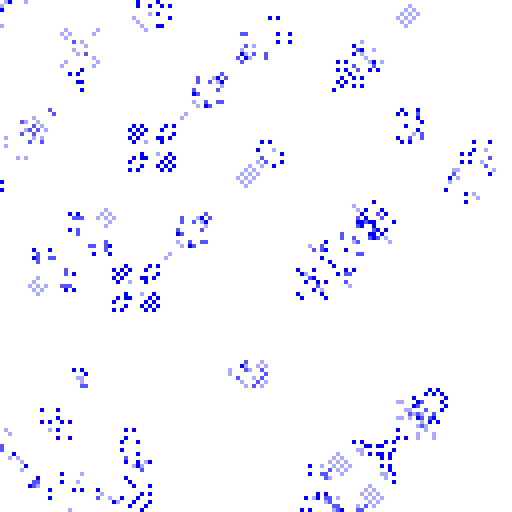}}
    \caption{\label{subfig:normal}Original CA}
  \end{subfigure}
  \hfill
  \begin{subfigure}{.32\linewidth}
    \fbox{\centering
    \includegraphics[width=\linewidth]{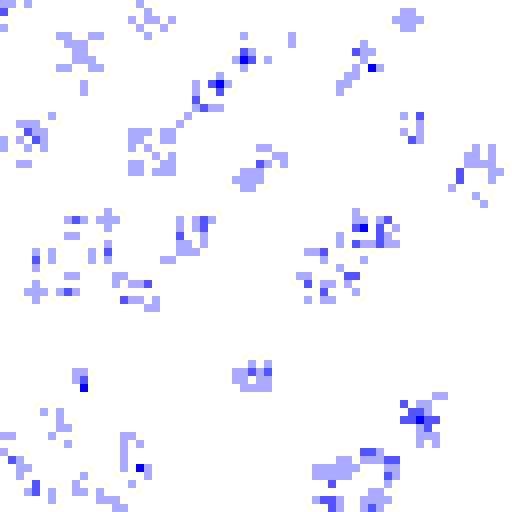}}
    \caption{\label{subfig:no-dithering}w/out dithering}
  \end{subfigure}
  \hfill
  \begin{subfigure}{.32\linewidth}
    \fbox{\centering
    \includegraphics[width=\linewidth]{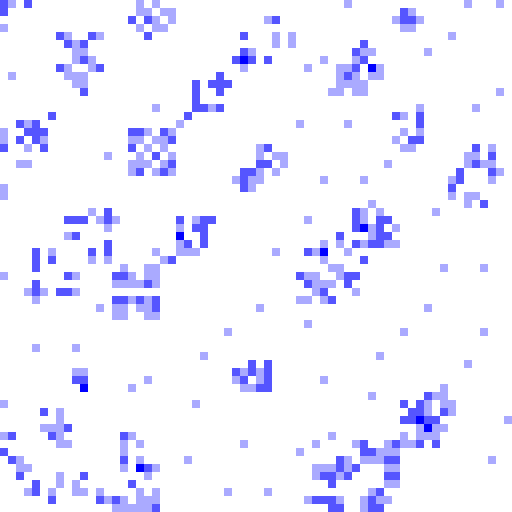}}
    \caption{\label{subfig:dithering}with dithering}
  \end{subfigure}

  \caption{\label{fig:close-up} Close-up view of coarse-graining effects on a
    4-states CA rule (1 shade of blue per state). Both coarse-graining methods
    conserve many of the interesting structures. Dithering introduces additional
    artifacts on regular backgrounds. Fig.~\ref{subfig:normal} shows actual
    states in the CA simulation on a $128 \times 128$ grid.
    Fig.~\ref{subfig:no-dithering} is a coarse-grained version
    of~\ref{subfig:normal} with histogram coarse-graining, the grid is
    $64 \times 64$ cells. \ref{subfig:dithering} is obtained with histogram
    coarse-graining and dithering (see \nameref{sec:dithering}).}
\end{figure}

Histogram partitioning introduces another set of parameters to be manually
tuned, adding complexity to the procedure. An alternative way to produce an
output image from the histogram is to use dithering. Dithering is an image
processing technique commonly used to reduce large visual artifacts induced by
quantization errors. Noise is added to the image during the quantization process
to make the average local value of a set of pixels as close to their target
continuous value as possible. The resulting image is created so as to match
target continuous values with discrete values only --- cell states in the grid.
It can be seen as another way of partitioning the histogram with variable
thresholds that depend on a running quantization error.
Figure~\ref{fig:close-up} shows a comparison of dithering and regular histogram
partitioning (Floyd–Steinberg's algorithm was used
\parencite{floydAdaptiveAlgorithmSpatial1976}).

\setlength{\fboxsep}{0pt}

\subsubsection{Visualization}
\begin{figure}[th]
  \centering
  \begin{subfigure}{.40\linewidth}
    \fbox{
    \centering
    \includegraphics[width=\linewidth]{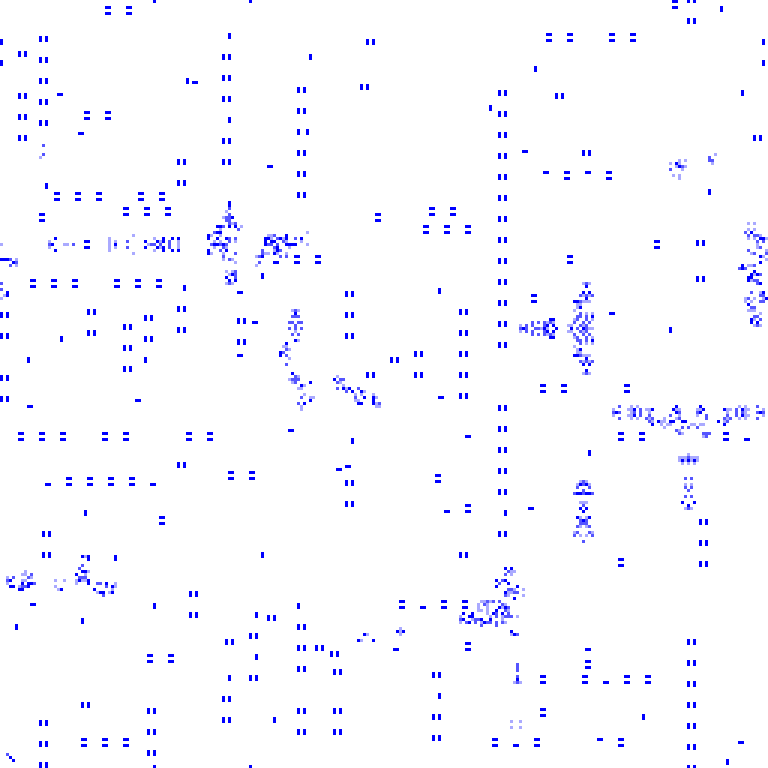}}
    \caption{\label{subfig:gol}Base grid}
  \end{subfigure}
  \hspace{10pt}
  \begin{subfigure}{.40\linewidth}
    \fbox{
    \centering
    \includegraphics[width=\linewidth]{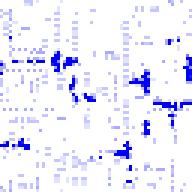}}
    \caption{\label{subfig:gol_cg}Coarse-grained}
  \end{subfigure}

  \caption{\label{fig:gol_comparison} Side-to-side comparison of a CA simulation
    and its coarse-grained version. The first simulation is $256 \times 256$
    cells and the second has been coarse-grained to $64\times 64$. Notice the
    interesting patterns on Figure~\ref{subfig:gol} are hardly distinguishable.
    They are highlighted by histogram-based coarse-graining in
    Figure~\ref{subfig:gol_cg}.}
\end{figure}

One advantage of this frequency histogram-based method is that it naturally
highlights rarer events in the simulation grid, creating a ``heatmap'' of the
simulation's activity. Since we sort supercells according to their observed
frequency, the right choice of colors --- e.g.\ progressively darker gradient ---
can lead to automatic highlighting of active regions of a cellular automaton.
Figure~\ref{fig:gol_comparison} shows the same simulation both unprocessed and
downscaled by a factor of 4 with coarse-graining. Although much coarser,
Figure~\ref{subfig:gol_cg} is more readable than the base version, which is
helpful when dealing with large grids\footnote{Several figures in this chapter
  have animated versions, accessible at the project page of our paper
  \parencite{cisnerosVisualizingComputationLargescale2020} \projecturl}.

\subsubsection{Hierarchical coarse-graining}
The above procedure can be applied recursively to the same cellular automaton or
with larger block sizes to get a progressively coarser representation. Since
information is systematically discarded in the process, it cannot be applied any
number of times. For this reason, many 2D CAs exhibiting interesting behaviors at
the micro-level but not at the macro-level have no remaining visible structure
after reducing their scale several times with this method.

Because a simple model like frequency counting can be estimated quickly,
hierarchical coarse-graining is easily applied to large grids, reducing the size
by a factor of $n$ (block size) every time. For instance, this property makes it
suitable to search the cellular automata rule space for CAs behaving
interestingly at multiple coarse-graining levels simultaneously.

\subsection{Clustering}

Another way to convert blocks of cells for coarse-graining is to distribute
these blocks into a small number of clusters, where each group becomes the new
coarse state.

Several distance functions may apply here, the most natural of which being
Hamming distance, which measures how many states differ between two
positions~\parencite{hammingErrorDetectingError1950}. It is defined for two strings
of equal length $n$, $s_1 = [s_{(1, 1)}, \ldots, s_{(1,n)}]$ and $s_2 =
[s_{(2,1)}, \ldots, s_{(2,n)}]$, as the number of positions where the two
strings differ:
\begin{align}
  \sum_{k = 1}^n \mathds{1}\left\{\ s_1^{(k)}  \neq s_2^{(k)}\right\}.
\end{align}

A supercell of $N\times N$ cells of a CA can be converted into a string to be
compared to other blocks with the Hamming distance. For CAs, we limit ourselves
to strings of digits representing states, i.e. $s_{(i,j)} \in \mathbb{N}$. We
use a vanilla implementation of the K-means algorithm where clusters' centers
are computed using a continuous average of position vectors rounded to nearest
integer values. Clusters are initialized with randomly selected observations.

\subsection{Autoencoders for coarse-graining}

Instead of just relying on the amount of information of a given supercell's
configuration, one can also try to automatically find a relevant representation
with dimensionality reduction methods. Autoencoders are neural networks composed
of an encoder part and a decoder part, originally designed to identify principal
components of a collection of data
points~\parencite{baldiNeuralNetworksPrincipal1989,
  hintonConnectionistLearningProcedures1989,
  kramerNonlinearPrincipalComponent1991}. An encoder neural network converts
data to a \emph{latent} vector of smaller dimension than the original input.
Then, a decoder neural network reconstructs a vector with the same dimension as
the input from this encoded \emph{latent} representation.These models can
automatically find an optimal constrained representation through minimizing a
reconstruction loss between the original input and the reconstructed output.

We denote the encoder network with $E$ and the decoder network with $D$. We frame
the reconstruction problem as a $N$ class classification problem with multiple
components --- one class per input state, one component for each cell of the $K$
cells in a block. The reconstruction loss is the component-wise cross-entropy
between the state of each input cell and the reconstructed state after $D \circ
E$ is applied.

\begin{figure}[th]
  \centering
  \includegraphics[width=.8\linewidth]{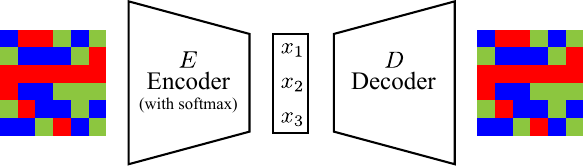}
  \caption{\label{fig:autoencoder} Diagram of the autoencoder architecture used
    for coarse-graining. A block of $6\times 6$ cells is encoded in a vector of
    fixed dimension. There are 3 components in the example. They can either
    represent a RGB color or a 3 states smaller automaton.}
\end{figure}

Figure~\ref{fig:autoencoder} illustrates the autoencoder layout for
coarse-graining. By adjusting the block size and dimension of the encoded
vector, one can influence the amount of information conserved during encoding.
Naturally, smaller blocks will be more easily represented in lower dimension.

The encoder has a softmax layer to ensure the coded state's components sum to
one. Therefore, one can view this coded supercell as a mixture of states which
can either be kept as is or converted to a discrete state by keeping the maximal
component only. They are trained with stochastic gradient descent until
convergence.

\section{\label{sec:results}Results}

We evaluate our proposed coarse-graining methods in the following two different ways:
\begin{itemize}
\item We compare our results on elementary cellular automata (ECA) to previous
  works on particle and domain filtering.
\item We use a metric which evaluates complexity of CAs introduced
  in~\parencite{cisnerosEvolvingStructuresComplex2019} in order to compare our
  methods' complexity metric scores of the coarse-grained systems and contrast
  the scores against a standard image processing baseline that computes local
  average of neighbouring cells followed by downscaling the grid. Using the
  complexity metric we measure to what extent the interesting behavior of
  cellular automata is conserved after coarse-graining compared to this image
  processing baseline.
\end{itemize}

In the following we begin by showing that a simple histogram-based
coarse-graining is effective at detecting structures (such as gliders) in ECA
space-time diagrams. Our method achieves results comparable with previous work,
while being simpler to apply.

\subsection{Domains and filtering}

In the space-time diagrams of cellular automata, moving structures such as
gliders are embedded in uniform or periodic backgrounds, or ``domains''. This
domain is different depending on the rule: some ECAs have uniform backgrounds,
checkerboard backgrounds or more complicated patterns (e.g.\ rule
110).~\parencite{crutchfieldTurbulentPatternBases1993} also identified chaotic
domains, which cannot support regular gliders but have ``walls'' and
``particles''. Those correspond, respectively, to boundaries between two chaotic
domains and propagating defects (localized structures with a pattern different
from the domain) within a domain.

Our proposed coarse-graining methods offer interesting perspectives to filter
cellular automata's space-time diagrams, which enables identifying gliders and
studying the formation of large-scale patterns. We find that a simple histogram
coarse-graining achieves results comparable to those reported
in~\parencite{hansonAttractorbasinPortraitCellular1992,
  elorantaKinkCellularAutomaton1992, hansonComputationalMechanicsCellular1997,
  wuenscheExploringDiscreteDynamics2011} for ECA rules 18 and 54. A similar
approach was undertaken in \parencite{wuenscheClassifyingCellularAutomata1999} in
which the authors used the entropy of rule table lookup frequencies to filter
out regular domains in the space-time diagrams of cellular automata and to
identify gliders and domain boundaries. However, Wuensche's approach described
in~\parencite{wuenscheClassifyingCellularAutomata1999} does not attempt to downscale
space-time diagrams.

\subsection{Results on elementary cellular automata}

\begin{figure}[htbp]
  \centering
  \includegraphics[width=\linewidth]{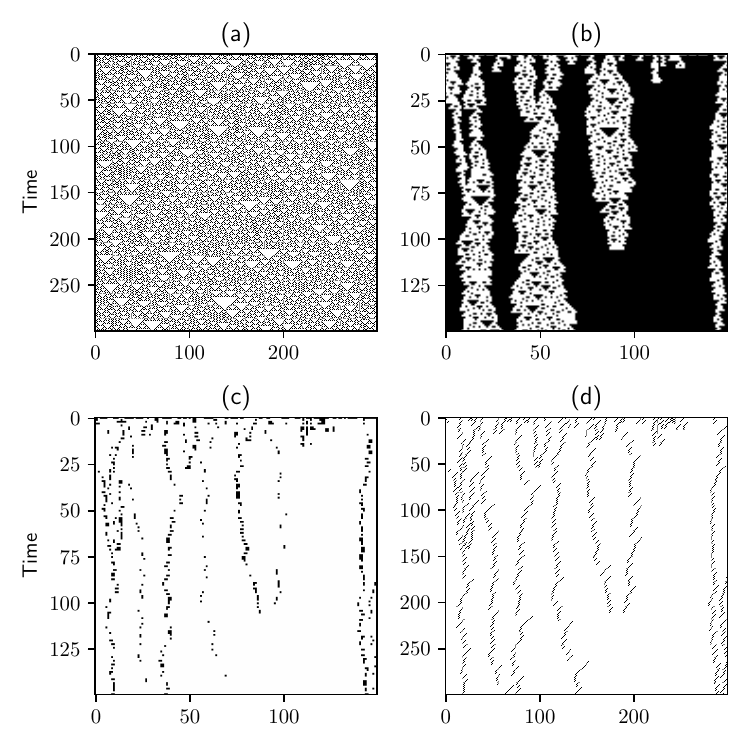}
  \caption{\label{fig:rule18} Space-time diagrams for rule 18 in elementary
    cellular automata. \textbf{(a)} Standard rule 18 space-time diagram,
    starting from a random position. \textbf{(b)} Filtered domain with our
    frequency coarse-graining (even). \textbf{(c)} Our domain boundaries
    extracted from the filtered domains in \textbf{(b)}. \textbf{(d)} Domain
    boundaries computed according to
    \parencite{hansonAttractorbasinPortraitCellular1992}. Note that \textbf{(a)}
    shows semi-chaotic behaviour, which is hard to interpret, whereas our method
    \textbf{(b)} highlights distinct domains within the disordered space-diagram
    in (a). The detected domains and domain boundaries from previous work (d)
    and ours (c) are very similar.}
\end{figure}

\begin{figure}[htbp]
  \centering
  \includegraphics[width=\linewidth]{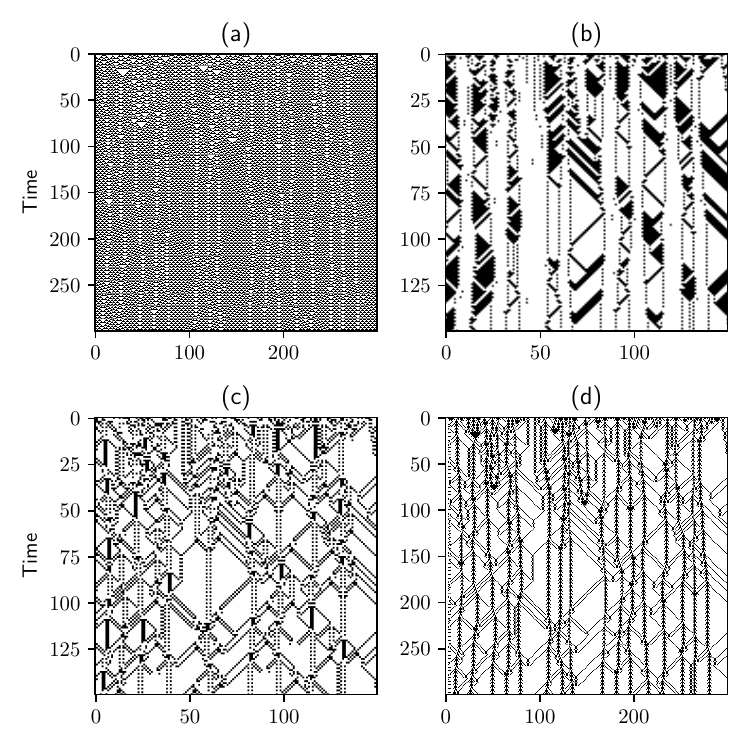}
  \caption{\label{fig:rule54} Space-time diagrams for rule 54. \textbf{(a)}
    Space-time diagram of standard rule 54, starting from a random position.
    \textbf{(b)} Filtered domain with our frequency coarse-graining (even).
    \textbf{(c)} Particles filtered from the domains in \textbf{(b) using our
      method}. \textbf{(d)} Domain boundaries computed using computational
    mechanics \parencite{hansonComputationalMechanicsCellular1997}. Please note
    that particles are detected equally well using computational mechanics (d)
    and our (simpler) frequency-based method (c). Some close-by particle trails
    are merged using our method.}
\end{figure}

We apply frequency histogram-based coarse-graining on elementary cellular
automata (ECA) and obtain space-time diagrams with suppressed background
domains. Resulting partitions of ECAs' space-time diagram are similar to results
reported by~\parencite{hansonAttractorbasinPortraitCellular1992,
  hansonComputationalMechanicsCellular1997}. Figure~\ref{fig:rule18}\textbf{(a)}
and~\ref{fig:rule54}\textbf{(a)} show the space-time diagrams of rules 18 and 54
with random initialization. Boundaries between different background patterns in
both Figures were obtained with coarse-graining; they are similar to boundaries
obtained by Hanson and Crutchfield. We also observe the propagation of many of
the same particles and defects without any prior information about the cellular
automaton rule.

Particles in Rule 54 have been used to implement
computations~\parencite{boccaraParticlelikeStructuresTheir1991,
  pivatoSpectralDomainBoundaries2007,
  martinezCompleteCharacterizationStructure2014}. Because the presented
reduction reduces the size of the grid, it can merge some of those particles,
sometimes resulting in ambiguities and gaps. However, our goal here is not to
precisely describe particle interactions in order to manipulate or construct
complex computations manually. Underlying computations described in the works
above are still happening within our reduced CA simulation. We consider the
\emph{apparent} destruction of some of these fine-scale details acceptable in
order to discover larger-scale complex behavior.

Our method is also arguably much simpler than computational mechanics (used by
Hanson and Crutchfield) which requires some reverse-engineering of the rule and
the construction of a finite-state transducer to generate output symbols.
Although full automation has been demonstrated, this method introduces
significant overhead~\parencite{rupeLocalCausalStates2018}. On the other hand, our
method is sensitive to the quality of statistical estimation of the frequency
histogram (see equation~\eqref{eq:stat_est}) and needs enough input examples to
achieve a reasonable result --- examples in Figure~\ref{fig:rule18}
and~\ref{fig:rule54} used simulations with the width of 3000 cells, ran for 6000
timesteps to obtain reliable pattern frequency estimates.

In the Figures, we used the coarse-graining method introduced
in~\nameref{sec:simple-hier-coarse}. Space-time diagrams are coarse-grained by a
factor 2 to a binary automaton --- each cell corresponds to a 2-cell block.
These binary coarse-graining results in the Figures are labeled \textbf{(c)}.
Because of the statistical nature of the domains of these 1D ECA's and the use of
blocks of size 2, filtered domains differ depending on the starting position of
coarse-graining. We distinguish an odd and even filtered domain.

\begin{figure}[htbp]
  \centering
  \includegraphics[width=\linewidth]{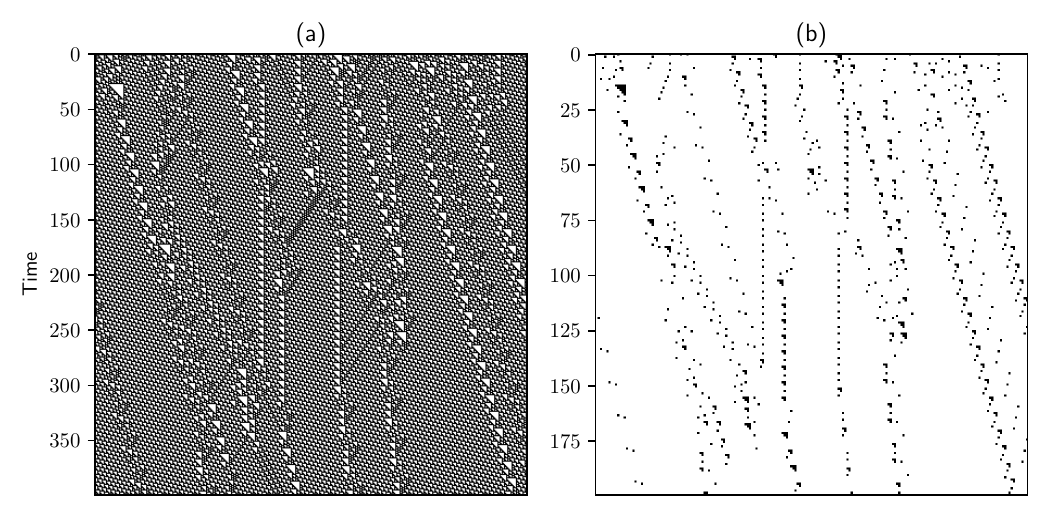}
  \caption{\label{fig:rule110} Spate-time diagram of rule 110 (\textbf{a}) and
    filtered particles using our histogram-based coarse-graining (\textbf{b}).
    Structures propagating in time (vertical axis) and space (horizontal axis)
    become clearly visible in \textbf{(b)} as vertical and diagonal lines.
  }
\end{figure}

Figure~\ref{fig:rule18}\textbf{(c)} is obtained by applying the element-wise
\texttt{OR} operator to both the even and odd domain diagrams to merge them into
a single space-time diagram. Figure~\ref{fig:rule54}\textbf{(c)} is obtained by
computing differences between neighboring cells after the filtering process to
highlight lines. Figure~\ref{fig:rule110} is another example showing filtering
of particles in rule 110.

\subsection{Complexity metrics and coarse-graining}

\setlength{\fboxsep}{0pt}
\begin{figure}[th]
  \centering
  \begin{subfigure}{.48\linewidth}
    \fbox{\centering
      \includegraphics[width=\linewidth]{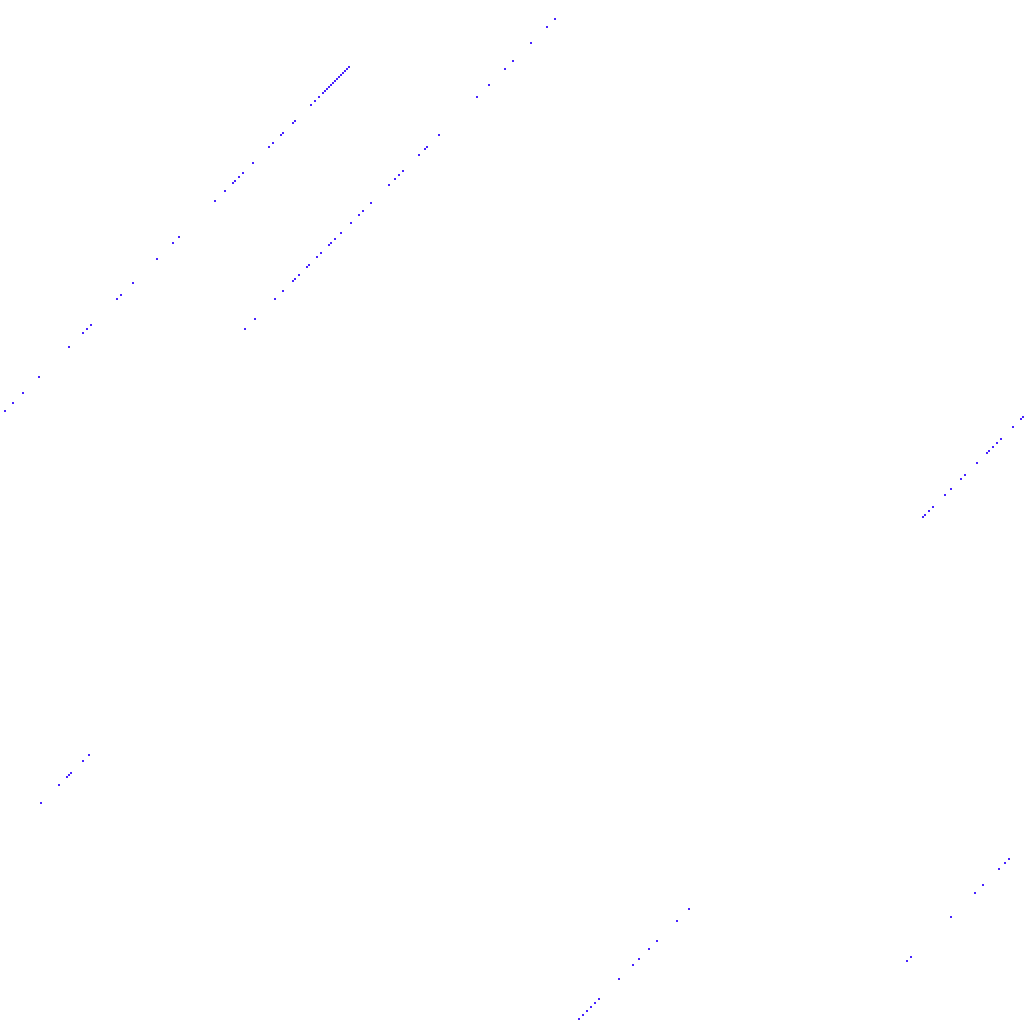}}
    \caption{\label{subfig:downscale3} Downscaling by averaging}
  \end{subfigure}
  \begin{subfigure}{.48\linewidth}
    \fbox{\centering
      \includegraphics[width=\linewidth]{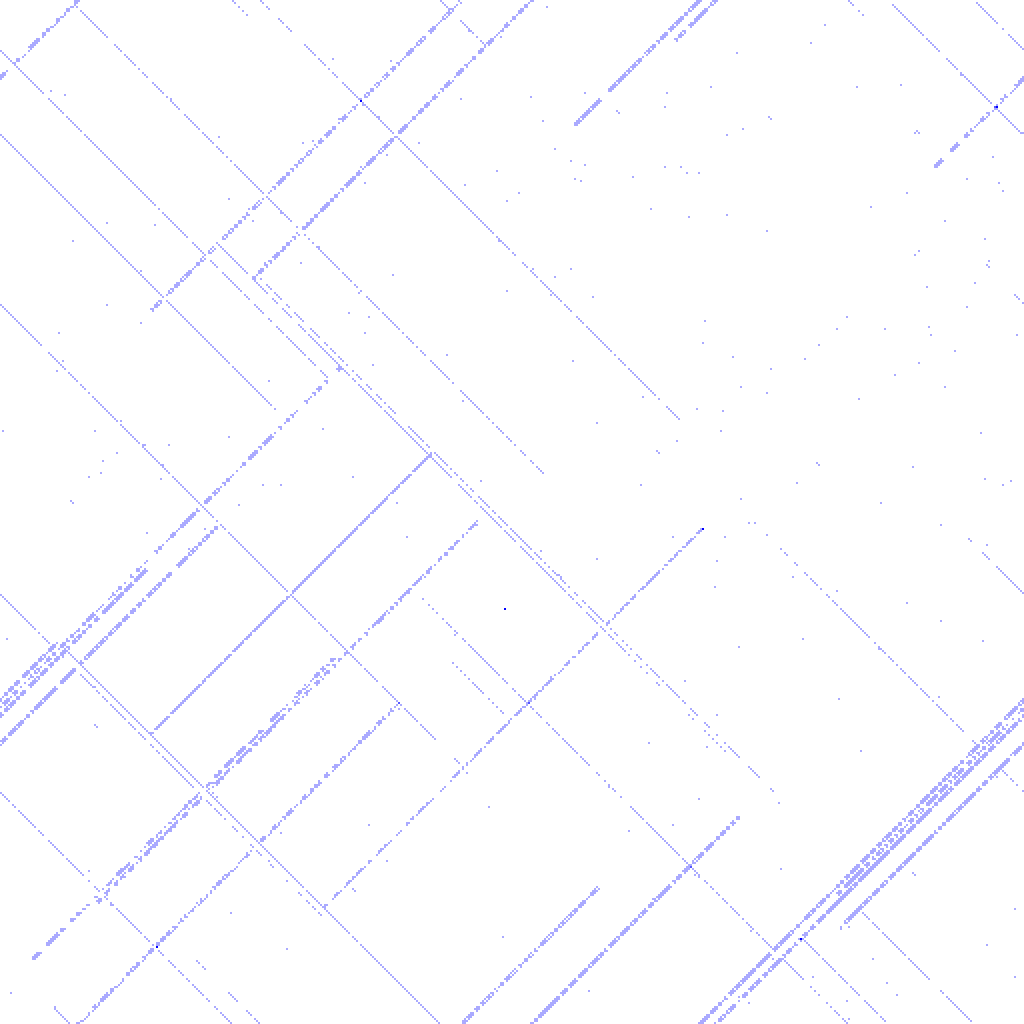}}
    \caption{\label{subfig:histogram3} Histogram}
  \end{subfigure}
  \begin{subfigure}{.48\linewidth}
    \fbox{\centering
      \includegraphics[width=\linewidth]{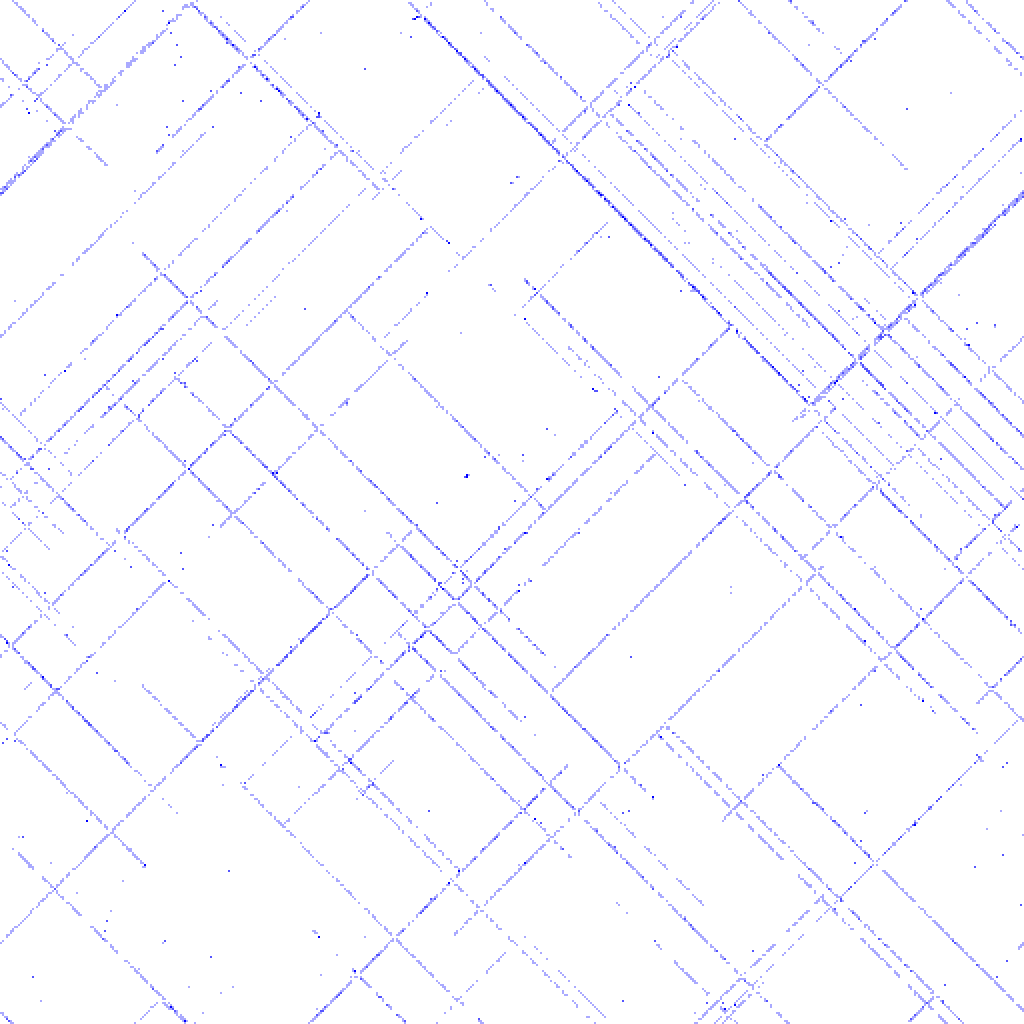}}
    \caption{\label{subfig:kmeans3} K-means}
  \end{subfigure}
  \begin{subfigure}{.48\linewidth}
    \fbox{\centering
    \includegraphics[width=\linewidth]
    {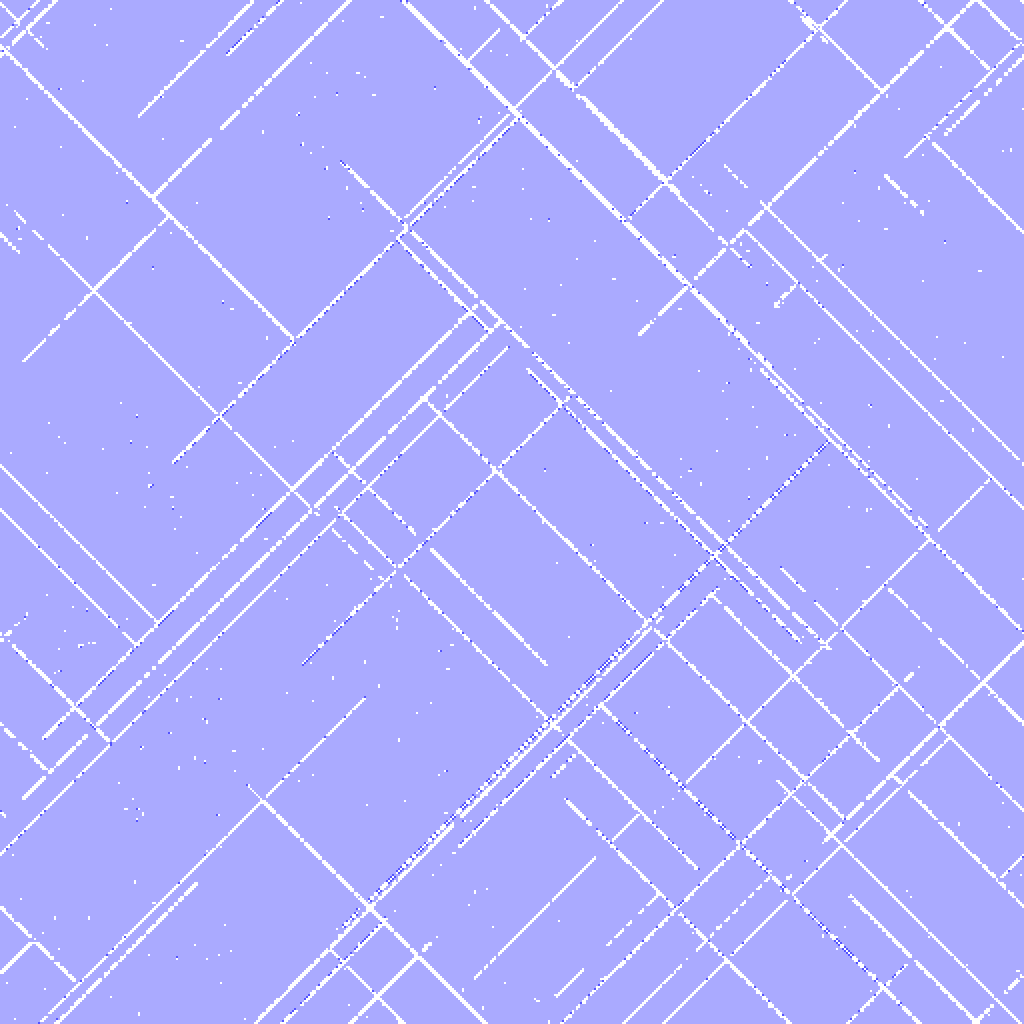}}
    \caption{\label{subfig:autoencoder3}Autoencoder}
  \end{subfigure}
  \caption{ Qualitative comparison of coarse-graining methods Simulations are on
    grids of $4096 \times 4096$ cells coarse-grained to $512 \times 512$. Most lines
    are barely visible with downscaling~\ref{subfig:downscale3}, but are visible
    in~\ref{subfig:histogram3}-\ref{subfig:autoencoder3}. Coarse-graining helps
    visualize linear structures that would be hard to see
    otherwise.\label{fig:qualitative}}
\end{figure}

Coarse-graining is not only useful for detecting gliders and domains in
space-time diagrams, but also as a tool to visualize large CAs. To evaluate the
quality of our proposed coarse-graining methods, we compare complexity scores
computed according to~\parencite{cisnerosEvolvingStructuresComplex2019} for different
coarse-graining methods. This metric was shown to correlate well with a user
study of interesting automata. It uses neural networks to estimate how easy it
is to learn a compressed representation of a CA\@. We also computed the scores on
downscaled CAs as a baseline. Local averaging is used for downscaling, with each
block of $N$ cells being replaced by its average value rounded to the nearest
integer state.

Experiments begin by sampling 3600 cellular automata rules with 3 or 4 states.
We apply the complexity metric on a randomly initialized simulation on a $512
\times 512$ grid of cells. The top 100 rules with the highest complexity scores,
which should correspond to rules with interesting behaviors, are then used for
coarse-graining. We apply coarse-graining on grids of $4096 \times 4096$ cells,
scaling the grid down by a factor of $8$, and compute the complexity metric also
on the reduced grid. Figures reported in Table~\ref{tab:experiments_table_vc} are
percentages of rules still considered interesting (above the selection threshold
for the first step of the process) after coarse-graining. The higher this number
is, the more a method is able to conserve complex and interesting behaviors
after the reduction.

\begin{table}[htbp]
  \centering
  \begin{tabular}{ccccc}
    \toprule
       Local-averaging & K-Means & Histogram & Autoencoder\\
     baseline &  & & \\
    \midrule
      19.3\%  & 40.4\% & 82.4\% & 84.2\%\\
    \bottomrule
  \end{tabular}
  \caption{Experimental results --- Percentage of rules classified as
      interesting after reduction with our 3 proposed
      methods (K-means, Histogram, Autoencoder), compared to a local averaging
      baseline.}\label{tab:experiments_table_vc}
\end{table}

Results in Table~\ref{tab:experiments_table_vc} suggest that using our proposed methods
seems largely beneficial for studying complexity in large systems.
Histogram and autoencoder methods are superior to downscaling using k-means and
local averaging. This could be attributed to the fact that contrary to the
latter two, the histogram and autoencoder both represent well anomalies (rare
events). This is because rare events are explicitly captured and kept by the
histogram method. They also represent useful information that may be kept for
reconstruction using the autoencoder.

\subsection{Discussion}\label{sec:discussion}

Downscaling by local averaging is not an effective solution to the
coarse-graining problem for several reasons. In particular, it tends to favor
the majority state in a supercell because of the averaging effect. Thin
structures spanning only a few cells placed on a uniform background are likely to
disappear after coarse-graining, although they may still be relevant with respect
to large-scale patterns. The histogram-based method explicitly encodes those
rare events in a supercell, even if their size is relatively small compared
to the supercell size.

\begin{figure}[htbp]
  \centering
  \begin{subfigure}{.052\linewidth}
    \fbox{\centering
    \includegraphics[width=\linewidth]{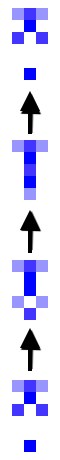}}
    \caption{\label{subfig:single}}
  \end{subfigure}
  \begin{subfigure}{.45\linewidth}
    \fbox{\centering
    \includegraphics[width=\linewidth]{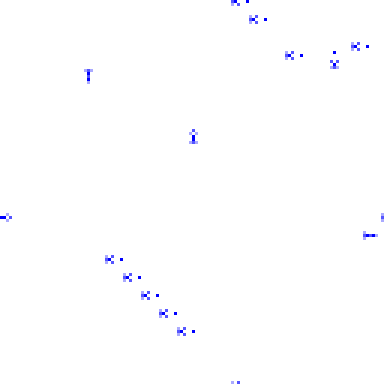}}
    \caption{\label{subfig:mult_glider} $128 \times 128$ cells}
  \end{subfigure}
  \begin{subfigure}{.45\linewidth}
    \fbox{\centering
    \includegraphics[width=\linewidth]{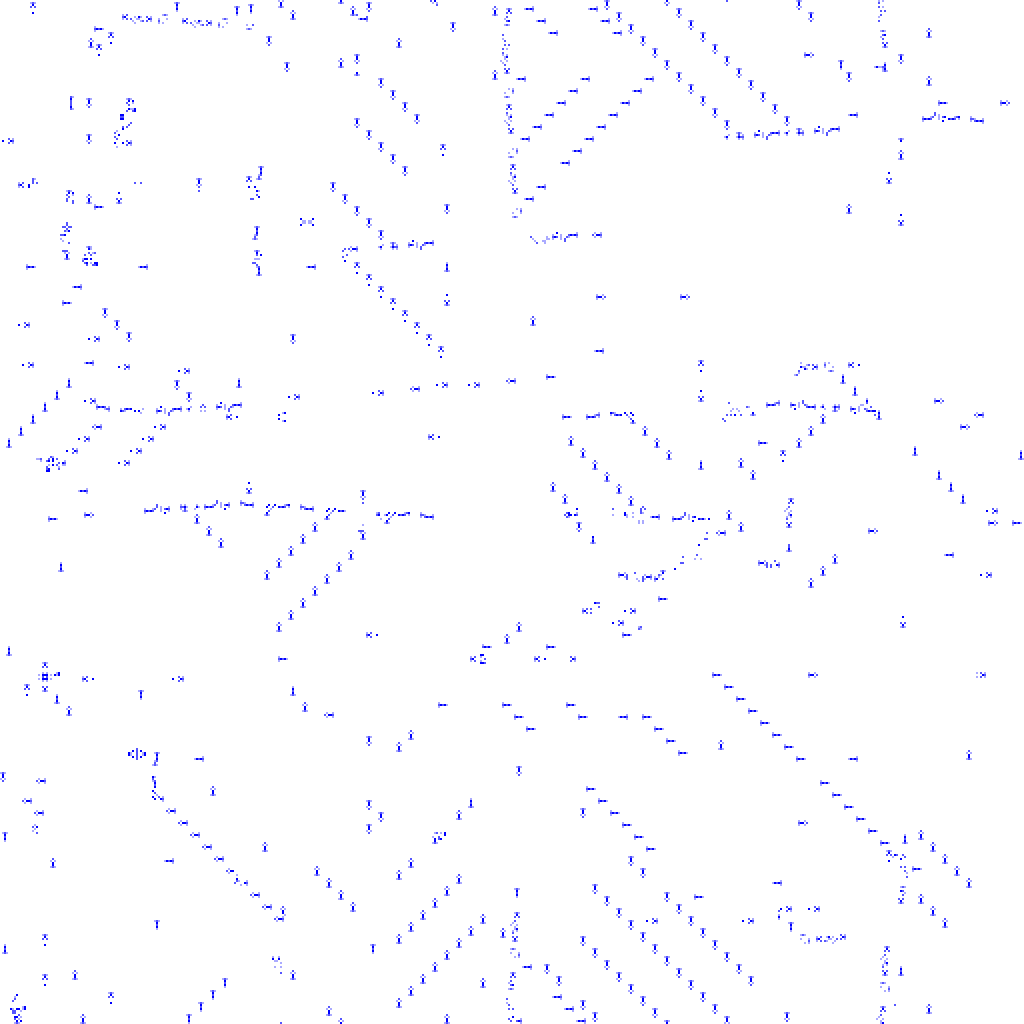}}
  \caption{\label{subfig:mult_glider_larger} $512 \times 512$ cells}
  \end{subfigure}
  \begin{subfigure}{.48\linewidth}
    \fbox{\centering
      \includegraphics[width=\linewidth]{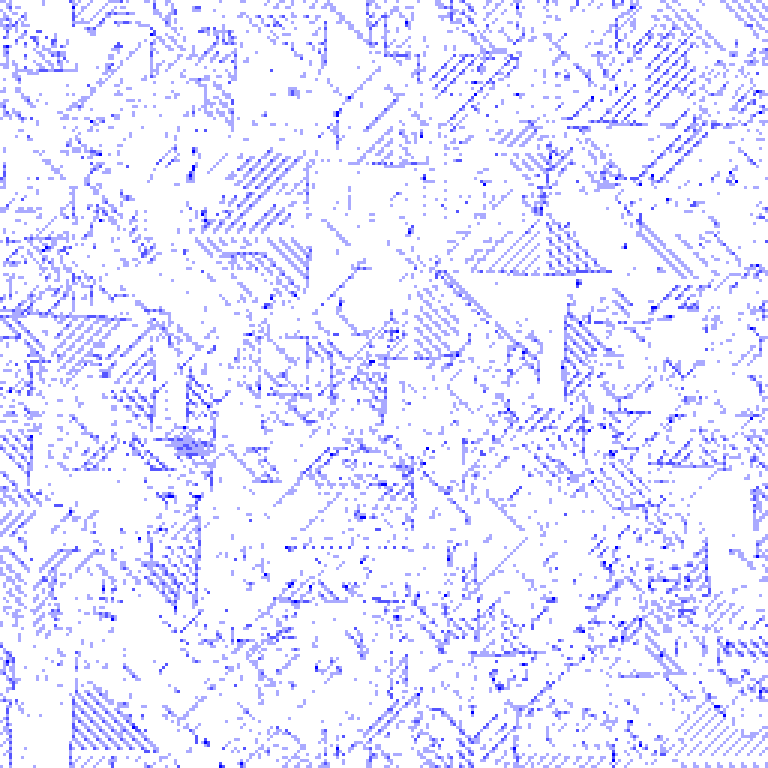}}
    \caption{\label{subfig:waves_b} $2048 \times 2048$ cells coarse-grained to $256 \times 256$.}
  \end{subfigure}
  \hfill
  \begin{subfigure}{.48\linewidth}
    \fbox{\centering
      \includegraphics[width=\linewidth]{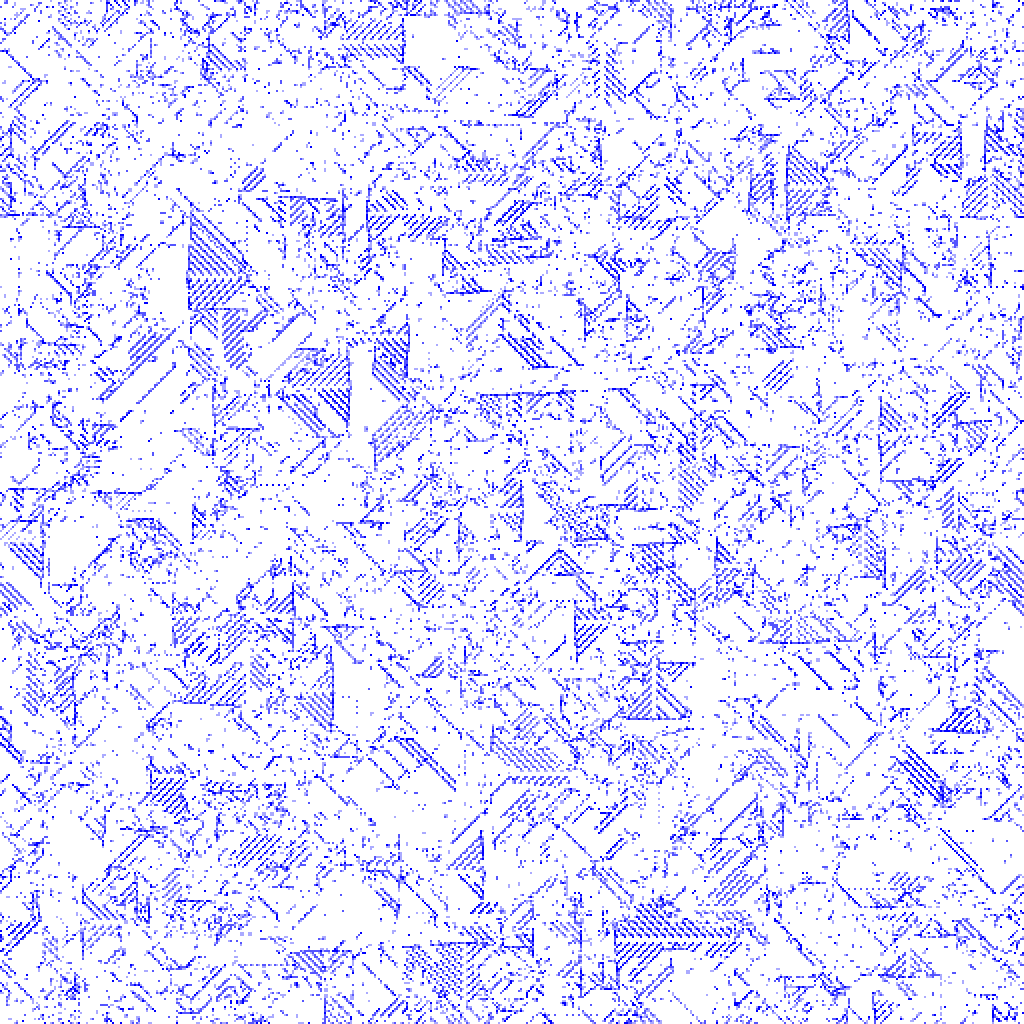}}
    \caption{\label{subfig:waves_c} $4096 \times 4096$ cells coarse-grained to $256 \times 256$.}
  \end{subfigure}
  \caption{\label{fig:dynamics_levels} Changing CA dynamics at multiple scales.
    (a) shows a single glider oscillating between 3 positions. Such gliders
    emerge spontaneously from a random initialization of a small grid, as shown
    in (b). When scaling the grid up, trails of gliders begin to appear,
    creating moving straight and diagonal lines, as shown in (c). Scaling up even
    more, individual gliders are not visible anymore, as shown in (d). In an
    even larger grid, shown in (e), many more triangular-shaped waves travel and
    collide with each other. Please note that (d) and (e) are coarse-grained to
    $256 \times 256$. Otherwise, the patterns would not be visible.}
\end{figure}

\begin{figure}[htbp]
\begin{subfigure}{.48\linewidth}
  \centering
  \includegraphics[width=\linewidth]
  {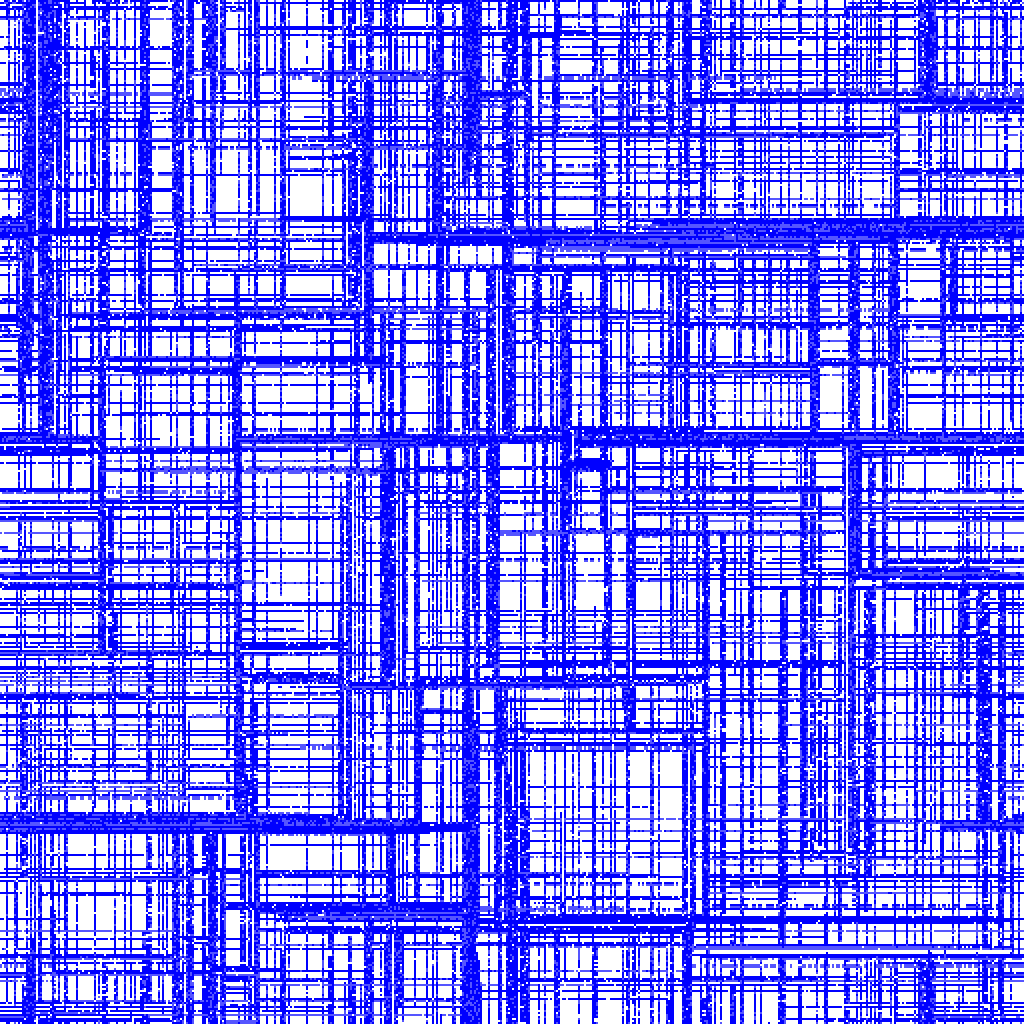}
  \caption{\label{fig:autoencoder_cg1}Autoencoder}
\end{subfigure}
\begin{subfigure}{.48\linewidth}
  \centering
  \includegraphics[width=\linewidth]{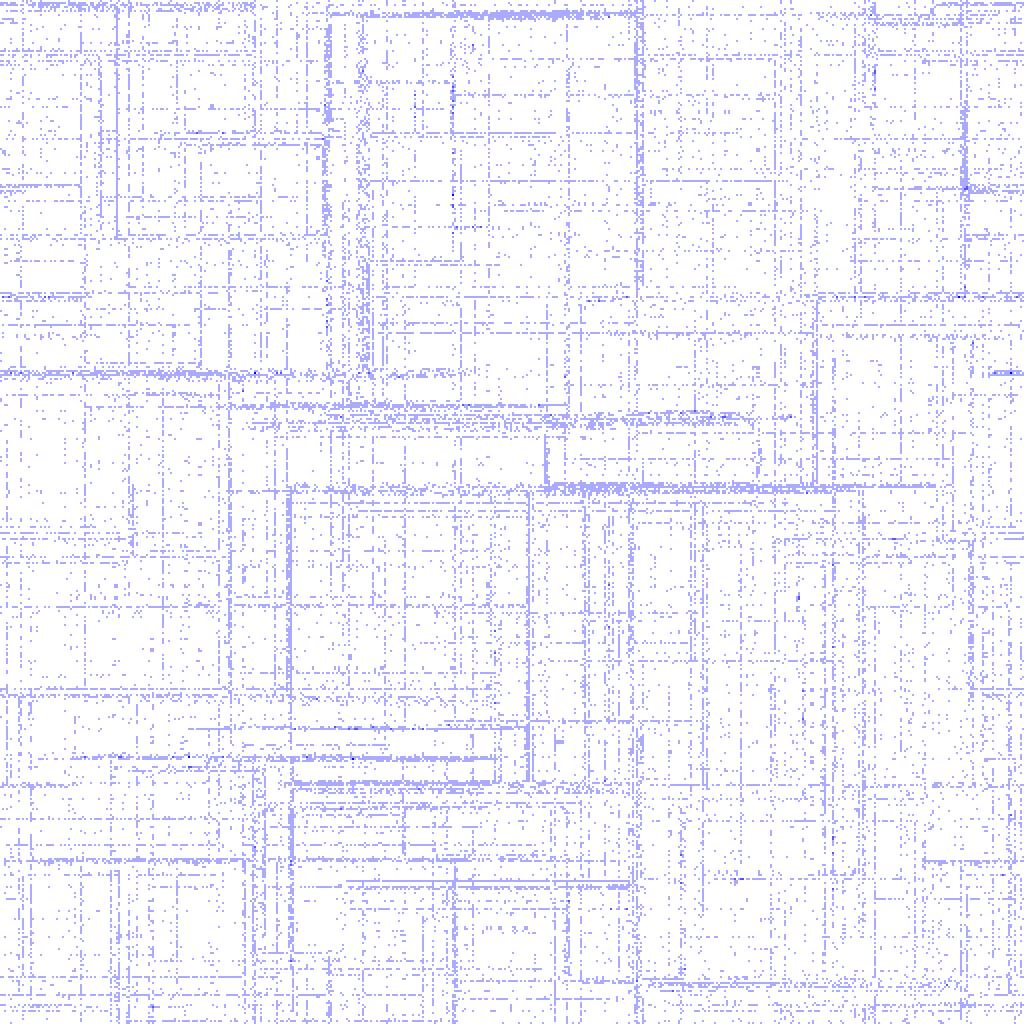}
  \caption{\label{fig:histogram_cg1}Histogram}
\end{subfigure}
\caption{\label{fig:coarse_graining_1} A \ac{CA} with $4096 \times 4096$ cells
  coarse-grained to $256 \times 256$. Comparison of the histogram
  (\ref{fig:histogram_cg1}) and autoencoder (\ref{fig:autoencoder_cg1})
  methods.}
      \label{fig:extra1}

\end{figure}

\begin{figure}[htbp]
  \begin{subfigure}{.48\linewidth}
    \centering
    \includegraphics[width=\linewidth]
    {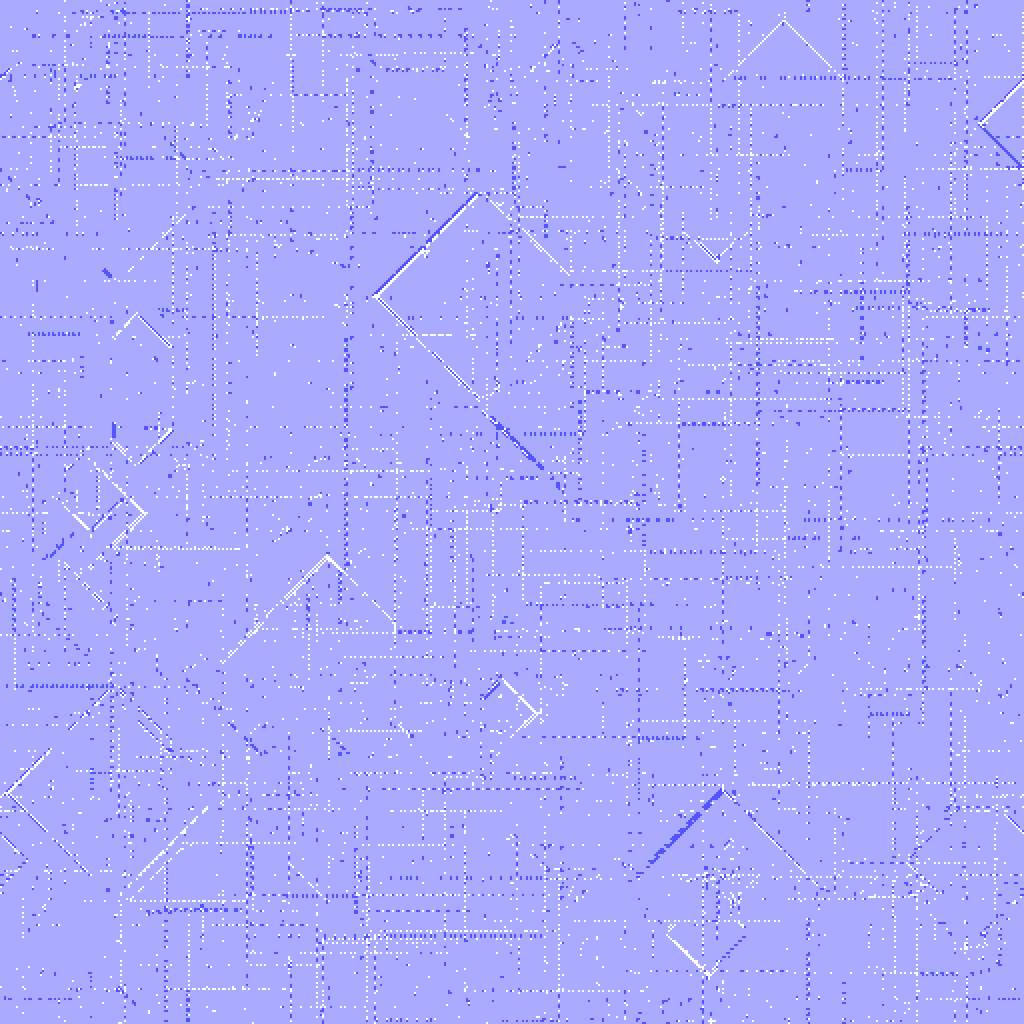}
    \caption{\label{fig:autoencoder_cg2}Autoencoder}
  \end{subfigure}
  \begin{subfigure}{.48\linewidth}
    \centering
    \includegraphics[width=\linewidth]
    {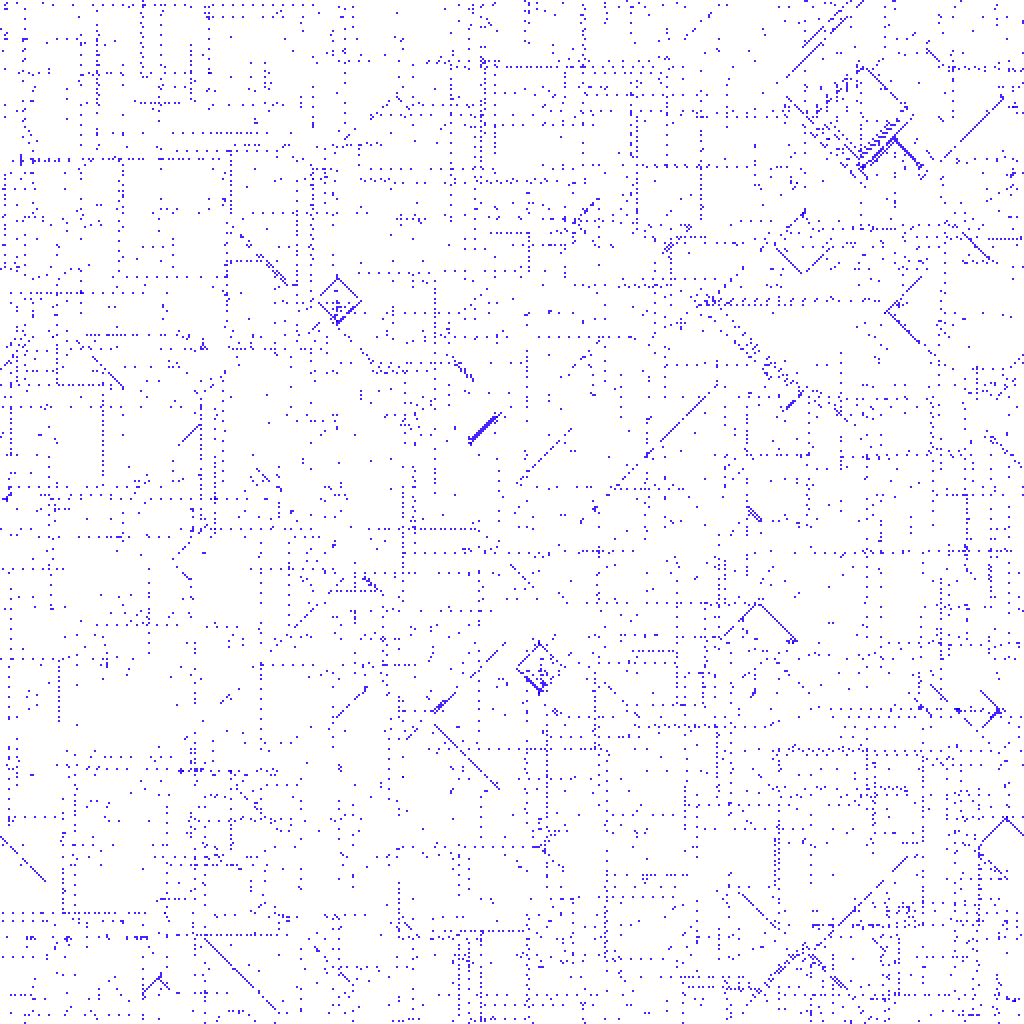}
    \caption{\label{fig:histogram_cg2} Histogram}
  \end{subfigure}
  \caption{\label{fig:coarse_graining_2} A \ac{CA} with $4096 \times 4096$ cells
    coarse-grained to $256 \times 256$. Comparison of the histogram
    (\ref{fig:histogram_cg2}) and autoencoder (\ref{fig:autoencoder_cg2})
    methods.}
    \label{fig:extra2}
\end{figure}

Figure~\ref{fig:qualitative} is a qualitative comparison of coarse-graining
methods. This cellular automaton was selected from the experimental dataset.
When simulated on large grids, it generates large linear structures that are
4-cells wide. These structures disappear after downscaling by averaging because
the background dominates the average. Other methods correctly highlight these
structures when downscaling the grids by a factor of 8. In
Figure~\ref{fig:dynamics_levels}, we show another rule that was selected for its
high complexity score on multiple coarse-graining scales from our dataset. The
CA has significantly different dynamics depending on the chosen scale.
Example~\ref{subfig:single} is a spontaneously occurring stable oscillating
glider with period 3. Large structures emerge from these simple gliders when
observing large grids. The online project page\footnote{\projecturl} shows
animated examples for Figure~\ref{fig:dynamics_levels}, emphasizing the advantage
of using coarse-graining for visualization. Figures \ref{fig:extra1} and \ref{fig:extra2} show more comparisons of large-scale coarse-graining of interesting \acp{CA}.

A crucial advantage of the frequency histogram method is its speed and ease of
implementation compared to autoencoders. Other than a few hyper-parameters for
partitioning the histogram, no training or tuning is needed to produce the
coarse-grained output.

\section{Conclusion\label{sec:conclusion-vc}}

We intend to use these coarse-graining methods to find cellular automata (CA) which
exhibit interesting behaviors at multiple scales.
Figure~\ref{fig:dynamics_levels} shows an example of such a CA\@. We observe
various dynamics depending on the scale, from simple oscillating gliders to
large wave-like patterns composed of thousands of gliders. It demonstrates that
observing multi-scale behaviors within those automata is possible. The existence
of 2D cellular automata with disordered behaviors at the smallest level but
organized at coarser scales, similar to hidden patterns in rule 18, would also
be of great interest.

Cellular automata are powerful computational models. Some of them have been
shown to be Turing-complete, and can thus be expected to support arbitrarily
complex computations~\parencite{berlekampWinningWaysYour2001,
  cookUniversalityElementaryCellular2004}. Naturally, most interesting CAs
spontaneously generate a fraction of available computations at a time, usually
supporting a few stable oscillators or moving structures. Proofs of universality
for these CAs required careful design of computational devices out of these
stable oscillators and structures, resulting in very brittle and inefficient
universal computers. In practise, only elementary functions --- such as density
classification, binary addition, etc. --- can be implemented. This requires
searching for CA rules specifically targeted at a particular
function~\parencite{mitchellEvolvingCellularAutomata1996, wolframNewKindScience2002,
  sapinResearchCellularAutomaton2003}. Hierarchies are central to naturally
occurring complex phenomena~\parencite{simonArchitectureComplexity1962}, and may be
required for robust and complex processes to emerge in CAs.

Viewing space-time diagrams of cellular automata is akin to visualizing a
foreign computer design. Cellular automata are manipulating information,
registers and instructions in parallel in the form of cell states. We believe
visualization tools proposed in this chapter can help understand computations in
those unconventional computers. By reducing available information to its
essential parts, we attempt to distill the content of the space-time diagram
with as little prior information as possible. Future work could focus on
identifying some known simple computational primitives within cellular automata
and understanding how our visualization can help to find them.

These methods also enable apprehending large grid sizes for which even image
processing algorithms begin to show limitations. Complexity metrics and CA
classification techniques can be extended to these reduced large grids and could
lead to the discovery of CAs with --- similar to life and physical processes ---
significantly different dynamics at multiple scales that could in turn be a
basis for artificial evolution.

\chapter{Learning efficiency in deep reservoir computing}
\label{cha:learn-effic-compl}

It is common to evaluate the performance of a machine learning model by
measuring its predictive power on a test dataset. This approach favors
complicated models that can smoothly fit complex functions and generalize well
from training data points. Although essential components of intelligence, speed, 
and data efficiency of this learning process are rarely reported or compared
between different candidate models. In this chapter, we introduce a benchmark 
of increasingly difficult tasks together with a data efficiency metric to measure 
how quickly machine learning models learn from training data. We also highlight 
that within complex systems, there are computations taking place that can be 
harnessed for various purposes. To that end, we compare the learning speed of 
some established sequential supervised models, such as RNNs, LSTMs, or Transformers, 
with relatively less-known alternative models based on reservoir computing, which 
is a tool for harvesting the computations happening within a complex system.
 The proposed tasks require a wide range of computational
primitives, such as memory or the ability to compute Boolean functions, to be
effectively solved. Surprisingly, we observe that reservoir computing systems
that rely on dynamically evolving feature maps learn faster than fully
supervised methods trained with stochastic gradient optimization while achieving
comparable accuracy scores. The code, benchmark, trained models, and results to
reproduce our experiments described in this chapter are available at
{\small\url{https://github.com/hugcis/benchmark_learning_efficiency/}}.

\section{Introduction}
Most machine learning models are evaluated by measuring performance on a
specific dataset or task. Learning efficiency -- the ability to learn,
generalize, and adapt quickly from a few examples -- is crucial for practical
intelligence \parencite{kanazawaGeneralIntelligenceDomainspecific2004} and
low-data machine learning applications, but rarely used to evaluate models.
Supervised learning systems are theoretically limited in their learning speed by
the optimization algorithms used for training. These algorithms, such as
stochastic gradient descent (SGD), have various speed guarantees depending on the
structure of the function to be
optimized~\parencite{bottouOptimizationMethodsLargescale2018}. However, when
intelligent beings learn, they appear to quickly reuse past knowledge and
progressively improve over time. Their learning speed depends on a dynamically
evolving internal state.
To measure the learning efficiency of various systems, in this work we propose the \ac{WADE}
metric based on the time taken to reach several test accuracy checkpoints. We
also design a simple modular benchmark composed of a set of sequential tasks.
They begin with the task of recognizing a simple periodic sequence in an input
string and end with elaborate question-answering tasks that require counting
occurrences of patterns and long-term memory.

Established sequential supervised models such as recurrent neural networks
\parencite[RNNs; ][]{elmanFindingStructureTime1990}, long short-term memory
networks \parencite[LSTMs;][]{hochreiterLongShortTermMemory1997} or Transformers
\parencite{vaswaniAttentionAllYou2017} lack essential properties such as
learning beyond the training phase or the ability to adapt over time after being
trained. These models can also be expensive to train, requiring a large number
of labeled training examples to reach reasonable performance, leading to poor
learning speeds. In this work, we use the newly proposed \ac{WADE} metric and
the benchmark dataset to experimentally compare the learning speed of these
well-established models, such as RNNs, LSTMs, and Transformers, with less
explored {\em reservoir computing models}.

Reservoir computing is a computational framework that aims to exploit the states
of a complex dynamical system. The simplest example of a reservoir computer is a
\ac{RNN} with frozen weights. This special \ac{RNN} performs random manipulation
on its hidden state in reaction to each new input. Interestingly, it has been
shown that with a specific initialization of the frozen weights, these \ac{RNN}s
(called Echo state networks) can keep a memory of past inputs
\parencite{jaegerEchoStateApproach2001}. Usually, a standard linear regression
is added as a decoder to extract valuable representation from the hidden state
for some downstream task. Freezing the weights of these recurrent models is
useful when available supervision is very limited or non-existent or for
reinforcement learning with sparse rewards since direct training would be
impossible. In such cases, a reservoir computer creates a continuously evolving
pool of random functions that can be combined using the last trainable layer.

When evolving in response to input stimuli, complex recurrent systems such as
\acp{RNN} are building dynamically changing representations of data within their
internal state \parencite{boccaraModelingComplexSystems2010}. We know that these
internal states can be interesting on their own because of their ability to
self-organize and exhibit increasingly complex behaviors
\parencite{koppelAlmostMachineindependentTheory1991,
  bennettLogicalDepthPhysical1995, allenEvolutionEmergenceLearning2003,
  goldsteinEmergenceComplexSystems2011, cisnerosEvolvingStructuresComplex2019}.
We wish to investigate whether complex dynamical systems --- in
particular RNNs with frozen weights (echo-state networks)
\parencite{jaegerEchoStateApproach2001} and reservoir cellular automata
\parencite{yilmazReservoirComputingUsing2014} --- create representations that
allow them to learn faster, as measured by our metric.

\paragraph{Contributions.} In this chapter, we make the following main
contributions: First, we introduce the \acf{WADE} metric to measure the learning
speed of various learning systems and use it to benchmark a few standard models
on the IMDB text classification task \parencite{maasLearningWordVectors2011}.
Second, we present a benchmark of language-based tasks of increasing difficulty
to evaluate the learning speed in different conditions. The proposed tasks
require a wide range of computational primitives, such as memory or the ability
to compute Boolean functions, to be effectively solved. Third, we study the
learning speed of reservoir computing learning models and compare them with more
standard supervised solutions.

\section{Related work}
The \ac{WADE} metric is a generalization of the \emph{Time-to-threshold} metric
\parencite{taylorCrossdomainTransferReinforcement2007,
  taylorTransferLearningInterTask2007} introduced for measuring transfer
learning in reinforcement learning contexts. In general, the Time-to-threshold
is simply defined as the number of training steps needed to reach a fixed
threshold performance. However, this definition leaves open the choice of
threshold or the definition of a training step. \ac{WADE} alleviates this issue
by aggregating several of these thresholds into a single number that summarizes
the learning speed.

Other metrics for measuring how quickly a model adapts to new tasks have been
introduced in the context of transfer learning, few-shot and zero-shot learning.
In few-shot learning, one tries to obtain the best performance for a particular
task using a small amount of labeled data compared to the task's fully
supervised equivalent~\parencite{wangGeneralizingFewExamples2020}. This correlates
with a model's learning speed, but these problems often measure how much prior
information about similar data has been encoded in the models. With our
benchmark and the \ac{WADE} metric, we explicitly measure the number of steps to reach
multiple test accuracy values using all the data needed, effectively emphasizing
data efficiency.

Sample efficiency has also been studied in the context of
reinforcement learning. \parencite{chevalier-boisvertBabyaiPlatformStudy2018} use the
number of demonstrations before a task is solved to measure sample efficiency.
This requires defining what solving the task means, which may vary from task to
task. Another approach is to measure performance (cumulated reward, accuracy,
\etc) after a fixed budget of training steps
\parencite{yaratsImprovingSampleEfficiency2019}. In this case, the most efficient
model is the one that achieves the best performance within the allocated budget.
In other cases, the sample efficiency is mentioned but not explicitly measured
and one has to examine the learning curves
\parencite{buckmanSampleefficientReinforcementLearning2018}. The \ac{WADE} metric is
a general approach to measure the learning efficiency of machine learning
models. We use it to benchmark a few standard models on the IMDB text
classification tasks \parencite{maasLearningWordVectors2011} and propose a set of
modular and extensible language-based tasks.

Synthetic tasks such as ours have played a vital role in a series of crucial
advances in machine learning algorithms. For example, the XOR problem has
partially motivated the development of neural networks
\parencite{minskyPerceptronsIntroductionComputational1972,
  rumelhartLearningInternalRepresentations1985}, and the \emph{circle and ring}
dataset has inspired the creation of novel clustering algorithms
\parencite{ngSpectralClusteringAnalysis2001}. The design of synthetic datasets has also been
an essential component of the development of learning algorithms with
memory and general computational capabilities
\parencite{hochreiterLongShortTermMemory1997,
  joulinInferringAlgorithmicPatterns2015, gravesNeuralTuringMachines2014,
  westonAICompleteQuestionAnswering2016, richardsonProbingNaturalLanguage2020}.

Other tasks are based on real datasets with artificial manipulations
\parencite{krizhevskyLearningMultipleLayers2009, srivastavaCompeteCompute2013a,
  goodfellowEmpiricalInvestigationCatastrophic2014,
  nguyenVariationalContinualLearning2017}. The goal of our dataset is to be
truly progressive in difficulty yet simple to understand and extend, to allow
applications in the field of online learning, and to easily understand a model's
basic computational capacities. Combined with our metric, it enables us to
measure learning speed across a range of conditions. In contrast to similar
synthetic datasets, we built this benchmark so that the last task is vastly more
complicated than the first and could still be extended to more complex examples.

\section{A benchmark for reservoir computing\label{sec:tasks}}
To measure the learning speed of candidate systems and their ability to improve
over time, we propose a performance metric and a standardized set of tasks. We
want to select those systems that quickly and reliably adapt and learn from new inputs.
For this purpose, we introduce the Weighted Average Data Efficiency (WADE)
metric. It aggregates the speed at which a model reaches several test accuracy
checkpoints. We describe the metric in more detail in Section
\ref{sec:performance-metric}.

To reliably compare learning speeds for various systems on a shared foundation,
we also introduce a novel dataset described in Table~\ref{tab:all-tasks}. It is
made up of sequential tasks that begin with straightforward pattern recognition
and progressively increase in complexity to approach the complexity of natural
language and other complex real-world tasks.

We do not focus on the prediction performance of our models but rather on their
data efficiency --- the number of example sequences they need to learn from before
reaching a target accuracy on a validation set.

\begin{table}[htbp]
  \centering
  \begin{tabular}{cp{.37\linewidth}p{.48\linewidth}}
    \toprule
    \bfseries Task id & \bfseries Name & \bfseries Description \\
    \midrule
    1 & Simple periodic pattern identification & Identify a simple periodic pattern. \\
    \arrayrulecolor{black!20}\specialrule{0.2pt}{.2em}{.4em}
    2 & Harder periodic pattern identification & Identify a periodic pattern with an arithmetically
                            increasing period. \\
    \arrayrulecolor{black}\midrule
    3 & Symbol counting & Count symbols from a sequence. \\
    \arrayrulecolor{black!20}\specialrule{0.2pt}{.2em}{.4em}
    4 & Pattern counting & Count patterns (delimited group of symbols) from a sequence. \\
    \arrayrulecolor{black}\midrule
    5 & Simple question answering & Answer simple YES/NO questions from a single prompt. \\
    \arrayrulecolor{black!20}\specialrule{0.2pt}{.2em}{.4em}
    6 & Harder question answering & Answer simple YES/NO questions from a single prompt
                                    with a more extensive vocabulary. \\
    \arrayrulecolor{black!20}\specialrule{0.2pt}{.2em}{.4em}
    7 & Question answering with world definition & Answer YES/NO questions from a
                                                   sequence of prompts. \\
    \arrayrulecolor{black!20}\specialrule{0.2pt}{.2em}{.4em}
    8 & Question answering with world definition and counting & Answer YES/NO
                                                                and counting questions from a
                                                                sequence of prompts. \\
    \arrayrulecolor{black!20}\specialrule{0.2pt}{.2em}{.4em}
    9 & Adjective question answering & Answer YES/NO and adjective questions from a
                                       sequence of prompts. \\
    \arrayrulecolor{black!20}\specialrule{0.2pt}{.2em}{.4em}
    10 & Adjective question answering and counting & Answer YES/NO, adjective,
                                                     and counting
                                                     questions from a
                                                     sequence of prompts. \\
    \arrayrulecolor{black}\bottomrule
  \end{tabular}
  \caption{General description of all the tasks in the benchmark.}
  \label{tab:all-tasks}
\end{table}
Standard benchmarks and metrics such as those introduced in this chapter have
always been essential in advancing various aspects of machine learning. For
example, the LSTM network demonstrated a superior memory capacity on a set of
synthetic tasks designed to challenge the memory of sequential learning systems
\parencite{hochreiterLongShortTermMemory1997}. Our goal with this benchmark is
to emphasize measuring learning speed across tasks of varying difficulties with
a range of computational requirements rather than focusing on performance only.
We describe the performance metric and the benchmark next.

\subsection{Performance metric\label{sec:performance-metric}}

We introduce the Weighted Average Data Efficiency (WADE) metric as a way to
measure how quickly a model learns using a weighted average of inverse times
taken to reach various test accuracy \emph{checkpoints} over time.

It is computed for an evenly distributed set of target accuracies $\sA$. They
represent the \emph{checkpoints} at which the speed of learning is estimated.
For example, we may choose $\sA = [0.1, 0.2, 0.3, 0.4, \ldots, 1.]$. The metric is
then calculated as
\begin{equation}
\text{WADE}(\va) = \frac{1}{\sum \alpha} \sum_{\alpha \in \mathbb{A}}\frac{\alpha}{\text{T}(\alpha, \va)},
\label{eq:wade}
\end{equation}
where $\va = (\eva_{0}, \eva_{1}, \ldots, \eva_{n})$ is a sequence of test accuracies achieved by the evaluated system
sampled at different training steps. The quantity $a_{i}$ typically
corresponds to the accuracy reached after seeing $i$ examples, and
$\text{T}(\alpha, \va)$ is the number of steps in the sequence $\va$ needed to reach
an accuracy of $\alpha$. It is defined as
\begin{equation}
  \text{T}(\alpha, \va) = \min\left\{ i \in \{1, \ldots, n, + \infty \}\; |\; a_{i} \geq \alpha \right\}.
\label{eq:tto}
\end{equation}

We also define $\text{T}(\alpha, \va) = +\infty$ if the accuracy value $\alpha$ is never
reached in $\va$. This is equivalent to appending an additional term $\eva_{+\infty}$
to $\va$, always set to the maximum accuracy 1. Note that by construction,
$\text{T}(\alpha, \va)$ is in $[1, + \infty [$.

\begin{figure}[htbp]
  \centering
 \includegraphics[width=.6\linewidth]{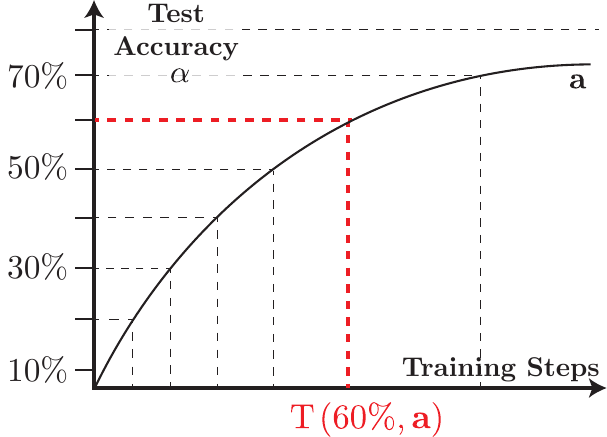}
 \caption{Illustration of the calculation of $\text{T}(\cdot, \cdot)$,
   representing the number of training steps (x axis) needed to reach a certain test
   accuracy $\alpha$ (y axis) from a learning curve. In this example,
   $\sA = [0.1, \ldots, 0.8]$ (y axis). $\text{T}(0.6, \va)$ is highlighted in red.
   $\text{T}(0.8, \va) = +\infty$ as the accuracy of 0.8 is never reached.}
  \label{fig:metric_tto}

\end{figure}

Since $\text{T}$ can be $+\infty$ we define
$\frac{1}{+\infty} = 0$ for the quantity in \eqref{eq:wade} to always exist. A visual
intuition of $\text{T}(\cdot, \cdot)$ is given in \figref{fig:metric_tto}.

The choice of checkpoints $\sA$ does not need to be tuned in any specific way
because $\text{WADE}(\va)$ quickly converges to a single value when $\sA$
approaches the continuous interval $[0, 1]$. The approximation is good enough as
long as $\sA$ is not too coarse (more than ten elements was enough in our
experiments), and the WADE values computed from the same set $\sA$ are
comparable.

The time-to-threshold $\text{T}$ is always greater than or equal to 1 step for any threshold and sequence of accuracy scores.
We have
$\forall\: \alpha \in [0, 1],\;  \forall\: \va = (\eva_{n})_{I \subset \{\mathbb{N}\; \cup \{+\infty\} \}}$, and we have
\begin{equation}
  \label{eq:wade-0-1}
\frac{1}{\text{T}(\alpha, \va)} \leq 1,
\end{equation}
and therefore we always get $0 \leq \text{WADE}(\va) \leq 1$. The metric is equal to $0$
for systems that never get past the smallest possible accuracy, while $1$
corresponds to reaching a perfect test accuracy in one single training step. Such
a system would also be considered to be performing well according to the
underlying performance metric with which it is usually evaluated. Therefore,
maximizing WADE also maximizes performance.

\subsection{Description of tasks in the benchmark\label{sec:descr-tasks-benchm}}

This section provides a more detailed description of each task in our benchmark.
The tasks are designed to be language modeling tasks, where the goal is to
predict some tokens from sequences of previously processed tokens. An overview
of the tasks is given in Table~\ref{tab:all-tasks}. The tasks are divided into three
groups: (i) binary tasks with only binary symbols, (ii) general symbolic tasks
-- symbolic manipulations with arbitrary symbols, and (iii) language-based tasks, where symbols represent words in English and behave like a language. We introduce
this benchmark together with the WADE metric, but both can be used in other
contexts as well to measure the learning speed of other systems. Individual
sentences are generated and divided into a training set and a test set for
periodic evaluation of the test accuracy. We give a more detailed description of
each task below\footnote{The tasks are also available as a
  \href{https://github.com/hugcis/incremental_tasks}{Python package on GitHub}.}:

\subsubsection{Binary}
\paragraph{Simple periodic pattern identification.}

The goal of the periodic binary task is to teach the model a fixed-length
regular pattern. As the system is presented with new binary input tokens, it has
to learn the periodic pattern on the fly and correctly predict the next token. A
pattern of size $n$ is chosen at random and repeated $k$ times to produce a
sequence of length $n \times k$. Examples include:
\begin{align*}
  \begin{split}
  \textbf{\texttt{01}}\texttt{01010101010101010101010101010101} & \quad\text{Pattern with period 2}\\
  \texttt{\textbf{0011}001100110011001100110011001100} & \quad\text{Pattern with period 4}\\
  \texttt{\textbf{011}0110110110110110110110110110110} & \quad\text{Pattern with period 3}
\end{split}\label{eq:1}
\end{align*}

\paragraph{Harder periodic pattern identification.}

For this task, we also draw a random binary pattern of size $n$. Each of its
symbols is repeated $k$ times, with $k$ increasing monotonically from 1. A
successful model must learn the pattern on the fly and correctly implement the
arithmetic increase in the size of the period. We set the pattern length to increase by
1 every period in our experiments, but this value can be changed.

\subsubsection{Symbolic counting}

These tasks consist of reading patterns from an input sequence and answering a
simple query about the number of patterns. Unlike previous tasks, these
require implementing a form of addressable memory that can be queried after the prompt has
ended.

\paragraph{Basic symbol counting.}
The first version of the counting task focuses on counting single symbols from
an input sequence. The sequence ends with a \emph{query} for the count of
one of the symbols. The goal is to predict the last token (in bold) of sequences
of the following form:

\begin{align*}
  \underbrace{\texttt{AABBCBABAAB}}_{\text{Input symbols}}
  & \underbrace{\texttt{x}}_{\text{QS}} \texttt{A}
    \underbrace{\textbf{\texttt{5}}}_{\text{Answer}} \\
\end{align*}

The symbol \texttt{x} is the query symbol (QS) that marks the beginning of the
query. In the first example above, the goal is to predict the token \texttt{5}
because the symbol \texttt{A} appears 5 times. As detailed in
Sect.~\ref{sec:compared-methods}, we represent these nonbinary symbols with
one hot encoding, so the numerical nature of some tokens is not encoded a priori.

\paragraph{Pattern counting.}
This aim of this task is to count the number of occurrences of delimited patterns
instead of single symbols. A sequence is still divided between a prompt --- before
\texttt{x} --- and a query --- after \texttt{x}. One has to predict the symbol
coming after each separator symbol (S) \texttt{y} in the query part of the
sentence. For example, sentences are of the form:

\begin{align*}
  &\underbrace{\texttt{AA}}_{\text{Pattern 1}}\underbrace{\texttt{y}}_{\text{S}}
    \underbrace{\texttt{BBC}}_{\text{Pattern 2}}\underbrace{\texttt{y}}_{\text{S}}\underbrace{\texttt{BAB}}_{\text{Pattern 3}}\underbrace{\texttt{y}}_{\text{S}}\underbrace{\texttt{AA}}_{\text{Pattern 4}}\underbrace{\texttt{y}}_{\text{S}}\underbrace{\texttt{B}}_{\text{Pattern 5}}
    \underbrace{\texttt{x}}_{\text{QS}}\underbrace{\texttt{AAy}}_{\text{Query 1}}\underbrace{\texttt{\textbf{2}}}_{\text{Answer 1}}
    \underbrace{\texttt{By}}_{\text{Query 2}}\underbrace{\texttt{\textbf{1}}}_{\text{Answer 2}} \\
\end{align*}

Multiple queries are presented successively, which requires keeping and being
able to retrieve several counts simultaneously. A query is composed of a
pattern, a separator symbol, and the pattern count that the system should
predict.

\subsubsection{Basic language understanding}
To make the tasks progressively more complex, we steer them towards general
language understanding tasks. The tasks described below are generated automatically, but gradually incorporate more complex skills
required for advanced language processing.
The last task is a step towards understanding the general language albeit with a limited vocabulary.

\paragraph{Elementary question answering (QA).}
This task introduces elements of natural language. Each example is composed of a
stated fact and a question about that fact. A sentence is constructed from a few
basic elements:
(i) Names (\eg\texttt{JOHN}, \texttt{JAMES}, \etc),
(ii) Verbs (\eg\texttt{HEAR}, \texttt{SEE}, \etc),
(iii) Answers (\texttt{YES} or \texttt{NO}),
(iv) Additional words and symbols (\texttt{I, DO, NOT, AND, BUT, ?, .}).

A random subset of names is selected, and we generate a random prompt/question
pair from it.
The question is drawn to ensure an equal proportion of
positive and negative answers. For example, sentences may look like this:

\begin{align*}
&  \texttt{I HEAR JOHN AND PAUL .}\texttt{ DO I HEAR PAUL ?} \texttt{\textbf{ YES}}\\
&    \texttt{I SEE JOHN BUT I DO NOT SEE PAUL AND TOM . DO I SEE TOM ? \textbf{NO}}
\end{align*}

The only token to predict is the binary answer \texttt{YES} or \texttt{NO}.

\paragraph{Question answering (QA) with adjectives.}

This task extends the previous task by adding adjectives and modifiers to the
object names. The queries may be about the subject-verb relation or the
subject-adjective relation.

\begin{align*}
  &\texttt{I SEE A SMALL BANANA .}\texttt{ WHAT IS THE SIZE OF THE BANANA I SEE ? } \texttt{\textbf{SMALL}}\\
  &  \texttt{I SEE A LARGE GREEN APPLE BUT I DO NOT SEE A RED APPLE .} \\
  & \quad\quad\texttt{DO I SEE A LARGE APPLE ? \textbf{YES}} \\
  &  \texttt{I SEE A SMALL GREEN APPLE BUT I DO NOT SEE A BANANA .}\\
  &\quad\quad\texttt{WHAT IS THE COLOR OF THE APPLE I SEE ? \textbf{GREEN}}
\end{align*}

Here the task output space is slightly larger because the model may be
predicting \texttt{YES}, \texttt{NO}, \texttt{SMALL}, \texttt{GREEN}, \etc

\paragraph{Question answering (QA) with world definition.}

This task introduces more complex configurations in which the state of the world is
defined in one or more sentences, and an unknown number of questions follow.
This is more akin to a real-world conversation where stored facts should be
remembered longer and accessed on demand.
For example, below we show a generated group of sentences followed by several questions:
\begin{align*}
  &\texttt{I SEE A SMALL BANANA .}\\
  &  \texttt{I SEE A LARGE GREEN APPLE BUT I DO NOT SEE A RED APPLE .} \\
  &  \texttt{I SEE A SMALL GREEN APPLE BUT I DO NOT SMELL A BANANA .}\\
  &\quad\quad\texttt{WHAT IS THE COLOR OF THE APPLE I SEE ? \textbf{GREEN}}\\
  &\quad\quad\texttt{HOW MANY THINGS DO I SMELL ? \textbf{ONE}}\\
  & \quad\quad\texttt{DO I SEE A LARGE APPLE ? \textbf{YES}}. \\
\end{align*}

The difficulty of each of these tasks can be modulated by changing the size of the
base vocabulary, the length of sequences, or the number of queries. Our particular setting of these parameters will be described in section~\ref{sec:task-gen-params}.

\section{Standard language classification task}

To show the usefulness of measuring \ac{WADE} on standard language
classification tasks, we train various text classifiers on the IMDB dataset and
compare their \ac{WADE} scores with a usual performance metric for
classification: accuracy. This classification task, first proposed by
\parencite{maassRealTimeComputingStable2002}, consists of deciding if a movie review
is positive or negative from its text. It contains 25000 training examples and
25000 test examples. The two labels are balanced in both the training and test
set. This dataset is a high dimensional language-based task with a binary
output.

We study five standard models: (i) an Elman recurrent neural network (RNN)
\parencite{elmanFindingStructureTime1990} with $\tanh$ activation functions trained
with backpropagation through time. (ii) A long-short term memory (LSTM)
recurrent neural network \parencite{hochreiterLongShortTermMemory1997}, also trained
with backpropagation through time. (iii) A gated recurrent unit (GRU) recurrent
neural network \parencite{choPropertiesNeuralMachine2014}. (iv) A standard
encoder-only transformer neural network model \parencite{vaswaniAttentionAllYou2017}.
(v) A logistic regression using a bag-of-words representation of each sentence
as input features. All the models are trained with batches of training data
using the Adam optimization algorithm \parencite{kingmaAdamMethodStochastic2015}.
Each model's hyperparameters are chosen to ensure they have a similar number of
trainable parameters except for the logistic regression whose parameter count is
solely determined by the input and output dimensions.

\begin{table}[htbp]
  \centering
  \begin{tabular}{c|cc}
\toprule
& \bfseries WADE $\times 10^{-2}$ (std.) $\uparrow$ & \bfseries Max test accuracy (std.) $\uparrow$ \\
\midrule
    RNN & 1.028 ${\scriptstyle \pm0.283}$ & 0.705 ${\scriptstyle \pm0.076 }$ \\
LSTM & 2.280 ${\scriptstyle \pm0.303}$ & 0.902 ${\scriptstyle \pm0.002 }$ \\
GRU & 2.711 ${\scriptstyle \pm0.318}$ & \bfseries 0.904 ${\scriptstyle \pm0.002 }$ \\
Linear & 3.737 ${\scriptstyle \pm0.745}$ & 0.862 ${\scriptstyle \pm0.000 }$ \\
Transformer & \bfseries 8.716 ${\scriptstyle \pm0.720}$ & 0.872 ${\scriptstyle \pm0.003 }$ \\

    \bottomrule
  \end{tabular}
  \caption{Comparison of our new WADE metric to assess learning speed
    and the standard maximum test accuracy on the IMDB
    classification dataset. Results are averaged over 50 separate runs. ($\uparrow$
    indicates that higher is better).
    \label{tab:comp-sup-acc}}
\end{table}

\begin{figure}

  \centering
  \includegraphics[width=.6\linewidth]{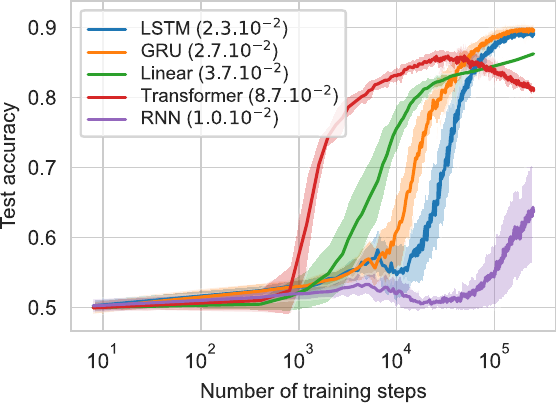}
  \caption{Test accuracy curves for each model. Shaded areas are
    $\pm 1\sigma$ around the average over 50 runs.\label{fig:wade-sup}}%
  \end{figure}

Table~\Ref{tab:comp-sup-acc} shows the results of all models on the IMDB dataset
reporting both the standard test accuracy as well as our new WADE metric
measuring the learning speed of the different models. We report the maximum test
accuracy observed during training. This corresponds to the model checkpoint that
would be selected with a validation set before using it on test data. The
transformer model learns the fastest of all, as shown by its higher WADE score,
but it does not reach test accuracy values as high as the GRU and LSTM model. We
think that this could be attributed to the transformer model over-fitting on the
available data as discussed below.

According to the WADE metric, the RNN is the worst (i.e., slowest to learn) model
with a score of $1.04 \times 10^{-2}$ whereas the linear model and the transformer
are the fastest with a score above $8 \times 10^{-2}$. The transformer seems to have
more reliable learning speeds than the linear model as shown by the lower
standard deviation ($0.7\times 10 ^{-2}$ versus $4.3 \times 10^{-2}$). The LSTM and GRU
models have intermediate and stable learning speeds.
The GRU has slightly better WADE score than the LSTM, which can be also seen by inspecting the learning curves in figure~\ref{fig:wade-sup} where the GRU is consistently above the LSTM.

However, when we look at the maximum test accuracy we obtain a different ordering,
with LSTM and GRU performing best with slightly above $90\%$ accuracy followed by the
transformer and the linear model (around $86\%$) and finally the RNN with the lowest
accuracy. The LSTM, GRU and linear models all seem to have converged at the end
of the experiment. The transformer's lower final accuracy despite it being
considered a state of the art model may be explained by over-fitting (apparent
on the graph with the test accuracy score going down). Moreover, the RNN clearly
hasn't converged at the end of the experiment which also explains the relatively
low max accuracy score.

Our new WADE metric gives a complementary view of a model's abilities, which may be significantly different from what can be obtained from the usual
performance metrics focused on the final accuracy of the model.

\section{Reservoir computing and reservoir cellular automata \label{sec:rc-and-reca}}

We use 
names adapted from \parencite{jaegerLongShortTermMemory2012} in
this section. An echo-state network is a random recurrent neural network with
frozen weights and skip-connections. Its random weights are sampled in a
specific way so that it performs random combinations of input vectors with its
current state while keeping a history of past inputs \parencite{jaegerEchoStateApproach2001}.

Reservoir cellular automaton (ReCA) is a model similar to echo-state networks
but where a cellular automaton (CA) replaces the random RNN. The cellular
automaton can be seen as a \ac{RNN} with additional weigth-sharing similar to
convolutional neural network. The special structure of a recurrent CA update
makes it more likely than random RNNs to generate complex structures
\parencite{wolframStatisticalMechanicsCellular1983, wolframNewKindScience2002}.
Because CAs were not designed to make use of inputs or produce outputs, we
extend the model to make it accept input vectors and to make reading from the CA
state possible (details in Section~\ref{sec:app-ca-res}).

We also propose two novel projection schemes that offer more flexibility than
existing ones. They extend the one-to-one projection and allow for a wider
variety of produced encodings:

\begin{description}
  \item[One-to-many.] Each input bit is assigned exactly to $k$ positions in $P$
        instead of one. This is equivalent to adding $k$ separate and mutually
        exclusive one-to-one projections into a single one.
  \item[One-to-pattern.] Each input bit is assigned to a random contiguous pattern
        of $k$ bits set in a fixed position.
\end{description}

\begin{figure}[htbp]
  \centering
  \includegraphics[width=\linewidth]{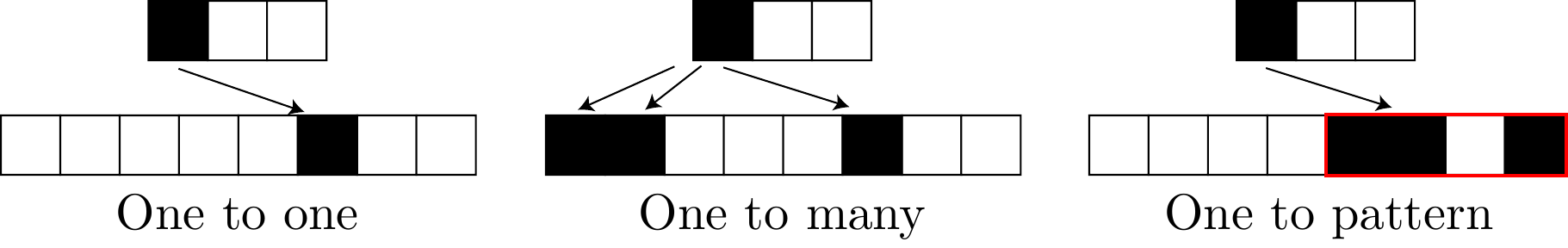}
  \caption{Three encoding methods. From left to right: \emph{one-to-one},
    \emph{one-to-many} and \emph{one-to-pattern}.}\label{fig:enc_meth}
\end{figure}

The effect of applying our three methods to a simple input with three components
is illustrated in Fig.~\ref{fig:enc_meth}. Projections are chosen to map
each input bit to a unique output configuration.

Following \parencite{yilmazReservoirComputingUsing2014,
  nicheleReservoirComputingUsing2017, nicheleDeepLearningCellular2017}, the
input is projected multiple times to add redundancy to the encoding. This was
experimentally observed to improve performance. We also use a projection vector
larger than the initial one. Projections are generated randomly $R$ times, applied,
and concatenated as a single large encoder to create the full \ac{CA} input
vector. The parameter $R$ is called \emph{redundancy}.

For cellular automata in 1D, the space $\mathcal{P}$ is one dimensional, and the
concatenation is performed on this dimension. In higher-dimensional spaces,
this concatenation must be defined in another way.

We write the $R$ projection functions $E_{1}, E_{2}, \ldots, E_{R}$. The final size
of the input vector and the \ac{CA} state grid is now $R \times L_{d}$ instead of $L_{d}$.
We have
\begin{align}
  \vp_{t} = P(\vx_{t}) = E_{1}(\vx_{t})\mathbin\Vert \ldots \mathbin \Vert E_{R}(\vx_{t}),
\end{align}
where $\mathbin\Vert$ is the concatenation operator.

For example, other projections have been proposed and implemented in
\parencite{yilmazReservoirComputingUsing2014}. We choose to present the three
above because they are simple to understand and implement, and their effects can
be explored thoroughly.

\paragraph{Input combination.}
There are multiple ways to combine the input vector with the current state of the \ac{CA}.
\textcite{gloverDynamicalLandscapeReservoir2021} propose to use a XOR
function between the projected input and the state of the \ac{CA}.

At each input step $t$, the projected input vector $\vp_{t}$ is XORed
element-wise with the current CA state $\vs_{t}$, so that each 1 bit of the
input switches the corresponding bit in the CA state to another value. This
creates a new state $\vs_{t}'$ from which the CA will evolve. $\vs_{t}'$ is
defined as
\begin{align}
  \vs_{t}' \coloneqq \vp_{t} \otimes \vs_{t},
\end{align}
where $\otimes$ is the element-wise XOR operator between two vectors of binary
values, and $\vs'_{t}$ is the temporary CA state resulting from the combination
of the CA state and the input vectors at time $t$.

From the state of the \ac{CA} at time $t$, $\vs'_{t}$, we then compute the next CA states
applying the update function $\Phi$,

\begin{equation}
  \label{eq:ca-state}
  \vs_{t + 1} = \Phi(\vs'_{t}).
\end{equation}

This is one of many possible ways to combine an input vector with the current
state of the CA that may affect its performance as a reservoir. The XOR-based
input combination gives an asymmetrical role to 0 and 1 in the input vectors,
which is natural considering how we encode and project categorical data points.
Different rules of \acp{CA} might benefit or lose performance due to these
encoding choices. The combination problem can be generalized to incorporate it
into the \ac{CA} rule search problem. We demonstrate this in the next paragraph.

\paragraph{Generalization.} We generalize the input combination method above by
incorporating it into the \ac{CA} update rule. Since the input vector $\vp$ is
binary, we can treat its value at position $i$ as just another virtual \ac{CA}
cell. For a standard \ac{ECA}, the update step takes into account the three
immediate neighbors $\vs^{(i - 1)}, \vs^{(i)}, \vs^{(i + 1)}$ plus the
additional virtual cell $\vp^{(i)}$. Instead of 8 possible input configurations
for a binary 1D \ac{CA}, we now have 16. In this extended \ac{CA} space, the
number of possible rules is $2^{16} = 65536$ instead of 256.

\begin{figure}[htbp]
  \centering
    \begin{subfigure}[b]{.387\linewidth}
    \centering
    \includegraphics[width=\linewidth]{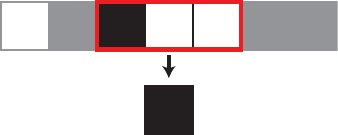}
    \caption{Regular CA update only takes neighbor states into
      account.}\label{fig:standard-ca-update}
  \end{subfigure}
  \hspace{10pt}
  \begin{subfigure}[b]{.55\linewidth}
    \centering
    \includegraphics[width=\linewidth]{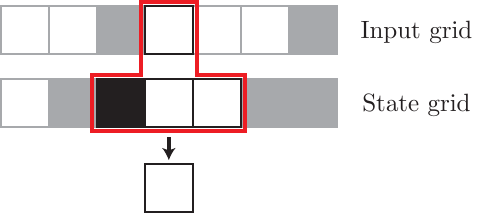}
    \caption{The extended neighborhood update also uses an input value to
      decide the next state of a cell.}\label{fig:generalized-ca-update}
  \end{subfigure}
  \caption{Comparison of standard CA update rule and our extended input
    combination rule.\label{fig:ca_update_rule}}
\end{figure}

This method is illustrated in Fig.~\ref{fig:ca_update_rule} during the regular CA
update. It generalizes the concept of input combination with the state $\vs$.
The XOR method described above is one of its special cases, but our method also
contains all possible Boolean functions with 4 inputs.

\section{New benchmark: compared methods\label{sec:compared-methods}}

In the following sections, we carry out experiments to measure \ac{WADE} on
the benchmark described in Table~\ref{tab:all-tasks} and introduced in section~\ref{sec:descr-tasks-benchm}. We compare several
baseline models to understand their learning efficiency: three of the most common
sequential machine learning models for which all parameters are trained, RNNs, LSTMs, and transforms, with two methods using reservoir computing: the echo-state networks
(ESN, with a random RNN reservoir) and reservoir cellular automata (ReCA, with a
cellular automaton reservoir) for which only a fraction of the parameters are
learned. Table~\ref{tab:methods-summary} presents all the methods we study in
our experiments.

\begin{table}[htbp]
  \centering
  \begin{tabular}{lll}
    \toprule
    \multicolumn{2}{l}{\bfseries Methods} & \bfseries Type \\
    \midrule
   RNN & Recurrent neural networks  & \multirow{3}{*}{Fully trained}\\
   LSTM & Long-short term memory networks  & \\
   Transformer &  & \\
    \midrule
   ESN & Echo-state networks  & \multirow{2}{*}{Reservoir-based}\\
   ReCA & Reservoir Cellular Automata  & \\
    \bottomrule
  \end{tabular}
  \caption{Summary of the compared models: three are fully trained, and two are
    reservoir-based.}
\label{tab:methods-summary}
\end{table}

In all our tasks, the input data is assumed to be categorical and sequential.
Tokens are observed one by one in sequence, we define this input as
$X = [X_{1}, \ldots, X_{t}, \ldots],\; t\in \mathbb{N}, \; \forall t \ X_{t} \in \mathcal{X} \subset \mathbb{N}$, where $t$ is the
time index of the sequence, each $X_{i}$ is a token corresponding to time index
$i$, and $\mathcal{X}$ is the set of numbered input categories --- or different tokens in the
vocabulary. First, each categorical input vector is one-hot encoded into a
vector of size $L$, where $L = |\mathcal{X}|$ is the size of the input vocabulary. We define
$\vx_{t}$ as the encoded vector form of $X_{t}$, and we have
$\forall t,\; \vx_{t} \in {\{0, 1\}}^{L}$, with $\sum_{i = 1}^{L}{(\vx_{t})}_{i} = 1$.
Since the tasks are designed to be generally compatible with language modeling,
the input and output vocabularies are the same.

\subsection{Fully trained sequential models (RNN, LSTM and Transformer)}
We first study two standard supervised recurrent models: (i) an Elman recurrent
neural network (RNN) \parencite{elmanFindingStructureTime1990} with $\tanh$
activation functions trained with backpropagation through time. (ii) A
long-short term memory (LSTM) recurrent neural network
\parencite{hochreiterLongShortTermMemory1997}, also trained with backpropagation
through time. (iii) An encoder-only transformer model with positional encoding
\parencite{vaswaniAttentionAllYou2017}.

The three models are trained with a batched Adam optimization algorithm
\parencite{kingmaAdamMethodStochastic2015} to minimize a cross-entropy loss function
between the predicted and target tokens. No other training device, such as
dropout, regularization, or normalization, is used for these fully trained baselines.

\subsection{Echo-state networks and reservoir cellular automata\label{sec:echo-state-networks}}

In our experiments, we also use two reservoir based models. The first is an echo
state network (ESN). We compare this model with reservoir cellular automata
(ReCA, see Sections \ref{sec:app-ca-res} and \ref{sec:rc-and-reca} for more
details).

\subsection{Experimental set-up}
We ran 100 separate experiments with different random seeds for each of the CA
rules, the RNN, LSTM, Transformer, and ESN on each task in the benchmark. These experiments
have a separate input projection matrix for the CA, different random weights
for the ESN, and different weight initialization for the supervised baselines.
The task inputs are also generated from a new seed for every experiment, but
we reuse the seeds for the same experiment on different models to ensure they
were trained with the same data. Intervals of one standard deviation for these multiple
experiments are reported in the result graphs.
The code to reproduce our experiments is
available on GitHub\footnote{\url{https://github.com/hugcis/benchmark_learning_efficiency}}.

\subsubsection{Training parameters}
For each experiment, we generate 1200 random examples from the task generator.
We split this set randomly into a training set with 80\% of the data --- 960
examples --- and a test set with the remaining 240 examples. The reservoir is run
on each training example for the reservoir-based models, which creates the
input features for training the decoder. The sequential supervised models use
batches of single sequences with the Adam algorithm
\parencite{kingmaAdamMethodStochastic2015}. They are trained for ten epochs in total. With
reservoir models, only the last linear layer (the decoder) is trained. We minimize the
cross-entropy loss with stochastic gradient descent (SGD), doing only a single
pass over the 960 training examples.

Every few training steps, we generate the output predictions on the testing set,
decode it, and compute the test accuracy for our WADE metric. Supervision is only
applied on tokens that can be predicted --- similar to masked language modeling,
\eg only the answer token is used in the symbol counting or question answering
tasks. Accuracy is also computed for these symbols only.

\section{Human performance evaluation}
\label{sec:human}
The proposed benchmark is relatively simple to understand, and hence one may ask
what human performance would be and what WADE values this would correspond
to. In particular, language tasks appear to be readily solvable from
one's understanding of the English language. However, similar to how words are
encoded as vectors before being processed by a neural network, we need to obfuscate the tokens to make the task equivalent. This
could be achieved in a number of ways. For example, if we map all available
tokens to random letters of the alphabet, the apparently trivial task
{\small \begin{align*}
          & \texttt{I DO NOT SEE PAUL . DO I SEE PAUL ? NO} \\
          & \texttt{I HEAR JAMES BUT I DO NOT HEAR PAUL AND JOHN . DO I HEAR JAMES ? YES}
\end{align*}}

would become significantly less obvious to humans if written
{\small
\begin{align*}
  & \texttt{w K k A D r K w A D l W} \\
  & \texttt{w Z t g w K k Z D G C r K w Z t l H}.
\end{align*}
}

One would need to read through several of these sentences to identify patterns
such as the fact that \texttt{W} and \texttt{H} represent Yes or No or that
\texttt{r} plays the role of a full stop.

We apply the following procedure in our human evaluation experiments:
\begin{enumerate}
  \item Choose a random task among 10.
  \item Apply a random mapping from the task token to letters of the alphabet.
  \item Present the sequences one by one with the tokens to be predicted hidden.
  \item The user enters his predicted answer.
  \item The valid sequence is shown so that the user can learn from their mistakes.
  \item Steps 3-5 are repeated until 10 sequences are correctly solved in a row.
  Then we go back to step 2 with a new task.
\end{enumerate}
We add the requirement that no external device such as pen and paper or a
note-taking program should be used during the experiment, so the user relies
solely on their memory. The accuracy is computed by counting the number of right
answers in a row. For example, three right answers in a row would correspond to
an accuracy of 30\%, while ten right answers in a row are 100\%, which also
corresponds to a change of task. This accuracy score is attributed to the
first correct answer of the series of correct answers. This means that a series
of five correct answers, starting from Question 6 to Question 11, will
correspond to reaching the accuracy of 50\% in Question 6. This ensures that ten
correct answers on the first attempt yield a WADE score of 1.

\section{Experimental parameters}
\label{sec:parameters}
We report all the parameters used in our experiments in table \ref{tab:appendix-params}:
\begin{table}[htbp]
  \centering
  \begin{tabular}{p{.38\linewidth}p{.2\linewidth}p{.32\linewidth}}
    \toprule
    & \bfseries Reservoir-based & \bfseries Fully trained \\
    \midrule
    \bfseries Total number of sequences per task & 1200 & 1200 \\
    \bfseries Number of training sequences & 960 & 960 \\
    \bfseries Number of testing sequences & 240 & 240 \\
    \bfseries Passes over the data (epochs) & 1 & 10 \\
    \bfseries Random runs per task & 100 & 100\\
    \bfseries Algorithm & SGD & Adam \parencite{kingmaAdamMethodStochastic2015}\\
    \bfseries Learning rate & $0.001$ & $0.001$ \\
    \bfseries Regularization & weight decay $0.001$ & None \\
    \bfseries Internal state (hidden) size & $1800$ & $h$ task-dependent \\
    \bfseries Number of trainable parameters & $1800 \times \text{dictionary\_size}$ &
      $h^{2} + 2 \times h \times \text{dictionary\_size}$ for the RNN\textsuperscript{2}\\
    \bfseries Output non-linearity & Softmax & Softmax \\
    \bfseries Internal non-linearity & Not applicable & $\tanh$ \\
    \bfseries Batch size  & 1 & 1 \\
    \bottomrule
  \end{tabular}
  \caption{Experimental parameters common to all tasks. Some of the sizes (such
    as $h$) vary from task to task.}
  \label{tab:appendix-params}
\end{table}

\subsection{Task generation parameters\label{sec:task-gen-params}}
The tasks are generated according to the procedures described in
Section~\ref{sec:tasks}. They have some parameters that allow the
difficulty of the task to be adjusted. We report results on ten tasks. The parameters
chosen for our experiments are listed below:

\begin{description}
  \item[Periodic (1) and Increasing period (2):] Sequences are generated from patterns of
        length between 1 and 10.
  \item[Easy symbol counting (3):] Available symbols are \texttt{A}, \texttt{B} and
        \texttt{C}. Generated sequences have between 1 and 10 of these symbols
        in the prompt, and the query has either one, two, or all three symbols.
  \item[Hard symbol counting (4):] Available symbols are the same as for the previous
        task. Generated sequences have between 1 and 45 symbols
        with separators in the prompt. There is a minimum of 1 query and can be
        as many queries as there are different patterns in the prompt.
  \item[Question answering (5):] There are five available names and two verbs. The
        prompt consists of between one and five names, and the question is about one
        of these names.
  \item[Harder question answering :] There are eleven available names and five
        verbs. The prompt is made of between one and five names, and the
        question is about one of these names.
  \item[Question answering with world (7), with counting (8):] There are thirteen available
        names and seven verbs.
  \item[Adjective question answering (9), with counting (10):] There are eight
        available names, six verbs, four color adjectives, and five size
        adjectives. The prompt consists of between one and six statements, and
        there are up to eight questions.
\end{description}

We chose relatively small values of these parameters because we are interested in simple tasks
so that models can learn from less than 100 examples. Moreover, the
number of possible sentences generated from these small values is already huge,
with more than $10^{35}$ possible sentences for task 4, for example. These
parameters could also be varied dynamically to change the task's difficulty when
needed or turn each task into multiple subtasks or levels of difficulty. We
chose the ten separate tasks/settings pairs listed above to get a broad overview of our
benchmarked models on a range of task difficulties.
\footnotetext{$h$ is chosen to match the number of parameters of the reservoir-based methods.}

\section{Results}

We report the final weighted average data efficiency (WADE) scores on our
benchmark in Table \ref{tab:summary} and the accuracy results for comparison in Table~\ref{tab:accuracy_all}. We include the best
elementary reservoir cellular automaton (ReCA) and the echo
state network (ESN) with the same internal state size for each task.

\begin{table}[htbp]
  \centering
    \begin{tabular}{p{7cm}ccc}
      \toprule
      \multirow{2}{*}{\bfseries Task ID - Name} & \multicolumn{2}{c}{Reservoir}
      & \multirow{2}{*}{\bfseries \shortstack{Human \\ subject}} \\

& \bfseries ReCA & \bfseries ESN &\\
\midrule

\bfseries 1 - Periodic & \bfseries 0.78 ${\scriptscriptstyle \pm0.05 }$ &  0.74 ${\scriptscriptstyle \pm0.05 }$  & 1.00 \\
\bfseries 2 - Incremental periodic &  0.57 ${\scriptscriptstyle \pm0.02 }$ & \bfseries 0.72 ${\scriptscriptstyle \pm0.02 }$ & 1.00 \\
\bfseries 3 - Symbol counting & \bfseries 0.06 ${\scriptscriptstyle \pm0.01 }$ & 0.04 ${\scriptscriptstyle \pm0.01 }$ & 0.11 \\
\bfseries 4 - Pattern counting &  0.12 ${\scriptscriptstyle \pm0.04 }$ & \bfseries 0.14 ${\scriptscriptstyle \pm0.03 }$ & 0.09 \\
\bfseries 5 - Basic question answering (QA) & \bfseries 0.31 ${\scriptscriptstyle \pm0.08 }$ &  0.26 ${\scriptscriptstyle \pm0.03 }$ & 0.15 \\
\bfseries 6 - Harder QA & \bfseries 0.35 ${\scriptscriptstyle \pm0.11 }$ &  0.26 ${\scriptscriptstyle \pm0.03 }$ & 0.12 \\
\bfseries 7 - QA with world def. & \bfseries 0.32 ${\scriptscriptstyle \pm0.10 }$ &  0.27 ${\scriptscriptstyle \pm0.04 }$ & --- \\
\bfseries 8 - QA with world def. \& counting &  0.09 ${\scriptscriptstyle \pm0.03 }$ & \bfseries 0.14 ${\scriptscriptstyle \pm0.06 }$ & --- \\
\bfseries 9 - Adjective QA &  0.04 ${\scriptscriptstyle \pm0.03 }$ &  0.04 ${\scriptscriptstyle \pm0.04 }$ & --- \\
\bfseries 10 - Adjective QA \& counting &  0.04 ${\scriptscriptstyle \pm0.02 }$ & \bfseries 0.06 ${\scriptscriptstyle \pm0.02 }$ & --- \\
\midrule
& \multicolumn{3}{c}{Fully supervised}  \\
& \bfseries RNN & \bfseries LSTM & \bfseries Transformer \\
    \midrule
     \bfseries 1 \dotfill &  0.28 ${\scriptscriptstyle \pm0.07 }$  &  0.31 ${\scriptscriptstyle \pm0.04 }$&  0.31 ${\scriptscriptstyle \pm0.06 }$\\
      \bfseries 2 \dotfill&  0.32 ${\scriptscriptstyle \pm0.11 }$  &  0.49 ${\scriptscriptstyle \pm0.15 }$&  0.42 ${\scriptscriptstyle \pm0.16 }$ \\
      \bfseries 3 \dotfill&  0.04 ${\scriptscriptstyle \pm0.02 }$  &  0.03 ${\scriptscriptstyle \pm0.01 }$&  0.05 ${\scriptscriptstyle \pm0.02 }$ \\
      \bfseries 4 \dotfill&  0.10 ${\scriptscriptstyle \pm0.05 }$  &  0.08 ${\scriptscriptstyle \pm0.03 }$&  0.13 ${\scriptscriptstyle \pm0.04 }$ \\
      \bfseries 5 \dotfill&  0.17 ${\scriptscriptstyle \pm0.08 }$  &  0.16 ${\scriptscriptstyle \pm0.07 }$&  0.20 ${\scriptscriptstyle \pm0.06 }$ \\
      \bfseries 6 \dotfill&  0.16 ${\scriptscriptstyle \pm0.07 }$  &  0.12 ${\scriptscriptstyle \pm0.06 }$&  0.24 ${\scriptscriptstyle \pm0.05 }$ \\
      \bfseries 7 \dotfill&  0.17 ${\scriptscriptstyle \pm0.07 }$  &  0.11 ${\scriptscriptstyle \pm0.06 }$&  0.24 ${\scriptscriptstyle \pm0.05 }$ \\
      \bfseries 8 \dotfill&  0.10 ${\scriptscriptstyle \pm0.05 }$ &  0.05 ${\scriptscriptstyle \pm0.02 }$  & 0.02 ${\scriptscriptstyle \pm0.01 }$ \\
      \bfseries 9 \dotfill& \bfseries 0.05 ${\scriptscriptstyle \pm0.03 }$ &  0.03 ${\scriptscriptstyle \pm0.02 }$ &  0.02 ${\scriptscriptstyle \pm0.01 }$\\
      \bfseries 10 \dotfill&  0.05 ${\scriptscriptstyle \pm0.02 }$ &  0.03 ${\scriptscriptstyle \pm0.02 }$  &  0.01 ${\scriptscriptstyle \pm0.01 }$\\
      \bottomrule
    \end{tabular}
    \caption{Comparison of WADE scores (also shown in parentheses in the legend
      in figure~\ref{fig:all_metrics}, higher is better) of the best cellular
      automaton rules \parencite[\textbf{ReCA},
      ][]{yilmazReservoirComputingUsing2014} for each task against an echo-state
      network \parencite[\textbf{ESN},][]{jaegerEchoStateApproach2001}, a
      \textbf{RNN}, a \textbf{LSTM} \parencite{hochreiterLongShortTermMemory1997},
      and a
      \textbf{Transformer} \parencite{vaswaniAttentionAllYou2017} with the same number of parameters.
      The dashed line ``---'' indicates that the task was too difficult to complete from memory alone for the human subject. The dots indicate that the task name is the same as in the first line. Accuracy scores are also reported in Table \ref{tab:accuracy_all}.}\label{tab:summary}
\end{table}

\begin{figure}[htbp]
  \centering
  \includegraphics[width=.98\linewidth]{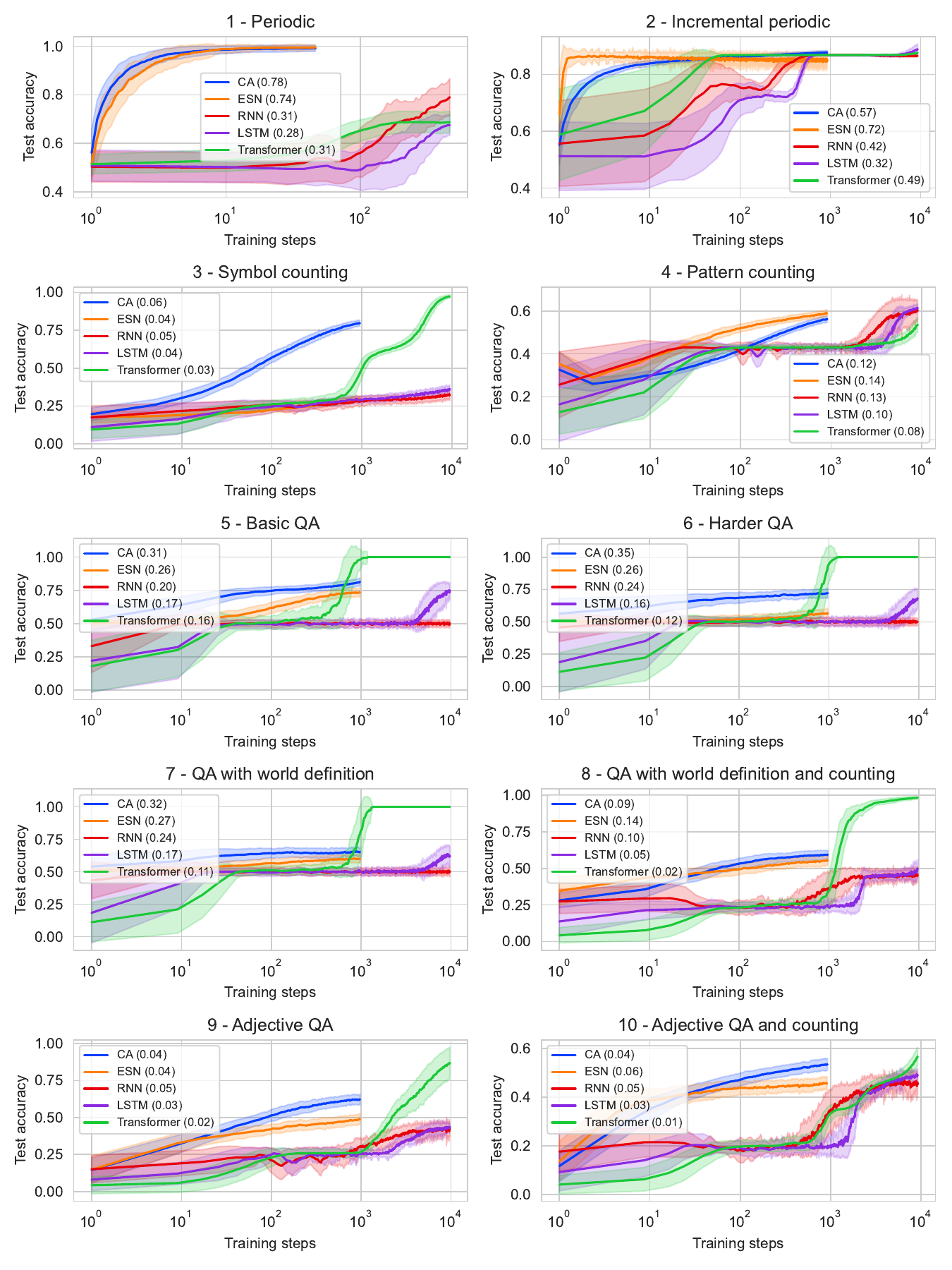}
  \caption{Average learning curves and WADE scores (shown in parentheses in the
    legend) for each task in the benchmark. The dark blue curves represent the
    best elementary cellular automata (CA) rule, and the green curves represent
    the echo-state networks (ESN). Note that the x-axis is logarithmic, showing
    ten times more training steps for the RNN and LSTM. Shaded areas represent
    one standard deviation around the average over the 100 different
    experiments.}
  \label{fig:all_metrics}
\end{figure}

Figure \ref{fig:all_metrics} shows the learning curves for the best reservoir cellular
automaton, as well as the echo-state network, the recurrent neural network (RNN), LSTM, and the Transformer for all ten tasks. Interestingly, the reservoir-based models are consistently more efficient
learners than the fully supervised methods, reaching better accuracy in much
fewer training steps. For example, for tasks 5, 6, and 7, the LSTM needs ten
times more steps to approach the accuracy that the reservoir CA model reached in less
than 1000 training steps.

Echo-state networks (ESN) and reservoir cellular automata (ReCA) appear to learn
at similar rates, with no clear advantage for one or the other, as each model
outperforms the others in five tasks out of ten. On tasks 2, 4, 8, and 10, the ESN
is visibly faster than the ReCA (see curves Figure \ref{fig:all_metrics}), which
explains the ESN's higher WADE scores even though the ReCA reaches higher accuracy
values after several more training steps.

\begin{table}[htbp]
  \centering
    \begin{tabular}{p{7cm}cccccc}
      \toprule
      \multirow{2}{*}{\bfseries Task ID - Name} & \multicolumn{6}{c}{Reservoir} \\
& \multicolumn{3}{c}{\bfseries ReCA}& \multicolumn{3}{c}{\bfseries ESN} \\
\midrule
\bfseries 1 - Periodic & \multicolumn{3}{c}{0.99 ${\scriptscriptstyle \pm 0.01}$} & \multicolumn{3}{c}{\bfseries 1.00 ${\scriptscriptstyle \pm 0.01}$}  \\
\bfseries 2 - Incremental periodic & \multicolumn{3}{c}{0.88 ${\scriptscriptstyle \pm 0.01}$} & \multicolumn{3}{c}{0.87 ${\scriptscriptstyle \pm 0.03}$}  \\
\bfseries 3 - Symbol counting & \multicolumn{3}{c}{0.80 ${\scriptscriptstyle \pm 0.02}$} & \multicolumn{3}{c}{0.30 ${\scriptscriptstyle \pm 0.02}$}  \\
\bfseries 4 - Pattern counting & \multicolumn{3}{c}{0.56
                                 ${\scriptscriptstyle \pm 0.01}$} & \multicolumn{3}{c}{ 0.59 ${\scriptscriptstyle \pm 0.02}$}\\
\bfseries 5 - Basic QA & \multicolumn{3}{c}{0.81 ${\scriptscriptstyle \pm 0.03}$} & \multicolumn{3}{c}{0.73 ${\scriptscriptstyle \pm 0.03}$} \\
\bfseries 6 - Harder QA & \multicolumn{3}{c}{0.72 ${\scriptscriptstyle \pm 0.04}$} & \multicolumn{3}{c}{0.57 ${\scriptscriptstyle \pm 0.03}$}\\
\bfseries 7 - QA with world def. & \multicolumn{3}{c}{0.66 ${\scriptscriptstyle \pm 0.03}$} & \multicolumn{3}{c}{0.60 ${\scriptscriptstyle \pm 0.03}$}\\
\bfseries 8 - QA with world def. \& counting & \multicolumn{3}{c}{0.59 ${\scriptscriptstyle \pm 0.03}$} & \multicolumn{3}{c}{0.56 ${\scriptscriptstyle \pm 0.03}$}\\
\bfseries 9 - Adjective QA & \multicolumn{3}{c}{0.62 ${\scriptscriptstyle \pm 0.04}$} & \multicolumn{3}{c}{0.49 ${\scriptscriptstyle \pm 0.03}$}\\
\bfseries 10 - Adjective QA \& counting & \multicolumn{3}{c}{0.54 ${\scriptscriptstyle \pm 0.02}$} & \multicolumn{3}{c}{0.46 ${\scriptscriptstyle \pm 0.03}$} \\
\midrule
      \multirow{2}{*}{} & \multicolumn{6}{c}{Fully supervised} \\
& \multicolumn{2}{c}{\bfseries RNN}& \multicolumn{2}{c}{\bfseries LSTM}  & \multicolumn{2}{c}{\bfseries Transformer} \\
      \midrule
\bfseries 1 \dotfill & \multicolumn{2}{c}{0.79 ${\scriptscriptstyle \pm 0.08}$} & \multicolumn{2}{c}{0.68 ${\scriptscriptstyle \pm 0.04}$} & \multicolumn{2}{c}{0.69 ${\scriptscriptstyle \pm 0.04}$} \\
\bfseries 2 \dotfill & \multicolumn{2}{c}{0.87 ${\scriptscriptstyle \pm 0.00}$} & \multicolumn{2}{c}{\bfseries 0.89 ${\scriptscriptstyle \pm 0.02}$} & \multicolumn{2}{c}{0.88 ${\scriptscriptstyle \pm 0.00}$}\\
\bfseries 3 \dotfill& \multicolumn{2}{c}{0.33 ${\scriptscriptstyle \pm 0.03}$} & \multicolumn{2}{c}{0.36 ${\scriptscriptstyle \pm 0.03}$} &
                                                                            \multicolumn{2}{c}{\bfseries 0.97 ${\scriptscriptstyle \pm 0.01}$} \\
 \bfseries 4  \dotfill    & \multicolumn{2}{c}{0.61 ${\scriptscriptstyle \pm 0.05}$} & \multicolumn{2}{c}{\bfseries 0.61
                                              ${\scriptscriptstyle \pm 0.02}$} &
                                                                              \multicolumn{2}{c}{0.54
                                                                               ${\scriptscriptstyle \pm 0.03}$} \\
 \bfseries 5  \dotfill    & \multicolumn{2}{c}{0.51 ${\scriptscriptstyle \pm 0.02}$} & \multicolumn{2}{c}{0.75 ${\scriptscriptstyle \pm 0.05}$} & \multicolumn{2}{c}{\bfseries 1.00 ${\scriptscriptstyle \pm 0.00}$} \\
      \bfseries 6 \dotfill     & \multicolumn{2}{c}{0.51 ${\scriptscriptstyle \pm 0.03}$} &
                            \multicolumn{2}{c}{0.68 ${\scriptscriptstyle \pm 0.07}$} &
                              \multicolumn{2}{c}{\bfseries 1.00 ${\scriptscriptstyle \pm 0.00}$} \\
   \bfseries 7  \dotfill  & \multicolumn{2}{c}{0.51 ${\scriptscriptstyle \pm 0.04}$} & \multicolumn{2}{c}{0.64 ${\scriptscriptstyle \pm 0.07}$}
                                    & \multicolumn{2}{c}{\bfseries 1.00
                                      ${\scriptscriptstyle \pm 0.00}$}  \\
 \bfseries 8 \dotfill & \multicolumn{2}{c}{0.47 ${\scriptscriptstyle \pm 0.05}$} & \multicolumn{2}{c}{0.50 ${\scriptscriptstyle \pm 0.06}$} &
                                                                             \multicolumn{2}{c}{\bfseries 0.98 ${\scriptscriptstyle \pm 0.01}$}  \\
      \bfseries 9 \dotfill & \multicolumn{2}{c}{0.43 ${\scriptscriptstyle \pm 0.04}$} &
                                    \multicolumn{2}{c}{0.44 ${\scriptscriptstyle \pm 0.03}$}
                                    & \multicolumn{2}{c}{\bfseries 0.87 ${\scriptscriptstyle \pm 0.11}$}   \\
\bfseries 10  \dotfill     & \multicolumn{2}{c}{0.47 ${\scriptscriptstyle \pm 0.04}$} & \multicolumn{2}{c}{0.49 ${\scriptscriptstyle \pm 0.03}$} & \multicolumn{2}{c}{\bfseries 0.57 ${\scriptscriptstyle \pm 0.04}$} \\

      \bottomrule
    \end{tabular}
    \caption{Comparison of accuracy scores (higher is better) of the
      \textbf{ReCA}, \textbf{ESN}, \textbf{RNN}, \textbf{LSTM}, and
      \textbf{Transformer} models with a similar number of parameters. In
      contrast to the results of table~\ref{tab:summary}, the fully-supervised
      models are more performant when we measure the accuracy, as shown here,
      but do not necessarily do as well when we measure the speed of learning,
      as shown in table~\ref{tab:summary}. }\label{tab:accuracy_all}
\end{table}

Even if they lack slightly in accuracy, as seen for example on the curves of task 4,
the two reservoir-based methods consistently outperform the fully supervised
sequential methods (RNNs and LSTMs) in terms of learning speeds. This better
learning efficiency could be explained by the internal state structure
introduced by the CA rules and the special form of the ESN random matrix which
favors memory retention whereas usual recurrent network initialization does not
--- we initialize the weights of the fully trained models from a uniform
distribution $\mathcal{U}( - \sqrt{h^{-1}} ,\sqrt{h^{-1}})$ in our experiments, where $h$
is the hidden size.

The human subject scores are estimated based on the authors' performance on
obfuscated versions of the tasks --- all input tokens are mapped to random
symbols. These scores may appear surprisingly low in some of these experiments.
Although the task examples presented in Section~\ref{sec:descr-tasks-benchm}
seem easy to understand. We rely heavily on our prior knowledge about the
symbols to understand the patterns --- the words in tasks 5--10 or the numbers in
tasks 3 and 4. These symbols are randomly remapped in our human experiments, and
more examples, as well as a good memory, are needed to understand the tasks and
learn the mapping itself, hence the slower learning.

It is interesting to note that RNNs also seem to perform better than LSTMs and
Transformers at the beginning of the training in all tasks but the first two.
Even though the Transformer is nowadays generally considered a superior model,
vanilla RNNs may still remain competitive in the very low data and computation
regimes. This is especially pronounced for tasks 8, 9, and 10, where the RNN
test accuracy curve starts increasing significantly 1000 steps before the LSTM.
After several epochs, the Transformer still outperforms other alternatives in
terms of test accuracy on most tasks.

We note that we implicitly assume a fixed cost per training iteration in our
experiments and that each training example is seen once. Without these
requirements, one could achieve higher WADE results at the cost of additional
computations and memory usage by using \emph{replay}-inspired methods
\parencite{hintonUsingFastWeights1987, robinsCatastrophicForgettingNeural1993a,
  gepperthBioInspiredIncrementalLearning2016,
  rebuffiIcarlIncrementalClassifier2017} that retrain the models with past
stored inputs. The results in Table~\ref{tab:summary} used each input sequence
once.

\section{Reservoir cellular automata (ReCA) results}

We present some detailed results from our experiments with reservoir cellular
automata. First, general statistics about \ac{ECA} performance on the five first
tasks (tasks with ID 1 to 5).

\begin{figure}[htbp]
  \centering
  \includegraphics[width=.8\linewidth]{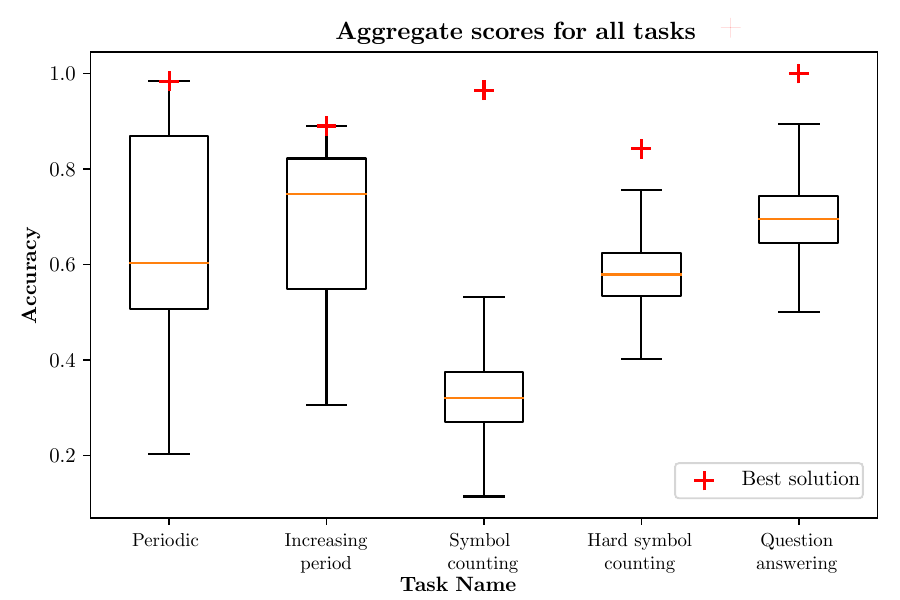}
  \caption{Average test accuracies across rules, hyperparameters, and runs for
    each task. The best solution for each task is shown in
    red.}\label{fig:aggregate-results}
\end{figure}

Fig.~\ref{fig:aggregate-results} shows a summary of the accuracy scores for each
task. Score distributions vary a lot from task to task because of the different
output spaces (binary for Periodic, Inc-per, and QA and multi-modal for the
others).

Next, we give a more in-depth analysis of the best performing \acp{ECA} on the
first two tasks.

\subsection{Binary sequences}

The binary sequence task should be the simplest to solve. Since it requires some
memory, we expect some CA rules with very chaotic behavior --- like ECA rule 30
--- to fail at it. We observed that the best rules are consistent with the
periodic and ordered types.

\begin{figure}[htbp]
  \centering
  \includegraphics[width=.7\linewidth]{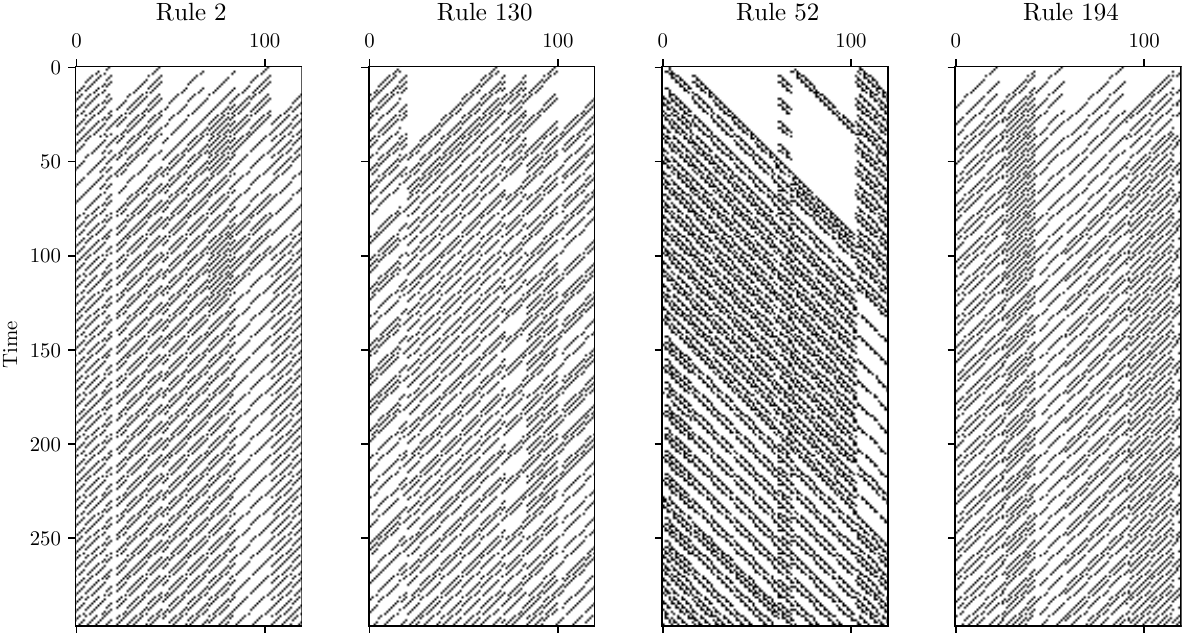}
  \caption{Top performing rules and hyperparameters for the binary periodic
    sequence task. All of the best rules seem to make use of a lateral
    translation, which is an effective memory mechanism. Rule 2, 130 and 194
    only differ by 1 transition in their rule
    function.}\label{fig:top-binary-seq}
\end{figure}

The four best rules displayed in figure \ref{fig:top-binary-seq} all implement
a similar mechanism, using translations of cells from one column to the next at
every step. This translation is a memory mechanism, because it shifts any cell
that was turned on in reaction to input before it could be overwritten by a
subsequent input that would be mapped to the same grid position.

\begin{table}[htbp]
  \centering
    \begin{tabular}{p{4cm}ccccc}
      \toprule
      & \multicolumn{5}{c}{\bfseries Top 10 rules}\\
      \midrule
      \bfseries Rule & 38 & 24 & 231 & 152 & 66 \\
      \bfseries Average accuracy & 0.969 & 0.97 & 0.97 & 0.97 & 0.97 \\
      \midrule
      \hrulefill & 189 & 194 & 52 & 130 & 2 \\
      \hrulefill & 0.97 & 0.97 & 0.97 & 0.971 & 0.971\\
      \bottomrule
    \end{tabular}
  \caption{Top 10 rules for the periodic binary sequence task (ordered left to
    right) --- best hyper-parameter combination. Many rules get close to the best
    possible accuracy.}\label{tab:top_periodic_rules}
\end{table}

\subsection{Symbol counting}

The easy version of this task yielded the most surprising results: some CA rules
unexpectedly reach very good scores. These top rules are reliably found to have
the best accuracy scores, even when using powerful decoders. The best performing
rule of all the possible \acp{ECA} is rule 37.

\begin{figure}[htbp]
  \centering
  \includegraphics[width=\linewidth]{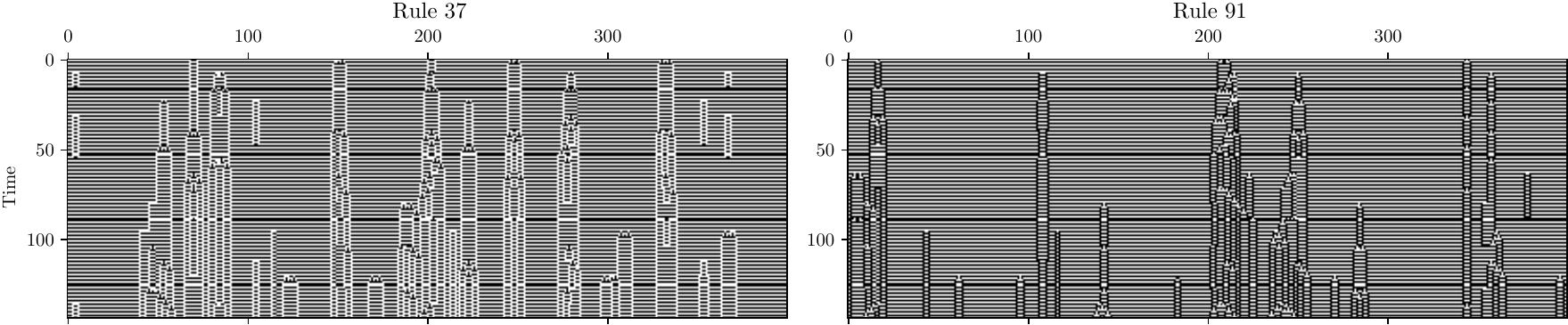}
  \caption{The two best performing rules for the easy symbol counting task (.82
    average accuracy).}\label{fig:best_sym_ct}
\end{figure}

As shown in Figure \ref{fig:best_sym_ct}, the two best rules seem to employ a
similar ``computing mechanism'' to solve the task. They seem to grow tree-shaped
structures that sometimes become wider. One hypothesis for the success of these
two rules is that they encode information about the number of symbols observed in the width of these tree structures. \ac{RC} is useful for this kind
of analysis because it has some level of interpretability. The weights of the
output decoding layer can inform us of which of the components of the \acp{CA}
internal state are the most important for the task being solved. However, even
with this tool, it can still be hard to disentangle the effect of each
component and their interactions.

\begin{table}[htbp]
  \centering
    \begin{tabular}{p{4cm}ccccc}
      \toprule
      & \multicolumn{5}{c}{\bfseries Top 10 rules}\\
      \midrule
      \bfseries Rule & 28 & 92 & 220 & 70 & 198\\
      \bfseries Average accuracy & 0.571 & 0.572 & 0.573 & 0.574 & 0.582 \\
      \midrule
      \hrulefill & 156 & 78 & 123 & 91 & 37\\
      \hrulefill & 0.594 & 0.598 & 0.643 & 0.721 & 0.807\\
      \bottomrule
    \end{tabular}
  \caption{Top 10 rules for the easy symbol counting task (ordered left to
    right) --- best hyper-parameter combination. Notice the three best rules are
    largely better than most of the others.}\label{tab:top_sym_ct_rules}
\end{table}

\section{Conclusions}

Learning speed and data efficiency are essential components of any learning
system. Our learning speed metric can offer a novel perspective on several
machine learning models. Instead of focusing on pure performance, measuring and
comparing the data efficiency of different models will hopefully lead to better
systems for online and continual learning.

Even with the right metric, it is difficult to thoroughly evaluate the ability of a model to learn efficiently. Our benchmark evaluates a range of problem
difficulties which would be challenging to construct by combining or
manipulating existing datasets. However, the tasks remain easy to understand and
use. Since the tasks are language-based, they can be further mixed or chained to
create continuous learning problems and may also be easily extended in a
follow-up work.

We study lesser-known machine learning models based on evolving states of
complex systems that can learn through self-organization. A complex dynamical
system comprises many interacting agents and evolves over time according to a
fixed update rule. These agents can be hidden neurons of a \ac{RNN} or nodes in
a graph. Such systems often exhibit emergent global dynamics resulting from the
actions of its parts rather than the decisions of a central controller. These
dynamics lead to surprisingly complex behavior, which can be random, chaotic, or
may lead to unbounded growth of complexity
\parencite{boccaraModelingComplexSystems2010}. Due to these
properties,
reservoir computing systems may be a promising alternative that addresses
the shortcomings of standard supervised models.

Surprisingly, such models achieve remarkable learning efficiency compared to the
more standard sequentially supervised models trained with stochastic
gradient-based learning. Complex systems-based models consistently
outperform sequential supervised methods and even achieve better learning
efficiency than humans on some tasks. They demonstrate more efficient learning
on our benchmark at a fraction of the computational and data cost of the
conventional models. We believe that more advanced models of this type could
lead to more robust and data-efficient machine learning in the future,
especially in low-data applications or problems where supervision is limited.
Complex systems are underexplored and seem worth further investigating for
building the next generation of learning algorithms.

\chapter{Conclusion}\label{cha:conclusion}

In this chapter, we summarize our contributions in the thesis.

\section{Contributions}

In this thesis, we have developed some tools to better understand and make use
of the computations that occur within complex systems.
We summarize our contributions in the following.

\begin{itemize}
  \item In Chapter \ref{cha:background}, we gave an overview of the \acf{CA}
        model and the specific challenges it poses. We review the deep
        connection between \acp{CA} and \acfp{RNN} and how this can be used to apply
        \acp{CA} in \acf{RC}. Next, in Chapter \ref{cha:literature-review}, we
        place our work within the vast body of literature on complex systems,
        complexity, emergence, and learning.

  \item In Chapter \ref{cha:meas-compl-evolv}, we define a novel metric of
        complexity for complex systems. The metric is inspired by principles of
        compression and algorithmic complexity. It uses the ability of small
        neural networks to learn a model of the local structures in a system,
        and the evolution of the predictive power of that model as the system
        evolves over time. We evaluated the quality of this metric by measuring
        its correlation with human perception of complexity. We find a
        higher correlation with human annotations than alternative methods on a
        dataset of \acp{CA} rules labeled as complex or not. We use the metric
        to discover new \acp{CA} rules with surprisingly complex behavior
        semi-automatically.

  \item In Chapter \ref{cha:visu-comp-large}, we built on the work of Chapter
        \ref{cha:meas-compl-evolv} to work on the specific challenges posed by
        working with large-scale systems such as cellular automata. In these
        systems, qualitatively different behavior may emerge at various scales.
        To allow dealing with these large systems and apply complexity measures
        and classification methods on a range of scales, we developed three
        coarse-graining methods based on simple statistical analysis, clustering
        algorithms and autoencoder neural networks that can reduce the size of
        a system while retaining useful information.

  \item In Chapter \ref{cha:learn-effic-compl}, we tackled the issue of using
        complex systems for general-purpose task solving, using the \ac{RC}
        paradigm. We also developed a metric for the speed of learning of
        machine learning systems. Using that metric, we looked at the data
        efficiency of some well-known algorithms rather than their task-based
        performance only. Surprisingly, \ac{RC} models using random frozen
        \acp{RNN} or \acp{CA} are significantly more efficient than other
        alternatives on a range of tasks. We evaluated this on some standard
        language datasets and introduced our own dataset of progressively more
        complex tasks. Efficiency is a crucial property in low data and compute
        settings, and our work showed that some overlooked methods may actually
        be competitive in these situations.
\end{itemize}

\include{future_work}

\setstretch{1.5}
\setlength\bibitemsep{0.5\baselineskip}
\addcontentsline{toc}{chapter}{Bibliography}
\printbibliography

\end{document}